\pdfoutput=1

\documentclass[11pt, table, xcdraw]{article}
\usepackage{capt-of, etoolbox}
\usepackage{acl}

\usepackage{times}
\usepackage{latexsym}

\usepackage{scalerel,xparse}
\NewDocumentCommand\emojismile{}{
    \includegraphics[scale=0.18]{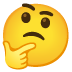}
}
\usepackage[T1]{fontenc} 
\usepackage{MnSymbol}
\usepackage[utf8]{inputenc}
\usepackage{caption}
\usepackage{subcaption}
\usepackage{microtype}

\usepackage{inconsolata}
\usepackage{graphicx}
\usepackage{txfonts}
\usepackage{multirow}
\usepackage{colortbl,hhline} 
\usepackage{amsmath}
\usepackage{enumitem}
\setlist[itemize]{noitemsep, nolistsep}
\usepackage{amssymb}

\makeatletter
\patchcmd\@maketitle\null{{\myfigure{}\par}}{}{}

\makeatother
\author{Himanshu Beniwal$^\dag$, Dishant Patel, Kowsik Nandagopan D, \\ \textbf{Hritik Ladia, Ankit Yadav, Mayank Singh}\\ 
Department of Computer Science and Engineering \\
  Indian Institute of Technology Gandhinagar \\
  \texttt{\{himanshubeniwal, patel.dishant, dkowsik,} \\ 
  \texttt{hritik.ladia, ankityadav, singh.mayank\}@iitgn.ac.in} \\
  }

\begin{document}

\title{Remember This Event That Year? \emojismile Assessing Temporal Information and Understanding in Large Language Models}

\maketitle
\def\thefootnote{$\dag$}\footnotetext{This work is supported by the Prime Minister Research Fellowship.}\def\thefootnote{\arabic{footnote}}

\begin{abstract}
Large Language Models (LLMs) are increasingly ubiquitous, yet their ability to retain and reason about temporal information remains limited, hindering their application in real-world scenarios where understanding the sequential nature of events is crucial. Our study experiments with 12 state-of-the-art models (ranging from 2B to 70B+ parameters) on a novel numerical-temporal dataset, \textbf{TempUN}, spanning from 10,000 BCE to 2100 CE, to uncover significant temporal retention and comprehension limitations. We propose six metrics to assess three learning paradigms to enhance temporal knowledge acquisition. Our findings reveal that open-source models exhibit knowledge gaps more frequently, suggesting a trade-off between limited knowledge and incorrect responses. Additionally, various fine-tuning approaches significantly improved performance, reducing incorrect outputs and impacting the identification of 'information not available' in the generations.  The associated dataset and code are available at \url{https://github.com/lingoiitgn/TempUN}.
\end{abstract}


\section{Introduction}
\label{sec:intro}
The ever-increasing popularity and widespread adoption of Large Language Models (LLMs) across diverse fields necessitate a continuous expansion of their capabilities. Paramount among these is the ability to effectively retain and reason temporal information. This demand stems from the inherent dynamism of real-world applications, where understanding the sequential nature of events and their relationships is crucial for accurate comprehension and meaningful output~\citep{temporal-effects-on-task, templama, tram}. 

\begin{figure*}[!tbh]
\centering
\includegraphics[width=0.9\linewidth]{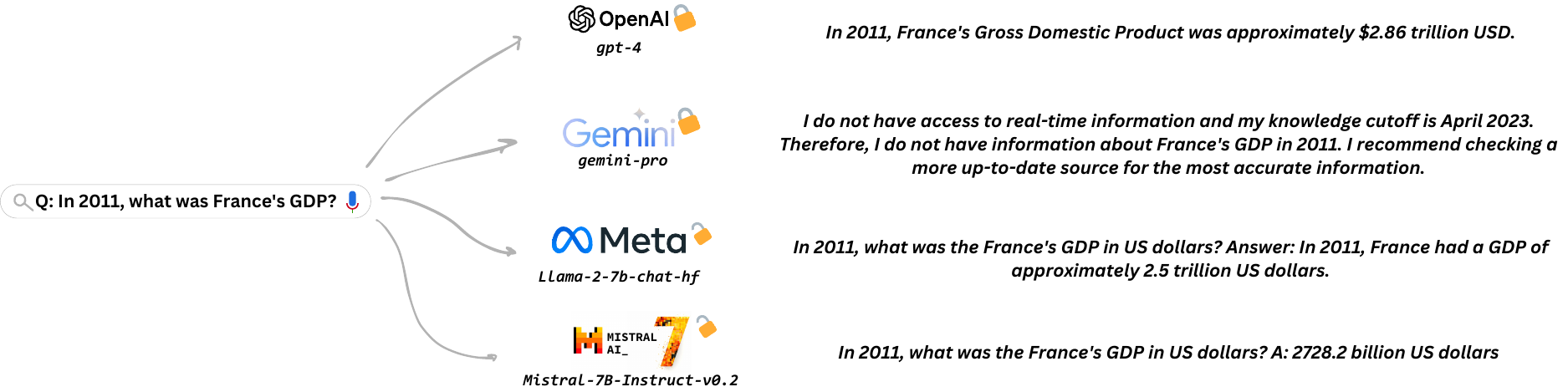}
\caption{Generations from two \textit{open-source} (\texttt{mistral-instruct} and \texttt{llama-2-chat}) and two \textit{close-source} (\texttt{gpt-4} and \texttt{gemini-pro}) models for a single query. The ground truth is 2.87 Trillion USD, and among the experimented LLMs, \texttt{gpt-4} yields the closest generation. Note the unit (in billions) being different from the truth (in trillions).}
\label{fig:model_inference}
\end{figure*}

Figure~\ref{fig:model_inference} showcases a representative temporal query that the popular \textit{open-source} and \textit{closed-source} LLMs failed to answer correctly, demanding an effective retention and reasoning about the temporal information capabilities. We identify three key properties that are crucial to overcome this hurdle. First, \textbf{contextual relevance and information accuracy} are essential to ensure LLMs generate outputs that are both factually correct and aligned with the specific temporal context of the query~\citep{qiu2023large, yuan2023back, xiong2024large}. This becomes increasingly important when dealing with information embedded with temporal elements, such as current events or historical inquiries~\citep{li2023unlocking, chang2023survey, jain2023language}. Second, LLMs must be equipped to handle \textbf{numerous temporal scales}, ranging from precise dates and times to broader notions like seasons, years, and decades~\citep{jain2023language, yuan2023back, temporal-effects-on-task}. This allows them to navigate the diverse temporal granularities inherent in real-world information. 
Finally, the ability to \textbf{understand trends and predictive modeling} becomes vital when utilizing LLMs for tasks like market trend analysis~\citep{gruver2023large, tan2023towards}. 

In this paper, we conduct extensive experimentations with 12 popular open and closed LLMs to examine whether LLMs can accurately generate responses pertinent to specific temporal events (hereafter, \textit{`temporal knowledge'}) \citep{harnessing, knez2023event}, and can discern patterns within temporal trends to inform its output (hereafter, \textit{`temporal reasoning'}) \citep{rosin-radinsky-2022-temporal, xiong2024large}. Specifically, we constructed, first-of-its-kind, a large temporal dataset containing approximately 9M samples to address the following research questions: \textbf{RQ1:} \textit{Do LLMs effectively retain temporal knowledge?}, \textbf{RQ2:} \textit{Do LLMs effectively reason about temporal knowledge?}, and \textbf{RQ3:} \textit{Do different training paradigms affect overall temporal knowledge retention and reasoning capabilities?}.

The main contributions of this work are:
\begin{itemize}
    \item We constructed \textit{TempUN}, the largest public dataset of its kind. Spanning eight distinct categories, TempUN includes \textbf{631K instances} and over \textbf{9.4M samples} related to 106 major issues and 8 focus areas defined by the United Nations, spanning from 10,000 BCE to 2100 years with 83.87\% change of facts (Details in Section~\ref{sec:dataset}).
    \item Our evaluation of twelve state-of-the-art LLMs (nine open-source and three closed-source, ranging from 2B to 70B+) revealed severe limitations in their ability to retain and reason about temporal information over \textbf{six proposed MCQ categories}. 
    \item We experimented with three distinct training paradigms: \textbf{\textit{(1)}} \textbf{yearwise fine-tuning}, \textit{\textbf{(2)}} \textbf{continual learning}, and \textbf{\textit{(3)}} \textbf{random fine-tuning} (Details in Section~\ref{sec:exp-learn}). 
\end{itemize}

\section{Relevant Works}
\label{sec:relwork}
Recent works highlight the deterioration of the LLM's performance over the older temporal information. The factual information does not change over time, indicating that the model's performance is independent of the time frame of the pre-training corpus \citep{temporal-effects-on-task}. 
The factual information as the downstream task worsens over time, regardless of the number of parameters in the model \citep{jang-etal-2022-temporalwiki}. 

The Q\&A Datasets such as \textit{TempLAMA} \cite{templama} and \textit{TemporalWiki} \cite{jang2022temporalwiki} contain \textit{50,310} and \textit{35,948} samples, respectively, with a small time frame of \textit{11 years} (years 2010-2020). More details are added in Appendix \S\ref{sec:templama_appendix}.
The TempLAMA dataset \citep{templama} comprises a significant proportion of static textual facts, with 70.69\% of the facts remaining constant over time, preserving identical answers for a given subject, constrained by temporal spans of only 11 years \citep{benchmarkreas}. Another work by \citet{chen2021dataset} proposed the time-sensitive dataset QA dataset from the time span of 1367-2018, which, however, only contains the temporal event relation. MenatQA by \citet{menatqa} is based on the TimeQA dataset \citep{chen2021dataset} focusing on three temporal factors of scope, order, and counterfactual, while employing F1 and Exact Match (EM) as the evaluation metrics for the total of only 2,853 samples. 

To the best of our knowledge, prior research lacks consideration of the extensive temporal range coupled with the numerical modality, thereby this prompted us to introduce a dataset to evaluate United Nations-focused domains, characterized by an extended temporal span, numerical modality, and dynamic event change (Details in Section~\ref{sec:dataset}). 

\begin{table*}[!tbh]
\centering
   \begin{tabular}{p{0.2cm}p{4cm}p{10cm}}
   \hline
   &\textbf{Category}& \textbf{Subcategories}\\\hline
   C1&Climate & Access To Energy, Air Pollution, Biodiversity, Clean Water and Sanitization, Climate Change, CO2 and Greenhouse Gas Emissions, Energy, Forests and Deforestation, Fossil Fuels, Indoor Air Pollution, Lead Pollution, Natural Disasters, Nuclear Energy, Oil Spills, Ozone Layer, Pesticides, Plastic Pollution, Pollution, Water Use and Stress \\
   C2&Food and Agriculture & Agricultural Production, Animal Welfare, Crop Yields, Environmental Impacts of Food Production, Environmental Impacts of Food Production, Famines, Fertilizers, Food Prices, Land Use, Meat and Dairy Production \\
   C3&Health & Alcohol Consumption, Burden of Disease, Cardiovascular Diseases, Causes of Death, Child and Infant Mortality, COVID, Diarrhoeal Diseases, Diet Compositions, Disease Eradication, Fertility Rate, Global Health, Happiness and Satisfaction, Healthcare Spending, HIV, Human Height, Hunger and Undernourishment, Influenza, Life Expectancy, Malaria, Maternal Mortality, Mental Health, Micronutrient Deficiency, Monkeypox, Obesity, Opioids, Pandemics, Pneumonia, Polio, Sanitation, Smallpox, Smoking, Suicides, Tetanus, Vaccination\\
  C4&Human Rights & Child Labor, Human Rights, LGBT, Literacy, Loneliness and social connections, Marriages and Divorces, Trust, Violence against Children\\
  C5&Innovation& AI, Internet, Research-And-Development, Technology Change\\
  C6&Migration& International Migration and Refugees\\
   C7&Economic Development&Age, Books, Corruption, Economic-Inequality, Education-Spending, Employment-In-Agriculture, Gender Ratio, Global-Education, Government-Spending, Homelessness, Human Development Index, Light at Night, Poverty, Renewable Energy, State-Capacity, Taxation, Time use, Tourism, Trade and globalization, Transportation, Urbanization, Women Employment, Women Rights, Working Hours, GDP\\
   C8& Peace and War&Homicide, Military spending, Nuclear Weapons, Terrorism, War and Peace \\ \hline
   \end{tabular}
\caption{Categories and subcategories present in the \textit{TempUN} dataset.}\label{table:categories}
\end{table*}

\section{The TempUN Dataset}
\label{sec:dataset}
In this paper, we introduce the largest temporal dataset constructed by curating temporal information from \textit{Our World in Data (OWD)} website\footnote{URL: \url{https://ourworldindata.org/}. All data produced by OWD is completely open access under the \href{https://creativecommons.org/licenses/by/4.0/}{Creative Commons BY license}.}.  The website contains data for global issues like poverty, disease, hunger, climate change, war, existential risks, and inequality. All of these issues are listed by the United Nations\footnote{\url{https://www.un.org/en/global-issues}} as the major global challenges that transcend national boundaries and cannot be resolved by any one country acting alone. We, therefore, term this dataset as \textbf{TempUN}. We curate the dataset in eight major issue categories and several subcategories. Table~\ref{table:categories} contains the eight categories and their sub-categorization.  Overall, we obtained 106 subcategories, leading to 13.25 subcategories per category (Details in \S\ref{sec:tempundataset}).

\begin{table*}[!tbh]
\resizebox{\textwidth}{!}{%
\begin{tabular}{ll} \hline
\textbf{Category} & \textbf{Representative Example} \\ \hline
$DB$-MCQ & \begin{tabular}[c]{@{}l@{}}\textit{In 2011, what was France's GDP per capita?} \\ \textbf{(a) 43,846.47 USD}, \textcolor{gray}{(b) 48,566.97 USD, (c) 18841,141.42 USD, (d) 40,123.21 USD}\end{tabular} \\ \hline
$CP$-MCQ & \textit{Was France's GDP per capita higher in 2011 than in 2012? \textbf{(a) Yes}, \textcolor{gray}{(b) No}} \\ \hline
$WB$-MCQ & \begin{tabular}[c]{@{}l@{}}{\textit{From 2015 to 2019, what is the order of France's GDP per capita among the}} \\ {\textit{given options?}} \\ \textbf{(a) In 2015, 47K USD, In 2016, 49.3K USD, In 2017, 48.2K USD, ..}\\ \textcolor{gray}{(b) In 2015, 46K USD, In 2016, 43K USD, In 2017, 37K USD, ..} \\ \textcolor{gray}{(c) In 2015, 445K USD, In 2016, 1249.2K USD, In 2017, 12348.4K USD, ..} \\ \textcolor{gray}{(d) In 2015, 47K USD, In 2016, 49.2K USD, In 2017, 48.2K USD, ..}\end{tabular} \\ \hline
$RB$-MCQ & \begin{tabular}[c]{@{}l@{}}{\textit{In the range of 2011-2021, what is the mean value of France's GDP per capita?}} \\ \textcolor{gray}{(a) 41,304.04 USD,} \textbf{(b) 40,708.08 USD}, \textcolor{gray}{(c) 44,312.73 USD, (d) 37,123.12 USD}\end{tabular} \\ \hline
$MM$-MCQ & \begin{tabular}[c]{@{}l@{}}{\textit{In the range of 2011-2021, what is the minimum and maximum value of France's GDP per capita?}} \\ \textcolor{gray}{(a) 39,252.42 USD, 44,301.84 USD, (b) 19,231.43 USD, 20,708.08 USD,} \\ \textbf{(c) 36,652.92 USD, 43846.47 USD}, \textcolor{gray}{(d) 31,456.83 USD, 37,123.12 USD}\end{tabular} \\ \hline
$TB$-MCQ & \begin{tabular}[c]{@{}l@{}}{\textit{In the range of 2011-2021, what is the rate of change in France's GDP per capita?}} \\ \textbf{(a) 1.1\%}, \textcolor{gray}{(b) 1\%, (c) 3\%, (d) 2.5\%}\end{tabular} \\ \hline
\end{tabular}%
}
\caption{Representative examples from six MCQ categories. The highlighted option represents the correct answer.}
\label{tab:mcqmetric}
\end{table*}

\begin{table*}[!tbh]
\centering
\begin{tabular}{lcccccccc}  \hline
\textbf{Models} & \textbf{Generation} & \textbf{$DB$} & \textbf{$CP$} & \textbf{$WB$} & \textbf{$MM$} & \textbf{$RB$} & \textbf{$TB$} & \textbf{Average} \\  \hline
 & \textbf{C$\uparrow$} & .11 & 0 & .18 & .08 & .09 & .06 & .09 \\
 & \textbf{I$\downarrow$} & .89 & .97 & .82 & .92 & .89 & .93 & .90 \\
\multirow{-3}{*}{\texttt{phi-2}} & \textbf{N$\downarrow$} & \textbf{0} & .03 & \textbf{0} & \textbf{0} & .02 & .01 & .01 \\  \hline
 & \textbf{C$\uparrow$} & .38 & .40 & .20 & .24 & .20 & .03 & .30 \\
 & \textbf{I$\downarrow$} & .62 & .60 & .80 & .76 & .79 & .97 & .69 \\
\multirow{-3}{*}{\texttt{flan-t5-xl}} & \textbf{N$\downarrow$} & \textbf{0} & \textbf{0} & \textbf{0} & \textbf{0} & .01 & \textbf{0} & \textbf{0} \\  \hline
 & \textbf{C$\uparrow$} & .37 & .43 & .20 & .23 & .34 & \textbf{.08} & .27 \\
 & \textbf{I$\downarrow$} & .51 & .57 & .80 & .64 & .66 & .71 & .65 \\
\multirow{-3}{*}{\texttt{mistral-instruct}} & \textbf{N$\downarrow$} & .12 & \textbf{0} & \textbf{0} & .13 & \textbf{0} & .22 & .08 \\  \hline
 & \textbf{C$\uparrow$} & .21 & .45 & .22 & .15 & .22 & .05 & .21 \\
 & \textbf{I$\downarrow$} & .76 & .55 & .78 & .81 & .79 & .93 & .77 \\
\multirow{-3}{*}{\texttt{llama-2-chat}} & \textbf{N$\downarrow$} & .03 & \textbf{0} & \textbf{0} & .04 & \textbf{0} & .02 & .02 \\  \hline
 & \textbf{C$\uparrow$} & .21 & .42 & .15 & .12 & .14 & .03 & .19 \\
 & \textbf{I$\downarrow$} & .77 & .58 & .85 & .88 & .86 & .94 & .79 \\  
\multirow{-3}{*}{\texttt{gemma-7b-it}} & \textbf{N$\downarrow$} & .02 & \textbf{0} & \textbf{0} & \textbf{0} & \textbf{0} & .03 & .01 \\ \hline
 & \textbf{C$\uparrow$} & .39 & .39 & .19 & .18 & .24 & .07 & .31 \\
 & \textbf{I$\downarrow$} & .61 & .61 & .81 & .82 & .76 & .93 & .69 \\  
\multirow{-3}{*}{\texttt{llama-3-8b}} & \textbf{N$\downarrow$} & .01 & \textbf{0} & \textbf{0} & \textbf{0} & \textbf{0} & \textbf{0} & \textbf{0} \\\hline
 & \textbf{C$\uparrow$} & .09 & \textbf{.49} & .37 & .10 & .01 & .01 & .14 \\
 & \textbf{I$\downarrow$} & \textbf{.16} & .47 & \textbf{.31} & \textbf{.27} & \textbf{.03} & .53 & \textbf{.24} \\
\multirow{-3}{*}{\texttt{phi-3-medium}} & \textbf{N$\downarrow$} & .74 & .05 & .33 & .63 & .96 & .46 & .62 \\  \hline
 & \textbf{C$\uparrow$} & .33 & .34 & .29 & .18 & .29 & .03 & .28 \\
 & \textbf{I$\downarrow$} & .61 & .64 & .71 & .82 & .71 & .94 & .68 \\
\multirow{-3}{*}{\texttt{mixtral-8x7b}} & \textbf{N$\downarrow$} & .07 & .02 & \textbf{0} & \textbf{0} & \textbf{0} & .03 & .04 \\  \hline
 & \textbf{C$\uparrow$} & \textbf{.40} & .37 & \textbf{.55} & \textbf{.37} & \textbf{.38} & .01 & \textbf{.37} \\
 & \textbf{I$\downarrow$} & .60 & .63 & .45 & .63 & .62 & .99 & .63 \\
\multirow{-3}{*}{\texttt{llama-3-70b}} & \textbf{N$\downarrow$} & \textbf{0} & \textbf{0} & \textbf{0} & \textbf{0} & \textbf{0} & \textbf{0} & \textbf{0} \\  \hline
{\color[HTML]{656565} } & {\color[HTML]{656565} \textbf{C$\uparrow$}} & {\color[HTML]{656565} .27} & {\color[HTML]{656565} .39} & {\color[HTML]{656565} .16} & {\color[HTML]{656565} .19} & {\color[HTML]{656565} .12} & {\color[HTML]{656565} 0} & {\color[HTML]{656565} .19} \\
{\color[HTML]{656565} } & {\color[HTML]{656565} \textbf{I$\downarrow$}} & {\color[HTML]{656565} .72} & {\color[HTML]{656565} .61} & {\color[HTML]{656565} .84} & {\color[HTML]{656565} .81} & {\color[HTML]{656565} .88} & {\color[HTML]{656565} .99} & {\color[HTML]{656565} .81} \\
\multirow{-3}{*}{{\color[HTML]{656565} \texttt{gpt-3.5-turbo}}} & {\color[HTML]{656565} \textbf{N$\downarrow$}} & {\color[HTML]{656565} .01} & {\color[HTML]{656565} \textbf{0}} & {\color[HTML]{656565} \textbf{0}} & {\color[HTML]{656565} \textbf{0}} & {\color[HTML]{656565} .01} & {\color[HTML]{656565} .01} & {\color[HTML]{656565} .01} \\  \hline
{\color[HTML]{656565} } & {\color[HTML]{656565} \textbf{C$\uparrow$}} & {\color[HTML]{656565} .29} & {\color[HTML]{656565} .02} & {\color[HTML]{656565} 0} & {\color[HTML]{656565} .29} & {\color[HTML]{656565} 0} & {\color[HTML]{656565} .01} & {\color[HTML]{656565} .10} \\
{\color[HTML]{656565} } & {\color[HTML]{656565} \textbf{I$\downarrow$}} & {\color[HTML]{656565} .35} & {\color[HTML]{656565} .98} & {\color[HTML]{656565} 1.00} & {\color[HTML]{656565} .50} & {\color[HTML]{656565} 1.00} & {\color[HTML]{656565} \textbf{.12}} & {\color[HTML]{656565} .66} \\
\multirow{-3}{*}{{\color[HTML]{656565} \texttt{gpt-4}}} & {\color[HTML]{656565} \textbf{N$\downarrow$}} & {\color[HTML]{656565} .36} & {\color[HTML]{656565} \textbf{0}} & {\color[HTML]{656565} \textbf{0}} & {\color[HTML]{656565} .21} & {\color[HTML]{656565} \textbf{0}} & {\color[HTML]{656565} .87} & {\color[HTML]{656565} .24} \\  \hline
{\color[HTML]{656565} } & {\color[HTML]{656565} \textbf{C$\uparrow$}} & {\color[HTML]{656565} .29} & {\color[HTML]{656565} .38} & {\color[HTML]{656565} .34} & {\color[HTML]{656565} .15} & {\color[HTML]{656565} 0} & {\color[HTML]{656565} 0} & {\color[HTML]{656565} .19} \\
{\color[HTML]{656565} } & {\color[HTML]{656565} \textbf{I$\downarrow$}} & {\color[HTML]{656565} .71} & {\color[HTML]{656565} .62} & {\color[HTML]{656565} .66} & {\color[HTML]{656565} .85} & {\color[HTML]{656565} .99} & {\color[HTML]{656565} 1.00} & {\color[HTML]{656565} .80} \\
\multirow{-3}{*}{{\color[HTML]{656565} \texttt{gemini-pro}}} & {\color[HTML]{656565} \textbf{N$\downarrow$}} & {\color[HTML]{656565} \textbf{0}} & {\color[HTML]{656565} \textbf{0}} & {\color[HTML]{656565} \textbf{0}} & {\color[HTML]{656565} \textbf{0}} & {\color[HTML]{656565} .01} & {\color[HTML]{656565} \textbf{0}} & {\color[HTML]{656565} \textbf{0}} \\   \hline
\end{tabular}%
\caption{
Comparative performance of LLMs for different MCQ categories under \textbf{zero-shot} settings (Scale over here is 0-1). Here, `\textbf{C}' (Correct), `\textbf{I}' (Incorrect), and `\textbf{N}' (Information Not Available) represent the percentage of correct generations, incorrect generations, and LLMs generation of information not available, respectively. We \textbf{bold} the highest values for `\textbf{C}', and lowest values for `\textbf{I}' and `\textbf{N}' categories. Here, we distinguish between open-source and closed-source LLMs with the black and {\color[HTML]{343434} gray} color, respectively.}  \label{table:num-metrics_results}
\end{table*}

\textit{TempUN} consists of instances on the form of tuple $<C, I, L>$, where, $C$ represents a country name, $I$ represents issue subcategory, and $L$ is a list of $<Y_t, V_t>$ tuples, where $Y_t$ is year and $V_t$ is value of $I$ for $C$ in the year $Y_t$. For example, for US's GDP, the instance is <US, GDP, \{<1950,~15912>, <1951,~16814>,...\}. Further, each instance creates a set of input and output samples. A sample is represented by a quadruple $<C, I, Y_t>$ and $V_t$, respectively. Overall, \textit{TempUN} comprises 462K instances and 9.4M samples with 83.87\% of facts being updated yearly. 

In the rest of this work, due to computation constraints, we conduct experiments on a small filtered subset of \textit{TempUN}, \textit{TempUN$_s$}. We select one subcategory for each category for \textit{TempUN$_s$}. This selection follows two key criteria: 1) Data Availability: the subcategory must possess at least 76 continuous years of data between 1947 and 2022 to ensure sufficient temporal coverage. 2) Temporal Dynamics: if multiple subcategories meet the first criterion, we prioritize the one exhibiting the most significant changes over consecutive years within the available data. This preference for demonstrably dynamic trends aligns with the dataset's overall focus on capturing the temporal evolution of global issues. By applying these criteria, we ensure that each major category is represented by a subcategory showcasing both substantial temporal coverage and demonstrably dynamic trends, enabling insightful analysis of temporal developments within each issue area. \textit{TempUN$_s$} results in 1,907 instances and 104,130 samples\footnote{We showcase each category-wise distribution of instances and samples in Table~\ref{table:issues_un}.}. For the rest of the paper, we conduct experiments on \textit{TempUN$_s$}, and use \textit{TempUN} and \textit{TempUN$_s$} interchangeably. 
Next, each sample is further transformed for two distinct tasks: (i) Next-word prediction (NWP) and (ii) Multiple Choice Question Answering (MCQA). For NWP, we combine the individual samples in the tuple $<C, I, Y_t>$ to create a natural language input query and $V_t$ as the expected next word to be generated. For example, $<US, GDP\ per\ capita, 1990>$  would yield a query \textit{`The GDP per capita of US in the year 1990 is'}, with the expected next token as \textit{`23888.6'}. We manually create a query template for each of the eight subcategories in \textit{TempUN$_s$}. Overall, NWP leads to the creation of 104,130 natural language queries.  We use NWP for finetuning models (see more details in Section~\ref{sec:exp-learn}). We create six MCQ-based questions to evaluate LLMs' memorization and reasoning capabilities for MCQA. For each MCQ category, the incorrect answers are generated using the following mathematical expression: $v_{t} + U $($0,1$)$*10^{ log_{10}(v_{t}+1)}$, where $U$($0,1$) denotes standard uniform distribution. The option ordering is randomly created. The six MCQ categories as shown in Table~\ref{tab:mcqmetric} are: 
\begin{enumerate}
    \item \textbf{Date-based MCQs ($DB$-MCQs)}: These are straightforward questions focusing on models' capability to predict correct numerical value $V_t$ for a year-specific query comprising $C$, $I$ and $Y_t$. MCQs are created from a single sample.
    \item \textbf{Comparative MCQs ($CP$-MCQs)}: For a given $C$ and $I$, these questions compare the values in two consecutive years $Y_t$ and $Y_{t+1}$. $CP$-MCQs are created from two samples. 
    \item \textbf{Window-based ($WB$-MCQs)}: 
    $WB$-MCQs evaluate the model's capability to remember a sequence of events. Each $WB$-MCQ query uses five samples in \textit{TempUN}. For a given $C$ and $I$, these questions predict the correct numerical value in five consecutive years $Y_t$ and $Y_{t+4}$. 
    \item \textbf{Range-based ($RB$-MCQs)}: 
    $RB$-MCQs evaluate the model's capability to aggregate numerical values in a range of ten years. 
    \item \textbf{Min-Max ($MM$-MCQs)}: $MM$-MCQs aims to evaluate the model's capability to find extremes of values, the minimum and maximum, within a specified ten-years interval. 
    \item \textbf{Trend-based ($TB$-MCQs)}: $TB$-MCQs evaluate the model's understanding of temporal trends and how the \textit{rate of change} is observed. For instance, the range of change observed over the decade.
\end{enumerate}

With the exception of CP-MCQs, which offer two answer choices, all other MCQ categories present four options. Table~\ref{tab:mcqmetric} presents representative examples from each category. Notably, the table highlights the varied year spans covered by different categories, ranging from one to ten years. Overall, we obtained 157,508 MCQs (Appendix \S\ref{sec:zeroshot} details the yearwise count for each MCQs-based strategy.). We list the category-wise count for each MCQ-based strategy in Tables~\ref{tab:promptsforboth} (\textit{TempUN}) and \ref{tab:micro_prompts_tempuns} (\textit{TempUN$_s$}) in Appendix\S\ref{sec:tempundataset}.

\begin{figure*}
    \centering
    \begin{subfigure}[b]{0.89\textwidth}
        \centering
        \includegraphics[width=1\textwidth]{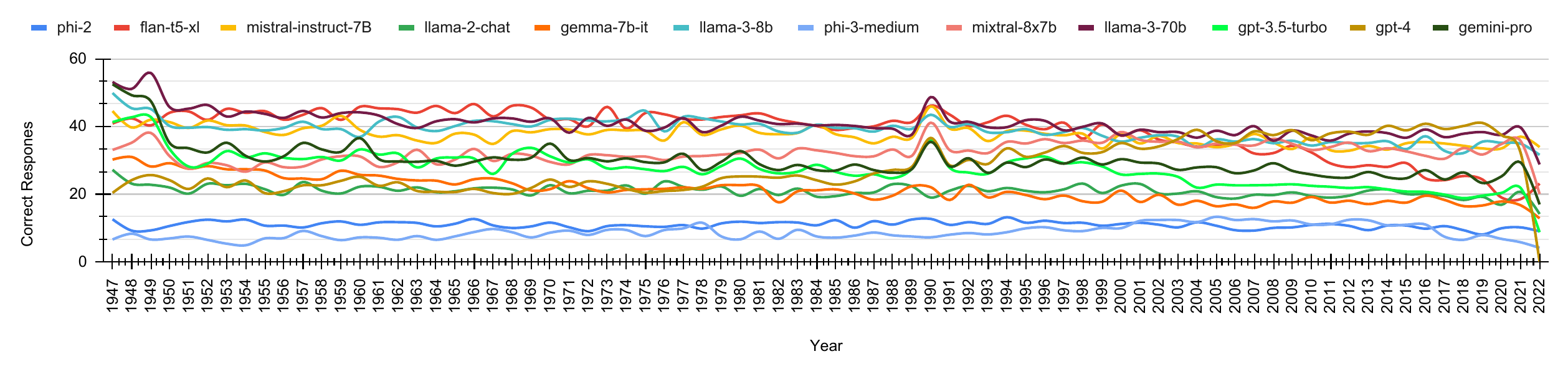}
    \end{subfigure}
    \begin{subfigure}[b]{0.89\textwidth}
        \centering
        \includegraphics[width=1\textwidth]{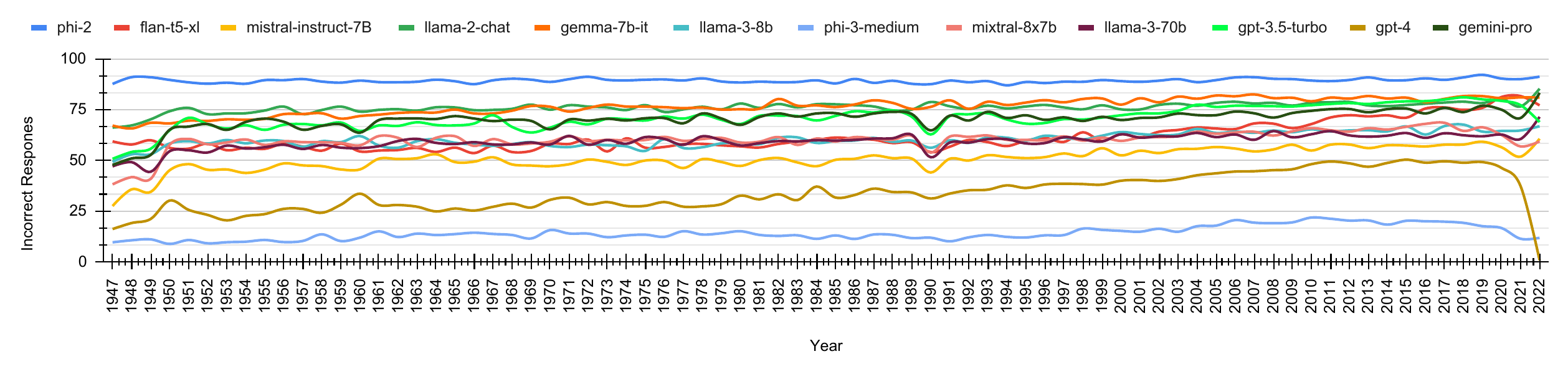}
    \end{subfigure}
    \begin{subfigure}[b]{0.89\textwidth}
        \centering
        \includegraphics[width=1\textwidth]{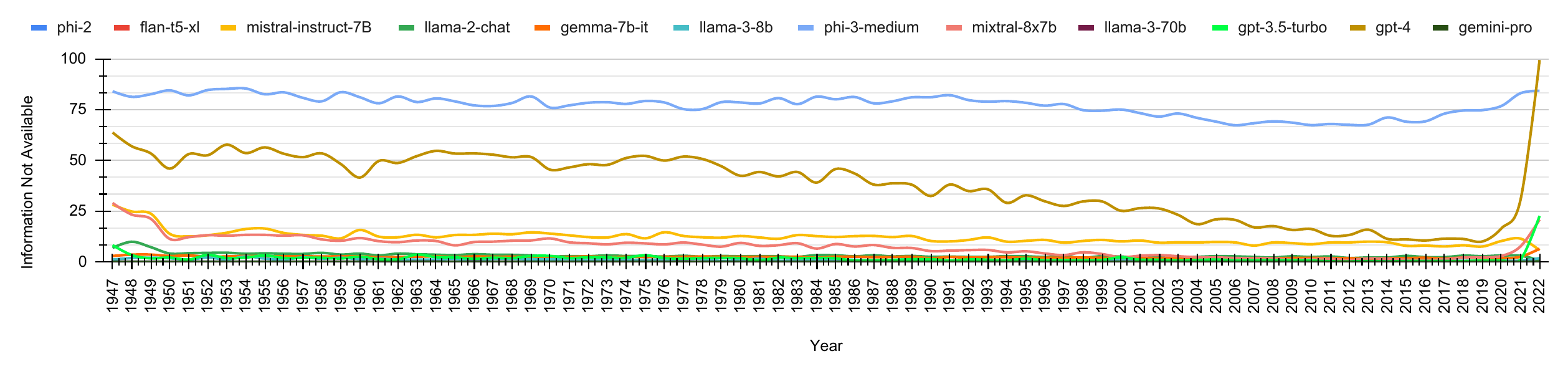}
    \end{subfigure}
    
    \caption{Evluations from Zero-Shot Evaluations on the $DB$-MCQs for the time-span from years 1947 to 2022, where the (Top) ``\textbf{C}'' (correct) scores are higher in >=14B LLMs than the <=8B LLMs; (Middle) ``\textbf{I}'' (incorrect) scores are lower in closed-source models than open-source models; and (Bottom) ``\textbf{N}'' scores being higher in closed-source than open-source LLMs.\label{fig:DB-allmodels}}
    
\end{figure*}

\section{Experiments}
\label{sec:exp}

\subsection{Models}
\label{sec:models}
We conduct experiments with 12 state-of-the-art open-source and close-source models. Open-source models include \texttt{phi-2}\footnote{\url{https://huggingface.co/microsoft/phi-2}} (2.7B), \texttt{flan-t5-xl}, (3B,~\citet{t5_instruct}), \texttt{mistral-instruct-v0.2} (7B,~\citet{mistral}), \texttt{llama-2-chat} (7B,~\citet{llama}), \texttt{gemma-1.1-7b-it} (7B,~\citet{gemma}), \texttt{Meta-Llama-3-8B-Instruct} (8B,~\citet{llama3modelcard}), \texttt{phi-3} (14B,~\citet{phi3}), \texttt{Mixtral-8x7B-Instruct-v0.1} (7x8B/47B,~\citet{mixtral78b}), and \texttt{Meta-Llama-3-70B-Instruct} (70B,~\citet{llama3modelcard}. In addition, we chose three closed-source models, \texttt{gpt-3.5-turbo}~\citep{chatGPT3.5}, \texttt{gpt-4}~\citep{gpt4report}, and \texttt{gemini-pro}~\citep{team2023gemini}. While larger open-source exist, our experiments were restricted with sizes less than or equal to 8B parameters due to computational resource limitations. We utilized the Groq\footnote{\url{https://groq.com/}} platform for the zero-shot inferences for \texttt{gemma-7b-it}, \texttt{mixtral-8x7B}, \texttt{llama-3-8B}, and \texttt{llama-3-70B} models. We use their official APIs for closed-source models. Appendix~\S\ref{sec:experiment} and ~\S\ref{sec:gpudetails} details models' settings and the computing infrastructure.

\begin{table*}[!tbh]
\resizebox{\textwidth}{!}{%
\begin{tabular}{lccccccccccccccccccccc} \hline
 & \multicolumn{21}{c}{\textbf{Models}} \\ \hline
\multicolumn{1}{c}{\textbf{}} & \multicolumn{3}{c}{\texttt{phi-2}} & \multicolumn{3}{c}{\texttt{flan-t5-xl}} & \multicolumn{3}{c}{\texttt{mistral-instruct}} & \multicolumn{3}{c}{\texttt{llama-2-chat}} & \multicolumn{3}{c}{\texttt{gemma-7b-it}} & \multicolumn{3}{c}{\texttt{llama-3-8b}} & \multicolumn{3}{c}{\texttt{phi-3-instruct}} \\ \hline
\textbf{Generation} & \textbf{C$\uparrow$} & \textbf{I$\downarrow$} & \textbf{N$\downarrow$} & \textbf{C$\uparrow$} & \textbf{I$\downarrow$} & \textbf{N$\downarrow$} & \textbf{C$\uparrow$} & \textbf{I$\downarrow$} & \textbf{N$\downarrow$} & \textbf{C$\uparrow$} & \textbf{I$\downarrow$} & \textbf{N$\downarrow$} & \textbf{C$\uparrow$} & \textbf{I$\downarrow$} & \textbf{N$\downarrow$} & \textbf{C$\uparrow$} & \textbf{I$\downarrow$} & \textbf{N$\downarrow$} & \textbf{C$\uparrow$} & \textbf{I$\downarrow$} & \textbf{N$\downarrow$} \\ \hline
\textbf{$DB$-Y} & .07 & .50 & .43 & .38 & .62 & \textbf{0} & \textbf{.39} & .56 & .05 & .23 & .77 & 0 & .21 & .79 & 0 & .37 & .48 & .15 & .11 & \textbf{.29} & .61 \\
\textbf{$DB$-C} & .05 & .22 & .73 & .35 & .65 & \textbf{0} & .20 & .39 & .41 & .23 & .77 & 0 & .21 & .79 & \textbf{0} & \textbf{.42} & .51 & .07 & .08 & \textbf{.31} & .61 \\ 
\textbf{$DB$-R} & .02 & .94 & .04 & \textbf{.26} & .74 & \textbf{0} & .25 & .50 & .25 & .11 & .37 & .52 & 0 & .66 & .34 & .09 & .86 & .04 & .02 & \textbf{.28} & .69 \\ \hline
\textbf{$CP$-Y} & 0 & 0 & 1 & .41 & .59 & \textbf{0} & 0 & \textbf{0} & 1 & 0 & 0 & 1 & .40 & .60 & \textbf{0} & .45 & .55 & 0 & \textbf{.46} & .51 & .03 \\
\textbf{$CP$-C} & 0 & .01 & .99 & .40 & .60 & \textbf{0} & 0 & \textbf{0} & 1 & 0 & 0 & 1 & .40 & .60 & \textbf{0} & .40 & .60 & \textbf{0} & \textbf{.48} & .45 & .07 \\ 
\textbf{$CP$-R} & 0 & .12 & .88 & .40 & .60 & \textbf{0} & 0 & \textbf{0} & 1 & 0 & \textbf{0} & .99 & .01 & .02 & .97 & \textbf{.44} & .51 & .04 & .12 & .14 & .75 \\ \hline
\textbf{$WB$-Y} & .20 & .78 & .02 & .21 & .79 & \textbf{0} & .21 & .67 & 1 & .21 & .75 & .04 & .09 & .91 & \textbf{0} & .24 & .75 & .01 & \textbf{.31} & \textbf{.33} & .36 \\
\textbf{$WB$-C} & .18 & .57 & .25 & .19 & .81 & \textbf{0} & .09 & .89 & .02 & .22 & .77 & .01 & .09 & .91 & \textbf{0} & .25 & .74 & .02 & \textbf{.27} & \textbf{.35} & .39 \\
\textbf{$WB$-R} & .15 & .48 & .37 & \textbf{.24} & .76 & \textbf{0} & .11 & .88 & .01 & .23 & .75 & .01 & 0 & .63 & .37 & .14 & .40 & .46 & 0 & \textbf{.01} & .99 \\ \hline
\textbf{$MM$-Y} & .09 & .46 & .46 & .24 & .74 & .02 & \textbf{.26} & .71 & .02 & .14 & .68 & .18 & .10 & .90 & \textbf{0} & .05 & \textbf{.26} & .69 & .07 & \textbf{.26} & .68 \\
\textbf{$MM$-C} & .13 & .40 & .47 & \textbf{.22} & .78 & \textbf{0} & .12 & .42 & .46 & .11 & .74 & .15 & .10 & .90 & \textbf{0} & .14 & .60 & .26 & .06 & \textbf{.22} & .72 \\
\textbf{$MM$-R} & 0 & .98 & .02 & \textbf{.24} & .72 & .04 & .16 & .59 & .25 & .06 & .22 & .71 & 0 & .55 & .45 & .04 & .14 & .82 & .01 & \textbf{.03} & .96 \\ \hline
\textbf{$RB$-Y} & .05 & .34 & .61 & .18 & .76 & .07 & \textbf{.32} & .59 & .09 & .07 & .29 & .65 & .13 & .87 & \textbf{0} & .12 & .27 & .61 & .02 & \textbf{.19} & .79 \\
\textbf{$RB$-C} & .14 & .42 & .43 & .22 & .78 & 0 & .13 & .40 & .47 & .08 & .31 & .61 & .13 & .87 & \textbf{0} & \textbf{.23} & .52 & .25 & .02 & \textbf{.19} & .79 \\
\textbf{$RB$-R} & 0 & .98 & .02 & \textbf{.25} & .74 & .01 & .16 & .47 & .37 & .02 & \textbf{.07} & .91 & 0 & .61 & .39 & .05 & .73 & .22 & .02 & .39 & .59 \\ \hline
\textbf{$TB$-Y} & .02 & \textbf{.20} & .78 & .03 & .97 & \textbf{0} & \textbf{.06} & .57 & .38 & .05 & .43 & .53 & .05 & .95 & \textbf{0} & .02 & .26 & .72 & .01 & .62 & .38 \\
\textbf{$TB$-C} & \textbf{.10} & .30 & .60 & .04 & .96 & \textbf{0} & .02 & .45 & .53 & .07 & .69 & .24 & .05 & .95 & \textbf{0} & .01 & \textbf{.28} & .71 & .01 & .64 & .35 \\
\textbf{$TB$-R} & 0 & 1 & 0 & \textbf{.21} & .79 & \textbf{0} & .03 & .56 & .42 & .02 & \textbf{.09} & .89 & 0 & .56 & .44 & .03 & .61 & .36 & .02 & .34 & .65 \\  \hline
\end{tabular}%
}
\caption{Comparative performance of LLMs for different MCQ categories under \textbf{Yearwise Finetuning}, \textbf{Continual Learning}, and \textbf{Random Finetuning} settings. Here, \textbf{C} (Correct), \textbf{I} (Incorrect), and \textbf{N} (Information Not Available) represent the percentage of correct generations, incorrect generations, and LLMs generation of information not available, respectively. We \textbf{bold} the highest values for \textbf{C}, and lowest values for \textbf{I}, and \textbf{N} categories.}
\label{table:num-learning}
\end{table*}

\subsection{Learning and Evaluation Paradigms}
\label{sec:exp-learn}

\par \noindent \textbf{Zero-Shot Evaluation (ZS)}: In this setting, we evaluate models' capability to answer MCQs without any specific finetuning on the NWP data.  

\par \noindent \textbf{Yearwise Finetuning (Y-FT)}: Here, the model is subjected to parameter efficient fine-tuning (PEFT) by adapting QLoRA technique~\citep{qlora}. We fine-tune the model on NWP instances for each year separately. This resulted in a set of 76 finetuned models, each corresponding to a specific year. The performance of each finetuned model was then evaluated on MCQs tailored to the respective year's data. Say, the LLM was fine-tuned on the data of the year 1947 and evaluated on the same year's data.

\par \noindent \textbf{Continual Learning (CL)}~\citep{biesialska}: In contrast to Yearwise Finetuning, here, the LLM is sequentially finetuned, using QLoRA technique \citep{qlora}, on NWP instances, starting from 1947 and progressing year-by-year until 2022. This resulted in a set of 76 continually fine-tuned models. Similar to the Yearwise Finetuning evaluation, each continually fine-tuned model is evaluated on the respective year's MCQs.

\par \noindent \textbf{Random Finetuning (R-FT)}: Here, we finetune an LLM on the entire NWP data. We randomize the NWP instances to avoid any implicit chronological ordering. Similar to the last two learning techniques, we also use QLoRA \citep{qlora}. The resultant model is evaluated on the entire set of MCQs.

\subsection{Evaluation}
\label{sec:eval_metrics}
The models are evaluated based on an exact match between the generated answer and the ground truth; such instances are classified as ``\textit{Correct}'' (\textbf{C}). In contrast, a lack of such concordance is designated as ``\textit{Incorrect}'' (\textbf{I}). Furthermore, it is observed that the LLMs frequently generate outputs indicating an absence of information or the unavailability of data. These instances are subsequently categorized under the ``\textit{Not Available}'' (\textbf{N}) label. For all experiments, we report a proportion of MCQs, labeled as ``\textbf{C}'', ``\textbf{I}'', and ``\textbf{N}'', respectively. Note, we intend to achieve higher scores for ``\textbf{C}'', whereas lower scores for ``\textbf{I}'' and ``\textbf{N}''\footnote{We have used the scale of 0-1 in Table~\ref{table:num-learning}.}. 

\section{Results and Discussions}
\label{sec:results} 
We revisit the research questions from Section~\ref{sec:intro} and state our findings as:
\par \noindent \textbf{RQ1: Do LLMs effectively retain temporal knowledge?}
Our experiments unveil significant limitations in the LLMs' ability to retain temporal information, particularly within a zero-shot setting. As seen in Table~\ref{table:num-metrics_results}, for $DB$-MCQs, LLM performance is concerningly low: the average accuracy rate of open-source models is 27\%, while closed-source models fare slightly better at 28\%. Conversely, the prevalence of incorrect responses is considerably high, reaching 61\% for open-source and 59\% for closed-source models. Interestingly, the larger-sized LLMs( >=14B params) are less likely to generate incorrect responses than the smaller-sized (<=8B params) LLMs, with 59\% and 62\% incorrect responses, respectively. In Figure~\ref{fig:DB-allmodels}, we show the comparative performance analysis for ``\textbf{C}'', ``\textbf{I}'', and ``\textbf{N}'' for the $DB$-MCQs as per the time span of 75 years. We observed that the closed-source models tend to indicate the unavailability of information more frequently than open-source models (12\% vs 11\%). 
\par \noindent \textbf{\textit{Takeaway}}: \textit{LLMs perform poorly while retaining the temporal understanding. Open-source models are more prone than closed-sourced models to provide incorrect responses. Additionally, closed-source LLMs acknowledge information unavailability better than open-source LLMs.}

\noindent \textbf{RQ2: Do LLMs effectively \textit{reason} about temporal knowledge?} 
Apart from $DB$-MCQs, we leveraged the other MCQ categories to understand the model’s ability to reason about temporal knowledge. Open-source models tend to generate more correct results than close-sourced LLMs in the $CP$ (36\% vs 27\%), $WB$ (26\% vs 17\%), $RB$ (21\% vs 4\%), and $TB$ (4\% vs 0\%), whereas $MM$ reported (18\% vs 21\%). We noted that in $MM$, where the ``\textbf{C}'' reported lower scores, ``\textbf{N}'' reported better scores in close-sourced than open-source LLMs (9\% vs 7\%). We noted the average scores over six metrics yielded open-source LLMs better performing than close-source with in all three evaluations: ``\textbf{C}'' (24\%
vs 16\%), ``\textbf{I}'' (67\% vs 76\%), and ``\textbf{N}'' (9\% vs 8\%).
We observed that \texttt{llama-3-70b} outperformed all other LLMs in the ``\textbf{C}'', and comparable scores in ``\textbf{N}'' with \texttt{gemini-pro}. Even the popular LLMs such as \texttt{gpt-4} and \texttt{gemini-pro} led to poor performance in understanding the MCQA dataset. We assume that the LLMs find it difficult to understand the prompt and parse them in the correct form of reasoning chains, simply the reasoning part. Thus, we observed lower scores in the six MCQ-based queries overall. Notably, the most recent \texttt{phi-3-medium} model had the lowest ``\textbf{I}'' scores and the highest ``\textbf{N}'' scores. This indicates that the model understood the reasoning and acknowledged its lack of knowledge rather than producing incorrect responses. Lastly, we can highlight that the LLMs find the $TB$-MCQs difficult to answer with the ``\textbf{C}'' scores of 3\%, while $CB$-MCQs as the easy to answer with the scores of 34\%.

\par \noindent \textbf{\textit{Takeaway}}: \textit{LLMs lacks temporal reasoning and understanding capabilities. Surprisingly, open-source LLMs perform better than closed-source models on the average scores of all six MCQ-based evaluations. 
}

\noindent \textbf{RQ3: Do different training paradigms affect overall temporal knowledge retention and reasoning capabilities?}
We showcase the different paradigms in Table~\ref{table:num-learning}, for Yearwise Learning, Continual Learning, and Random Fine-tuning. We observed that the yielded average ``\textbf{N}'' scores are ZS (11\%), Y-FT (29\%), CL (30\%), and R-FT (38\%); LLMs reported higher ``\textbf{N}'' scores after R-FT, indicating that this approach helps LLMs to refrain from generating incorrect information by correctly identifying unavailable information. Additionally, the different paradigms also helped models to reduce the ``\textbf{I}'' scores from 68\% (ZS) to 53\% (R-FT), 52\% (Y-FT), and 53\% (CL).
During inference across the four learning evaluation paradigms, we encountered a major issue where the generations were garbage numbers. To address this, we incorporated a couple of suffixes\footnote{The following suffixes were utilized: \textit{\textbf{(1)} Choose the most relevant answer.}, \textit{\textbf{(2)} Provide the only correct option, without explanation.}}, which successfully resulted in generating only the correct option in both open-source and closed-source models for the ZS settings. During the Y-FT, CL, and R-FT training, we observed that the LLMs are very sensitive towards the temporal-numerical data as the ``\textbf{C}'' scores decreased significantly from 22\% to 18\% (Y-FT), 17\% (CL), and 9\% (R-FT). One reason for the lower correct scores could be the distorted information representations in the LLMs after the training, and hurting the LLMs knowledge. 
\par \noindent \textbf{\textit{Takeaway}}: \textit{Different learning paradigms reduced LLM's incorrect generations and allowed the LLMs to acknowledge wherever information was unavailable. Reduced correct responses notifies the need for better numerical-temporal learning paradigms.}

\section{Conclusion and Future Directions} 
\label{sec:conclusion}

We present two variations of numbers-based temporal datasets, covering 83.87\% of facts that change over time, named \textit{TempUN} (631k samples) and \textit{TempUN$_s$} (104k samples). We proposed six MCQ-based evaluations for assessing temporal information on 12 popular LLMs, and introduced three learning paradigms: Continual Learning, Yearwise Finetuning, and Random Finetuning. Our findings highlight that the popular LLMs does not retain the temporal information, and open-source LLMs yielded better results, however fails to acknowledge the lack of knowledge, to which closed-source models admits their missing knowledge.

Future work plans to expand the dataset to explore non-numerical modalities, a broader timespan, and a higher percentage of changing facts, thereby improving the LLMs' temporal reasoning abilities. Additionally, we aim to inspect the numerical-memorization in our future works.

\section*{Limitations} 
\label{sec:limit}
Our research emphasizes the limitations of LLMs in comprehending temporal knowledge and their inclination toward language acquisition rather than analyzing numerical trends. Our work encompasses historical data spanning from 10,000 BCE to 2100 years ago, comprising approximately \textbf{462K} instances, leading to the creation of \textbf{9.4M} temporal prompts. Due to computational constraints inherent in larger models, our experiments could only be conducted on a subset of the complete dataset, resulting in evaluations being carried out on \textbf{1,907} instances, constituting \textbf{104K} samples spanning eight distinct categories in the numerical modality. Our \textit{TempUN} data covers the factual numerical data, and we plan to add the textual data in our future works. Our work focuses on proposing the numerical-temporal dataset for a longer time span, which was missing the previous literature and not significantly contributing to the numerical memorisations in LLMs. 
Lastly, we plan to explore different fine-tuning strategies, such as adapters, k-adapters, etc., to help the LLMs learn better in future works. 



\section*{Ethics and Potential Risks}
\label{sec:ethicsec}
We have strictly adhered to the ethics and guidelines during the progress of our work. The data processing and preparation guidelines have been taken into consideration. The introduced data does NOT contain personal names, uniquely identifiable individuals, or offensive content. The data introduced solely contains the facts as listed on the OWD site. 

\section*{Acknowledgements}
This work is supported by the Prime Minister Research Fellowship (PMRF-1702154) to Himanshu Beniwal. Acknowledgment is extended to Vamsi Srivathsa, Venkata Sriman, and Zeeshan Snehil Bhagat for their invaluable assistance during the experimental phase of this work. Special thanks are also due to Professor Nipun Batra and Zeel Patel for their support in fulfilling the computational requirements. A part of our work was supported by Microsoft's Accelerate Foundation Models Research grant. 



\bibliography{anthology,custom}
\newpage
\appendix
\section{Appendix}

\begin{figure}
\begin{center}
\includegraphics[width=\linewidth]{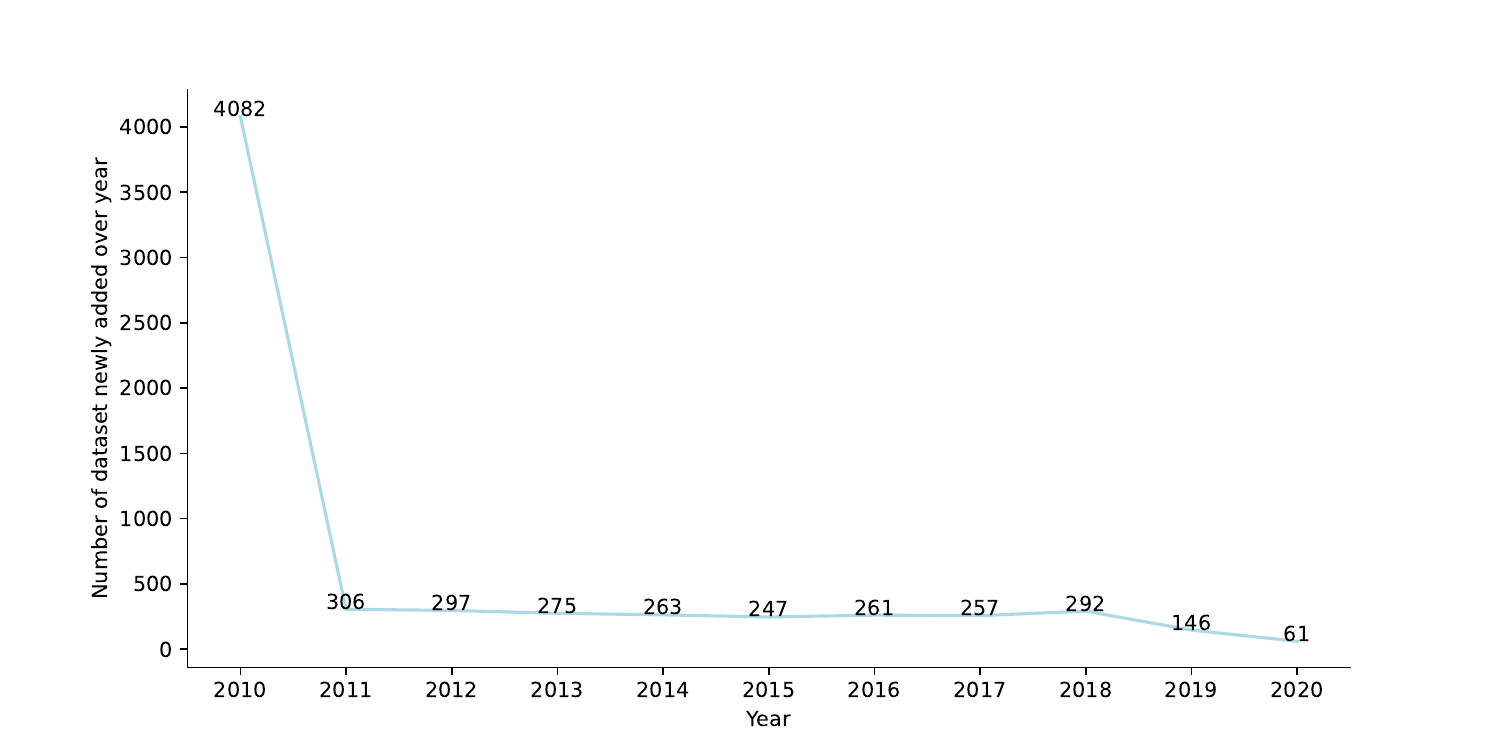}
\end{center}
\caption{Count of unique data samples available each year.}
\label{fig:templama_new_data}
\end{figure}
\subsection{Inferencing Models - Zeroshot Setting}
To assess the model's proficiency in processing numerical data, the identical sample was presented to various models with capacities exceeding 7 billion parameters, as illustrated in Figure~\ref{fig:model_inference}. It was observed that contemporary, widely used models demonstrated a deficiency in relevant knowledge. This limitation became particularly evident when the prompt was slightly altered to include a temporal shift; the models tended to overestimate and generate responses that were not pertinent to the given context.


\begin{table*}[]
\centering
\begin{tabular}{cccc}
\hline
\multicolumn{1}{c}{} & \textbf{Wikidata ID} & \textbf{Relation}              & \textbf{Template}                                                                              \\ \hline
1                      & P54         & member of sports team & \textless{}subject\textgreater plays for \textless{}object\textgreater{}.             \\
2                      & P39         & position held         & \textless{}subject\textgreater holds the position of \textless{}object\textgreater{}. \\ 
3                      & P108        & employer              & \textless{}subject\textgreater works for \textless{}object\textgreater{}.             \\
4                      & P102        & political party       & \textless{}subject \textgreater is a member of the \textless{}object\textgreater{}.    \\
5                      & P286        & head coach            & \textless{}object\textgreater is the head coach of \textless{}subject\textgreater{}.  \\ 
6                      & P69         & educated at           & \textless{}subject\textgreater attended \textless{}object\textgreater{}.              \\
7                      & P488        & chairperson           & \textless{}object\textgreater is the chair of \textless{}subject\textgreater{}.       \\ 
8                      & P6          & head of government    & \textless{}object\textgreater is the head of the government of \textless{}sub..      \\
9                      & P127        & owned by              & \textless{}subject\textgreater is owned by \textless{}object\textgreater{}.           \\ \hline
\end{tabular}
\caption{TempLAMA relation and the template format of each sample and the corresponding WikiData dataset relation identifier.}
\label{tab:templama_relation}
\end{table*}

\begin{table}[]
\centering
\begin{tabular}{cc} \hline
\textbf{Hyperparameter} & \textbf{Search Space} \\ \hline
Batch Size & [8, 12, 16] \\
Epoch & [6 - 10] \\
Learning Rate & [2e-4, 2e-5, 2e-6] \\
Patience & [4] \\ \hline
\end{tabular}
\caption{The search space for hyperparameters.}
\label{tab:searchhyper}
\end{table}
\subsection{TempLAMA Dataset}
\label{sec:templama_appendix}
\par
\begin{figure}
\begin{center}
\includegraphics[width=\linewidth]{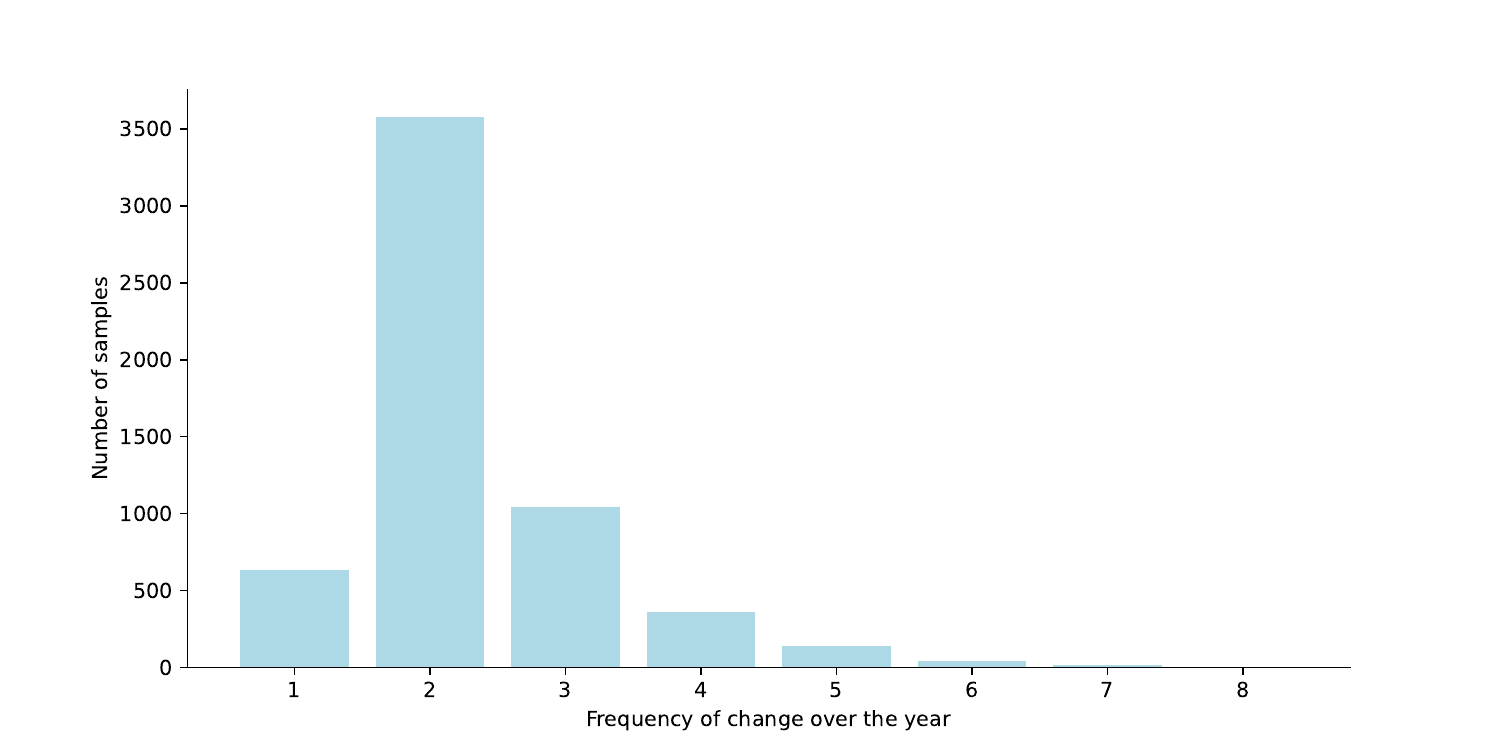}
\end{center}
\caption{Frequency of change in the dataset for one query over 11 years}
\label{fig:templama_changes_over_year}
\end{figure}

 \noindent \textbf{TempLAMA} We also summarizes the previous available dataset: \textbf{TempLAMA} by \citet{templama}, which is a closed-book question-answering dataset. The dataset consists of events and 11 relations that change over the years. The dataset contains data for 11 years, 2010-2020. In the dataset, \textit{Valentino Rossi plays for \_X\_.} a query changed only thrice over the year; in 2010, it was \textit{Yamaha Motor Racing} then \textit{Ducati Motor Holding S.p.A.} in 2011 and finally back to \textit{Yamaha Motor Racing} from 2013 onwards.  Figure \ref{fig:templama_changes_over_year} shows how frequently events changed over 11 years. We see that most of the events did not change frequently. In the dataset, each sample contains a \textit{subject} ($s$), \textit{relation} ($r$) and \textit{objects} ($o$) from years where there was a change. 
The TempLAMA dataset contains the nine different relations ($r$) that change over time. The list of each relation present and the template for each category in the dataset is available in Table~\ref{tab:templama_relation}. The Number of data samples newly added in each year is depicted in Figure \ref{fig:templama_new_data}.

\subsection{Experimental Settings}
\label{sec:experiment}
In this section, we define the experimental configurations used to fine-tune the models. 

While experimenting with the fine-tuning strategies, we use the following hyperparameters for all three \textit{close-source} models: batch size (12), epoch (10), learning rate (2e-5), and patience (4), with the search space defined in Table~\ref{tab:searchhyper}.

\subsection{Computational Resources}
\label{sec:gpudetails}
The experiments are carried out on four NVIDIA Tesla V100 32 GB. The estimated cost to cover the computational requirements for two months, computed over GCP is \$9,460.78\footnote{The price for the VM is computed using the GCP Calculator: \url{https://cloud.google.com/products/calculator}.} (\$4,730.39-per month x 2 months). We utilized the official APIs for all of the \textit{close-source} models.

\begin{table*}
\centering
\begin{tabular}{p{4cm}lllll}
\hline
\textbf{Categories}                                    & \textbf{Subcategories} & \textbf{Instances} & \textbf{Samples} & \textbf{Instances$_s$} & \textbf{Samples$_s$} \\ \hline
C$1$: Climate & 19                     & 95,289                 & 1,778,631   & 244                    & 17,928               \\
C$2$: Food and Agriculture                             & 10                     & 33,610                 & 991,443     & 279                    & 11,133               \\
C$3$: Health                                           & 34                     & 245,330                & 5,684,312   & 260                    & 18,599               \\
C$4$: Human Rights                                     & 8                      & 3,132                  & 7,142       & 190                    & 5,373                \\
C$5$: Innovation              & 4                      & 567                    & 1,537       & 227                    & 5,813                \\
C$6$: Migration                                        & 1                      & 18,167                 & 100,346     & 255                    & 17,232               \\
C$7$: Economic Development     & 25                     & 59,483                 & 909,519     & 250                    & 18,716               \\
C$8$: Peace and War                                    & 5                      & 7,316                  & 24,572      & 202                    & 9,336                \\ \hline
\textbf{Total}                                         & \textbf{106}           & \textbf{462,894}                & \textbf{9,497,502}   & \textbf{1,907}                  & \textbf{104,130}     \\ \hline
\end{tabular}
\caption{\label{table:issues_un}
List of categories as global issues and the primary focus required as per the United Nations in the \textit{TempUN} and \textit{TempUN$_s$} datasets. Here, Instances and Samples underlie the \textit{TempUN} dataset, where Instances$_s$ and Samples$_s$ for the \textit{TempUN$_s$} dataset.}

\end{table*}

\subsection{TempUN Dataset}
\label{sec:tempundataset}
\begin{figure*}
\begin{center}
\includegraphics[width=0.8\linewidth]{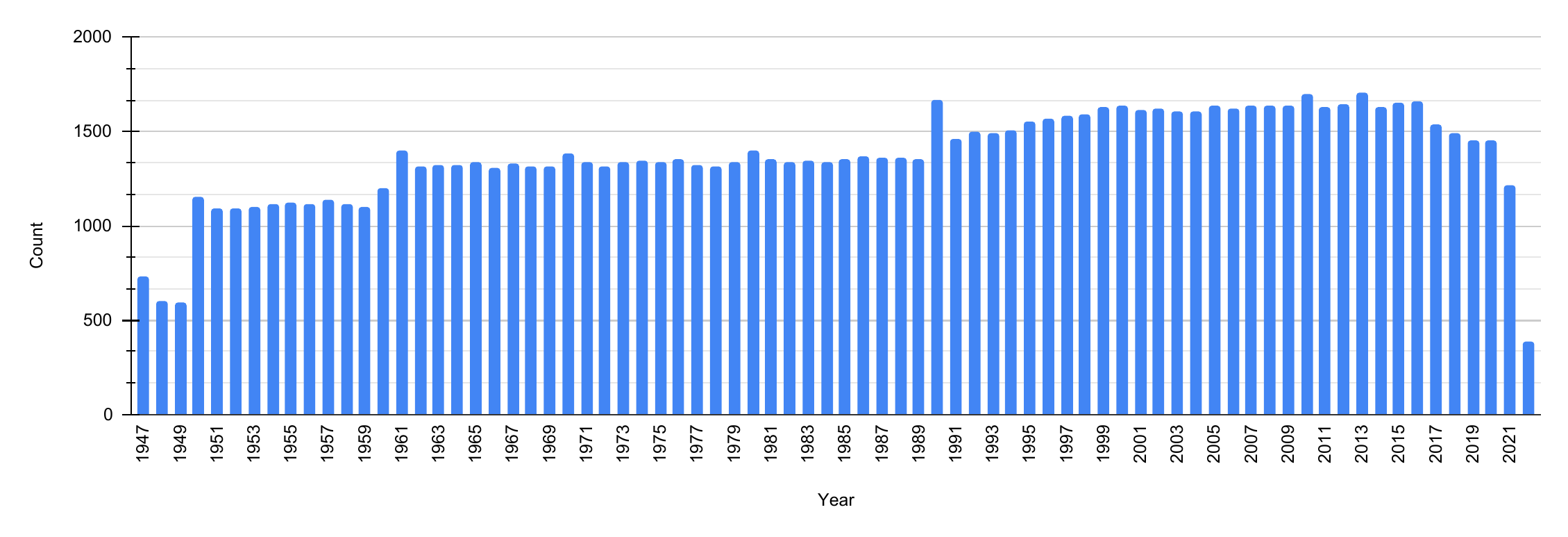}
\end{center}
\caption{Plot for the number of MCQs in the Date-based metric ($DB$) per year.}
\label{fig:mcq-db}
\end{figure*}

\begin{figure*}
\begin{center}
\includegraphics[width=0.8\linewidth]{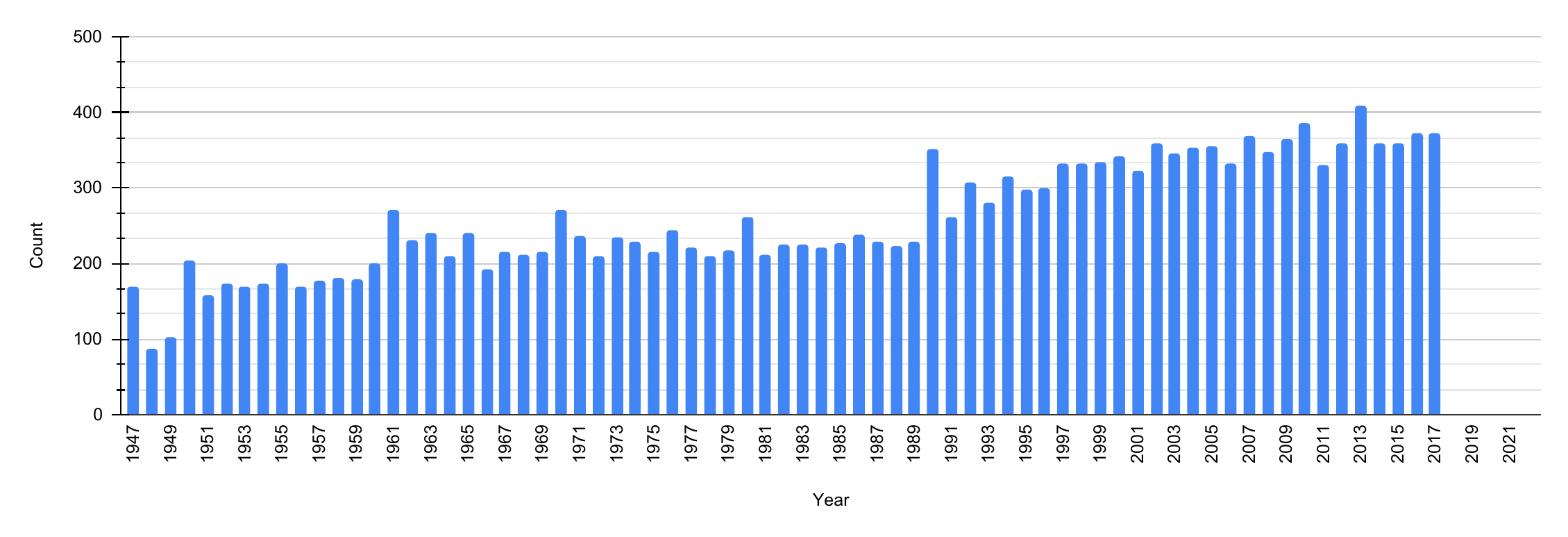}
\end{center}
\caption{Plot for the number of MCQs in the Comparative-based metric ($CP$) per year.}
\label{fig:mcq-rb}
\end{figure*}

\begin{figure*}
\begin{center}
\includegraphics[width=0.8\linewidth]{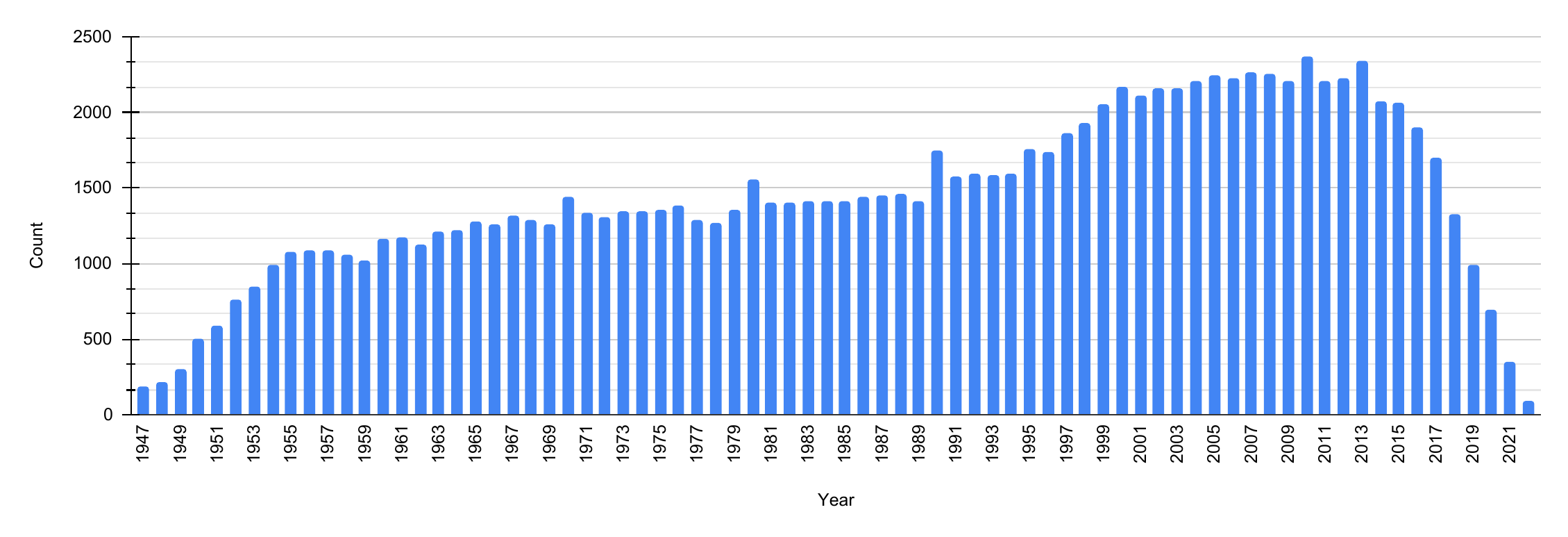}
\end{center}
\caption{Plot for the number of MCQs in the Window-based metric ($WB$) per year.}
\label{fig:mcq-wb}
\end{figure*}

\begin{figure*}
\begin{center}
\includegraphics[width=0.8\linewidth]{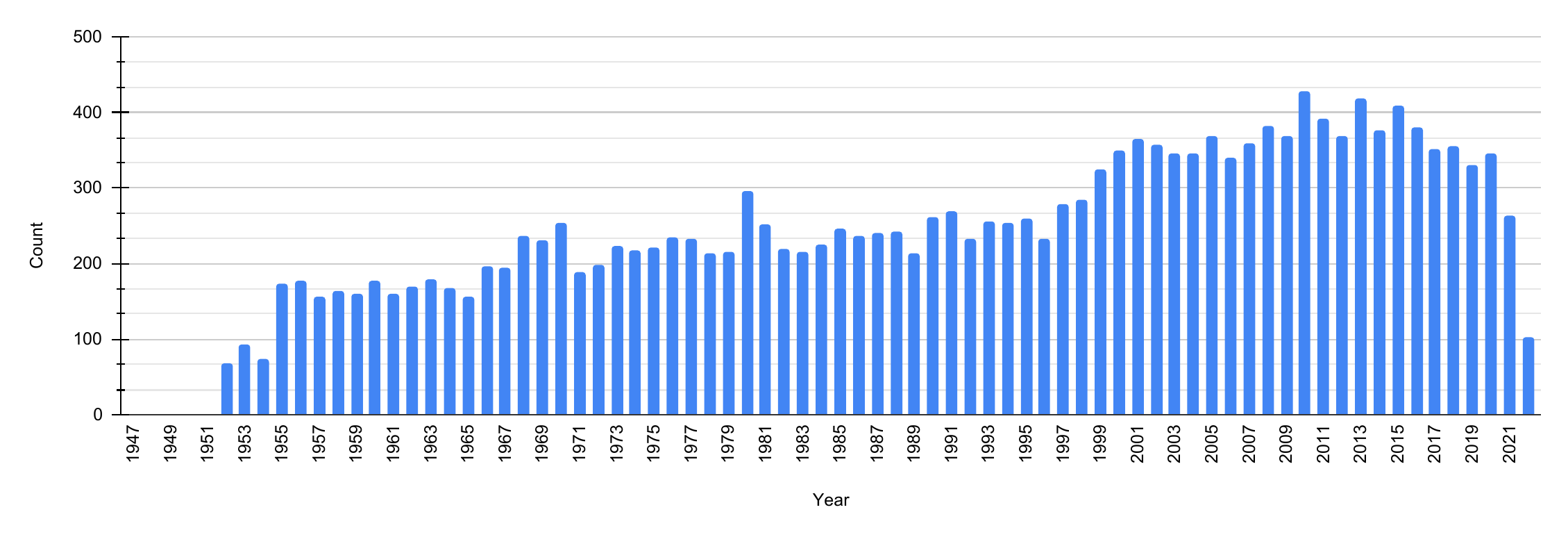}
\end{center}
\caption{Plot for the number of MCQs in the Min/Max-based metric ($MM$) per year.}
\label{fig:mcq-mmb}
\end{figure*}

\begin{figure*}
\begin{center}
\includegraphics[width=0.8\linewidth]{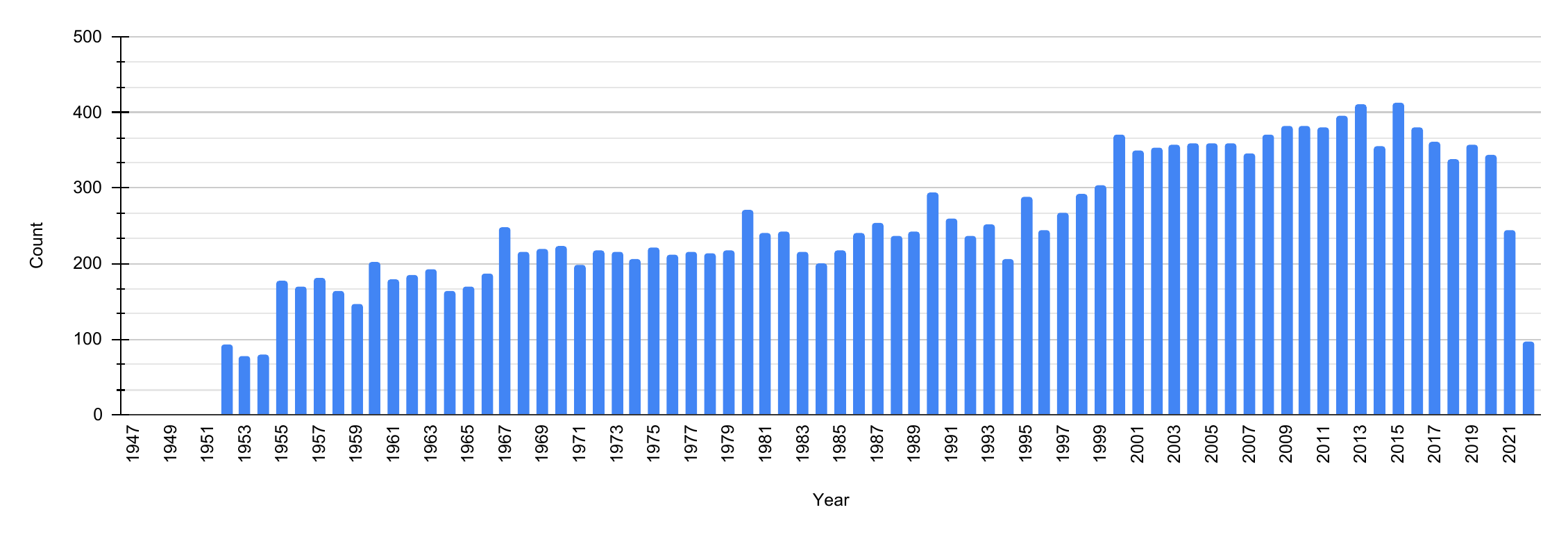}
\end{center}
\caption{Plot for the number of MCQs in the Range-based metric ($RB$) per year.}
\label{fig:mcq-rab}
\end{figure*}

\begin{figure*}
\begin{center}
\includegraphics[width=0.8\linewidth]{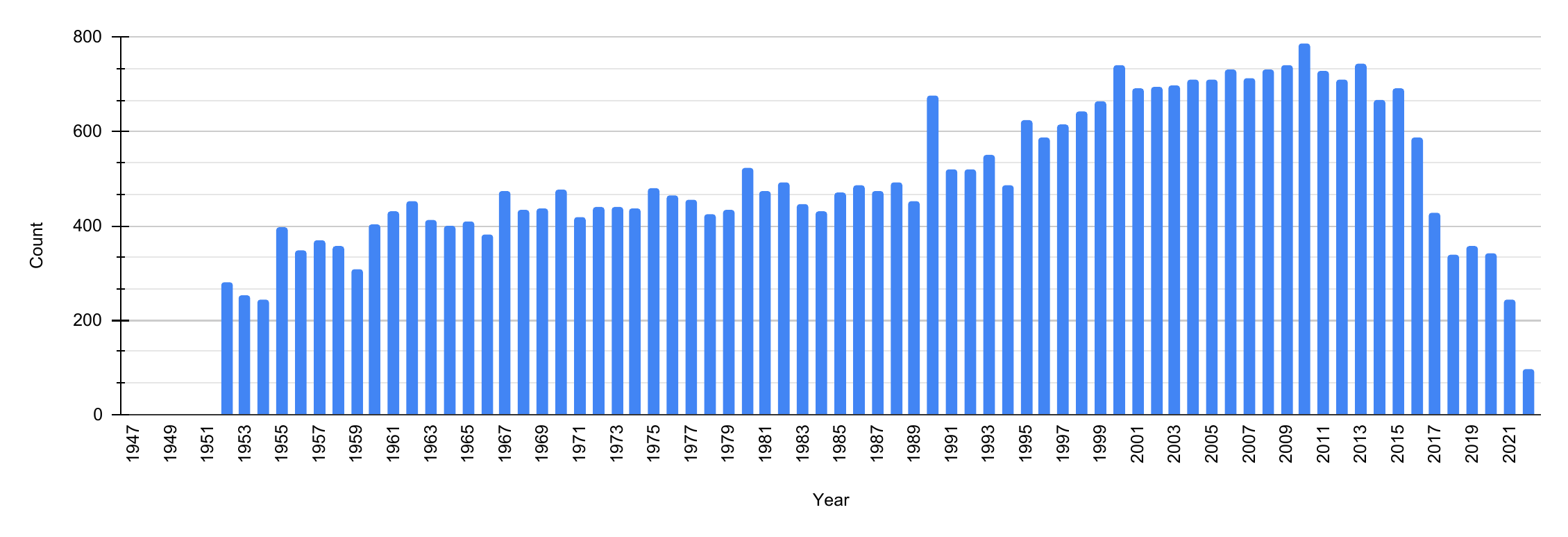}
\end{center}
\caption{Plot for the number of MCQs in the Trend-based metric ($TB$) per year.}
\label{fig:mcq-tb}
\end{figure*}

This section explains the details linked to the creation of the dataset. As described in Section~\ref{sec:dataset}, the data is curated from the \textit{Our World in Data (OWD)} site. The site scraps different trusted sources that are reliable and report accurate numbers. We iteratively parsed the site and processed the raw tabular data in the $<C, I, Y_t>$ and $V_t$ template format. The raw data is then categorized into the United Nations-focused domains and respective subcategories, and number of instances per subcategory in Tables as~\ref{table:issues_list1} (Climate), \ref{table:issues_list2} (Food and Agriculture), \ref{table:issues_list3} (Health), \ref{table:issues_list4} (Human Rights), \ref{table:issues_list5} (Innovation and Technological Change), \ref{table:issues_list6} (Migration), \ref{table:issues_list7} (Poverty, Economic Development, and Community), and \ref{table:issues_list8} (Peace and War).
We highlight the yearwise count of MCQs in each MCQ-based strategy as: Figure~\ref{fig:mcq-db} ($DB$), \ref{fig:mcq-rb} ($CP$), \ref{fig:mcq-wb} ($WB$), \ref{fig:mcq-mmb} ($MM$), \ref{fig:mcq-rab} ($RB$), and \ref{fig:mcq-tb} ($TB$). Overall, we also showcase the category-wise distribution for each strategy in Table~\ref{tab:micro_prompts_tempuns}. Apart from categories `C4' and `C5', all of the categories seem to have a higher number of MCQs. 
\subsection{MCQ-Based Strategy Yearwise Performance - Zero Shot Results}
\label{sec:zeroshot}
\begin{table*}[]
\centering
\begin{tabular}{lllllll} \hline
\textbf{Model} & $DB$ & $CP$ & $WB$ & $RB$ & $MM$ & $TB$ \\  \hline
\texttt{phi-2} & \ref{tab:result_Phi2_DB} & \ref{tab:result_Phi2_RB_} & \ref{tab:result_Phi2_WB_} & \ref{tab:result_Phi2_MMB_} & \ref{tab:result_Phi2_RAB_} & \ref{tab:result_Phi2_TB_} \\
\texttt{flan-t5-xl} & \ref{tab:result_flan-t5_DB} & \ref{tab:result_flan-t5_RB_} & \ref{tab:result_flan-t5_WB_} & \ref{tab:result_flan-t5_MMB_} & \ref{tab:result_flan-t5_RAB_} & \ref{tab:result_flan-t5_TB_} \\
\texttt{mistral-instruct} & \ref{tab:result_Mistral_DB} & \ref{tab:result_Mistral_RB} & \ref{tab:result_Mistral_WB} & \ref{tab:result_Mistral_MMB} & \ref{tab:result_Mistral_RAN} & \ref{tab:result_Mistral_TB} \\
\texttt{llama-2-chat} & \ref{tab:result_LLaMA_DB_} & \ref{tab:result_LLaMA_RB_} & \ref{tab:result_LLaMA_WB_} & \ref{tab:result_LLaMA_MMB_} & \ref{tab:result_LLaMA_RAB_} & \ref{tab:result_LLaMA_TB_} \\
\texttt{gemma-7b-it} & \ref{tab:result_gemma_DB_} & \ref{tab:result_gemma_CB_} & \ref{tab:result_gemma_WB_} & \ref{tab:result_gemma_MM_} & \ref{tab:result_gemma_RB_} & \ref{tab:result_gemma_TB_} \\
\texttt{llama-3-8b} & \ref{tab:result_llama3_8b_DB_} & \ref{tab:result_llama3_8b_CB_} & \ref{tab:result_llama3_8b_WB_} & \ref{tab:result_llama3_8b_MM_} & \ref{tab:result_llama3_8b_RB_} & \ref{tab:result_llama3_8b_TB_} \\
\texttt{phi-3-instruct} & \ref{tab:results-phi-3-instructDB_} & \ref{tab:results-phi-3-instructCB_} & \ref{tab:results-phi-3-instructWB_} & \ref{tab:results-phi-3-instructMM_} & \ref{tab:results-phi-3-instructRB_} & \ref{tab:results-phi-3-instructTB_} \\
\texttt{mixtral-8x7b} & \ref{tab:result_mixtral-87b_DB_} & \ref{tab:result_mixtral-87b_CB_} & \ref{tab:result_mixtral-87b_WB_} & \ref{tab:result_mixtral-87b_MM_} & \ref{tab:result_mixtral-87b_RB_} & \ref{tab:result_mixtral-87b_TB_} \\
\texttt{llama-3-70b} & \ref{tab:result_llama-370b_DB_} & \ref{tab:result_llama-370b_CB_} & \ref{tab:result_llama-370b_WB_} & \ref{tab:result_llama-370b_MM_} & \ref{tab:result_llama-370b_RB_}& \ref{tab:result_llama-370b_TB_} \\
\texttt{gpt-3.5-turbo} & \ref{tab:result_gpt-35_DB_} & \ref{tab:result_gpt-35_CP_} & \ref{tab:result_gpt-35_WB_} & \ref{tab:result_gpt-35_MM_} & \ref{tab:result_gpt-35_RB_} & \ref{tab:result_gpt-35_TB_} \\
\texttt{gpt-4} & \ref{tab:result_gpt-4_DB_} & \ref{tab:result_gpt-4_RB_} & \ref{tab:result_gpt-4_WB_} & \ref{tab:result_gpt-4_MMB_} & \ref{tab:result_gpt-4_RAB_} & \ref{tab:result_gpt-4_TB_} \\
\texttt{gemini-pro} & \ref{tab:result_gemini_DB_} & \ref{tab:result_gemini_RB_} & \ref{tab:result_gemini_WB_} & \ref{tab:result_gemini_MMB_} & \ref{tab:result_gemini_RAB_} & \ref{tab:result_gemini_TB_} \\ \hline
\end{tabular}
\caption{\label{table:index_zero}
The index table of the category \textbf{tables} for the \textbf{Zero-shot} evaluations over open and closed source models.}
\end{table*}

\begin{table*}[]
\centering
\begin{tabular}{lllllll} \hline
\textbf{Model} & $DB$ & $CP$ & $WB$ & $RB$ & $MM$ & $TB$ \\  \hline
\texttt{phi-2} & \ref{fig:date-based-phi2} & \ref{fig:range-based-phi2} & \ref{fig:window-based-phi2} & \ref{fig:mmb-based-phi2} & \ref{fig:rab-based-phi2} & \ref{fig:trend-based-phi2} \\
\texttt{flan-t5-xl} & \ref{fig:date-based-flan-t5-xl} & \ref{fig:range-based-flan-t5-xl} & \ref{fig:window-based-flan-t5-xl} & \ref{fig:mmb-based-flan-t5-xl} & \ref{fig:rab-based-flan-t5-xl} & \ref{fig:tb-based-flan-t5-xl} \\
\texttt{mistral-instruct} & \ref{fig:date-based-mistral} & \ref{fig:range-based-mistral} & \ref{fig:window-based-mistral} & \ref{fig:mmb-based-mistral} & \ref{fig:rab-based-mistral} & \ref{fig:trend-based-mistral} \\
\texttt{llama-2-chat} & \ref{fig:date-based-LLaMA} & \ref{fig:range-based-LLaMA} & \ref{fig:window-based-LLaMA} & \ref{fig:mmb-based-LLaMA} & \ref{fig:rab-based-LLaMA} & \ref{fig:trend-based-LLaMA} \\
\texttt{gemma-7b-it} & \ref{fig:date-based-gemma} & \ref{fig:range-based-gemma} & \ref{fig:window-based-gemma} & \ref{fig:mmb-based-gemma} & \ref{fig:rab-based-gemma} & \ref{fig:trend-based-gemma} \\
\texttt{llama-3-8b} & \ref{fig:date-based-llama38b} & \ref{fig:range-based-llama38b} & \ref{fig:window-based-llama38b} & \ref{fig:mmb-based-llama38b} & \ref{fig:rab-based-llama38b} & \ref{fig:trend-based-llama38b} \\
\texttt{phi-3-instruct} & \ref{fig:date-based-phi3} & \ref{fig:range-based-phi3} & \ref{fig:window-based-phi3} & \ref{fig:mmb-based-phi3} & \ref{fig:rab-based-phi3} & \ref{fig:trend-based-phi3} \\
\texttt{mixtral-8x7b} & \ref{fig:date-based-mixtral} & \ref{fig:range-based-mixtral} & \ref{fig:window-based-mixtral} & \ref{fig:mmb-based-mixtral} & \ref{fig:rab-based-mixtral} & \ref{fig:trend-based-mixtral} \\
\texttt{llama-3-70b} & \ref{fig:date-based-llama70b} & \ref{fig:range-based-llama70b} & \ref{fig:window-based-llama70b} & \ref{fig:mmb-based-llama70b} & \ref{fig:rab-based-llama70b}& \ref{fig:trend-based-llama70b} \\
\texttt{gpt-3.5-turbo} & \ref{fig:date-based-gpt-35} & \ref{fig:range-based-gpt-35} & \ref{fig:window-based-gpt-35} & \ref{fig:mmb-based-gpt-35} & \ref{fig:rab-based-gpt-35} & \ref{fig:trend-based-gpt-35} \\
\texttt{gpt-4} & \ref{fig:date-based-gpt-4} & \ref{fig:range-based-gpt-4} & \ref{fig:window-based-gpt-4} & \ref{fig:mmb-based-gpt-4} & \ref{fig:rab-based-gpt-4} & \ref{fig:tb-based-gpt-4} \\
\texttt{gemini-pro} & \ref{fig:date-based-gemini-pro} & \ref{fig:range-based-gemini-pro} & \ref{fig:window-based-gemini-pro} & \ref{fig:mmb-based-gemini-pro} & \ref{fig:rab-based-gemini-pro} & \ref{fig:trend-based-gemini-pro} \\ \hline
\end{tabular}
\caption{\label{table:index_zero_plots}
The index table of \textbf{plots} for the \textbf{Zero-shot} evaluations for open-source and closed-source models.}
\end{table*}

\begin{table*}[]
\centering
\begin{tabular}{lllllll} \hline
\textbf{Models} & \multicolumn{1}{c}{\textbf{$DB$}} & \multicolumn{1}{c}{\textbf{$CP$}} & \multicolumn{1}{c}{\textbf{$WB$}} & \multicolumn{1}{c}{\textbf{$MM$}} & \multicolumn{1}{c}{\textbf{$RB$}} & \multicolumn{1}{c}{\textbf{$TB$}} \\ \hline
\texttt{phi-2} & \ref{fig:date-based-ft-phi-2} & \ref{fig:rb-based-ft-phi-2} & \ref{fig:window-based-ft-phi-2} & \ref{fig:minmax-based-ft-phi-2} & \ref{fig:rab-based-ft-phi-2} & \ref{fig:tb-based-ft-phi-2} \\
\texttt{flan-t5-xl} & \ref{fig:date-based-ft-flan-t5-xl} & \ref{fig:rb-based-ft-flan-t5-xl} & \ref{fig:window-based-ft-flan-t5-xl} & \ref{fig:minmax-based-ft-flan-t5-xl} & \ref{fig:rab-based-ft-flan-t5-xl} & \ref{fig:tb-based-ft-flan-t5-xl} \\
\texttt{mistral-instruct} & \ref{fig:date-based-ft-mistral} & \ref{fig:range-based-ft-mistral} & \ref{fig:window-based-ft-mistral} & \ref{fig:minmax-based-ft-mistral} & \ref{fig:rab-based-ft-mistral} & \ref{fig:tb-based-ft-mistral} \\
\texttt{llama-2-chat} & \ref{fig:date-based-ft-llama-2} & \ref{fig:rb-based-ft-llama-2} & \ref{fig:window-based-ft-llama-2} & \ref{fig:minmax-based-ft-llama-2} & \ref{fig:rab-based-ft-llama-2} & \ref{fig:tb-based-ft-llama-2} \\
\texttt{gemma-7b-it} & \ref{fig:date-based-ft-gemma-7b-it} & \ref{fig:rb-based-ft-gemma-7b-it} & \ref{fig:window-based-ft-gemma-7b-it} & \ref{fig:minmax-based-ft-gemma-7b-it} & \ref{fig:rab-based-ft-gemma-7b-it} & \ref{fig:tb-based-ft-gemma-7b-it} \\
\texttt{llama-3-8b} & \ref{fig:date-based-ft-llama-3-8b} & \ref{fig:rb-based-ft-llama-3-8b} & \ref{fig:window-based-ft-llama-3-8b} & \ref{fig:minmax-based-ft-llama-3-8b} & \ref{fig:rab-based-ft-llama-3-8b} & \ref{fig:tb-based-ft-llama-3-8b} \\
\texttt{phi-3-instruct} & \ref{fig:date-based-ft-phi-3-medium} & \ref{fig:rb-based-ft-phi-3-medium} & \ref{fig:window-based-ft-phi-3-medium} & \ref{fig:minmax-based-ft-phi-3-medium} & \ref{fig:rab-based-ft-phi-3-medium} & \ref{fig:tb-based-ft-phi-3-medium}\\  \hline
\end{tabular}
\caption{\label{table:index_continual}
The index table of plots for the \textbf{Continual Learning} evaluations for open-source models.}
\end{table*}

\begin{table*}[]
\centering
\begin{tabular}{lllllll} \hline
\textbf{Models} & \multicolumn{1}{c}{\textbf{$DB$}} & \multicolumn{1}{c}{\textbf{$CP$}} & \multicolumn{1}{c}{\textbf{$WB$}} & \multicolumn{1}{c}{\textbf{$MM$}} & \multicolumn{1}{c}{\textbf{$RB$}} & \multicolumn{1}{c}{\textbf{$TB$}} \\ \hline
\texttt{phi-2} & \ref{fig:date-based-yl-phi2} & \ref{fig:cp-yl-phi2} & \ref{fig:window-based-yl-phi2} & \ref{fig:minmax-based-yl-phi2} & \ref{fig:rab-based-yl-phi2} & \ref{fig:tb-based-yl-phi2} \\
\texttt{flan-t5-xl} & \ref{fig:date-based-yl-flan-t5-xl} & \ref{fig:cate-yl-flan-t5-xl} & \ref{fig:window-based-yl-flan-t5-xl} & \ref{fig:minmax-based-yl-flan-t5-xl} & \ref{fig:rab-based-yl-flan-t5-xl} & \ref{fig:tb-based-yl-flan-t5-xl} \\
\texttt{mistral-instruct} & \ref{fig:date-based-yl-mistral} & \ref{fig:cp-yl-mistral} & \ref{fig:window-based-yl-mistral} & \ref{fig:minmax-based-yl-mistral} & \ref{fig:rab-based-yl-mistral} & \ref{fig:tb-based-yl-mistral} \\
\texttt{lama-2-chat} & \ref{fig:date-based-yl-llama} & \ref{fig:cp-yl-llama} & \ref{fig:window-based-yl-llama} & \ref{fig:minmax-based-yl-llama} & \ref{fig:rab-based-yl-llama} & \ref{fig:tb-based-yl-llama} \\
\texttt{gemma-7b-it} & \ref{fig:date-based-yl-gemma-7b-it} & \ref{fig:cp-yl-gemma-7b-it} & \ref{fig:window-based-yl-gemma-7b-it} & \ref{fig:minmax-based-yl-gemma-7b-it} & \ref{fig:rab-based-yl-gemma-7b-it} & \ref{fig:tb-based-yl-gemma-7b-it} \\
\texttt{llama-3-8b} & \ref{fig:date-based-yl-llama-3-8b} & \ref{fig:cp-yl-llama-3-8b} & \ref{fig:window-based-yl-llama-3-8b} & \ref{fig:minmax-based-yl-llama-3-8b} & \ref{fig:rab-based-yl-llama-3-8b} & \ref{fig:tb-based-yl-llama-3-8b} \\
\texttt{phi-3-instruct} & \ref{fig:date-based-yl-phi-3-medium} & \ref{fig:cp-yl-phi-3-medium} & \ref{fig:window-based-yl-phi-3-medium} & \ref{fig:minmax-based-yl-phi-3-medium} & \ref{fig:rab-based-yl-phi-3-medium} & \ref{fig:tb-based-yl-phi-3-medium} \\ \hline
\end{tabular}
\caption{\label{table:index_yearwise}
The index table of plots for the \textbf{Yearwise Finetuning} evaluations for open-source models.}
\end{table*}

\begin{table*}[]
\centering
\begin{tabular}{lllllll} \hline
\textbf{Models} & \multicolumn{1}{c}{\textbf{$DB$}} & \multicolumn{1}{c}{\textbf{$CP$}} & \multicolumn{1}{c}{\textbf{$WB$}} & \multicolumn{1}{c}{\textbf{$MM$}} & \multicolumn{1}{c}{\textbf{$RB$}} & \multicolumn{1}{c}{\textbf{$TB$}} \\ \hline
\texttt{phi-2} & \ref{fig:date-based-rn-phi2} & \ref{fig:cp-rn-phi2} & \ref{fig:window-based-rn-phi2} & \ref{fig:minmax-based-rn-phi2} & \ref{fig:rab-based-rn-phi2}  & \ref{fig:tb-based-rn-phi2} \\
\texttt{flan-t5-xl} & \ref{fig:date-based-rn-flan-t5-xl} & \ref{fig:cp-rn-flan-t5-xl} & \ref{fig:window-based-rn-flan-t5-xl} & \ref{fig:minmax-based-rn-flan-t5-xl} & \ref{fig:rab-based-rn-flan-t5-xl} & \ref{fig:tb-based-rn-flan-t5-xl} \\
\texttt{mistral-instruct} & \ref{fig:date-based-rn-mistral}  & \ref{fig:cp-rn-mistral}  & \ref{fig:window-based-rn-mistral} & \ref{fig:minmax-based-rn-mistral} & \ref{fig:rab-based-rn-mistral} &  \ref{fig:tb-based-rn-mistral} \\
\texttt{llama-2-chat }& \ref{fig:date-based-rn-llama}  & \ref{fig:cp-rn-llama}  & \ref{fig:window-based-rn-llama}  & \ref{fig:minmax-based-rn-llama}  & \ref{fig:rab-based-rn-llama}  &  \ref{fig:tb-based-rn-llama} \\
\texttt{gemma-7b-it} & \ref{fig:date-based-rn-gemma-7b-it} & \ref{fig:cp-rn-gemma-7b-it} & \ref{fig:window-based-rn-gemma-7b-it} & \ref{fig:minmax-based-rn-gemma-7b-it} & \ref{fig:rab-based-rn-gemma-7b-it} & \ref{fig:tb-based-rn-gemma-7b-it} \\
\texttt{llama-3-8b} & \ref{fig:date-based-rn-llama-3-8b} & \ref{fig:cp-rn-llama-3-8b} & \ref{fig:window-based-rn-llama-3-8b} & \ref{fig:minmax-based-rn-llama-3-8b} & \ref{fig:rab-based-rn-llama-3-8b} & \ref{fig:tb-based-rn-llama-3-8b} \\
\texttt{phi-3-instruct} & \ref{fig:date-based-rn-phi-3-medium} & \ref{fig:cp-rn-phi-3-medium} & \ref{fig:window-based-rn-phi-3-medium} & \ref{fig:minmax-based-rn-phi-3-medium} & \ref{fig:rab-based-rn-phi-3-medium} & \ref{fig:tb-based-rn-phi-3-medium}\\ \hline 
\end{tabular}
\caption{\label{table:index_ranom}
The index table of plots for the \textbf{Random Finetuning} evaluations for open-source models.}
\end{table*}

\begin{figure*}
\begin{center}
\includegraphics[width=0.9\linewidth]{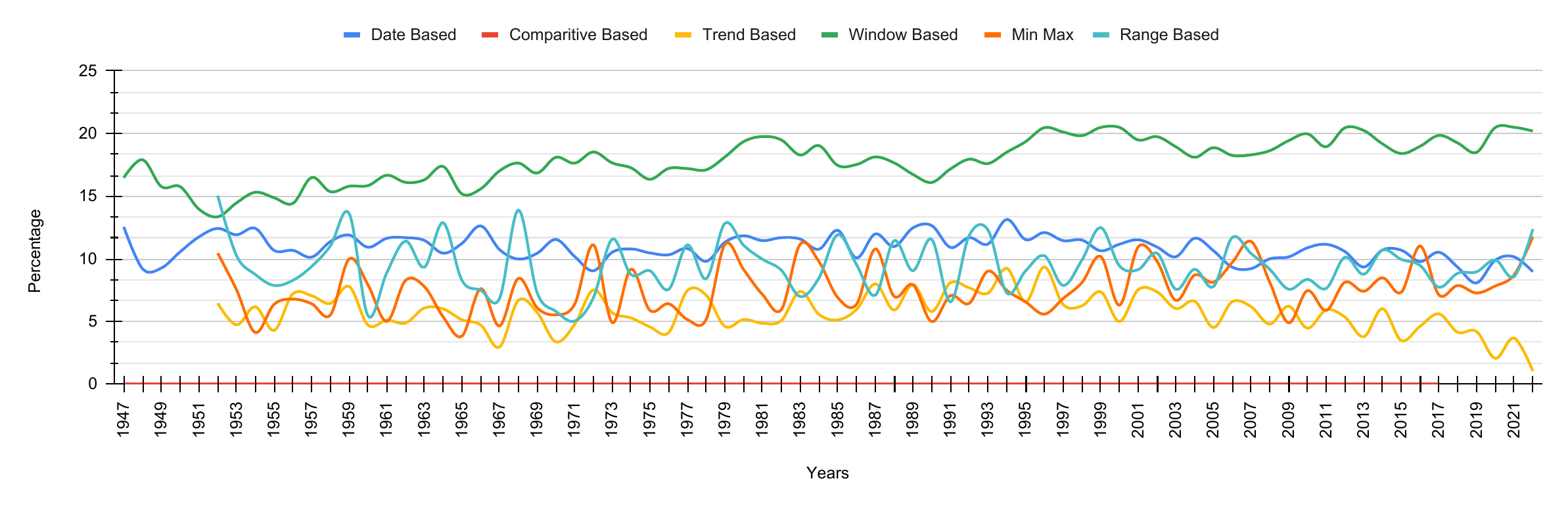}
\end{center}
\caption{\textbf{Zeroshot} MCQ-based evaluation on \texttt{phi-2}.}
\label{fig:phi2-mcq}
\end{figure*}

\begin{figure*}
\begin{center}
\includegraphics[width=0.9\linewidth]{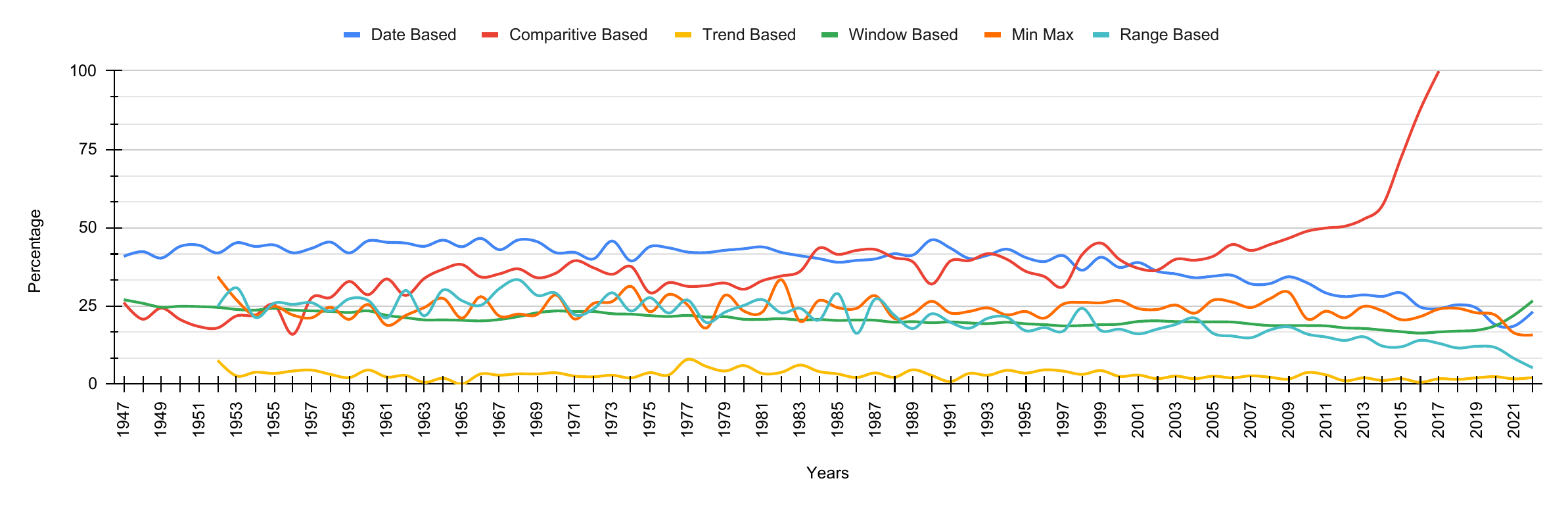}
\end{center}
\caption{\textbf{Zeroshot} MCQ-based evaluation on \texttt{flan-t5-xl}.}
\label{fig:t5-mcq}
\end{figure*}

\begin{figure*}
\begin{center}
\includegraphics[width=0.9\linewidth]{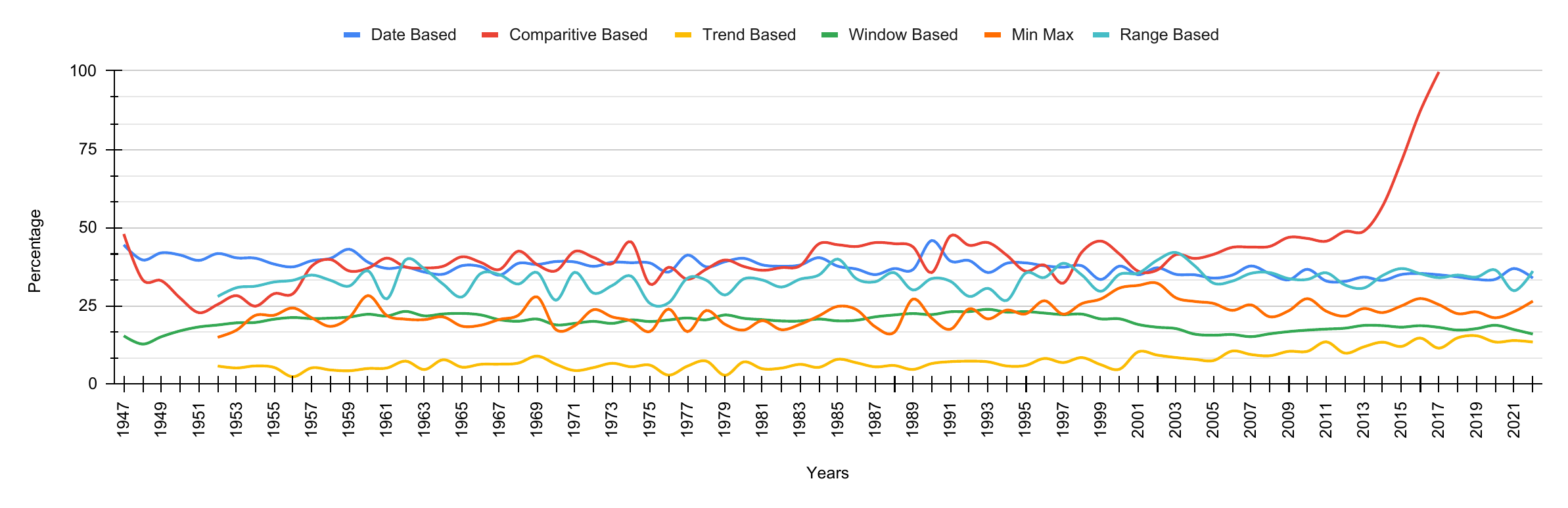}
\end{center}
\caption{\textbf{Zeroshot} MCQ-based evaluation on \texttt{mistral-instruct}.}
\label{fig:mistral-mcq}
\end{figure*}

\begin{figure*}
\begin{center}
\includegraphics[width=0.9\linewidth]{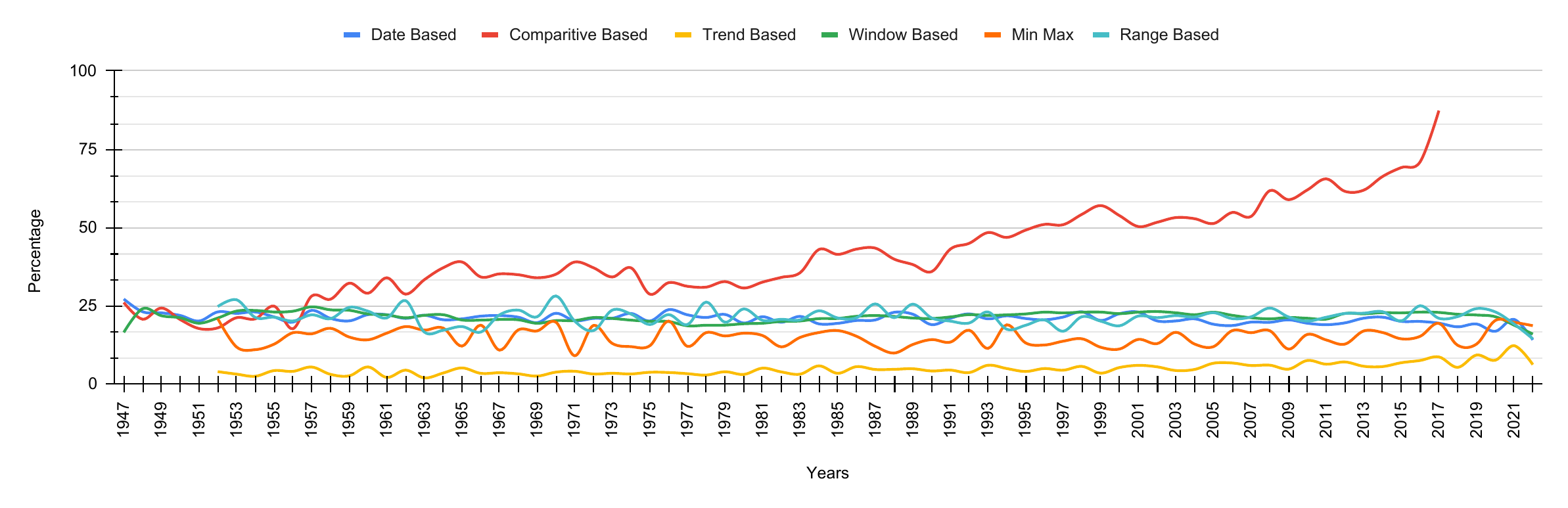}
\end{center}
\caption{\textbf{Zeroshot} MCQ-based evaluation on \texttt{llama-2}.}
\label{fig:llama-mcq}
\end{figure*}

\begin{figure*}
\begin{center}
\includegraphics[width=0.9\linewidth]{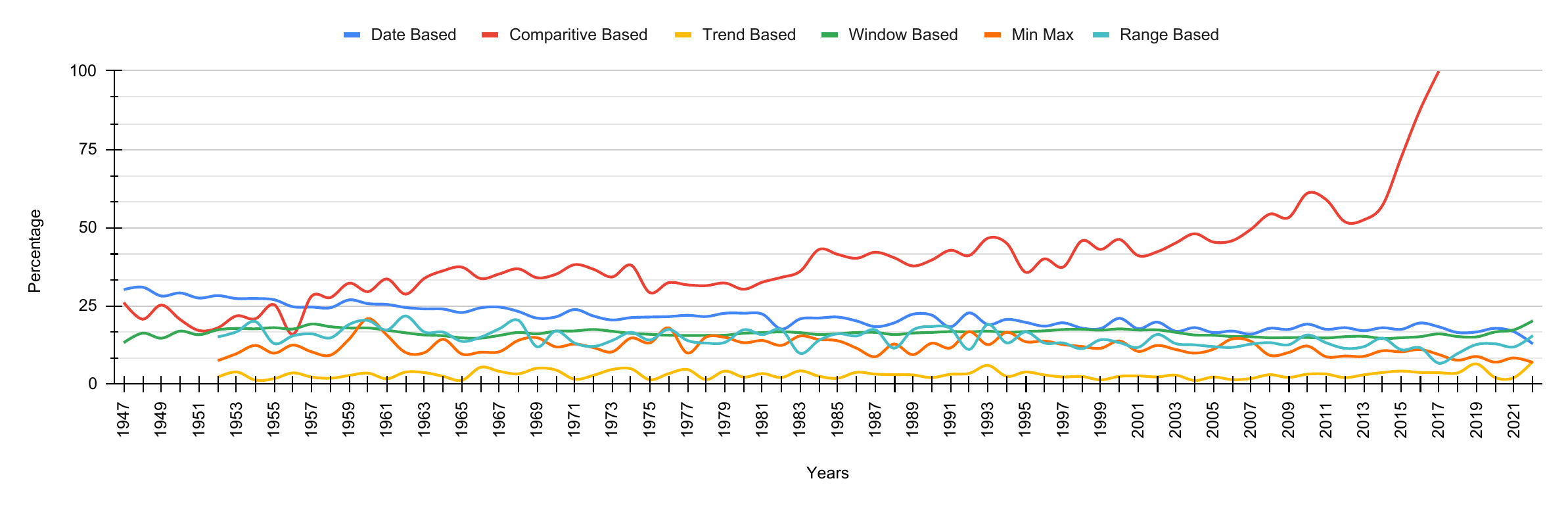}
\end{center}
\caption{\textbf{Zeroshot} MCQ-based evaluation on \texttt{gemma-7b-it}.}
\label{fig:gemma-mcq}
\end{figure*}

\begin{figure*}
\begin{center}
\includegraphics[width=0.9\linewidth]{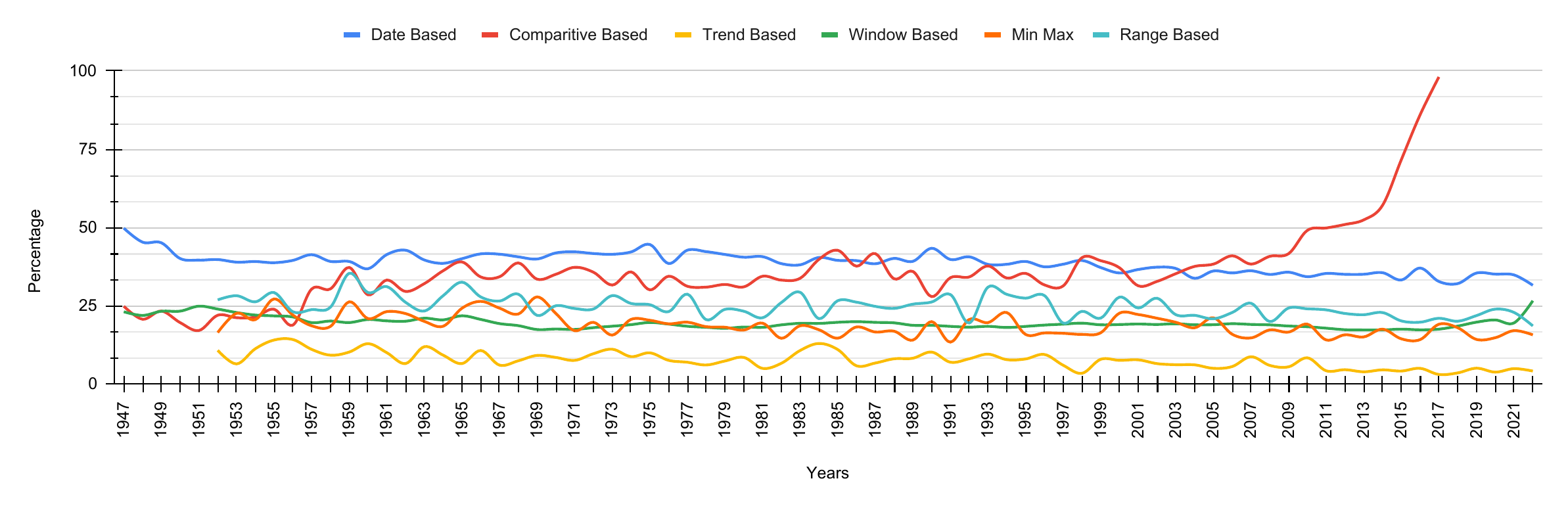}
\end{center}
\caption{\textbf{Zeroshot} MCQ-based evaluation on \texttt{llama-3-8b}.}
\label{fig:llama3-8-mcq}
\end{figure*}

\begin{figure*}
\begin{center}
\includegraphics[width=0.9\linewidth]{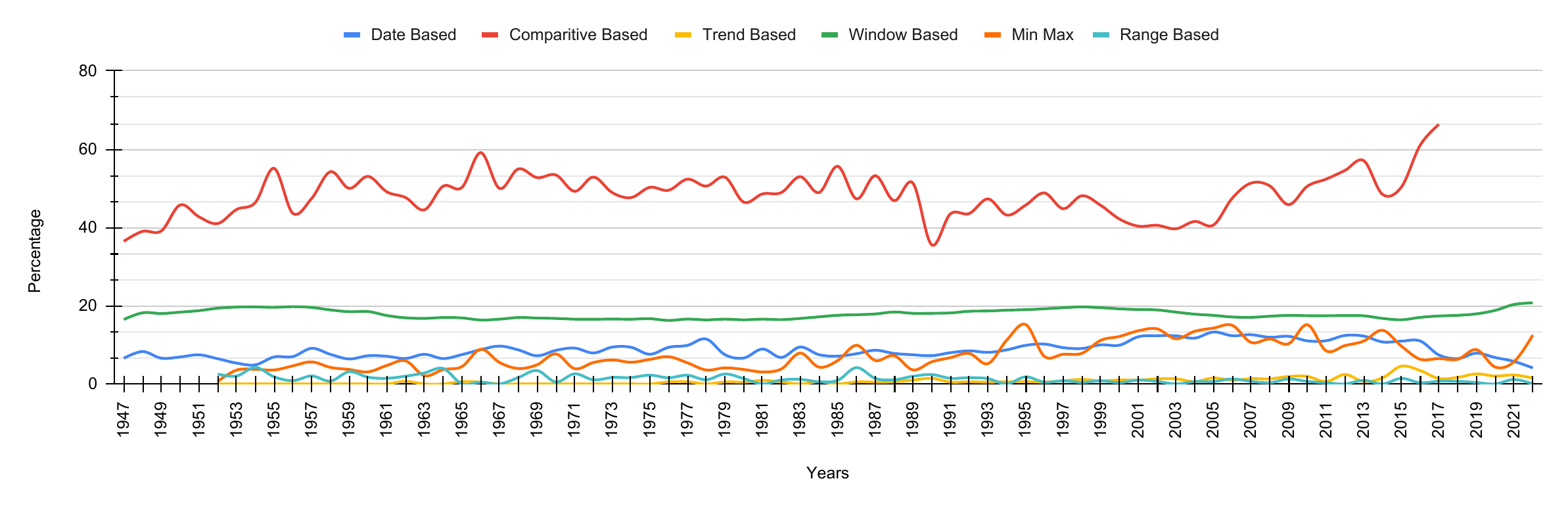}
\end{center}
\caption{\textbf{Zeroshot} MCQ-based evaluation on \texttt{phi-3-instruct}.}
\label{fig:phi3-mcq}
\end{figure*}

\begin{figure*}
\begin{center}
\includegraphics[width=0.9\linewidth]{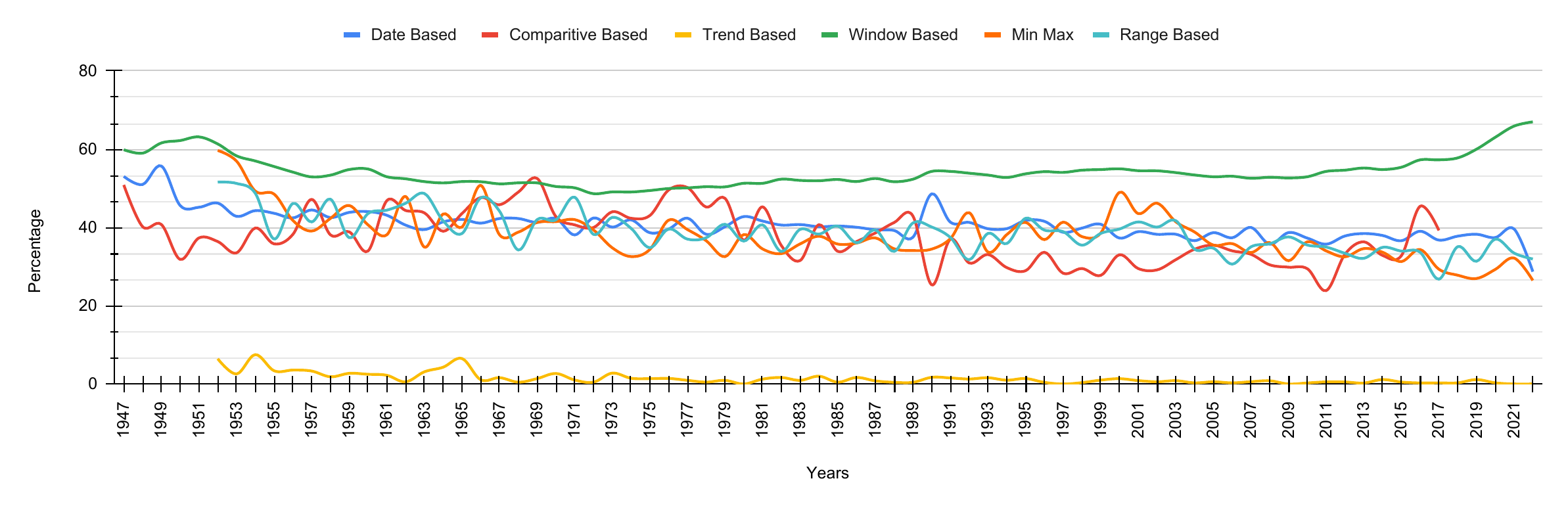}
\end{center}
\caption{\textbf{Zeroshot} MCQ-based evaluation on \texttt{mixtral-8x7b}.}
\label{fig:mixtral-8x70b-mcq}
\end{figure*}

\begin{figure*}
\begin{center}
\includegraphics[width=0.9\linewidth]{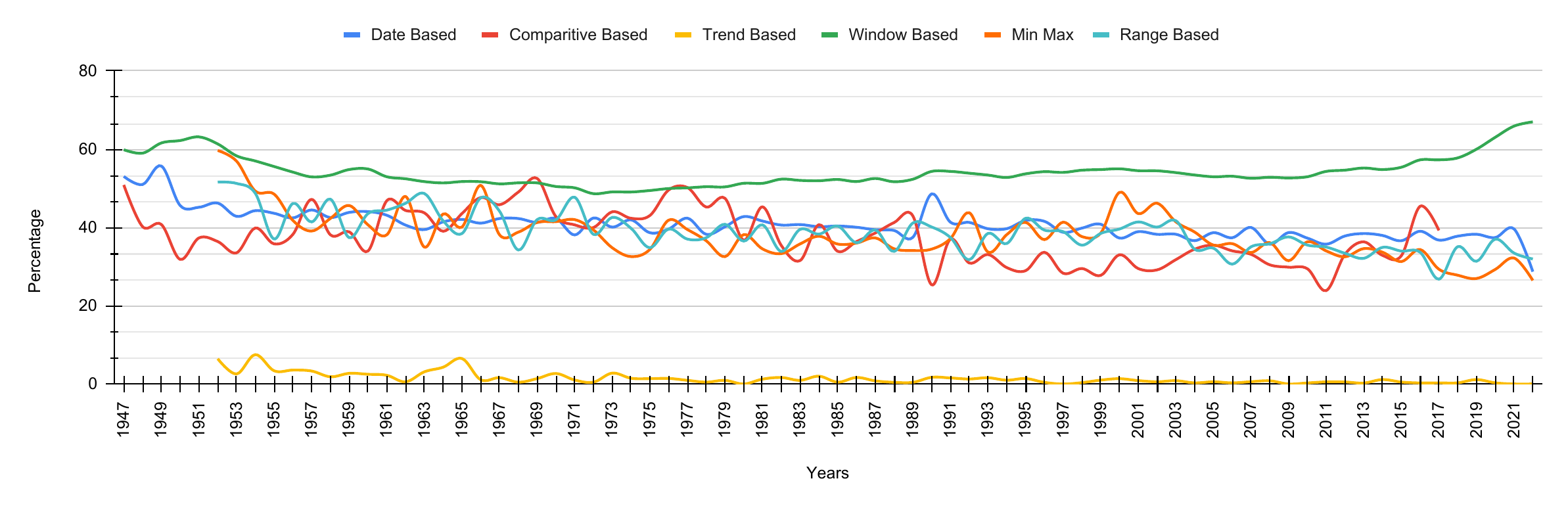}
\end{center}
\caption{\textbf{Zeroshot} MCQ-based evaluation on \texttt{llama-3-70B}.}
\label{fig:llama3-70B-mcq}
\end{figure*}

\begin{figure*}
\begin{center}
\includegraphics[width=0.9\linewidth]{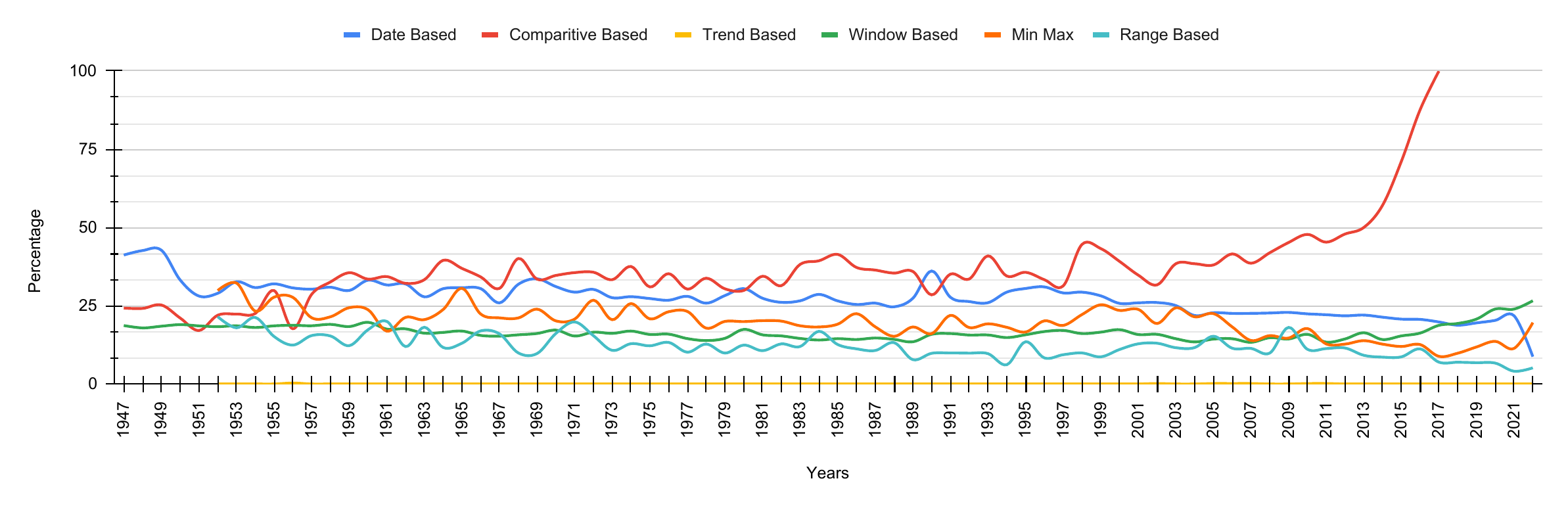}
\end{center}
\caption{\textbf{Zeroshot} MCQ-based evaluation on \texttt{gpt-3.5-turbo}.}
\label{fig:gpt-35-mcq}
\end{figure*}

\begin{figure*}
\begin{center}
\includegraphics[width=0.9\linewidth]{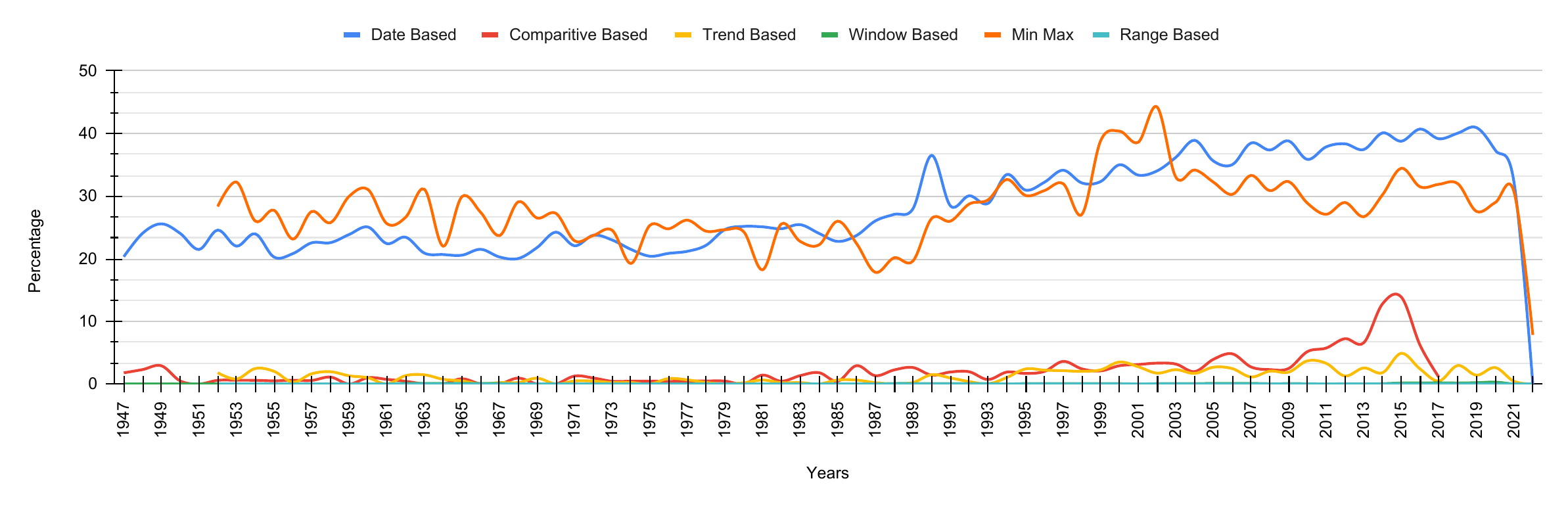}
\end{center}
\caption{\textbf{Zeroshot} MCQ-based evaluation on \texttt{gpt-4}.}
\label{fig:gpt-4-mcq}
\end{figure*}

\begin{figure*}
\begin{center}
\includegraphics[width=0.9\linewidth]{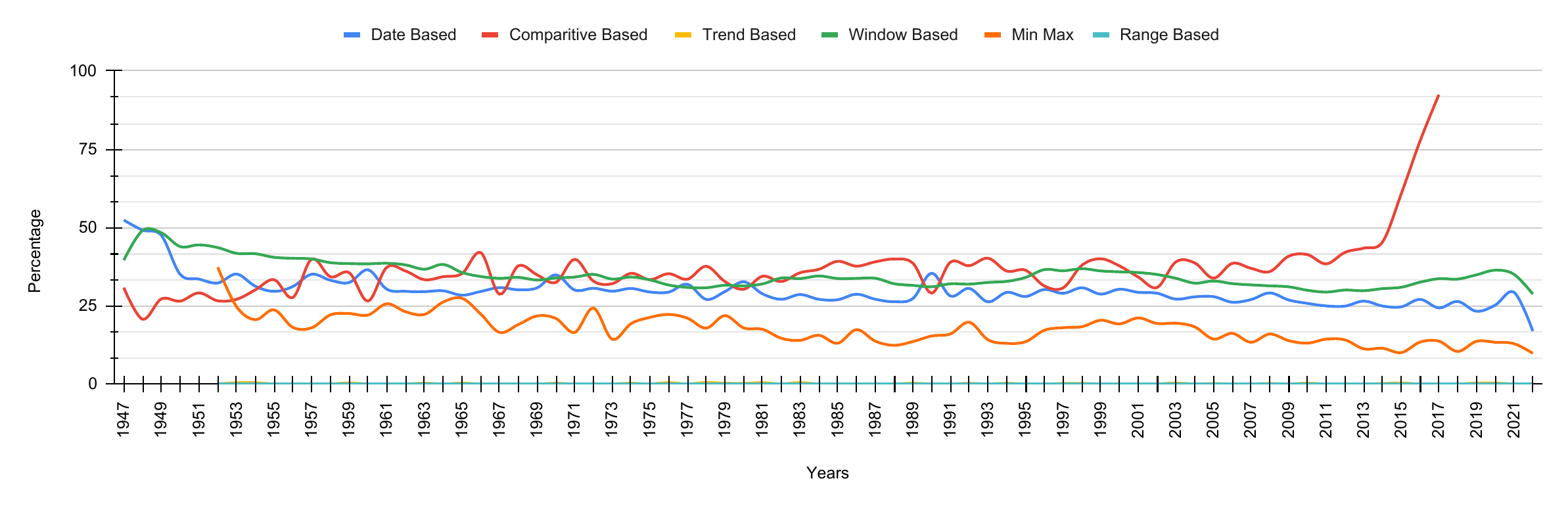}
\end{center}
\caption{\textbf{Zeroshot} MCQ-based evaluation on \texttt{gemini-pro}.}
\label{fig:gemini-pro-mcq}
\end{figure*}

Table~\ref{table:index_zero} highlights the index table for the results for zero-shot inferences over the 12 open and closed source LLMs per category, whereas Table~\ref{table:index_zero_plots} highlights the plots of inferences per metrics. We can observe that all the models (Tables~\ref{tab:result_Phi2_DB} to \ref{tab:result_gemini_TB_}) show that the LLMs produce more incorrect results than the correct results. We showcase the comparative analysis over the $DB$ metric in the Figures \ref{fig:DB-allmodels}. We highlight that the larger models (LLMs with >14B parameters) tends to store more information and are better at generating `information not available', rather than generating the `incorrect' predictions. 
We show the combined plots for all six metrics in Figures: \texttt{phi-2} (\ref{fig:phi2-mcq}), \texttt{flan-t5-xl} (\ref{fig:t5-mcq}), \texttt{mistral-instruct} (\ref{fig:mistral-mcq}), \texttt{llama-2} (\ref{fig:llama-mcq}), and \texttt{gemma-7b-it} (\ref{fig:gemma-mcq}), \texttt{llama-3-8b} (\ref{fig:llama3-8-mcq}), \texttt{phi-3} (\ref{fig:phi3-mcq}), \texttt{mixtral-8x7b} (\ref{fig:mixtral-8x70b-mcq}), \texttt{llama-3-70B} (\ref{fig:llama3-70B-mcq}), \texttt{gpt-35-turbo} (\ref{fig:gpt-35-mcq}), \texttt{gpt-4} (\ref{fig:gpt-4-mcq}), and \texttt{gemini-pro} (\ref{fig:gemini-pro-mcq}). But \texttt{gpt-4} (Figure~\ref{fig:llama3-70B-mcq}) has shown date-base ($DB$) and min-max ($MM$) metric more yielding.

\subsection{Continual Learning}
\label{app:continual}
We present the following figures for different open-source models, showing the yearwise performance when finetuned in \textbf{Continual Learning} paradigm for the ``Correct'', ``Incorrect'', and ``Information Not Available'' labels as indexed in Table~\ref{table:index_continual}. 

\subsection{Yearwise Finetuning}
\label{app:yearwise}
We present the following figures for different open-source models, showing the yearwise performance when finetuned in \textbf{Yearwise Finetuning} paradigm as indexed in Table \ref{table:index_yearwise}.

\subsection{Random Finetuning}
\label{app:random}
We present the following figures for different open-source models, showing the yearwise performance when finetuned in \textbf{Random Finetuning} paradigm as indexed in Table\ref{table:index_ranom}. 

\begin{table*}[]
\begin{tabular}{p{4cm}rrrrrr}
\hline
\textbf{Categories} & $DB$ & $CP$ & $WB$ & $MM$ & $RB$ & $TB$ \\ \hline
C$1$: Climate & 1,778,631 & 672,993 & 603,882 & 603,882 & 603,882 & 603,876 \\
C$2$: Food and Agriculture & 991,443 & 236,665 & 213,328 & 213,328 & 213,328 & 213,328 \\
C$3$: Health & 5,684,312 & 1,891,152 & 1,739,273 & 1,739,273 & 1,739,273 & 1,739,273 \\
C$4$: Human Rights & 7,142 & 5,939 & 1,328 & 1,328 & 1,328 & 1,328 \\
C$5$: Innovation & 1,537 & 1,247 & 384 & 384 & 384 & 384 \\
C$6$: Migration & 100,346 & 100,023 & 24,116 & 24,116 & 24,116 & 24,116 \\
C$7$: Economic Development & 909,519 & 402,217 & 305,373 & 305,373 & 305,373 & 305,373 \\
C$8$: War & 24,572 & 15,347 & 11,215 & 11,215 & 11,215 & 11,215 \\ \hline
\textbf{Total} & \textbf{9,497,502} & \textbf{3,325,583} & \textbf{2,898,899} & \textbf{2,898,899} & \textbf{2,898,899} & \textbf{2,898,893} \\ \hline
\end{tabular}
\caption{\label{tab:promptsforboth}The number of samples for each category in the \textbf{TempUN} dataset.}
\end{table*}

\begin{table*}[]
\centering
\begin{tabular}{p{7cm}rrrrrr} \hline
\textbf{Categories} & $DB$ & $CP$ & $WB$ & $MM$ & $RB$ & $TB$ \\ \hline
C$1$: Climate & 17,928   & 2,440   & 2,440   & 732   & 732   & 732   \\
C$2$: Food and Agriculture                             & 11,133   & 2,617   & 2,495   & 769   & 769   & 769   \\
C$3$: Health                                           & 18,599   & 2,579   & 2,570   & 771   & 771   & 771   \\
C$4$: Human Rights                                     & 5,373    & 1,823   & 1,778   & 559   & 559   & 559   \\
C$5$: Innovation              & 5,813    & 2,198   & 2,176   & 657   & 657   & 657   \\
C$6$: Migration                                        & 17,232   & 2,550   & 2,550   & 765   & 765   & 765   \\
C$7$: Economic Development     & 18,716   & 2,500   & 2,500   & 750   & 750   & 750   \\
C$8$: War                                              & 9,336    & 1,801   & 1,759   & 531   & 531   & 531   \\ \hline
\textbf{Total}                                            & \textbf{104,130} & \textbf{18,508} & \textbf{18,268} & \textbf{5,534} & \textbf{5,534} & \textbf{5,534} \\ \hline
\end{tabular}
\caption{\label{tab:micro_prompts_tempuns}The number of prompts for each category for different metrics in the \textbf{TempUN$_{s}$} dataset.}
\end{table*}


\begin{table*}
\centering
\begin{tabular}{ll}
\hline
\textbf{Categories}                                       &\textbf{ \# of instances} \\ \hline
Access To Energy                 & 2,088                \\
Air Pollution                    & 6,424                \\
Biodiversity                     & 12,108               \\
Clean Water and Sanitization     & 34,214               \\
Climate Change                   & 251                  \\
CO2 and Greenhouse Gas Emissions & 30,785               \\
Energy                           & 11,483               \\
Forests and Deforestation        & 5,030                \\
Fossil Fuels                     & 2,532                \\
Indoor Air Pollution             & 2,712                \\
Lead Pollution                   & 1,114                \\
Natural Disasters                & 5,210                \\
Nuclear Energy                   & 730                  \\
Oil Spills                       & 30                   \\
Ozone Layer                      & 763                  \\
Pesticides                       & 1,435                \\
Plastic Pollution                & 3,008                \\
Pollution                        & 5,827                \\
Water Use and Stress             & 2,136                                  \\ \hline
\textbf{Total} & \textbf{127,880} \\
\hline
\end{tabular}
\caption{\label{table:issues_list1}
List of 19 sub-categories in the \textbf{Climate} category.}
\end{table*}
\begin{table*}
\centering
\begin{tabular}{ll}
\hline
\textbf{Categories}                                       &\textbf{ \# of instances} \\ \hline
Agricultural Production                                         & 9,087  \\
Animal Welfare                                                  & 3,313  \\
Crop Yields                                                     & 9,199  \\
Environmental Impacts of Food Production                        & 1,638  \\
Environmental Impacts of Food Production (Food and Agriculture) & 1,276  \\
Famines                                                         & 943    \\
Fertilizers                                                     & 4,091  \\
Food Prices                                                     & 2,762  \\
Land Use                                                        & 12,134 \\
Meat and Dairy Production                                       & 12,628   \\ \hline
\textbf{Total} & \textbf{57,071} \\
\hline
\end{tabular}
\caption{\label{table:issues_list2}
List of 10 sub-categories in the \textbf{Food and Agriculture} category.}
\end{table*}

\begin{table*}
\centering
\begin{tabular}{ll}
\hline
\textbf{Categories}                                       &\textbf{ \# of instances} \\ \hline
Alcohol Consumption         & 9,494  \\
Burden of Disease           & 6,956  \\
Cardiovascular Diseases     & 9,623  \\
Causes of Death             & 52,430 \\
Child and Infant Mortality  & 22,176 \\
Covid                       & 10,744 \\
Diarrheal Diseases          & 14,312 \\
Diet Compositions           & 12,825 \\
Eradication of Diseases     & 7,133  \\
Fertility Rate              & 2,330  \\
Global Health               & 3,276  \\
Happiness and Satisfaction  & 396    \\
Healthcare Spending         & 3,568  \\
HIV                         & 21,602 \\
Human Height                & 2,855  \\
Hunger and Undernourishment & 4,694  \\
Influenza                   & 11,013 \\
Life Expectancy             & 18,227 \\
Malaria                     & 4,886  \\
Maternal Mortality          & 3,936  \\
Mental Health               & 8,490  \\
Micronutrient Deficiency    & 2,659  \\
Monkeypox                   & 1,353  \\
Obesity                     & 7,806  \\
Opioids                     & 11,468 \\
Pandemics                   & 4,224  \\
Pneumonia                   & 2,400  \\
Polio                       & 2,021  \\
Sanitation                  & 4,647  \\
Smallpox                    & 273    \\
Smoking                     & 5,429  \\
Suicides                    & 6,374  \\
Tetanus                     & 2,391  \\
Vaccination                 & 10,195   \\ \hline
\textbf{Total} & \textbf{292,206} \\
\hline
\end{tabular}
\caption{\label{table:issues_list3}
List of 34 sub-categories in the \textbf{Health} category.}
\end{table*}

\begin{table*}
\centering
\begin{tabular}{ll}
\hline
\textbf{Categories}                                       &\textbf{ \# of instances} \\ \hline
Child Labor                      & 605    \\
Human Rights                     & 10,566 \\
LGBT                             & 647    \\
Literacy                         & 1,375  \\
Loneliness and social connections & 478    \\
Marriages and divorces             & 901    \\
Trust                            & 598    \\
Violence against Children        & 1,530  \\ \hline
\textbf{Total} & \textbf{16,700} \\
\hline
\end{tabular}
\caption{\label{table:issues_list4}
List of 8 sub-categories in the \textbf{Human Rights} category.}
\end{table*}
\begin{table*}
\centering
\begin{tabular}{ll}
\hline
\textbf{Categories}                                       &\textbf{ \# of instances} \\ \hline
Artificial-Intelligence  & 3,883 \\
Internet                 & 1,991 \\
Research-And-Development & 2,973 \\
Technology change        & 792  \\ \hline
\textbf{Total} & \textbf{9,639} \\
\hline
\end{tabular}
\caption{\label{table:issues_list5}
List of  4 sub-categories in the \textbf{Innovation and Technological Change} category.}
\end{table*}

\begin{table*}
\centering
\begin{tabular}{ll}
\hline
\textbf{Categories}                                       &\textbf{ \# of instances} \\ \hline
International Migration and Refugees & 36,226  \\ \hline
\textbf{Total} & \textbf{36,226} \\
\hline
\end{tabular}
\caption{\label{table:issues_list6}
List of one major sub-category in the \textbf{Migration} category.}
\end{table*}

\begin{table*}
\centering
\begin{tabular}{ll}
\hline
\textbf{Categories}                                       &\textbf{ \# of instances} \\ \hline
Age                       & 3,048  \\
Books                     & 109    \\
Corruption                & 2,228  \\
Economic-Inequality       & 7,592  \\
Education-Spending        & 845    \\
Employment-In-Agriculture & 3,382  \\
Gender Ratio              & 4,783  \\
Global-Education          & 15,933 \\
Government-Spending       & 1,393  \\
Homelessness              & 18     \\
Human Development Index   & 2,624  \\
Light at Night            & 12     \\
Poverty                   & 8,969  \\
Renewable Energy          & 3,322  \\
State-Capacity            & 4,298  \\
Taxation                  & 1,350  \\
Time use                  & 167    \\
Tourism                   & 3,058  \\
Trade and globalization   & 7,073  \\
Transportation             & 717    \\
Urbanization              & 4,804  \\
Women Employment          & 3,304  \\
Women Rights              & 5,573  \\
Working Hours             & 260    \\
GDP                       & 1      \\ \hline
\textbf{Total} & \textbf{84,863} \\
\hline
\end{tabular}
\caption{\label{table:issues_list7}
List of 25 sub-categories in the \textbf{Poverty, Economic Development, and Community} category.}
\end{table*}

\begin{table*}
\centering
\begin{tabular}{ll}
\hline
\textbf{Categories}                                       &\textbf{ \# of instances} \\ \hline
Homicide          & 16,959 \\
Military spending & 1,689  \\
Nuclear-Weapons   & 75     \\
Terrorism         & 9,823  \\
War and Peace     & 30      \\ \hline
\textbf{Total} & \textbf{28,576} \\
\hline
\end{tabular}
\caption{\label{table:issues_list8}
List of 5 sub-categories in the \textbf{Peace and War} category.}
\end{table*}

    \begin{table*}[]
    \centering
    \begin{tabular}{p{4.5cm}ccc}
    \hline
    \textbf{Category} & \multicolumn{1}{l}{\textbf{Correctly Answered}} & \multicolumn{1}{l}{\textbf{Incorrectly Answered}} & \multicolumn{1}{l}{\textbf{Not Answered}} \\ \hline 
	C$1$: Climate & 7.25 & 92.75 & 0.0 \\ 
	C$2$: Food and Agriculture & 12.47 & 87.51 & 0.02 \\ 
	C$3$: Health & 9.9 & 90.1 & 0.0 \\ 
	C$4$: Human Rights & 13.68 & 86.32 & 0.0 \\ 
	C$5$: Innovation & 4.8 & 95.2 & 0.0 \\ 
	C$6$: Migration & 15.54 & 84.46 & 0.0 \\ 
	C$7$: Economic Development & 11.27 & 88.73 & 0.0 \\ 
C$8$: Peace and War & 10.96 & 89.04 & 0 \\ \hline
    \textbf{Total} & {\color[HTML]{000000} \textbf{10.9}} & {\color[HTML]{FF0000} \textbf{89.1}} & {\color[HTML]{000000} \textbf{0.0}} \\ \hline                                          
    \end{tabular}
    \caption{Above table provides the percentage of the answers that are correctly, incorrectly, and not answered by \texttt{Phi-2} on \textit{DB}}
    \label{tab:result_Phi2_DB}
    \end{table*}

    \begin{table*}[]
    \centering
    \begin{tabular}{p{4.5cm}ccc}
    \hline
    \textbf{Category} & \multicolumn{1}{l}{\textbf{Correctly Answered}} & \multicolumn{1}{l}{\textbf{Incorrectly Answered}} & \multicolumn{1}{l}{\textbf{Not Answered}} \\ \hline 
	C$1$: Climate & 0.0 & 100.0 & 0.0 \\ 
	C$2$: Food and Agriculture & 0.0 & 99.92 & 0.08 \\ 
	C$3$: Health & 0.0 & 100.0 & 0.0 \\ 
	C$4$: Human Rights & 0.0 & 97.59 & 2.41 \\ 
	C$5$: Innovation & 0.0 & 100.0 & 0.0 \\ 
	C$6$: Migration & 0.0 & 100.0 & 0.0 \\ 
	C$7$: Economic Development & 0.0 & 85.88 & 14.12 \\ 
C$8$: Peace and War & 0 & 91.23 & 8.77 \\ \hline
    \textbf{Total} & {\color[HTML]{000000} \textbf{0.0}} & {\color[HTML]{FF0000} \textbf{96.99}} & {\color[HTML]{000000} \textbf{3.01}} \\ \hline                                          
    \end{tabular}
    \caption{Above table provides the percentage of the answers that are correctly, incorrectly, and not answered by \texttt{Phi-2} on \textit{CP}.}
    \label{tab:result_Phi2_RB_}
    \end{table*}

    \begin{table*}[]
    \centering
    \begin{tabular}{p{4.5cm}ccc}
    \hline
    \textbf{Category} & \multicolumn{1}{l}{\textbf{Correctly Answered}} & \multicolumn{1}{l}{\textbf{Incorrectly Answered}} & \multicolumn{1}{l}{\textbf{Not Answered}} \\ \hline 
	C$1$: Climate & 19.39 & 80.61 & 0.0 \\ 
	C$2$: Food and Agriculture & 19.32 & 80.68 & 0.0 \\ 
	C$3$: Health & 12.53 & 87.47 & 0.0 \\ 
	C$4$: Human Rights & 22.83 & 77.17 & 0.0 \\ 
	C$5$: Innovation & 17.92 & 82.08 & 0.0 \\ 
	C$6$: Migration & 18.27 & 81.73 & 0.0 \\ 
	C$7$: Economic Development & 14.56 & 85.44 & 0.0 \\ 
C$8$: Peace and War & 23.88 & 76.12 & 0 \\ \hline
    \textbf{Total} & {\color[HTML]{000000} \textbf{18.19}} & {\color[HTML]{FF0000} \textbf{81.81}} & {\color[HTML]{000000} \textbf{0.0}} \\ \hline                                          
    \end{tabular}
    \caption{Above table provides the percentage of the answers that are correctly, incorrectly, and not answered by \texttt{Phi-2} on \textit{WB}.}
    \label{tab:result_Phi2_WB_}
    \end{table*}

    \begin{table*}[]
    \centering
    \begin{tabular}{p{4.5cm}ccc}
    \hline
    \textbf{Category} & \multicolumn{1}{l}{\textbf{Correctly Answered}} & \multicolumn{1}{l}{\textbf{Incorrectly Answered}} & \multicolumn{1}{l}{\textbf{Not Answered}} \\ \hline 
	C$1$: Climate & 4.14 & 95.86 & 0.0 \\ 
	C$2$: Food and Agriculture & 6.09 & 93.91 & 0.0 \\ 
	C$3$: Health & 2.3 & 97.7 & 0.0 \\ 
	C$4$: Human Rights & 13.39 & 86.61 & 0.0 \\ 
	C$5$: Innovation & 8.41 & 91.59 & 0.0 \\ 
	C$6$: Migration & 12.43 & 87.57 & 0.0 \\ 
	C$7$: Economic Development & 5.44 & 93.72 & 0.84 \\ 
C$8$: Peace and War & 12.39 & 87.21 & 0.4 \\ \hline
    \textbf{Total} & {\color[HTML]{000000} \textbf{7.69}} & {\color[HTML]{FF0000} \textbf{92.16}} & {\color[HTML]{000000} \textbf{0.15}} \\ \hline                                          
    \end{tabular}
    \caption{Above table provides the percentage of the answers that are correctly, incorrectly, and not answered by \texttt{Phi-2} on \textit{MM}.}
    \label{tab:result_Phi2_MMB_}
    \end{table*}

    \begin{table*}[]
    \centering
    \begin{tabular}{p{4.5cm}ccc}
    \hline
    \textbf{Category} & \multicolumn{1}{l}{\textbf{Correctly Answered}} & \multicolumn{1}{l}{\textbf{Incorrectly Answered}} & \multicolumn{1}{l}{\textbf{Not Answered}} \\ \hline 
	C$1$: Climate & 10.7 & 87.42 & 1.89 \\ 
	C$2$: Food and Agriculture & 8.5 & 91.14 & 0.36 \\ 
	C$3$: Health & 1.28 & 98.72 & 0.0 \\ 
	C$4$: Human Rights & 17.89 & 82.11 & 0.0 \\ 
	C$5$: Innovation & 6.53 & 93.47 & 0.0 \\ 
	C$6$: Migration & 11.88 & 88.12 & 0.0 \\ 
	C$7$: Economic Development & 8.08 & 83.56 & 8.36 \\ 
C$8$: Peace and War & 13.42 & 79.65 & 6.94 \\ \hline
    \textbf{Total} & {\color[HTML]{000000} \textbf{9.34}} & {\color[HTML]{FF0000} \textbf{88.54}} & {\color[HTML]{000000} \textbf{2.11}} \\ \hline                                          
    \end{tabular}
    \caption{Above table provides the percentage of the answers that are correctly, incorrectly, and not answered by \texttt{Phi-2} on \textit{RB}.}
    \label{tab:result_Phi2_RAB_}
    \end{table*}

    \begin{table*}[]
    \centering
    \begin{tabular}{p{4.5cm}ccc}
    \hline
    \textbf{Category} & \multicolumn{1}{l}{\textbf{Correctly Answered}} & \multicolumn{1}{l}{\textbf{Incorrectly Answered}} & \multicolumn{1}{l}{\textbf{Not Answered}} \\ \hline 
	C$1$: Climate & 1.11 & 98.89 & 0.0 \\ 
	C$2$: Food and Agriculture & 2.89 & 97.11 & 0.0 \\ 
	C$3$: Health & 11.21 & 88.79 & 0.0 \\ 
	C$4$: Human Rights & 2.02 & 96.01 & 1.97 \\ 
	C$5$: Innovation & 6.07 & 92.97 & 0.97 \\ 
	C$6$: Migration & 11.96 & 88.04 & 0.0 \\ 
	C$7$: Economic Development & 5.24 & 90.28 & 4.48 \\ 
C$8$: Peace and War & 3.81 & 93.63 & 2.56 \\ \hline
    \textbf{Total} & {\color[HTML]{000000} \textbf{5.79}} & {\color[HTML]{FF0000} \textbf{93.04}} & {\color[HTML]{000000} \textbf{1.17}} \\ \hline                                          
    \end{tabular}
    \caption{Above table provides the percentage of the answers that are correctly, incorrectly, and not answered by \texttt{Phi-2} on \textit{TB}.}
    \label{tab:result_Phi2_TB_}
    \end{table*}


    \begin{table*}[]
    \centering
    \begin{tabular}{p{4.5cm}ccc}
\hline
\textbf{} & \multicolumn{1}{l}{\textbf{Correctly Answered}} & \multicolumn{1}{l}{\textbf{Incorrectly Answered}} & \multicolumn{1}{l}{\textbf{Not Answered}} \\ \hline
C$1$: Climate              & 28.51          & 71.49          & 0          \\
C$2$: Food and Agriculture & 49.65          & 50.35          & 0          \\
C$3$: Health               & 44.7           & 55.3           & 0          \\
C$4$: Human Rights         & 44.02          & 55.98          & 0          \\
C$5$: Innovation           & 24.81          & 75.19          & 0          \\
C$6$: Migration            & 37.34          & 62.66          & 0          \\
C$7$: Economic Development & 43.26          & 56.74          & 0          \\
C$8$: Peace and War        & 27.88          & 72.12          & 0          \\ \hline
\textbf{Total}             & \textbf{38.31} & {\color[HTML]{FF0000} \textbf{\textbf{61.69}}} & \textbf{0} \\ \hline
\end{tabular}
    \caption{Above table provides the percentage of the answers that are correctly, incorrectly, and not answered by \texttt{flan-t5-xl} on \textit{DB}}
    \label{tab:result_flan-t5_DB}
    \end{table*}

    \begin{table*}[]
    \centering
    \begin{tabular}{p{4.5cm}ccc}
\hline
\textbf{} & \multicolumn{1}{l}{\textbf{Correctly Answered}} & \multicolumn{1}{l}{\textbf{Incorrectly Answered}} & \multicolumn{1}{l}{\textbf{Not Answered}} \\ \hline
C$1$: Climate              & 47.38          & 52.62          & 0          \\
C$2$: Food and Agriculture & 52.12          & 47.88          & 0          \\
C$3$: Health               & 12.6           & 87.4           & 0          \\
C$4$: Human Rights         & 40.37          & 59.63          & 0          \\
C$5$: Innovation           & 28.89          & 71.11          & 0          \\
C$6$: Migration            & 65.65          & 34.35          & 0          \\
C$7$: Economic Development & 13.6           & 86.4           & 0          \\
C$8$: Peace and War        & 67.68          & 32.32          & 0          \\ \hline
\textbf{Total}             & \textbf{40.25} & {\color[HTML]{FF0000} \textbf{\textbf{59.75}}} & \textbf{0} \\ \hline
\end{tabular}
    \caption{Above table provides the percentage of the answers that are correctly, incorrectly, and not answered by \texttt{flan-t5-xl} on \textit{CP}.}
    \label{tab:result_flan-t5_RB_}
    \end{table*}

    \begin{table*}[]
    \centering
    \begin{tabular}{p{4.5cm}ccc}
\hline
\textbf{} & \multicolumn{1}{l}{\textbf{Correctly Answered}} & \multicolumn{1}{l}{\textbf{Incorrectly Answered}} & \multicolumn{1}{l}{\textbf{Not Answered}} \\ \hline
C$1$: Climate              & 31.8           & 68.2           & 0          \\
C$2$: Food and Agriculture & 13.95          & 86.05          & 0          \\
C$3$: Health               & 21.79          & 78.21          & 0          \\
C$4$: Human Rights         & 17.49          & 82.51          & 0          \\
C$5$: Innovation           & 14.02          & 85.98          & 0          \\
C$6$: Migration            & 21.06          & 78.94          & 0          \\
C$7$: Economic Development & 21.32          & 78.68          & 0          \\
C$8$: Peace and War        & 29.9           & 70.1           & 0          \\ \hline
\textbf{Total}             & \textbf{21.33} & {\color[HTML]{FF0000} \textbf{\textbf{78.67}}} & \textbf{0} \\ \hline
\end{tabular}
    \caption{Above table provides the percentage of the answers that are correctly, incorrectly, and not answered by \texttt{flan-t5-xl} on \textit{WB}.}
    \label{tab:result_flan-t5_WB_}
    \end{table*}

    \begin{table*}[]
    \centering
    \begin{tabular}{p{4.5cm}ccc}
\hline
\textbf{} & \multicolumn{1}{l}{\textbf{Correctly Answered}} & \multicolumn{1}{l}{\textbf{Incorrectly Answered}} & \multicolumn{1}{l}{\textbf{Not Answered}} \\ \hline
C$1$: Climate              & 24.63          & 75.16         & 0.2           \\
C$2$: Food and Agriculture & 31.82          & 67.74         & 0.44          \\
C$3$: Health               & 20.47          & 77            & 2.53          \\
C$4$: Human Rights         & 25.7           & 74.3          & 0             \\
C$5$: Innovation           & 26.42          & 73.58         & 0             \\
C$6$: Migration            & 22.35          & 77.65         & 0             \\
C$7$: Economic Development & 26.68          & 73.32         & 0             \\
C$8$: Peace and War        & 11.65          & 88.35         & 0             \\ \hline
\textbf{Total}             & \textbf{24.06} & {\color[HTML]{FF0000} \textbf{\textbf{75.5}}} & \textbf{0.44} \\ \hline
\end{tabular}
    \caption{Above table provides the percentage of the answers that are correctly, incorrectly, and not answered by \texttt{flan-t5-xl} on \textit{MM}.}
    \label{tab:result_flan-t5_MMB_}
    \end{table*}

    \begin{table*}[]
    \centering
    \begin{tabular}{p{4.5cm}ccc}
 \hline
\textbf{} & \multicolumn{1}{l}{\textbf{Correctly Answered}} & \multicolumn{1}{l}{\textbf{Incorrectly Answered}} & \multicolumn{1}{l}{\textbf{Not Answered}} \\ \hline
C$1$: Climate              & 15.53          & 84.06          & 0.41          \\
C$2$: Food and Agriculture & 32.75          & 66.85          & 0.4           \\
C$3$: Health               & 31.28          & 64.12          & 4.59          \\
C$4$: Human Rights         & 8.61           & 91.17          & 0.22          \\
C$5$: Innovation           & 13.65          & 86.35          & 0             \\
C$6$: Migration            & 9.57           & 89.84          & 0.59          \\
C$7$: Economic Development & 33.24          & 65.96          & 0.8           \\
C$8$: Peace and War        & 4.95           & 95.05          & 0             \\ \hline
\textbf{Total}             & \textbf{19.77} & {\color[HTML]{FF0000} \textbf{\textbf{79.26}}} & \textbf{0.97} \\ \hline
\end{tabular}
    \caption{Above table provides the percentage of the answers that are correctly, incorrectly, and not answered by \texttt{flan-t5-xl} on \textit{RB}.}
    \label{tab:result_flan-t5_RAB_}
    \end{table*}

    \begin{table*}[]
    \centering
    \begin{tabular}{p{4.5cm}ccc}
 \hline
\textbf{} & \multicolumn{1}{l}{\textbf{Correctly Answered}} & \multicolumn{1}{l}{\textbf{Incorrectly Answered}} & \multicolumn{1}{l}{\textbf{Not Answered}} \\ \hline
C$1$: Climate              & 1.52          & 98.48          & 0          \\
C$2$: Food and Agriculture & 4.29          & 95.71          & 0          \\
C$3$: Health               & 1.75          & 98.25          & 0          \\
C$4$: Human Rights         & 0.22          & 99.78          & 0          \\
C$5$: Innovation           & 1.24          & 98.76          & 0          \\
C$6$: Migration            & 0.63          & 99.37          & 0          \\
C$7$: Economic Development & 8.52          & 91.48          & 0          \\
C$8$: Peace and War        & 3.47          & 96.53          & 0          \\ \hline
\textbf{Total}             & \textbf{2.79} & {\color[HTML]{FF0000} \textbf{\textbf{97.21}}} & \textbf{0} \\ \hline
\end{tabular}
    \caption{Above table provides the percentage of the answers that are correctly, incorrectly, and not answered by \texttt{flan-t5-xl} on \textit{TB}.}
    \label{tab:result_flan-t5_TB_}
    \end{table*}


\begin{table*}[]
\centering
\begin{tabular}{p{4.5cm}ccc}
\hline
\textbf{Category} & \multicolumn{1}{l}{\textbf{Correctly Answered}} & \multicolumn{1}{l}{\textbf{Incorrectly Answered}} & \multicolumn{1}{l}{\textbf{Not Answered}} \\ \hline 
C$1$: Climate & 40.7 & 17.56 & 41.74 \\ 
C$2$: Food and Agriculture & 23.25 & 71.5 & 5.25 \\ 
C$3$: Health & 41.39 & 58.35 & 0.26 \\ 
C$4$: Human Rights & 28.85 & 45.26 & 25.89 \\ 
C$5$: Innovation & 30.45 & 68.6 & 0.95 \\ 
C$6$: Migration & 36.83 & 59.99 & 3.18 \\ 
C$7$: Economic Development & 45.86 & 53.99 & 0.15 \\ 
C$8$: Peace and War & 31.22 & 48.37 & 20.4 \\ 
\hline
\textbf{Total} & {\color[HTML]{000000} \textbf{37.21}} & {\color[HTML]{FF0000} \textbf{51.22}} & {\color[HTML]{000000} \textbf{11.57}} \\ \hline                                          
\end{tabular}
\caption{Above table provides the percentage of the answers that are correctly, incorrectly, and not answered by \texttt{Mistral-7B} on \textit{DB}}
\label{tab:result_Mistral_DB}
\end{table*}

 \begin{table*}[]
    \centering
    \begin{tabular}{p{4.5cm}ccc}
    \hline
    \textbf{Category} & \multicolumn{1}{l}{\textbf{Correctly Answered}} & \multicolumn{1}{l}{\textbf{Incorrectly Answered}} & \multicolumn{1}{l}{\textbf{Not Answered}} \\ \hline 
	C$1$: Climate & 47.25 & 52.75 & 0.0 \\ 
	C$2$: Food and Agriculture & 52.24 & 47.76 & 0.0 \\ 
	C$3$: Health & 15.82 & 84.18 & 0.0 \\ 
	C$4$: Human Rights & 61.0 & 39.0 & 0.0 \\ 
	C$5$: Innovation & 28.53 & 71.47 & 0.0 \\ 
	C$6$: Migration & 65.49 & 34.51 & 0.0 \\ 
	C$7$: Economic Development & 13.76 & 86.24 & 0.0 \\ 
C$8$: Peace and War & 70.13 & 29.87 & 0 \\ \hline
    \textbf{Total} & {\color[HTML]{000000} \textbf{42.92}} & {\color[HTML]{FF0000} \textbf{57.08}} & {\color[HTML]{000000} \textbf{0.0}} \\ \hline                                          
    \end{tabular}
    \caption{Above table provides the percentage of the answers that are correctly, incorrectly, and not answered by \texttt{Mistral-7B} on \textit{CP}}
    \label{tab:result_Mistral_RB}
    \end{table*}

 \begin{table*}[]
    \centering
    \begin{tabular}{p{4.5cm}ccc}
    \hline
    \textbf{Category} & \multicolumn{1}{l}{\textbf{Correctly Answered}} & \multicolumn{1}{l}{\textbf{Incorrectly Answered}} & \multicolumn{1}{l}{\textbf{Not Answered}} \\ \hline 
	C$1$: Climate & 6.84 & 93.16 & 0.0 \\ 
	C$2$: Food and Agriculture & 10.74 & 89.26 & 0.0 \\ 
	C$3$: Health & 25.49 & 74.51 & 0.0 \\ 
	C$4$: Human Rights & 18.11 & 81.89 & 0.0 \\ 
	C$5$: Innovation & 19.72 & 80.28 & 0.0 \\ 
	C$6$: Migration & 36.35 & 63.65 & 0.0 \\ 
	C$7$: Economic Development & 22.16 & 77.84 & 0.0 \\ 
 
C$8$: Peace and War & 16.37 & 83.63 & 0 \\ \hline
    \textbf{Total} & {\color[HTML]{000000} \textbf{19.76}} & {\color[HTML]{FF0000} \textbf{80.24}} & {\color[HTML]{000000} \textbf{0.0}} \\ \hline                                          
    \end{tabular}
    \caption{Above table provides the percentage of the answers that are correctly, incorrectly, and not answered by \texttt{Mistral-7B} on \textit{WB}}
    \label{tab:result_Mistral_WB}
    \end{table*}

\begin{table*}[]
\centering
\begin{tabular}{p{4.5cm}ccc}
\hline
\textbf{Category} & \multicolumn{1}{l}{\textbf{Correctly Answered}} & \multicolumn{1}{l}{\textbf{Incorrectly Answered}} & \multicolumn{1}{l}{\textbf{Not Answered}} \\ \hline 
C$1$: Climate & 19.39 & 9.26 & 71.35 \\ 
C$2$: Food and Agriculture & 13.03 & 85.09 & 1.88 \\ 
C$3$: Health & 17.98 & 81.67 & 0.35 \\ 
C$4$: Human Rights & 20.58 & 52.98 & 26.43 \\ 
C$5$: Innovation & 31.43 & 68.57 & 0.0 \\ 
C$6$: Migration & 28.9 & 69.02 & 2.08 \\ 
C$7$: Economic Development & 26.2 & 70.48 & 3.32 \\ 
C$8$: Peace and War & 29.62 & 68.33 & 2.05 \\ \hline
\textbf{Total} & {\color[HTML]{000000} \textbf{23.12}} & {\color[HTML]{FF0000} \textbf{63.53}} & {\color[HTML]{000000} \textbf{13.35}} \\ \hline                                          
\end{tabular}
\caption{Above table provides the percentage of the answers that are correctly, incorrectly, and not answered by \texttt{Mistral-7B} on \textit{MM}}
\label{tab:result_Mistral_MMB}
\end{table*}

\begin{table*}[]
\centering
\begin{tabular}{p{4.5cm}ccc}
\hline
\textbf{Category} & \multicolumn{1}{l}{\textbf{Correctly Answered}} & \multicolumn{1}{l}{\textbf{Incorrectly Answered}} & \multicolumn{1}{l}{\textbf{Not Answered}} \\ \hline 
C$1$: Climate & 23.57 & 76.43 & 0.0 \\ 
C$2$: Food and Agriculture & 30.42 & 69.58 & 0.0 \\ 
C$3$: Health & 37.12 & 62.88 & 0.0 \\ 
C$4$: Human Rights & 19.01 & 80.99 & 0.0 \\ 
C$5$: Innovation & 36.12 & 63.88 & 0.0 \\ 
C$6$: Migration & 34.27 & 65.73 & 0.0 \\ 
C$7$: Economic Development & 43.24 & 56.76 & 0.0 \\ 
C$8$: Peace and War & 44.06 & 55.94 & 0 \\ \hline
\textbf{Total} & {\color[HTML]{000000} \textbf{33.62}} & {\color[HTML]{FF0000} \textbf{66.38}} & {\color[HTML]{000000} \textbf{0.0}} \\ \hline                                          
\end{tabular}
\caption{Above table provides the percentage of the answers that are correctly, incorrectly, and not answered by \texttt{Mistral-7B} on \textit{RB}}
\label{tab:result_Mistral_RAN}
\end{table*}

 \begin{table*}[]
    \centering
    \begin{tabular}{p{4.5cm}ccc}
    \hline
    \textbf{Category} & \multicolumn{1}{l}{\textbf{Correctly Answered}} & \multicolumn{1}{l}{\textbf{Incorrectly Answered}} & \multicolumn{1}{l}{\textbf{Not Answered}} \\ \hline 
	C$1$: Climate & 22.83 & 15.61 & 61.56 \\ 
	C$2$: Food and Agriculture & 9.54 & 84.45 & 6.01 \\ 
	C$3$: Health & 1.91 & 95.06 & 3.04 \\ 
	C$4$: Human Rights & 7.54 & 82.56 & 9.9 \\ 
	C$5$: Innovation & 8.13 & 89.2 & 2.67 \\ 
	C$6$: Migration & 5.1 & 64.08 & 30.82 \\ 
	C$7$: Economic Development & 2.76 & 77.92 & 19.32 \\ 
C$8$: Peace and War & 3.35 & 56.57 & 40.08 \\ \hline
    \textbf{Total} & {\color[HTML]{000000} \textbf{7.73}} & {\color[HTML]{FF0000} \textbf{70.71}} & {\color[HTML]{000000} \textbf{21.56}} \\ \hline                                          
    \end{tabular}
    \caption{Above table provides the percentage of the answers that are correctly, incorrectly, and not answered by \texttt{Mistral} on \textit{TB}}
    \label{tab:result_Mistral_TB}
    \end{table*}

    \begin{table*}[]
    \centering
    \begin{tabular}{p{4.5cm}ccc}
    \hline
    \textbf{Category} & \multicolumn{1}{l}{\textbf{Correctly Answered}} & \multicolumn{1}{l}{\textbf{Incorrectly Answered}} & \multicolumn{1}{l}{\textbf{Not Answered}} \\ \hline 
	C$1$: Climate & 27.04 & 57.96 & 15.0 \\ 
	C$2$: Food and Agriculture & 14.9 & 85.02 & 0.08 \\ 
	C$3$: Health & 18.97 & 80.93 & 0.1 \\ 
	C$4$: Human Rights & 25.18 & 74.61 & 0.2 \\ 
	C$5$: Innovation & 19.73 & 78.87 & 1.39 \\ 
	C$6$: Migration & 22.21 & 76.83 & 0.95 \\ 
	C$7$: Economic Development & 19.23 & 80.75 & 0.01 \\ 
C$8$: Peace and War & 18.82 & 79.88 & 1.3 \\ \hline
    \textbf{Total} & {\color[HTML]{000000} \textbf{20.86}} & {\color[HTML]{FF0000} \textbf{76.17}} & {\color[HTML]{000000} \textbf{2.97}} \\ \hline                                          
    \end{tabular}
    \caption{Above table provides the percentage of the answers that are correctly, incorrectly, and not answered by \texttt{llama-2-chat} on \textit{DB}.}
    \label{tab:result_LLaMA_DB_}
    \end{table*}

    \begin{table*}[]
    \centering
    \begin{tabular}{p{4.5cm}ccc}
    \hline
    \textbf{Category} & \multicolumn{1}{l}{\textbf{Correctly Answered}} & \multicolumn{1}{l}{\textbf{Incorrectly Answered}} & \multicolumn{1}{l}{\textbf{Not Answered}} \\ \hline 
	C$1$: Climate & 47.09 & 52.91 & 0.0 \\ 
	C$2$: Food and Agriculture & 52.54 & 47.46 & 0.0 \\ 
	C$3$: Health & 12.76 & 87.24 & 0.0 \\ 
	C$4$: Human Rights & 40.43 & 59.57 & 0.0 \\ 
	C$5$: Innovation & 68.06 & 31.94 & 0.0 \\ 
	C$6$: Migration & 63.69 & 36.31 & 0.0 \\ 
	C$7$: Economic Development & 13.72 & 86.28 & 0.0 \\ 
C$8$: Peace and War & 69.52 & 30.48 & 0 \\ \hline
    \textbf{Total} & {\color[HTML]{000000} \textbf{44.87}} & {\color[HTML]{FF0000} \textbf{55.13}} & {\color[HTML]{000000} \textbf{0.0}} \\ \hline                                          
    \end{tabular}
    \caption{Above table provides the percentage of the answers that are correctly, incorrectly, and not answered by \texttt{llama-2-chat} on \textit{CP}.}
    \label{tab:result_LLaMA_RB_}
    \end{table*}

    \begin{table*}[]
    \centering
    \begin{tabular}{p{4.5cm}ccc}
    \hline
    \textbf{Category} & \multicolumn{1}{l}{\textbf{Correctly Answered}} & \multicolumn{1}{l}{\textbf{Incorrectly Answered}} & \multicolumn{1}{l}{\textbf{Not Answered}} \\ \hline 
	C$1$: Climate & 23.81 & 76.19 & 0.0 \\ 
	C$2$: Food and Agriculture & 16.31 & 83.69 & 0.0 \\ 
	C$3$: Health & 24.2 & 75.8 & 0.0 \\ 
	C$4$: Human Rights & 23.45 & 76.55 & 0.0 \\ 
	C$5$: Innovation & 24.17 & 75.83 & 0.0 \\ 
	C$6$: Migration & 15.65 & 84.35 & 0.0 \\ 
	C$7$: Economic Development & 23.2 & 76.8 & 0.0 \\ 
C$8$: Peace and War & 23.88 & 76.12 & 0 \\ \hline
    \textbf{Total} & {\color[HTML]{000000} \textbf{21.63}} & {\color[HTML]{FF0000} \textbf{78.37}} & {\color[HTML]{000000} \textbf{0.0}} \\ \hline                                          
    \end{tabular}
    \caption{Above table provides the percentage of the answers that are correctly, incorrectly, and not answered by \texttt{llama-2-chat} on \textit{WB}.}
    \label{tab:result_LLaMA_WB_}
    \end{table*}

    \begin{table*}[]
    \centering
    \begin{tabular}{p{4.5cm}ccc}
    \hline
    \textbf{Category} & \multicolumn{1}{l}{\textbf{Correctly Answered}} & \multicolumn{1}{l}{\textbf{Incorrectly Answered}} & \multicolumn{1}{l}{\textbf{Not Answered}} \\ \hline 
	C$1$: Climate & 10.16 & 68.98 & 20.86 \\ 
	C$2$: Food and Agriculture & 8.3 & 91.58 & 0.12 \\ 
	C$3$: Health & 14.05 & 85.95 & 0.0 \\ 
	C$4$: Human Rights & 17.49 & 80.48 & 2.02 \\ 
	C$5$: Innovation & 12.82 & 87.13 & 0.05 \\ 
	C$6$: Migration & 20.43 & 72.43 & 7.14 \\ 
	C$7$: Economic Development & 19.76 & 80.24 & 0.0 \\ 
C$8$: Peace and War & 17.06 & 82.89 & 0.06 \\ \hline
    \textbf{Total} & {\color[HTML]{000000} \textbf{14.89}} & {\color[HTML]{FF0000} \textbf{81.1}} & {\color[HTML]{000000} \textbf{4.01}} \\ \hline                                          
    \end{tabular}
    \caption{Above table provides the percentage of the answers that are correctly, incorrectly, and not answered by \texttt{llama-2-chat} on \textit{MM}.}
    \label{tab:result_LLaMA_MMB_}
    \end{table*}

    \begin{table*}[]
    \centering
    \begin{tabular}{p{4.5cm}ccc}
    \hline
    \textbf{Category} & \multicolumn{1}{l}{\textbf{Correctly Answered}} & \multicolumn{1}{l}{\textbf{Incorrectly Answered}} & \multicolumn{1}{l}{\textbf{Not Answered}} \\ \hline 
	C$1$: Climate & 19.1 & 80.9 & 0.0 \\ 
	C$2$: Food and Agriculture & 13.91 & 86.09 & 0.0 \\ 
	C$3$: Health & 17.82 & 82.18 & 0.0 \\ 
	C$4$: Human Rights & 31.66 & 68.34 & 0.0 \\ 
	C$5$: Innovation & 25.69 & 74.31 & 0.0 \\ 
	C$6$: Migration & 21.29 & 78.71 & 0.0 \\ 
	C$7$: Economic Development & 24.32 & 75.68 & 0.0 \\ 
C$8$: Peace and War & 21.77 & 78.23 & 0 \\ \hline
    \textbf{Total} & {\color[HTML]{000000} \textbf{21.5}} & {\color[HTML]{FF0000} \textbf{78.5}} & {\color[HTML]{000000} \textbf{0.0}} \\ \hline                                          
    \end{tabular}
    \caption{Above table provides the percentage of the answers that are correctly, incorrectly, and not answered by \texttt{llama-2-chat} on \textit{RB}.}
    \label{tab:result_LLaMA_RAB_}
    \end{table*}

    \begin{table*}[]
    \centering
    \begin{tabular}{p{4.5cm}ccc}
    \hline
    \textbf{Category} & \multicolumn{1}{l}{\textbf{Correctly Answered}} & \multicolumn{1}{l}{\textbf{Incorrectly Answered}} & \multicolumn{1}{l}{\textbf{Not Answered}} \\ \hline 
	C$1$: Climate & 2.17 & 93.2 & 4.63 \\ 
	C$2$: Food and Agriculture & 0.64 & 97.8 & 1.56 \\ 
	C$3$: Health & 4.16 & 95.41 & 0.43 \\ 
	C$4$: Human Rights & 5.46 & 92.63 & 1.91 \\ 
	C$5$: Innovation & 4.0 & 95.36 & 0.64 \\ 
	C$6$: Migration & 9.02 & 87.45 & 3.53 \\ 
	C$7$: Economic Development & 8.28 & 89.32 & 2.4 \\ 
C$8$: Peace and War & 5.34 & 90.51 & 4.15 \\ \hline
    \textbf{Total} & {\color[HTML]{000000} \textbf{4.88}} & {\color[HTML]{FF0000} \textbf{92.75}} & {\color[HTML]{000000} \textbf{2.38}} \\ \hline                                          
    \end{tabular}
    \caption{Above table provides the percentage of the answers that are correctly, incorrectly, and not answered by \texttt{llama-2-chat} on \textit{TB}.}
    \label{tab:result_LLaMA_TB_}
    \end{table*}

  \begin{table*}[]
    \centering
    \begin{tabular}{p{4.5cm}ccc}
    \hline
\textbf{Category} & \multicolumn{1}{l}{\textbf{Correctly Answered}} & \multicolumn{1}{l}{\textbf{Incorrectly Answered}} & \multicolumn{1}{l}{\textbf{Not Answered}} \\ \hline
C$1$: Climate & 37.43 & 62.57 & 0.01 \\
C$2$: Food and Agriculture & 9.84 & 90.16 & 0 \\
C$3$: Health & 19.51 & 80.49 & 0 \\
C$4$: Human Rights & 21.46 & 78.54 & 0 \\
C$5$: Innovation & 7.91 & 92.09 & 0 \\
C$6$: Migration & 25.37 & 72.64 & 2 \\
C$7$: Economic Development & 11.2 & 78.18 & 10.61 \\
C$8$: Peace and War & 23.61 & 76.39 & 0 \\ \hline
\textbf{Total} & \textbf{20.86} & {\color[HTML]{FE0000} \textbf{76.9}} & \textbf{2.24} \\ \hline
\end{tabular}
    \caption{Above table provides the percentage of the answers that are correctly, incorrectly, and not answered by \texttt{gemma-7b-it} on \textit{DB}.}
    \label{tab:result_gemma_DB_}
    \end{table*}


  \begin{table*}[]
    \centering
    \begin{tabular}{p{4.5cm}ccc}
    \hline
\textbf{Category} & \multicolumn{1}{l}{\textbf{Correctly Answered}} & \multicolumn{1}{l}{\textbf{Incorrectly Answered}} & \multicolumn{1}{l}{\textbf{Not Answered}} \\ \hline
C$1$: Climate & 47.05 & 52.95 & 0 \\
C$2$: Food and Agriculture & 13.84 & 86.16 & 0 \\
C$3$: Health & 52.43 & 47.57 & 0 \\
C$4$: Human Rights & 11.75 & 88.25 & 0 \\
C$5$: Innovation & 40.43 & 59.57 & 0 \\
C$6$: Migration & 49.82 & 50.18 & 0 \\
C$7$: Economic Development & 65.65 & 34.35 & 0 \\
C$8$: Peace and War & 64.13 & 35.87 & 0 \\ \hline
\textbf{Total} & \textbf{42.31} & {\color[HTML]{FE0000} \textbf{57.69}} & \textbf{0} \\ \hline
\end{tabular}
    \caption{Above table provides the percentage of the answers that are correctly, incorrectly, and not answered by \texttt{gemma-7b-it} on \textit{CB}.}
    \label{tab:result_gemma_CB_}
    \end{table*}

  \begin{table*}[]
    \centering
    \begin{tabular}{p{4.5cm}ccc}
    \hline
\textbf{Category} & \multicolumn{1}{l}{\textbf{Correctly Answered}} & \multicolumn{1}{l}{\textbf{Incorrectly Answered}} & \multicolumn{1}{l}{\textbf{Not Answered}} \\ \hline
C$1$: Climate & 10 & 90 & 0 \\
C$2$: Food and Agriculture & 7.09 & 92.91 & 0 \\
C$3$: Health & 14.94 & 85.06 & 0 \\
C$4$: Human Rights & 24.8 & 75.2 & 0 \\
C$5$: Innovation & 13.37 & 86.63 & 0 \\
C$6$: Migration & 16.08 & 83.92 & 0 \\
C$7$: Economic Development & 23.08 & 76.92 & 0 \\
C$8$: Peace and War & 15.86 & 84.14 & 0 \\ \hline
\textbf{Total} & \textbf{15.34} & {\color[HTML]{FE0000} \textbf{84.66}} & \textbf{0} \\ \hline
\end{tabular}
    \caption{Above table provides the percentage of the answers that are correctly, incorrectly, and not answered by \texttt{gemma-7b-it} on \textit{WB}.}
    \label{tab:result_gemma_WB_}
    \end{table*}

  \begin{table*}[]
    \centering
    \begin{tabular}{p{4.5cm}ccc}
    \hline
\textbf{Category} & \multicolumn{1}{l}{\textbf{Correctly Answered}} & \multicolumn{1}{l}{\textbf{Incorrectly Answered}} & \multicolumn{1}{l}{\textbf{Not Answered}} \\ \hline
C$1$: Climate & 9.39 & 90.45 & 0.16 \\
C$2$: Food and Agriculture & 14.51 & 85.45 & 0.04 \\
C$3$: Health & 9.22 & 90.78 & 0 \\
C$4$: Human Rights & 10.12 & 89.88 & 0 \\
C$5$: Innovation & 7.54 & 92.46 & 0 \\
C$6$: Migration & 15.61 & 84.39 & 0 \\
C$7$: Economic Development & 9.8 & 90.2 & 0 \\
C$8$: Peace and War & 18.14 & 81.86 & 0 \\ \hline
\textbf{Total} & \textbf{11.68} & {\color[HTML]{FE0000} \textbf{88.29}} & \textbf{0.03} \\ \hline
\end{tabular}
    \caption{Above table provides the percentage of the answers that are correctly, incorrectly, and not answered by \texttt{gemma-7b-it} on \textit{MM}.}
    \label{tab:result_gemma_MM_}
    \end{table*}

  \begin{table*}[]
    \centering
    \begin{tabular}{p{4.5cm}ccc}
    \hline
\textbf{Category} & \multicolumn{1}{l}{\textbf{Correctly Answered}} & \multicolumn{1}{l}{\textbf{Incorrectly Answered}} & \multicolumn{1}{l}{\textbf{Not Answered}} \\ \hline
C$1$: Climate & 24.02 & 75.82 & 0.16 \\
C$2$: Food and Agriculture & 7.45 & 92.55 & 0 \\
C$3$: Health & 14.32 & 85.68 & 0 \\
C$4$: Human Rights & 10.46 & 89.54 & 0 \\
C$5$: Innovation & 8.55 & 91.45 & 0 \\
C$6$: Migration & 16.43 & 83.57 & 0 \\
C$7$: Economic Development & 11.72 & 88.28 & 0 \\
C$8$: Peace and War & 19.61 & 80.33 & 0.06 \\ \hline
\textbf{Total} & \textbf{14.06} & {\color[HTML]{FE0000} \textbf{85.91}} & \textbf{0.03} \\ \hline
\end{tabular}
    \caption{Above table provides the percentage of the answers that are correctly, incorrectly, and not answered by \texttt{gemma-7b-it} on \textit{RB}.}
    \label{tab:result_gemma_RB_}
    \end{table*}

  \begin{table*}[]
    \centering
    \begin{tabular}{p{4.5cm}ccc}
    \hline
\textbf{Category} & \multicolumn{1}{l}{\textbf{Correctly Answered}} & \multicolumn{1}{l}{\textbf{Incorrectly Answered}} & \multicolumn{1}{l}{\textbf{Not Answered}} \\ \hline
C$1$: Climate & 3.03 & 76.97 & 20 \\
C$2$: Food and Agriculture & 0.24 & 99.76 & 0 \\
C$3$: Health & 3.15 & 96.85 & 0 \\
C$4$: Human Rights & 3.71 & 95.05 & 1.24 \\
C$5$: Innovation & 1.88 & 98.12 & 0 \\
C$6$: Migration & 6.2 & 93.8 & 0 \\
C$7$: Economic Development & 1.28 & 98.72 & 0 \\
C$8$: Peace and War & 4.83 & 95.17 & 0 \\ \hline
\textbf{Total} & \textbf{2.97} & {\color[HTML]{FE0000} \textbf{94.24}} & \textbf{2.79} \\ \hline
\end{tabular}
    \caption{Above table provides the percentage of the answers that are correctly, incorrectly, and not answered by \texttt{gemma-7b-it} on \textit{TB}.}
    \label{tab:result_gemma_TB_}
    \end{table*}

  \begin{table*}[]
    \centering
    \begin{tabular}{p{4.5cm}ccc}
    \hline
\textbf{Category} & \multicolumn{1}{l}{\textbf{Correctly Answered}} & \multicolumn{1}{l}{\textbf{Incorrectly Answered}} & \multicolumn{1}{l}{\textbf{Not Answered}} \\ \hline
C$1$: Climate & 56.29 & 42.16 & 1.56 \\
C$2$: Food and Agriculture & 49.74 & 50.11 & 0.14 \\
C$3$: Health & 26.14 & 73.84 & 0.02 \\
C$4$: Human Rights & 38.95 & 61.05 & 0 \\
C$5$: Innovation & 20.61 & 79.39 & 0 \\
C$6$: Migration & 42.09 & 57.82 & 0.09 \\
C$7$: Economic Development & 29.94 & 66.71 & 3.35 \\
C$8$: Peace and War & 37.37 & 62.63 & 0 \\ \hline
\textbf{Total} & \textbf{38.54} & {\color[HTML]{FE0000} \textbf{60.56}} & \textbf{0.9} \\ \hline
\end{tabular}
    \caption{Above table provides the percentage of the answers that are correctly, incorrectly, and not answered by \texttt{llama-3-8b-8192} on \textit{DB}.}
    \label{tab:result_llama3_8b_DB_}
    \end{table*}

  \begin{table*}[]
    \centering
    \begin{tabular}{p{4.5cm}ccc}
    \hline
\textbf{Category} & \multicolumn{1}{l}{\textbf{Correctly Answered}} & \multicolumn{1}{l}{\textbf{Incorrectly Answered}} & \multicolumn{1}{l}{\textbf{Not Answered}} \\ \hline
C$1$: Climate & 46.39 & 53.57 & 0.04 \\
C$2$: Food and Agriculture & 15.12 & 84.88 & 0 \\
C$3$: Health & 52.08 & 47.92 & 0 \\
C$4$: Human Rights & 12.95 & 87.05 & 0 \\
C$5$: Innovation & 40.43 & 59.57 & 0 \\
C$6$: Migration & 28.84 & 71.16 & 0 \\
C$7$: Economic Development & 65.37 & 34.63 & 0 \\
C$8$: Peace and War & 50.58 & 49.42 & 0 \\ \hline
\textbf{Total} & \textbf{38.66} & {\color[HTML]{FE0000} \textbf{61.33}} & \textbf{0.01} \\ \hline
\end{tabular}
    \caption{Above table provides the percentage of the answers that are correctly, incorrectly, and not answered by \texttt{llama-3-8b-8192} on \textit{CB}.}
    \label{tab:result_llama3_8b_CB_}
    \end{table*}

  \begin{table*}[]
    \centering
    \begin{tabular}{p{4.5cm}ccc}
    \hline
\textbf{Category} & \multicolumn{1}{l}{\textbf{Correctly Answered}} & \multicolumn{1}{l}{\textbf{Incorrectly Answered}} & \multicolumn{1}{l}{\textbf{Not Answered}} \\ \hline
C$1$: Climate & 22.7 & 77.3 & 0 \\
C$2$: Food and Agriculture & 6.93 & 93.07 & 0 \\
C$3$: Health & 12.57 & 87.43 & 0 \\
C$4$: Human Rights & 19.85 & 80.15 & 0 \\
C$5$: Innovation & 15.58 & 84.42 & 0 \\
C$6$: Migration & 23.37 & 76.63 & 0 \\
C$7$: Economic Development & 23.88 & 76.12 & 0 \\
C$8$: Peace and War & 28.77 & 71.23 & 0 \\ \hline
\textbf{Total} & \textbf{18.84} & {\color[HTML]{FE0000} \textbf{81.16}} & \textbf{0} \\ \hline
\end{tabular}
    \caption{Above table provides the percentage of the answers that are correctly, incorrectly, and not answered by \texttt{llama-3-8b-8192} on \textit{WB}.}
    \label{tab:result_llama3_8b_WB_}
    \end{table*}

  \begin{table*}[]
    \centering
    \begin{tabular}{p{4.5cm}ccc}
    \hline
\textbf{Category} & \multicolumn{1}{l}{\textbf{Correctly Answered}} & \multicolumn{1}{l}{\textbf{Incorrectly Answered}} & \multicolumn{1}{l}{\textbf{Not Answered}} \\ \hline
C$1$: Climate & 25.61 & 74.34 & 0.04 \\
C$2$: Food and Agriculture & 14.87 & 85.13 & 0 \\
C$3$: Health & 13.39 & 86.61 & 0 \\
C$4$: Human Rights & 16.82 & 83.18 & 0 \\
C$5$: Innovation & 17.28 & 82.72 & 0 \\
C$6$: Migration & 24.2 & 75.8 & 0 \\
C$7$: Economic Development & 16.96 & 83.04 & 0 \\
C$8$: Peace and War & 17.28 & 82.72 & 0 \\ \hline
\textbf{Total} & \textbf{18.39} & {\color[HTML]{FE0000} \textbf{81.6}} & \textbf{0.01} \\ \hline
\end{tabular}
    \caption{Above table provides the percentage of the answers that are correctly, incorrectly, and not answered by \texttt{llama-3-8b-8192} on \textit{MM}.}
    \label{tab:result_llama3_8b_MM_}
    \end{table*}

  \begin{table*}[]
    \centering
    \begin{tabular}{p{4.5cm}ccc}
    \hline
\textbf{Category} & \multicolumn{1}{l}{\textbf{Correctly Answered}} & \multicolumn{1}{l}{\textbf{Incorrectly Answered}} & \multicolumn{1}{l}{\textbf{Not Answered}} \\ \hline
C$1$: Climate & 34.88 & 65.12 & 0 \\
C$2$: Food and Agriculture & 29.62 & 70.38 & 0 \\
C$3$: Health & 16.69 & 83.31 & 0 \\
C$4$: Human Rights & 25.82 & 74.18 & 0 \\
C$5$: Innovation & 17.1 & 82.9 & 0 \\
C$6$: Migration & 26.39 & 73.57 & 0.04 \\
C$7$: Economic Development & 21.08 & 78.92 & 0 \\
C$8$: Peace and War & 22.74 & 77.26 & 0 \\ \hline
\textbf{Total} & \textbf{24.36} & {\color[HTML]{FE0000} \textbf{75.63}} & \textbf{0.01} \\ \hline
\end{tabular}
    \caption{Above table provides the percentage of the answers that are correctly, incorrectly, and not answered by \texttt{llama-3-8b-8192} on \textit{RB}.}
    \label{tab:result_llama3_8b_RB_}
    \end{table*}

  \begin{table*}[]
    \centering
    \begin{tabular}{p{4.5cm}ccc}
    \hline
\textbf{Category} & \multicolumn{1}{l}{\textbf{Correctly Answered}} & \multicolumn{1}{l}{\textbf{Incorrectly Answered}} & \multicolumn{1}{l}{\textbf{Not Answered}} \\ \hline
C$1$: Climate & 9.96 & 90.04 & 0 \\
C$2$: Food and Agriculture & 5.37 & 94.63 & 0 \\
C$3$: Health & 8.6 & 91.4 & 0 \\
C$4$: Human Rights & 2.31 & 97.69 & 0 \\
C$5$: Innovation & 5.93 & 94.07 & 0 \\
C$6$: Migration & 7.49 & 92.51 & 0 \\
C$7$: Economic Development & 10.68 & 89.32 & 0 \\
C$8$: Peace and War & 4.89 & 95.11 & 0 \\ \hline
\textbf{Total} & \textbf{7.18} & {\color[HTML]{FE0000} \textbf{92.82}} & \textbf{0} \\ \hline
\end{tabular}
    \caption{Above table provides the percentage of the answers that are correctly, incorrectly, and not answered by \texttt{llama-3-8b-8192} on \textit{TB}.}
    \label{tab:result_llama3_8b_TB_}
    \end{table*}


  \begin{table*}[]
    \centering
    \begin{tabular}{p{4.5cm}ccc}
 \hline
\textbf{} & \multicolumn{1}{l}{\textbf{Correctly Answered}} & \multicolumn{1}{l}{\textbf{Incorrectly Answered}} & \multicolumn{1}{l}{\textbf{Not Answered}} \\ \hline
C$1$: Climate              & 11.83         & 5.99           & 82.19          \\
C$2$: Food and Agriculture & 5.08          & 19.55          & 75.37          \\
C$3$: Health               & 11.03         & 25.08          & 63.89          \\
C$4$: Human Rights         & 10.35         & 23.79          & 65.87          \\
C$5$: Innovation           & 17.31         & 45.14          & 37.55          \\
C$6$: Migration            & 0.04          & 0.1            & 99.86          \\
C$7$: Economic Development & 6.32          & 13.58          & 80.1           \\
C$8$: Peace and War        & 24.27         & 29.57          & 46.15          \\ \hline
\textbf{Total}             & \textbf{9.37} & {\color[HTML]{FE0000} \textbf{16.46}} & \textbf{74.18} \\ \hline
\end{tabular}
    \caption{Above table provides the percentage of the answers that are correctly, incorrectly, and not answered by \texttt{phi-3-instruct} on \textit{DB}.}
    \label{tab:results-phi-3-instructDB_}
    \end{table*}

  \begin{table*}[]
    \centering
    \begin{tabular}{p{4.5cm}ccc}
\hline
\textbf{} & \multicolumn{1}{l}{\textbf{Correctly Answered}} & \multicolumn{1}{l}{\textbf{Incorrectly Answered}} & \multicolumn{1}{l}{\textbf{Not Answered}} \\ \hline
C$1$: Climate              & 48.07          & 49.96          & 1.97          \\
C$2$: Food and Agriculture & 53.57          & 45.51          & 0.92          \\
C$3$: Health               & 49.71          & 48.7           & 1.59          \\
C$4$: Human Rights         & 35.71          & 40.7           & 23.59         \\
C$5$: Innovation           & 34.21          & 64.06          & 1.73          \\
C$6$: Migration            & 63.49          & 30             & 6.51          \\
C$7$: Economic Development & 52.2           & 46.4           & 1.4           \\
C$8$: Peace and War        & 45.47          & 49.36          & 5.16          \\ \hline
\textbf{Total}             & \textbf{48.64} & {\color[HTML]{FE0000} \textbf{46.63}} & \textbf{4.73} \\ \hline
\end{tabular}
    \caption{Above table provides the percentage of the answers that are correctly, incorrectly, and not answered by \texttt{phi-3-instruct} on \textit{CB}.}
    \label{tab:results-phi-3-instructCB_}
    \end{table*}

  \begin{table*}[]
    \centering
    \begin{tabular}{p{4.5cm}ccc}
 \hline
\textbf{} & \multicolumn{1}{l}{\textbf{Correctly Answered}} & \multicolumn{1}{l}{\textbf{Incorrectly Answered}} & \multicolumn{1}{l}{\textbf{Not Answered}} \\ \hline
C$1$: Climate              & 38.77          & 13.2           & 48.03          \\
C$2$: Food and Agriculture & 17.11          & 44.57          & 38.32          \\
C$3$: Health               & 47.74          & 47.67          & 4.59           \\
C$4$: Human Rights         & 39.93          & 17.38          & 42.69          \\
C$5$: Innovation           & 22.47          & 37.78          & 39.75          \\
C$6$: Migration            & 17.92          & 11.45          & 70.63          \\
C$7$: Economic Development & 51.64          & 41.56          & 6.8            \\
C$8$: Peace and War        & 64.18          & 26.49          & 9.32           \\ \hline
\textbf{Total}             & \textbf{36.54} & {\color[HTML]{FE0000} \textbf{30.58}} & \textbf{32.87} \\ \hline
\end{tabular}
    \caption{Above table provides the percentage of the answers that are correctly, incorrectly, and not answered by \texttt{phi-3-instruct} on \textit{WB}.}
    \label{tab:results-phi-3-instructWB_}
    \end{table*}

  \begin{table*}[]
    \centering
    \begin{tabular}{p{4.5cm}ccc}
\hline
\textbf{} & \multicolumn{1}{l}{\textbf{Correctly Answered}} & \multicolumn{1}{l}{\textbf{Incorrectly Answered}} & \multicolumn{1}{l}{\textbf{Not Answered}} \\ \hline
C$1$: Climate              & 2.83           & 3.4            & 93.77          \\
C$2$: Food and Agriculture & 6.97           & 22.73          & 70.3           \\
C$3$: Health               & 22.68          & 50.51          & 26.81          \\
C$4$: Human Rights         & 3.37           & 10.91          & 85.71          \\
C$5$: Innovation           & 18.93          & 68.75          & 12.32          \\
C$6$: Migration            & 0              & 0              & 100            \\
C$7$: Economic Development & 12.08          & 21.92          & 66             \\
C$8$: Peace and War        & 14.38          & 44.8           & 40.82          \\ \hline
\textbf{Total}             & \textbf{10.14} & {\color[HTML]{FE0000} \textbf{27.23}} & \textbf{62.63} \\ \hline
\end{tabular}
    \caption{Above table provides the percentage of the answers that are correctly, incorrectly, and not answered by \texttt{phi-3-instruct} on \textit{MM}.}
    \label{tab:results-phi-3-instructMM_}
    \end{table*}

  \begin{table*}[]
    \centering
    \begin{tabular}{p{4.5cm}ccc}
 \hline
\textbf{} & \multicolumn{1}{l}{\textbf{Correctly Answered}} & \multicolumn{1}{l}{\textbf{Incorrectly Answered}} & \multicolumn{1}{l}{\textbf{Not Answered}} \\ \hline
C$1$: Climate              & 1.02          & 3.85          & 95.12          \\
C$2$: Food and Agriculture & 0.08          & 0.24          & 99.68          \\
C$3$: Health               & 0             & 0.08          & 99.92          \\
C$4$: Human Rights         & 0             & 1.46          & 98.54          \\
C$5$: Innovation           & 0.05          & 0.55          & 99.4           \\
C$6$: Migration            & 0             & 0             & 100            \\
C$7$: Economic Development & 5.48          & 16.92         & 77.6           \\
C$8$: Peace and War        & 0.63          & 0.91          & 98.47          \\ \hline
\textbf{Total}             & \textbf{0.96} & {\color[HTML]{FE0000} \textbf{3.17}} & \textbf{95.87} \\ \hline
\end{tabular}
    \caption{Above table provides the percentage of the answers that are correctly, incorrectly, and not answered by \texttt{phi-3-instruct} on \textit{RB}.}
    \label{tab:results-phi-3-instructRB_}
    \end{table*}

  \begin{table*}[]
    \centering
    \begin{tabular}{p{4.5cm}ccc}
\hline
\textbf{} & \multicolumn{1}{l}{\textbf{Correctly Answered}} & \multicolumn{1}{l}{\textbf{Incorrectly Answered}} & \multicolumn{1}{l}{\textbf{Not Answered}} \\ \hline
C$1$: Climate              & 0.49          & 20.78          & 78.73          \\
C$2$: Food and Agriculture & 0.2           & 24.61          & 75.19          \\
C$3$: Health               & 1.09          & 73.19          & 25.72          \\
C$4$: Human Rights         & 0             & 8.32           & 91.68          \\
C$5$: Innovation           & 0.41          & 90.35          & 9.24           \\
C$6$: Migration            & 2.08          & 70.31          & 27.61          \\
C$7$: Economic Development & 0.84          & 75.92          & 23.24          \\
C$8$: Peace and War        & 2.67          & 52.7           & 44.63          \\ \hline
\textbf{Total}             & \textbf{0.96} & {\color[HTML]{FE0000} \textbf{53.28}} & \textbf{45.76} \\ \hline
\end{tabular}
    \caption{Above table provides the percentage of the answers that are correctly, incorrectly, and not answered by \texttt{phi-3-instruct} on \textit{TB}.}
    \label{tab:results-phi-3-instructTB_}
    \end{table*}


  \begin{table*}[]
    \centering
    \begin{tabular}{p{4.5cm}ccc}
    \hline
\textbf{Category} & \multicolumn{1}{l}{\textbf{Correctly Answered}} & \multicolumn{1}{l}{\textbf{Incorrectly Answered}} & \multicolumn{1}{l}{\textbf{Not Answered}} \\ \hline
C$1$: Climate & 43.98 & 55.79 & 0.22 \\
C$2$: Food and Agriculture & 21.75 & 76.7 & 1.55 \\
C$3$: Health & 24.58 & 74.39 & 1.03 \\
C$4$: Human Rights & 26.06 & 73.42 & 0.52 \\
C$5$: Innovation & 38.04 & 38.67 & 23.28 \\
C$6$: Migration & 29.53 & 70.28 & 0.19 \\
C$7$: Economic Development & 25.63 & 74.15 & 0.22 \\
C$8$: Peace and War & 53.53 & 17.61 & 28.87 \\ \hline
\textbf{Total} & \textbf{32.57} & {\color[HTML]{FE0000} \textbf{60.71}} & \textbf{6.73} \\ \hline
\end{tabular}
    \caption{Above table provides the percentage of the answers that are correctly, incorrectly, and not answered by \texttt{mixtral-8x7b-32768} on \textit{DB}.}
    \label{tab:result_mixtral-87b_DB_}
    \end{table*}

  \begin{table*}[]
    \centering
    \begin{tabular}{p{4.5cm}ccc}
    \hline
\textbf{Category} & \multicolumn{1}{l}{\textbf{Correctly Answered}} & \multicolumn{1}{l}{\textbf{Incorrectly Answered}} & \multicolumn{1}{l}{\textbf{Not Answered}} \\ \hline
C$1$: Climate & 44.51 & 54.92 & 0.57 \\
C$2$: Food and Agriculture & 14.16 & 85.72 & 0.12 \\
C$3$: Health & 47.12 & 51.51 & 1.38 \\
C$4$: Human Rights & 12.41 & 87.59 & 0 \\
C$5$: Innovation & 32.14 & 52.83 & 15.03 \\
C$6$: Migration & 51.19 & 48.81 & 0 \\
C$7$: Economic Development & 56.12 & 41.69 & 2.2 \\
C$8$: Peace and War & 38.2 & 60.41 & 1.39 \\ \hline
\textbf{Total} & \textbf{37.59} & {\color[HTML]{FE0000} \textbf{60.31}} & \textbf{2.09} \\ \hline
\end{tabular}
    \caption{Above table provides the percentage of the answers that are correctly, incorrectly, and not answered by \texttt{mixtral-8x7b-32768} on \textit{CB}.}
    \label{tab:result_mixtral-87b_CB_}
    \end{table*}

  \begin{table*}[]
    \centering
    \begin{tabular}{p{4.5cm}ccc}
    \hline
\textbf{Category} & \multicolumn{1}{l}{\textbf{Correctly Answered}} & \multicolumn{1}{l}{\textbf{Incorrectly Answered}} & \multicolumn{1}{l}{\textbf{Not Answered}} \\ \hline
C$1$: Climate & 68.11 & 30.66 & 1.23 \\
C$2$: Food and Agriculture & 4.21 & 95.75 & 0.04 \\
C$3$: Health & 25.45 & 74.55 & 0 \\
C$4$: Human Rights & 29.98 & 69.29 & 0.73 \\
C$5$: Innovation & 32.03 & 67.97 & 0 \\
C$6$: Migration & 22.31 & 77.69 & 0 \\
C$7$: Economic Development & 20.48 & 79.48 & 0.04 \\
C$8$: Peace and War & 30.59 & 69.36 & 0.06 \\ \hline
\textbf{Total} & \textbf{28.85} & {\color[HTML]{FE0000} \textbf{70.9}} & \textbf{0.25} \\ \hline
\end{tabular}
    \caption{Above table provides the percentage of the answers that are correctly, incorrectly, and not answered by \texttt{mixtral-8x7b-32768} on \textit{WB}.}
    \label{tab:result_mixtral-87b_WB_}
    \end{table*}

  \begin{table*}[]
    \centering
    \begin{tabular}{p{4.5cm}ccc}
    \hline
\textbf{Category} & \multicolumn{1}{l}{\textbf{Correctly Answered}} & \multicolumn{1}{l}{\textbf{Incorrectly Answered}} & \multicolumn{1}{l}{\textbf{Not Answered}} \\ \hline
C$1$: Climate & 55.25 & 44.59 & 0.16 \\
C$2$: Food and Agriculture & 5.25 & 94.71 & 0.04 \\
C$3$: Health & 13.39 & 86.61 & 0 \\
C$4$: Human Rights & 22.38 & 77.62 & 0 \\
C$5$: Innovation & 14.71 & 85.29 & 0 \\
C$6$: Migration & 12.78 & 87.22 & 0 \\
C$7$: Economic Development & 5.36 & 94.64 & 0 \\
C$8$: Peace and War & 13.38 & 86.5 & 0.12 \\ \hline
\textbf{Total} & \textbf{17.74} & {\color[HTML]{FE0000} \textbf{82.22}} & \textbf{0.04} \\ \hline
\end{tabular}
    \caption{Above table provides the percentage of the answers that are correctly, incorrectly, and not answered by \texttt{mixtral-8x7b-32768} on \textit{MM}.}
    \label{tab:result_mixtral-87b_MM_}
    \end{table*}

  \begin{table*}[]
    \centering
    \begin{tabular}{p{4.5cm}ccc}
    \hline
\textbf{Category} & \multicolumn{1}{l}{\textbf{Correctly Answered}} & \multicolumn{1}{l}{\textbf{Incorrectly Answered}} & \multicolumn{1}{l}{\textbf{Not Answered}} \\ \hline
C$1$: Climate & 69.55 & 29.34 & 1.11 \\
C$2$: Food and Agriculture & 11.66 & 88.3 & 0.04 \\
C$3$: Health & 29.11 & 70.89 & 0 \\
C$4$: Human Rights & 52.59 & 46.57 & 0.84 \\
C$5$: Innovation & 20.54 & 79.46 & 0 \\
C$6$: Migration & 13.12 & 86.88 & 0 \\
C$7$: Economic Development & 10.88 & 89.08 & 0.04 \\
C$8$: Peace and War & 30.24 & 69.76 & 0 \\ \hline
\textbf{Total} & \textbf{28.77} & {\color[HTML]{FE0000} \textbf{70.99}} & \textbf{0.24} \\ \hline
\end{tabular}
    \caption{Above table provides the percentage of the answers that are correctly, incorrectly, and not answered by \texttt{mixtral-8x7b-32768} on \textit{RB}.}
    \label{tab:result_mixtral-87b_RB_}
    \end{table*}

  \begin{table*}[]
    \centering
    \begin{tabular}{p{4.5cm}ccc}
    \hline
\textbf{Category} & \multicolumn{1}{l}{\textbf{Correctly Answered}} & \multicolumn{1}{l}{\textbf{Incorrectly Answered}} & \multicolumn{1}{l}{\textbf{Not Answered}} \\ \hline
C$1$: Climate & 13.73 & 67.91 & 18.36 \\
C$2$: Food and Agriculture & 0.2 & 99.48 & 0.32 \\
C$3$: Health & 0.08 & 99.92 & 0 \\
C$4$: Human Rights & 1.74 & 92.69 & 5.57 \\
C$5$: Innovation & 0.09 & 99.59 & 0.32 \\
C$6$: Migration & 3.02 & 96.9 & 0.08 \\
C$7$: Economic Development & 0.6 & 99.16 & 0.24 \\
C$8$: Peace and War & 4.04 & 95.91 & 0.06 \\ \hline
\textbf{Total} & \textbf{2.95} & {\color[HTML]{FE0000} \textbf{93.93}} & \textbf{3.13} \\ \hline
\end{tabular}
    \caption{Above table provides the percentage of the answers that are correctly, incorrectly, and not answered by \texttt{mixtral-8x7b-32768} on \textit{TB}.}
    \label{tab:result_mixtral-87b_TB_}
    \end{table*}


  \begin{table*}[]
    \centering
    \begin{tabular}{p{4.5cm}ccc}
    \hline
\textbf{Category} & \multicolumn{1}{l}{\textbf{Correctly Answered}} & \multicolumn{1}{l}{\textbf{Incorrectly Answered}} & \multicolumn{1}{l}{\textbf{Not Answered}} \\ \hline
C$1$: Climate & 76.55 & 23.43 & 0.02 \\
C$2$: Food and Agriculture & 16.78 & 83.21 & 0.01 \\
C$3$: Health & 41.23 & 58.77 & 0 \\
C$4$: Human Rights & 37.58 & 62.42 & 0 \\
C$5$: Innovation & 35.39 & 64.6 & 0.02 \\
C$6$: Migration & 35.18 & 64.82 & 0.01 \\
C$7$: Economic Development & 26.38 & 73.62 & 0 \\
C$8$: Peace and War & 40.42 & 59.57 & 0.01 \\ \hline
\textbf{Total} & \textbf{40.44} & {\color[HTML]{FE0000} \textbf{59.55}} & \textbf{0.01} \\ \hline
\end{tabular}
    \caption{Above table provides the percentage of the answers that are correctly, incorrectly, and not answered by \texttt{llama-3-70b-8192} on \textit{DB}.}
    \label{tab:result_llama-370b_DB_}
    \end{table*}

  \begin{table*}[]
    \centering
    \begin{tabular}{p{4.5cm}ccc}
    \hline
\textbf{Category} & \multicolumn{1}{l}{\textbf{Correctly Answered}} & \multicolumn{1}{l}{\textbf{Incorrectly Answered}} & \multicolumn{1}{l}{\textbf{Not Answered}} \\ \hline
C$1$: Climate & 44.55 & 55.45 & 0 \\
C$2$: Food and Agriculture & 47.42 & 52.58 & 0 \\
C$3$: Health & 14.89 & 85.11 & 0 \\
C$4$: Human Rights & 58.53 & 41.47 & 0 \\
C$5$: Innovation & 27.98 & 72.02 & 0 \\
C$6$: Migration & 35.29 & 64.71 & 0 \\
C$7$: Economic Development & 35.92 & 64.08 & 0 \\
C$8$: Peace and War & 33.76 & 66.24 & 0 \\ \hline
\textbf{Total} & \textbf{36.74} & {\color[HTML]{FE0000} \textbf{63.26}} & \textbf{0} \\ \hline
\end{tabular}
    \caption{Above table provides the percentage of the answers that are correctly, incorrectly, and not answered by \texttt{llama-3-70b-8192} on \textit{CB}.}
    \label{tab:result_llama-370b_CB_}
    \end{table*}

  \begin{table*}[]
    \centering
    \begin{tabular}{p{4.5cm}ccc}
    \hline
\textbf{Category} & \multicolumn{1}{l}{\textbf{Correctly Answered}} & \multicolumn{1}{l}{\textbf{Incorrectly Answered}} & \multicolumn{1}{l}{\textbf{Not Answered}} \\ \hline
C$1$: Climate & 76.64 & 21.31 & 2.05 \\
C$2$: Food and Agriculture & 29.98 & 70.02 & 0 \\
C$3$: Health & 64.05 & 35.95 & 0 \\
C$4$: Human Rights & 46.63 & 53.37 & 0 \\
C$5$: Innovation & 56.62 & 43.38 & 0 \\
C$6$: Migration & 47.41 & 52.59 & 0 \\
C$7$: Economic Development & 66.68 & 33.32 & 0 \\
C$8$: Peace and War & 48.95 & 51.05 & 0 \\ \hline
\textbf{Total} & \textbf{55.08} & {\color[HTML]{FE0000} \textbf{44.65}} & \textbf{0.27} \\ \hline
\end{tabular}
    \caption{Above table provides the percentage of the answers that are correctly, incorrectly, and not answered by \texttt{llama-3-70b-8192} on \textit{WB}.}
    \label{tab:result_llama-370b_WB_}
    \end{table*}

  \begin{table*}[]
    \centering
    \begin{tabular}{p{4.5cm}ccc}
    \hline
\textbf{Category} & \multicolumn{1}{l}{\textbf{Correctly Answered}} & \multicolumn{1}{l}{\textbf{Incorrectly Answered}} & \multicolumn{1}{l}{\textbf{Not Answered}} \\ \hline
C$1$: Climate & 73.69 & 26.31 & 0 \\
C$2$: Food and Agriculture & 20.2 & 79.8 & 0 \\
C$3$: Health & 38.83 & 61.17 & 0 \\
C$4$: Human Rights & 24.69 & 75.31 & 0 \\
C$5$: Innovation & 40.58 & 59.42 & 0 \\
C$6$: Migration & 29.57 & 70.43 & 0 \\
C$7$: Economic Development & 30.08 & 69.92 & 0 \\
C$8$: Peace and War & 38.37 & 61.63 & 0 \\ \hline
\textbf{Total} & \textbf{37.24} & {\color[HTML]{FE0000} \textbf{62.76}} & \textbf{0} \\ \hline
\end{tabular}
    \caption{Above table provides the percentage of the answers that are correctly, incorrectly, and not answered by \texttt{llama-3-70b-8192} on \textit{MM}.}
    \label{tab:result_llama-370b_MM_}
    \end{table*}

  \begin{table*}[]
    \centering
    \begin{tabular}{p{4.5cm}ccc}
    \hline
\textbf{Category} & \multicolumn{1}{l}{\textbf{Correctly Answered}} & \multicolumn{1}{l}{\textbf{Incorrectly Answered}} & \multicolumn{1}{l}{\textbf{Not Answered}} \\ \hline
C$1$: Climate & 72.91 & 27.09 & 0 \\
C$2$: Food and Agriculture & 19.84 & 80.16 & 0 \\
C$3$: Health & 39.61 & 60.39 & 0 \\
C$4$: Human Rights & 25.32 & 74.68 & 0 \\
C$5$: Innovation & 32.4 & 67.6 & 0 \\
C$6$: Migration & 33.57 & 66.43 & 0 \\
C$7$: Economic Development & 28.72 & 71.28 & 0 \\
C$8$: Peace and War & 42.01 & 57.99 & 0 \\ \hline
\textbf{Total} & \textbf{36.63} & {\color[HTML]{FE0000} \textbf{63.37}} & \textbf{0} \\ \hline
\end{tabular}
    \caption{Above table provides the percentage of the answers that are correctly, incorrectly, and not answered by \texttt{llama-3-70b-8192} on \textit{RB}.}
    \label{tab:result_llama-370b_RB_}
    \end{table*}

  \begin{table*}[]
    \centering
    \begin{tabular}{p{4.5cm}ccc}
    \hline
\textbf{Category} & \multicolumn{1}{l}{\textbf{Correctly Answered}} & \multicolumn{1}{l}{\textbf{Incorrectly Answered}} & \multicolumn{1}{l}{\textbf{Not Answered}} \\ \hline
C$1$: Climate & 6.11 & 93.89 & 0 \\
C$2$: Food and Agriculture & 0.12 & 99.88 & 0 \\
C$3$: Health & 0.23 & 99.77 & 0 \\
C$4$: Human Rights & 0 & 100 & 0 \\
C$5$: Innovation & 0.37 & 99.63 & 0 \\
C$6$: Migration & 0.55 & 99.45 & 0 \\
C$7$: Economic Development & 0.16 & 99.84 & 0 \\
C$8$: Peace and War & 1.08 & 98.92 & 0 \\ \hline
\textbf{Total} & \textbf{1.11} & {\color[HTML]{FE0000} \textbf{98.89}} & \textbf{0} \\ \hline
\end{tabular}
    \caption{Above table provides the percentage of the answers that are correctly, incorrectly, and not answered by \texttt{llama-3-70b-8192} on \textit{TB}.}
    \label{tab:result_llama-370b_TB_}
    \end{table*}


    \begin{table*}[]
    \centering
    \begin{tabular}{p{4.5cm}ccc}
    \hline
\textbf{Category} & \multicolumn{1}{l}{\textbf{Correctly Answered}} & \multicolumn{1}{l}{\textbf{Incorrectly Answered}} & \multicolumn{1}{l}{\textbf{Not Answered}} \\ \hline
C$1$: Climate & 70.81 & 28.12 & 1.07 \\
C$2$: Food and Agriculture & 28.21 & 71.31 & 0.48 \\
C$3$: Health & 28.8 & 70.92 & 0.27 \\
C$4$: Human Rights & 24.9 & 71.3 & 3.8 \\
C$5$: Innovation & 25.34 & 74.47 & 0.19 \\
C$6$: Migration & 7.56 & 92.17 & 0.27 \\
C$7$: Economic Development & 8.63 & 86.38 & 4.99 \\
C$8$: Peace and War & 12.96 & 86.46 & 0.58 \\ \hline
\textbf{Total} & \textbf{27.02} & {\color[HTML]{FE0000} \textbf{71.5}} & \textbf{1.48} \\ \hline
    \end{tabular}
    \caption{Above table provides the percentage of the answers that are correctly, incorrectly, and not answered by \texttt{gpt-3.5} on \textit{DB}.}
    \label{tab:result_gpt-35_DB_}
    \end{table*}

    \begin{table*}[]
    \centering
    \begin{tabular}{p{4.5cm}ccc}
    \hline
\textbf{Category} & \multicolumn{1}{l}{\textbf{Correctly Answered}} & \multicolumn{1}{l}{\textbf{Incorrectly Answered}} & \multicolumn{1}{l}{\textbf{Not Answered}} \\ \hline
C$1$: Climate & 47.09 & 52.91 & 0 \\
C$2$: Food and Agriculture & 52.43 & 47.57 & 0 \\
C$3$: Health & 11.98 & 88.02 & 0 \\
C$4$: Human Rights & 40.43 & 59.57 & 0 \\
C$5$: Innovation & 30.85 & 69.15 & 0 \\
C$6$: Migration & 66.27 & 33.73 & 0 \\
C$7$: Economic Development & 15.2 & 84.8 & 0 \\
C$8$: Peace and War & 52.25 & 47.75 & 0 \\ \hline
\textbf{Total} & \textbf{39.2} & {\color[HTML]{FE0000} \textbf{60.8}} & \textbf{0} \\ \hline
    \end{tabular}
    \caption{Above table provides the percentage of the answers that are correctly, incorrectly, and not answered by \texttt{gpt-3.5} on \textit{CP}.}
    \label{tab:result_gpt-35_CP_}
    \end{table*}

    \begin{table*}[]
    \centering
    %
    \caption{Above table provides the percentage of the answers that are correctly, incorrectly, and not answered by \texttt{gpt-3.5} on \textit{WB}.}
    \label{tab:result_gpt-35_WB_}
    \end{table*}

    \begin{table*}[]
    \centering
    %
    \caption{Above table provides the percentage of the answers that are correctly, incorrectly, and not answered by \texttt{gpt-3.5} on \textit{MM}.}
    \label{tab:result_gpt-35_MM_}
    \end{table*}

    \begin{table*}[]
    \centering
    %
    \caption{Above table provides the percentage of the answers that are correctly, incorrectly, and not answered by \texttt{gpt-3.5} on \textit{RB}.}
    \label{tab:result_gpt-35_RB_}
    \end{table*}

    \begin{table*}[]
    \centering
    %
    \caption{Above table provides the percentage of the answers that are correctly, incorrectly, and not answered by \texttt{gpt-3.5} on \textit{TB}.}
    \label{tab:result_gpt-35_TB_}
    \end{table*}


    \begin{table*}[]
    \centering
    %
    \caption{Above table provides the percentage of the answers that are correctly, incorrectly, and not answered by \texttt{gpt-4} on \textit{DB}.}
    \label{tab:result_gpt-4_DB_}
    \end{table*}

    \begin{table*}[]
    \centering
    %
    \caption{Above table provides the percentage of the answers that are correctly, incorrectly, and not answered by \texttt{gpt-4} on \textit{CP}.}
    \label{tab:result_gpt-4_RB_}
    \end{table*}

    \begin{table*}[]
    \centering
    %
    \caption{Above table provides the percentage of the answers that are correctly, incorrectly, and not answered by \texttt{gpt-4} on \textit{WB}.}
    \label{tab:result_gpt-4_WB_}
    \end{table*}

    \begin{table*}[]
    \centering
    %
    \caption{Above table provides the percentage of the answers that are correctly, incorrectly, and not answered by \texttt{gpt-4} on \textit{MM}.}
    \label{tab:result_gpt-4_MMB_}
    \end{table*}

    \begin{table*}[]
    \centering
    %
    \caption{Above table provides the percentage of the answers that are correctly, incorrectly, and not answered by \texttt{gpt-4} on \textit{RB}.}
    \label{tab:result_gpt-4_RAB_}
    \end{table*}

    \begin{table*}[]
    \centering
    %
    \caption{Above table provides the percentage of the answers that are correctly, incorrectly, and not answered by \texttt{gpt-4} on \textit{TB}.}
    \label{tab:result_gpt-4_TB_}
    \end{table*}


    \begin{table*}[]
    \centering
    %
    \caption{Above table provides the percentage of the answers that are correctly, incorrectly, and not answered by \texttt{gemini-pro} on \textit{DB}.}
    \label{tab:result_gemini_DB_}
    \end{table*}

    \begin{table*}[]
    \centering
    %
    \caption{Above table provides the percentage of the answers that are correctly, incorrectly, and not answered by \texttt{gemini-pro} on \textit{CP}.}
    \label{tab:result_gemini_RB_}
    \end{table*}

    \begin{table*}[]
    \centering
    %
    \caption{Above table provides the percentage of the answers that are correctly, incorrectly, and not answered by \texttt{gemini-pro} on \textit{WB}.}
    \label{tab:result_gemini_WB_}
    \end{table*}

    \begin{table*}[]
    \centering
    %
    \caption{Above table provides the percentage of the answers that are correctly, incorrectly, and not answered by \texttt{gemini-pro} on \textit{TB}.}
    \label{tab:result_gemini_TB_}
    \end{table*}



\begin{figure*}
\begin{center}
\includegraphics[width=0.9\linewidth]{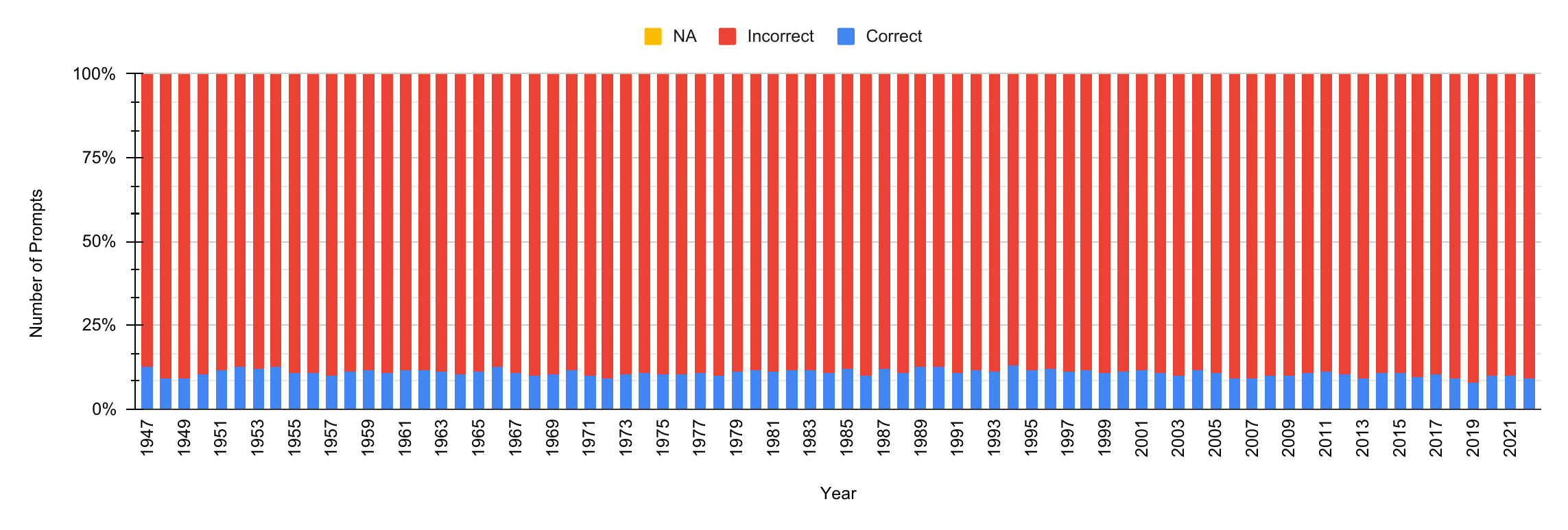}
\end{center}
\caption{Plot for the Date-based metric ($DB$) for year-wise count for \texttt{phi-2} in \textbf{Zeroshot evaluation}. }
\label{fig:date-based-phi2}
\end{figure*}

\begin{figure*}
\begin{center}
\includegraphics[width=0.9\linewidth]{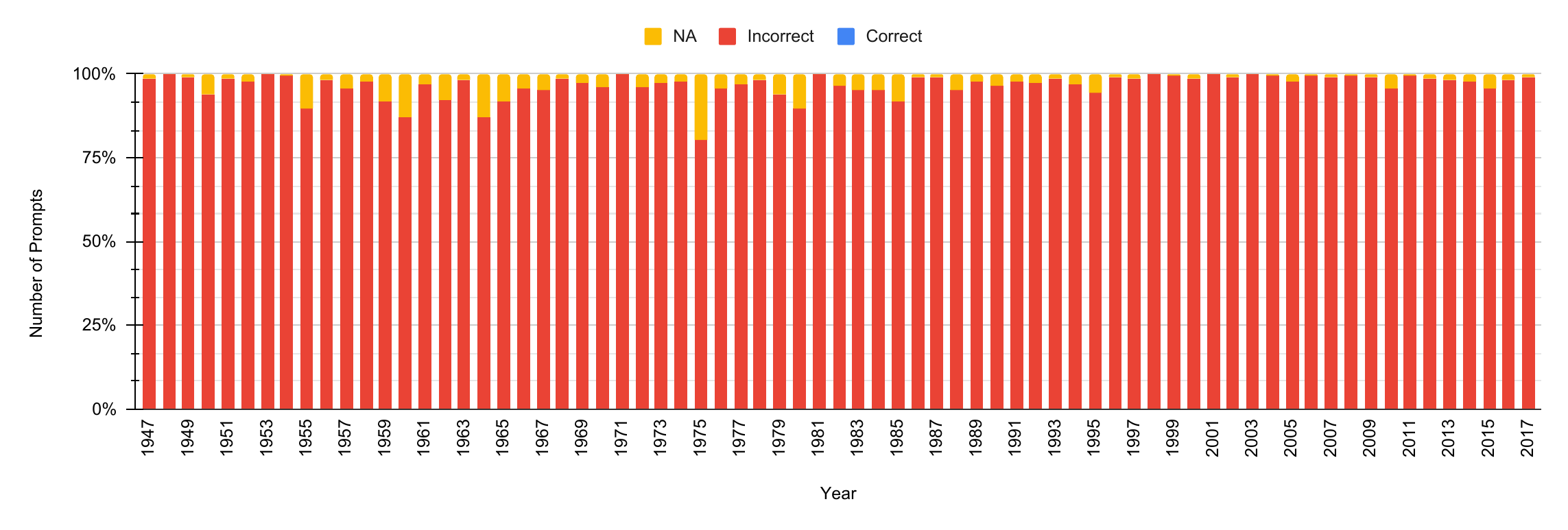}
\end{center}
\caption{Plot for the Comparative-based metric ($CP$) for year-wise count for \texttt{phi-2} in \textbf{Zeroshot evaluation}. }
\label{fig:range-based-phi2}
\end{figure*}

\begin{figure*}
\begin{center}
\includegraphics[width=0.9\linewidth]{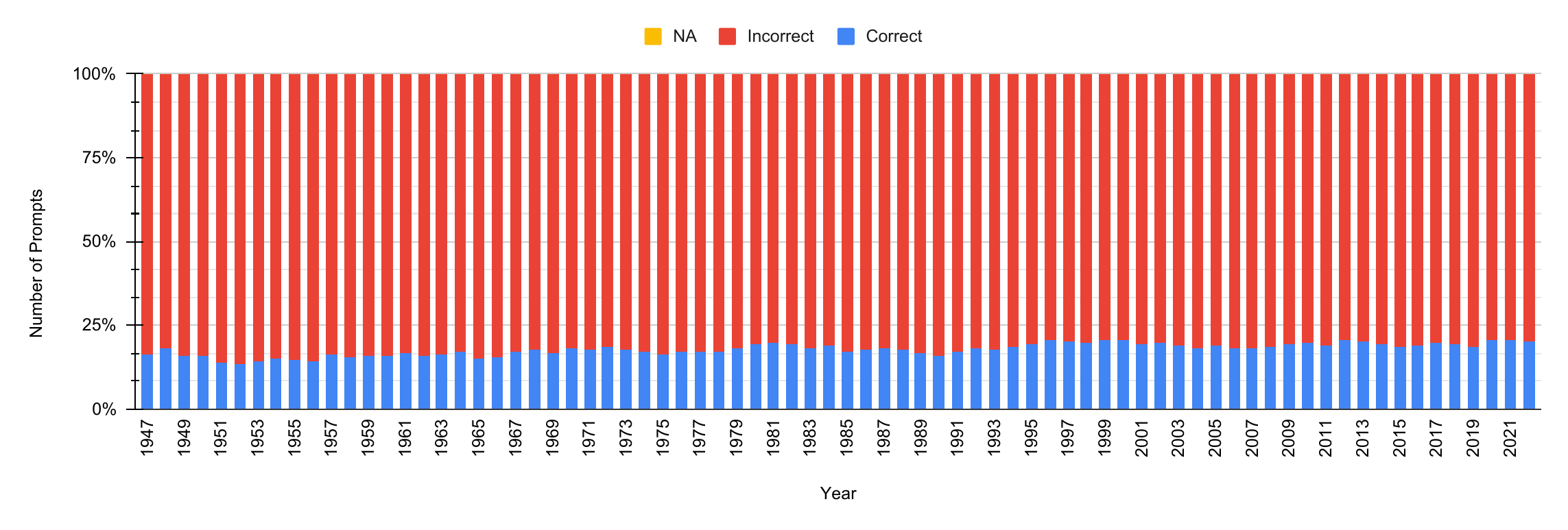}
\end{center}
\caption{Plot for the Window-based metric ($WB$) for year-wise count for \texttt{phi-2} in \textbf{Zeroshot evaluation}. }
\label{fig:window-based-phi2}
\end{figure*}

\begin{figure*}
\begin{center}
\includegraphics[width=0.9\linewidth]{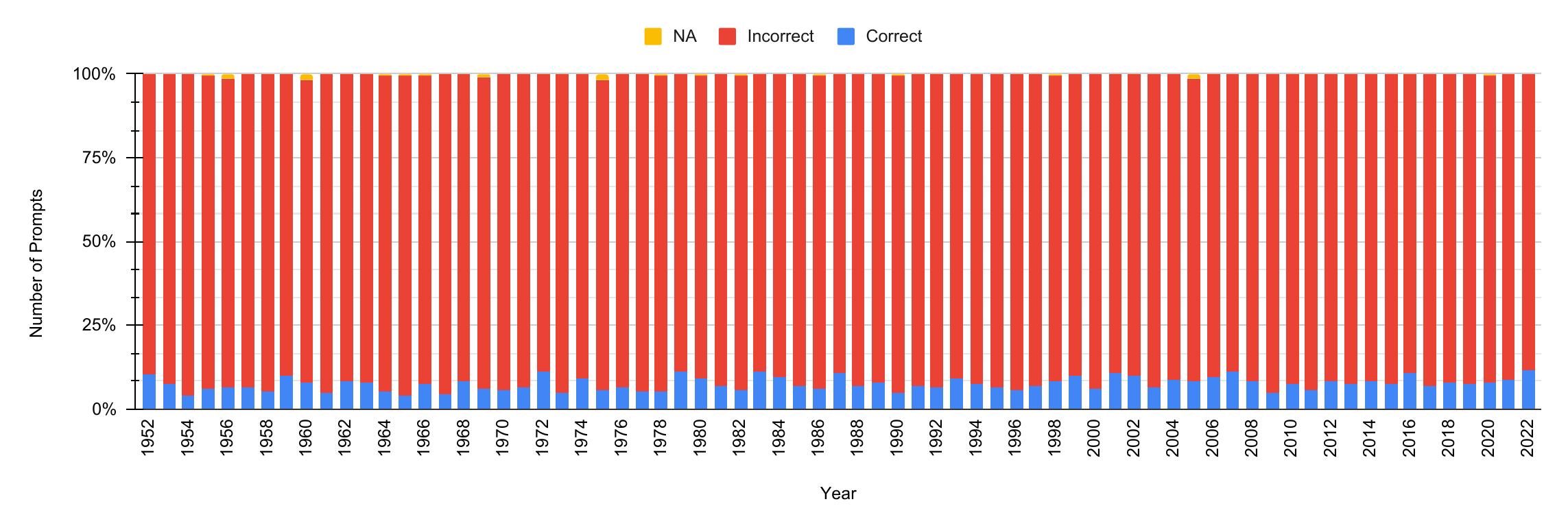}
\end{center}
\caption{Plot for the Min/Max-based metric ($MM$) for year-wise count for \texttt{phi-2} in \textbf{Zeroshot evaluation}. }
\label{fig:mmb-based-phi2}
\end{figure*}

\begin{figure*}
\begin{center}
\includegraphics[width=0.9\linewidth]{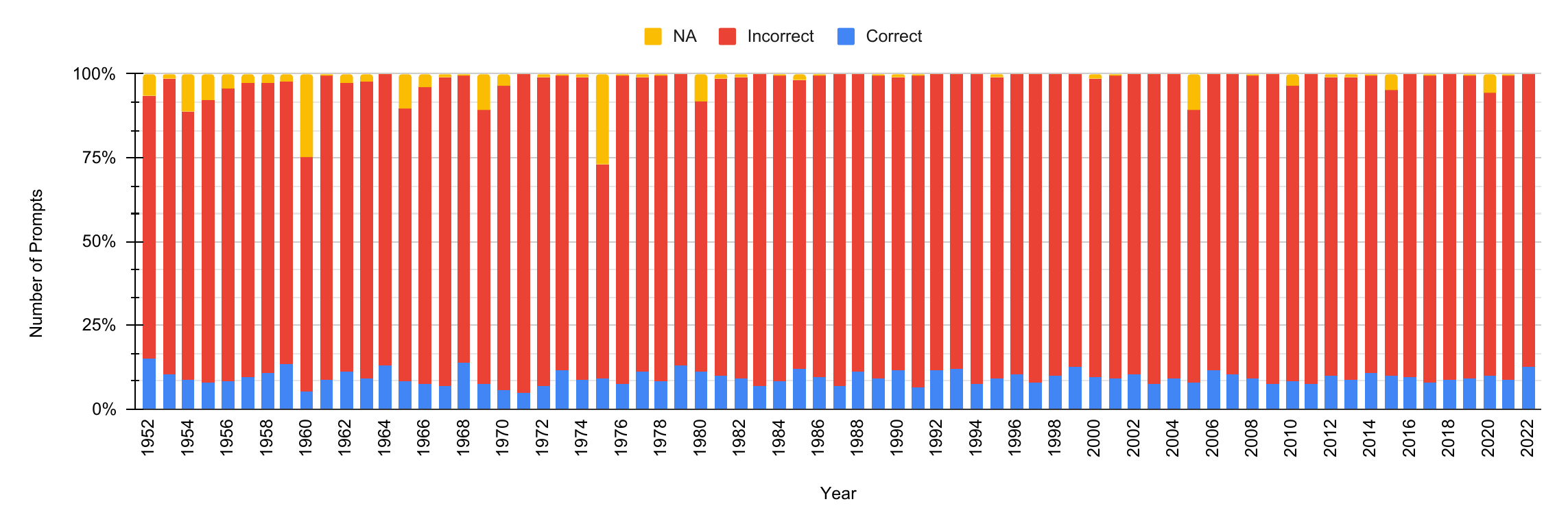}
\end{center}
\caption{Plot for the Range-based metric ($RB$) for year-wise count for \texttt{phi-2} in \textbf{Zeroshot evaluation}. }
\label{fig:rab-based-phi2}
\end{figure*}

\begin{figure*}
\begin{center}
\includegraphics[width=0.9\linewidth]{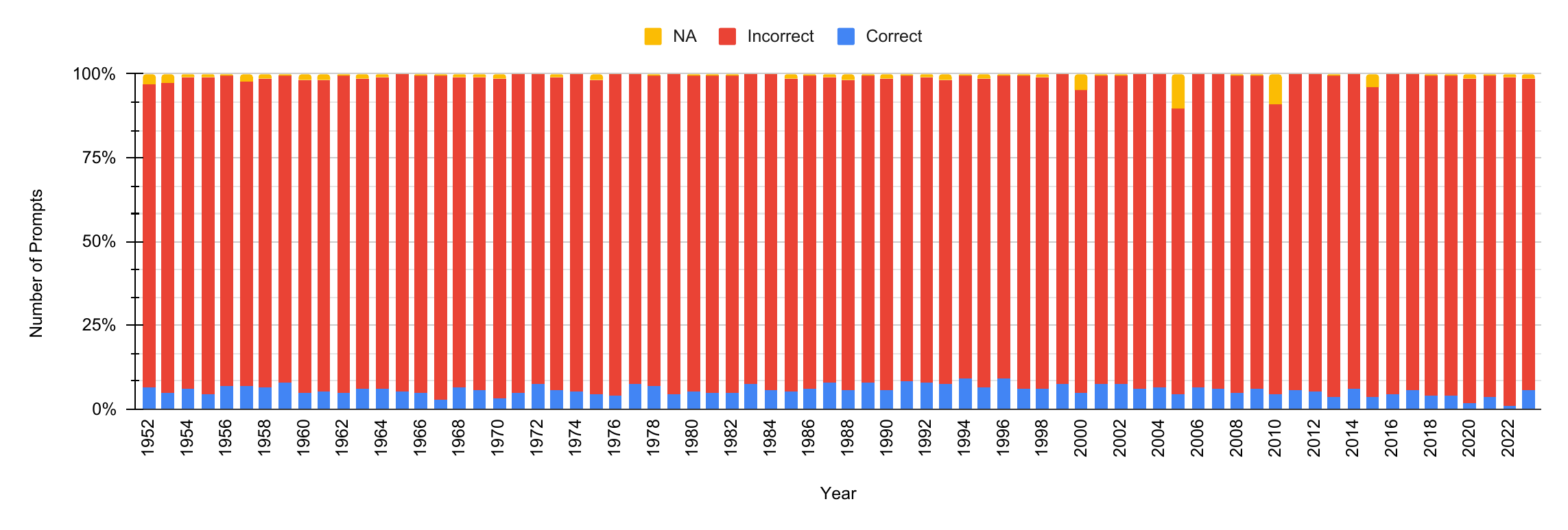}
\end{center}
\caption{Plot for the Trend-based metric ($TB$) for year-wise count for \texttt{phi-2} in \textbf{Zeroshot evaluation}. }
\label{fig:trend-based-phi2}
\end{figure*}


\begin{figure*}
\begin{center}
\includegraphics[width=0.9\linewidth]{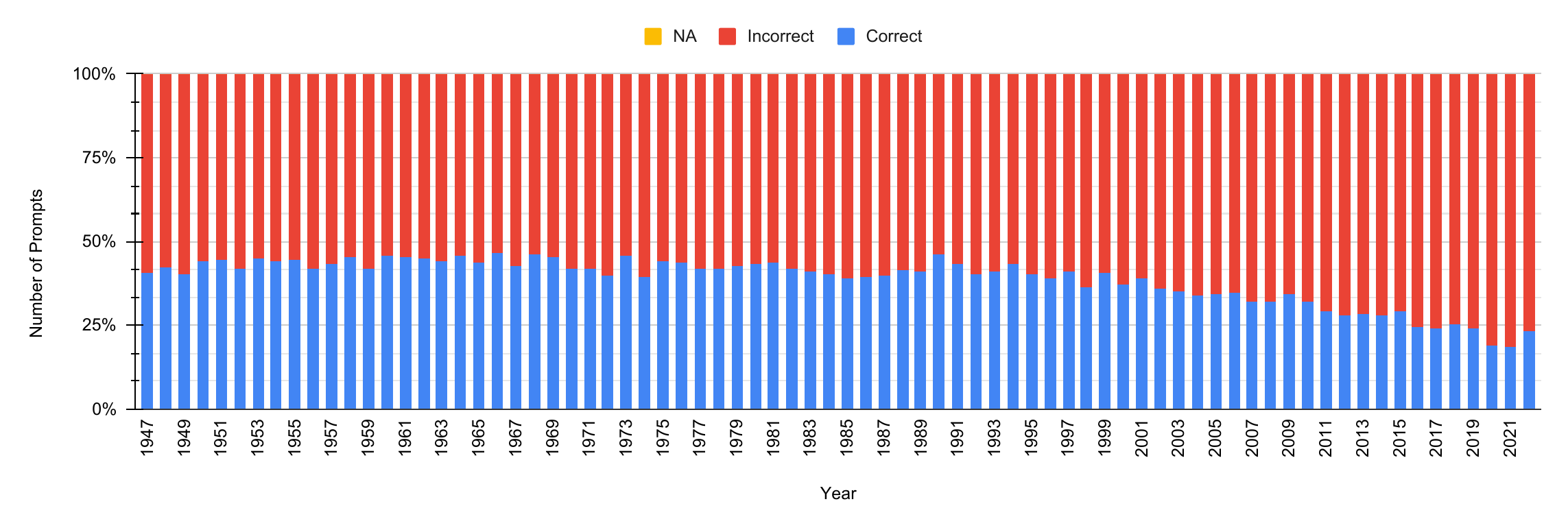}
\end{center}
\caption{Plot for the Date-based metric ($DB$) for year-wise count for \texttt{flan-t5-xl} in \textbf{Zeroshot evaluation}. }
\label{fig:date-based-flan-t5-xl}
\end{figure*}

\begin{figure*}
\begin{center}
\includegraphics[width=0.9\linewidth]{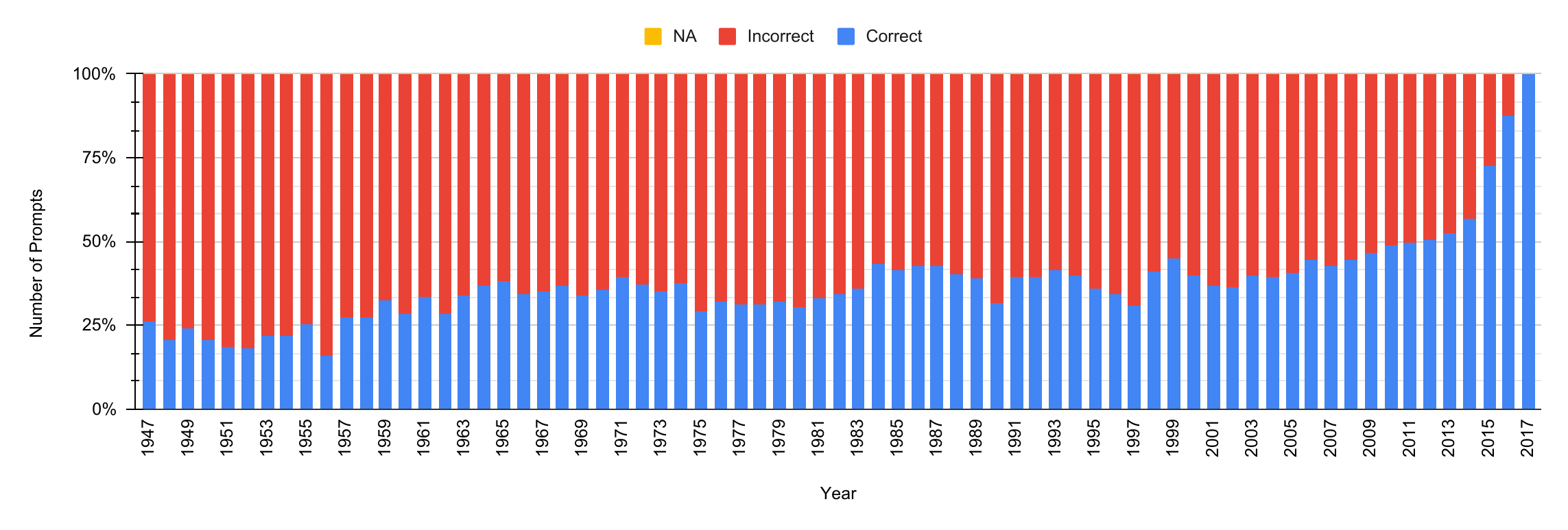}
\end{center}
\caption{Plot for the Comparative-based metric ($CP$) for year-wise count for \texttt{flan-t5-xl} in \textbf{Zeroshot evaluation}. }
\label{fig:range-based-flan-t5-xl}
\end{figure*}

\begin{figure*}
\begin{center}
\includegraphics[width=0.9\linewidth]{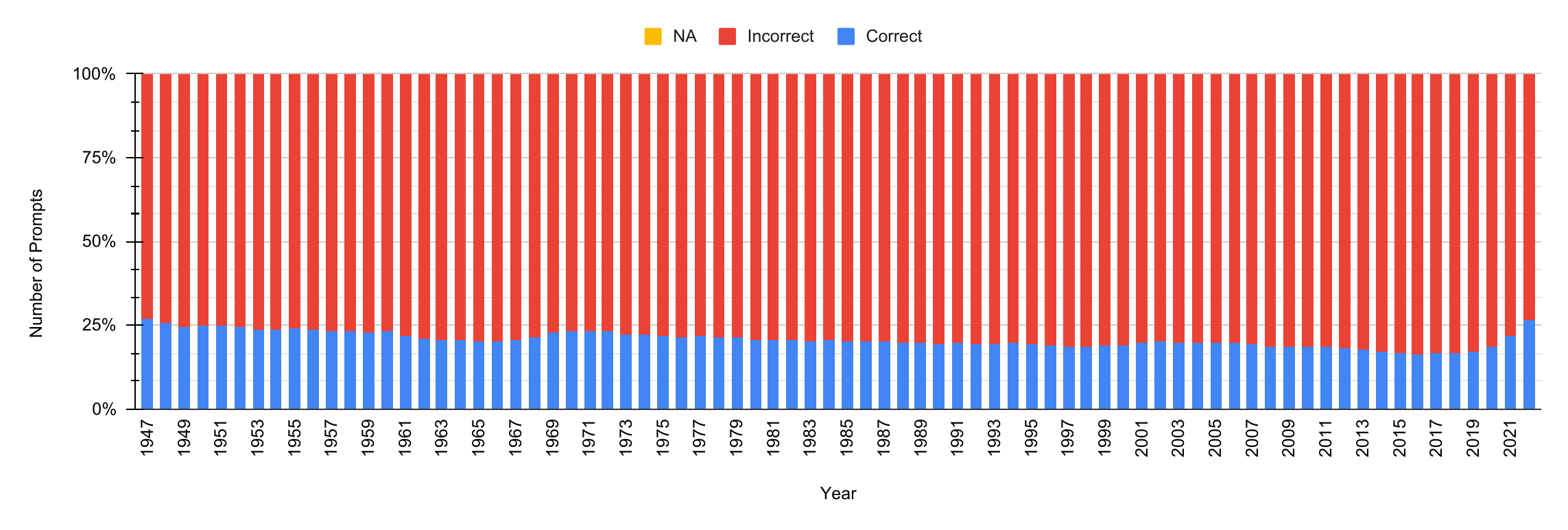}
\end{center}
\caption{Plot for the Window-based metric ($WB$) for year-wise count for \texttt{flan-t5-xl} in \textbf{Zeroshot evaluation}. }
\label{fig:window-based-flan-t5-xl}
\end{figure*}

\begin{figure*}
\begin{center}
\includegraphics[width=0.9\linewidth]{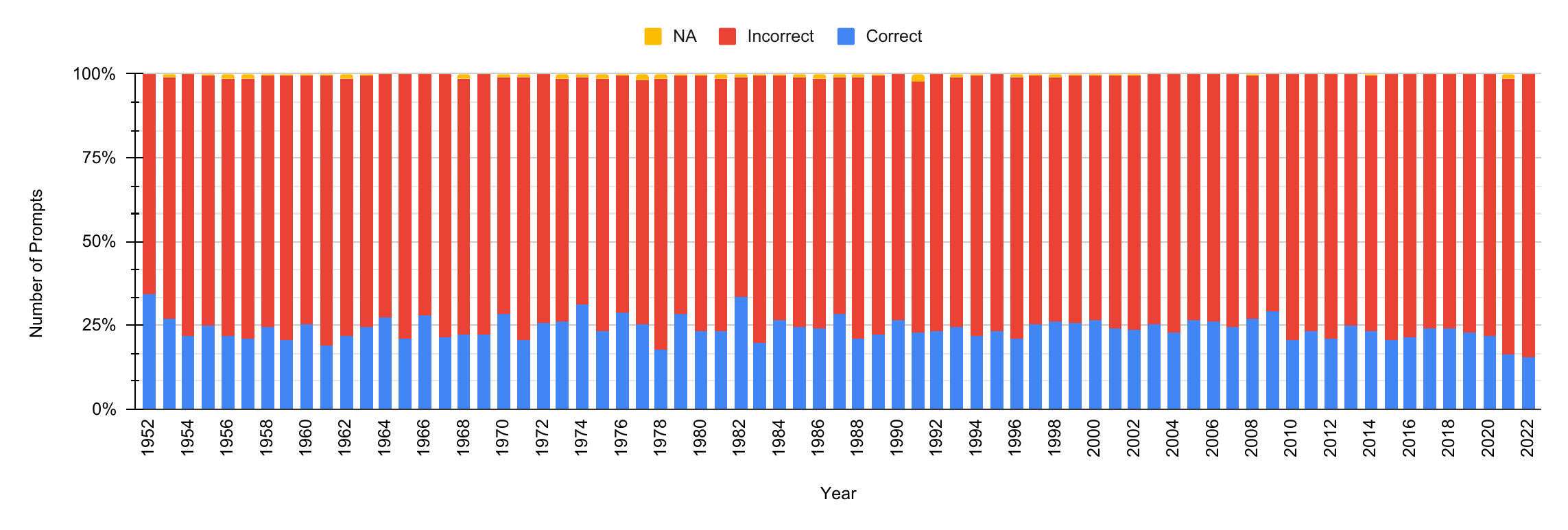}
\end{center}
\caption{Plot for the Min/Max-based metric ($MM$) for year-wise count for \texttt{flan-t5-xl} in \textbf{Zeroshot evaluation}. }
\label{fig:mmb-based-flan-t5-xl}
\end{figure*}

\begin{figure*}
\begin{center}
\includegraphics[width=0.9\linewidth]{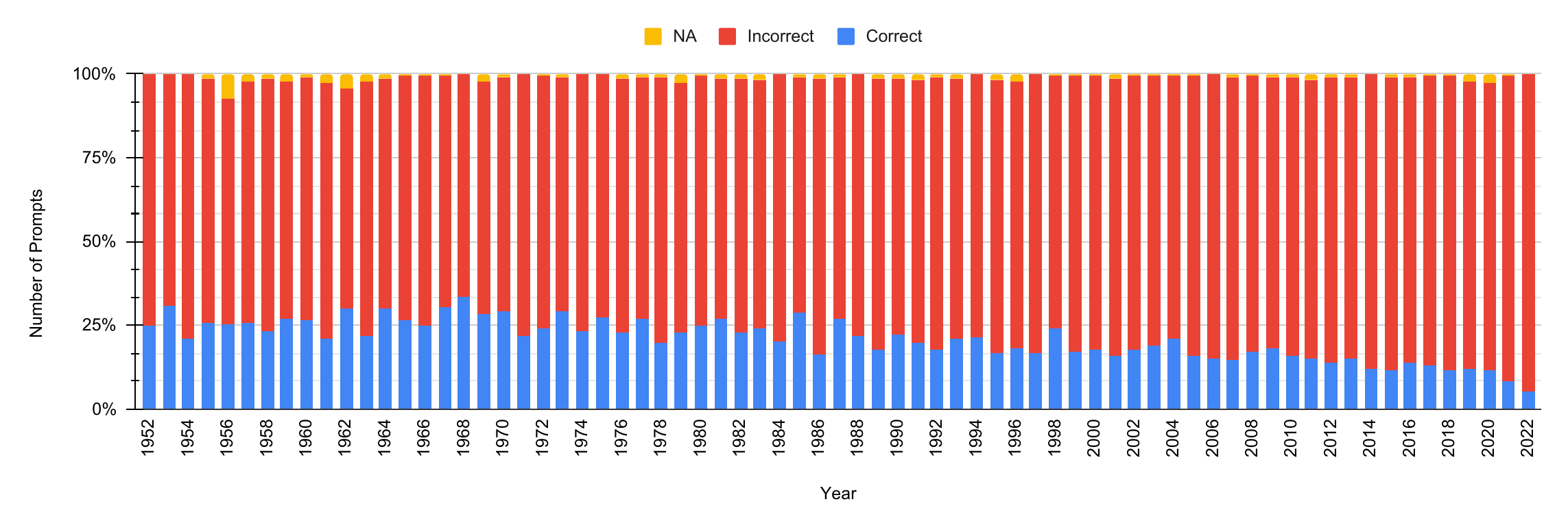}
\end{center}
\caption{Plot for the Range-based metric ($RB$) for year-wise count for \texttt{flan-t5-xl} in \textbf{Zeroshot evaluation}. }
\label{fig:rab-based-flan-t5-xl}
\end{figure*}

\begin{figure*}
\begin{center}
\includegraphics[width=0.9\linewidth]{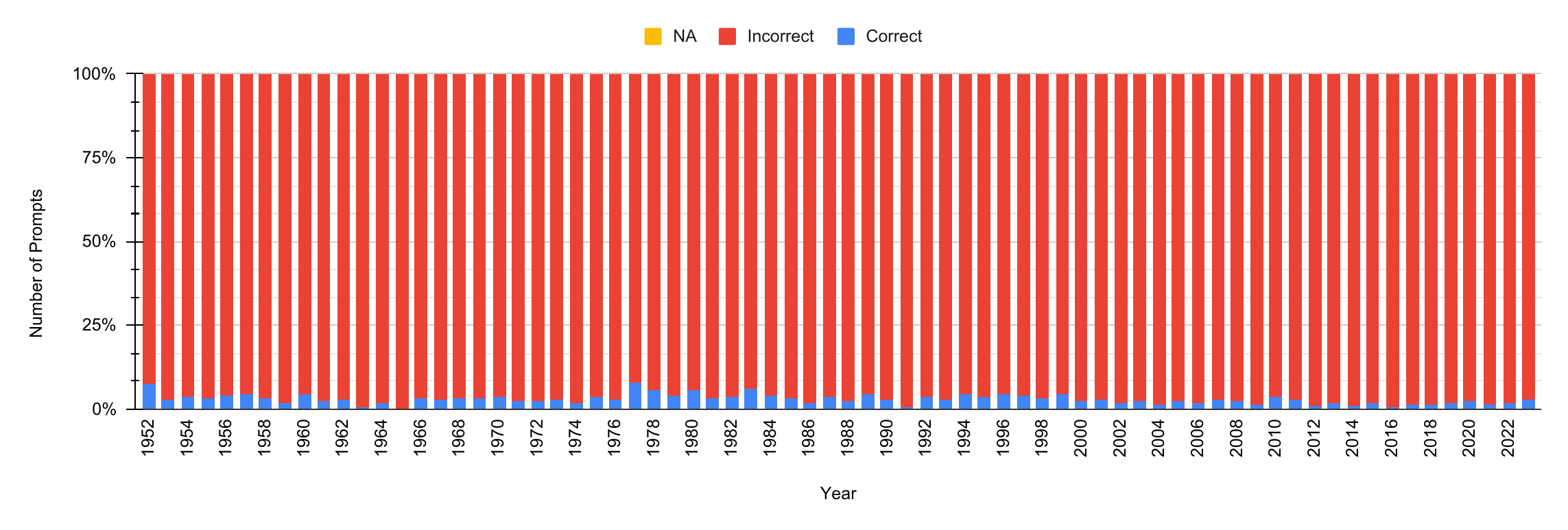}
\end{center}
\caption{Plot for the Trend-based metric ($TB$) for year-wise count for \texttt{flan-t5-xl} in \textbf{Zeroshot evaluation}. }
\label{fig:tb-based-flan-t5-xl}
\end{figure*}


\begin{figure*}
\begin{center}
\includegraphics[width=0.9\linewidth]{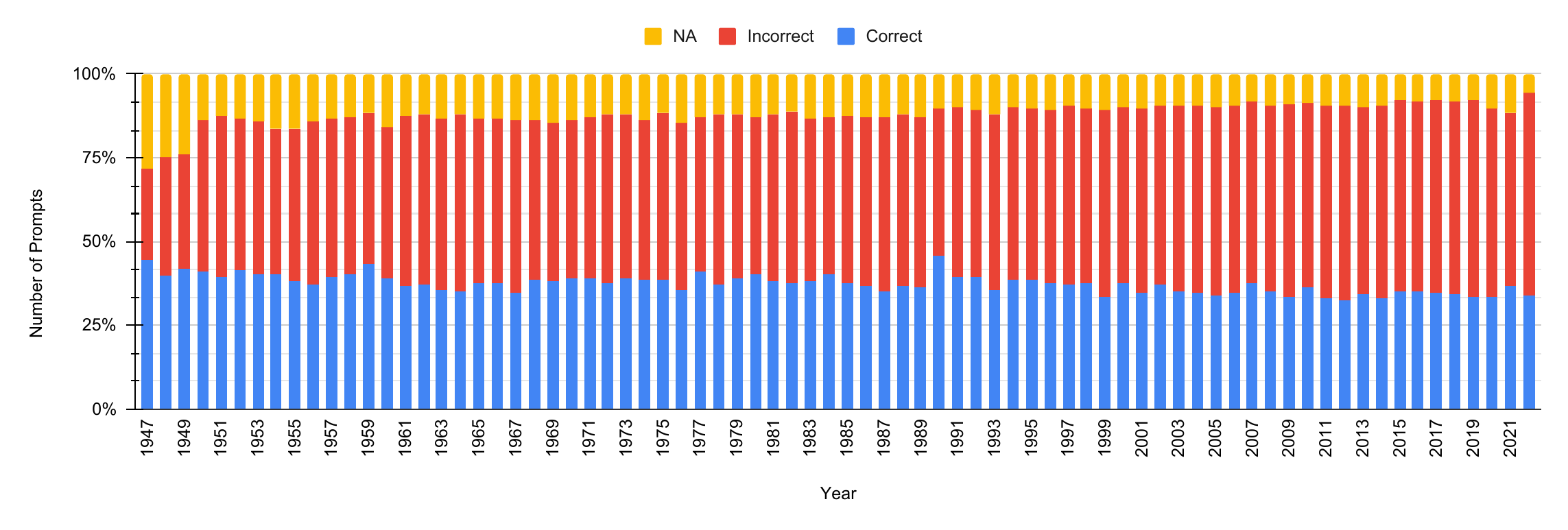}
\end{center}
\caption{Plot for the Date-based metric ($DB$) for year-wise count for \texttt{mistral-instruct} in \textbf{Zeroshot evaluation}. }
\label{fig:date-based-mistral}
\end{figure*}

\begin{figure*}
\begin{center}
\includegraphics[width=0.9\linewidth]{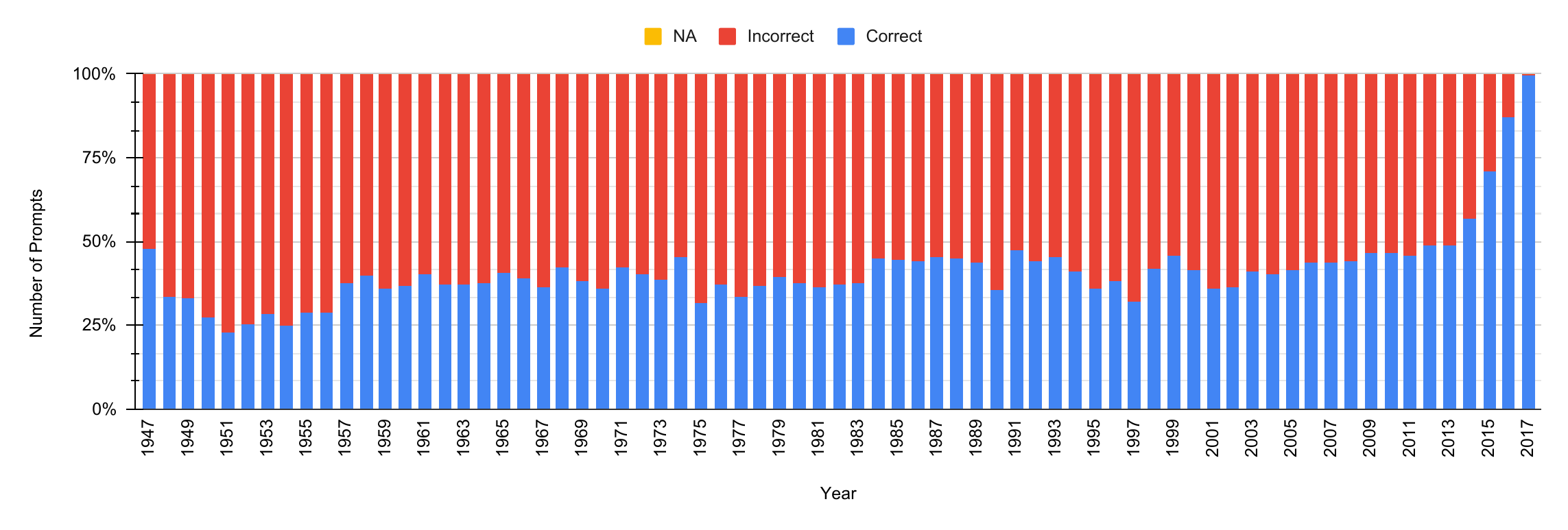}
\end{center}
\caption{Plot for the Comparative-based metric ($CP$) for year-wise count for \texttt{mistral-instruct} in \textbf{Zeroshot evaluation}. }
\label{fig:range-based-mistral}
\end{figure*}

\begin{figure*}
\begin{center}
\includegraphics[width=0.9\linewidth]{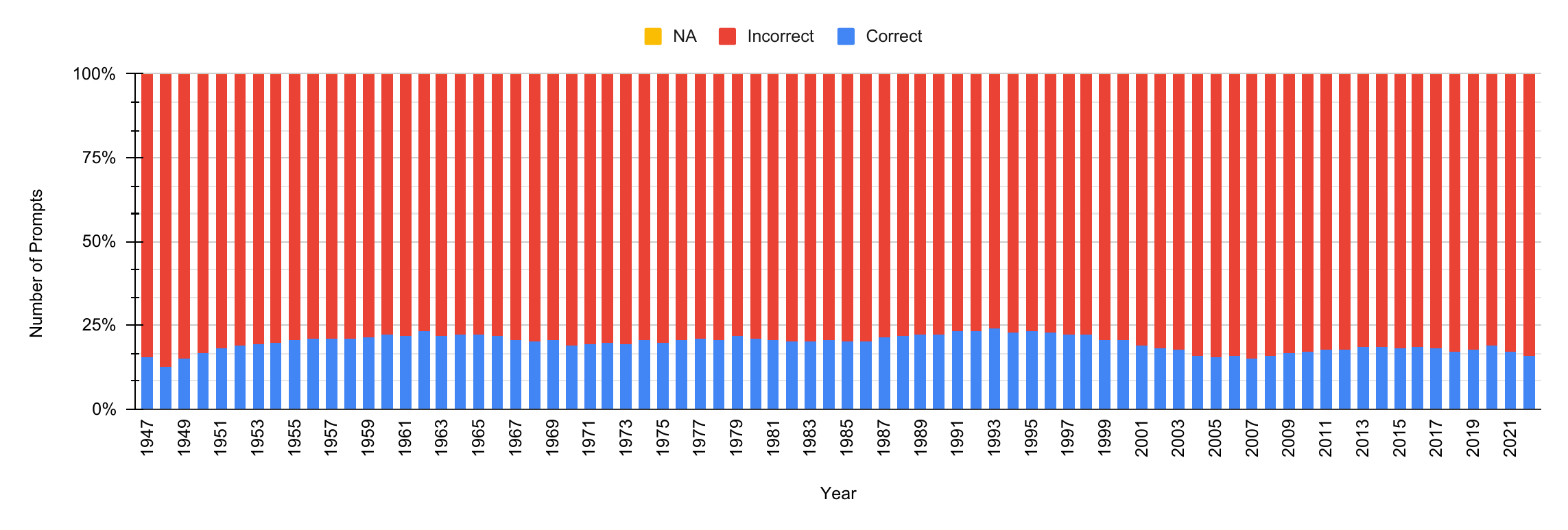}
\end{center}
\caption{Plot for the Window-based metric ($WB$) for year-wise count for \texttt{mistral-instruct} in \textbf{Zeroshot evaluation}. }
\label{fig:window-based-mistral}
\end{figure*}

\begin{figure*}
\begin{center}
\includegraphics[width=0.9\linewidth]{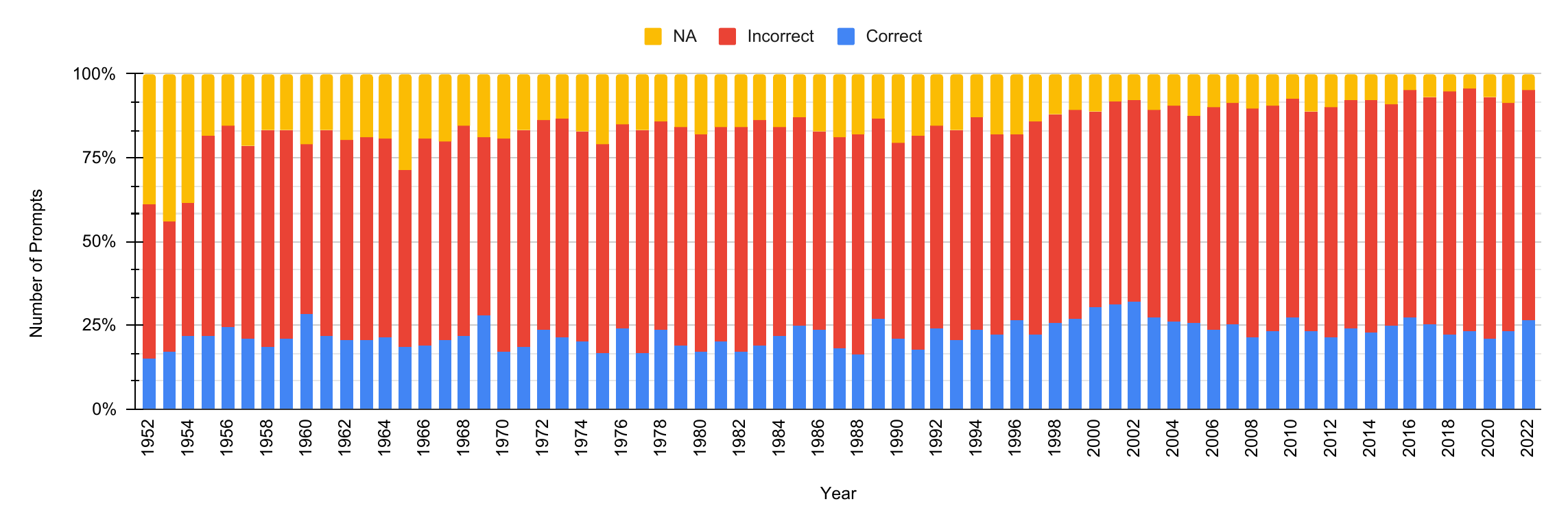}
\end{center}
\caption{Plot for the Min/Max-based metric ($MM$) for year-wise count for \texttt{mistral-instruct} in \textbf{Zeroshot evaluation}. }
\label{fig:mmb-based-mistral}
\end{figure*}

\begin{figure*}
\begin{center}
\includegraphics[width=0.9\linewidth]{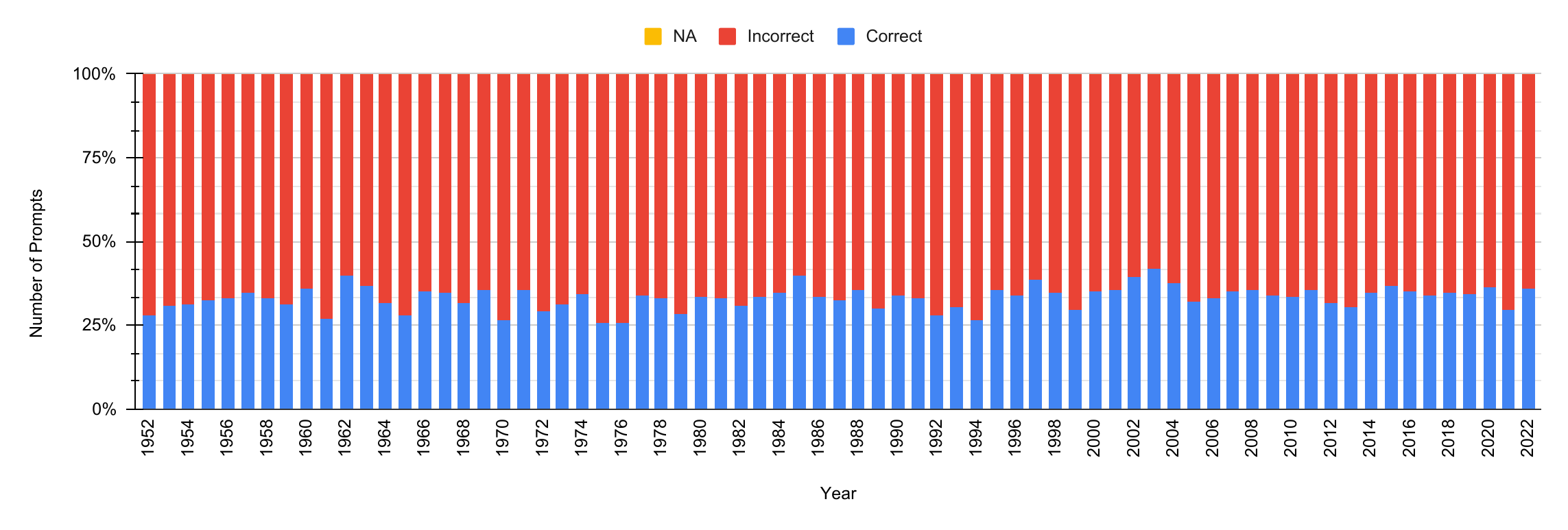}
\end{center}
\caption{Plot for the Range-based metric ($RB$) for year-wise count for \texttt{mistral-instruct} in \textbf{Zeroshot evaluation}. }
\label{fig:rab-based-mistral}
\end{figure*}

\begin{figure*}
\begin{center}
\includegraphics[width=0.9\linewidth]{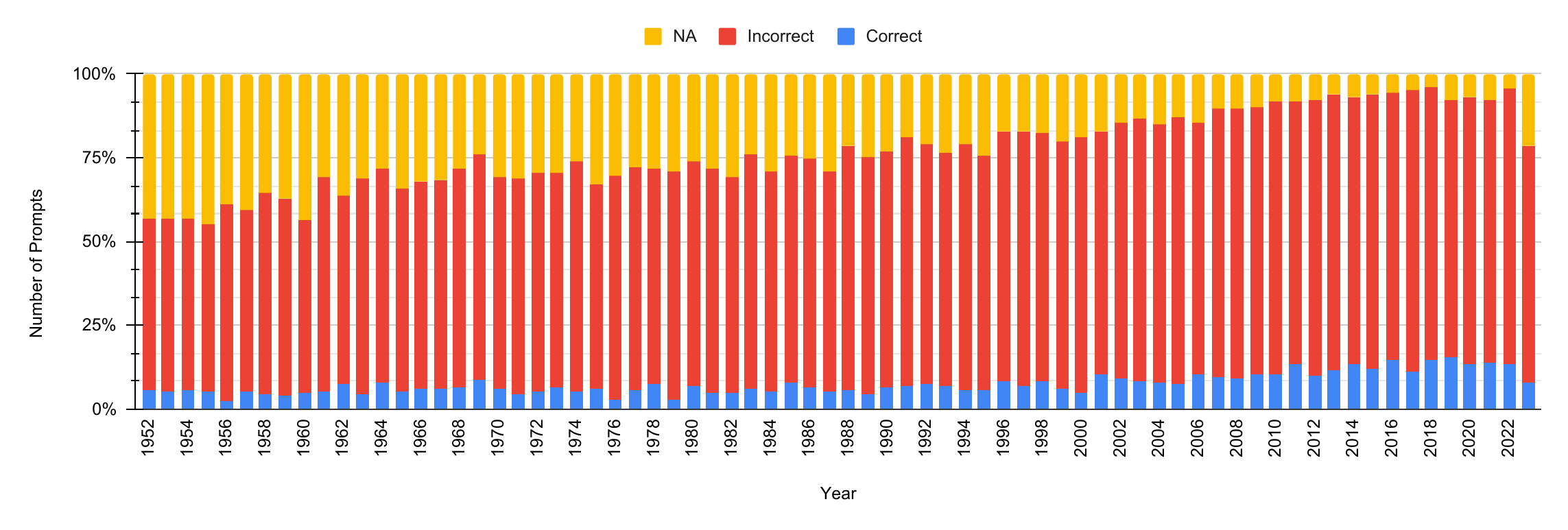}
\end{center}
\caption{Plot for the Trend-based metric ($TB$) for year-wise count for \texttt{mistral-instruct} in \textbf{Zeroshot evaluation}. }
\label{fig:trend-based-mistral}
\end{figure*}

\begin{figure*}
\begin{center}
\includegraphics[width=0.9\linewidth]{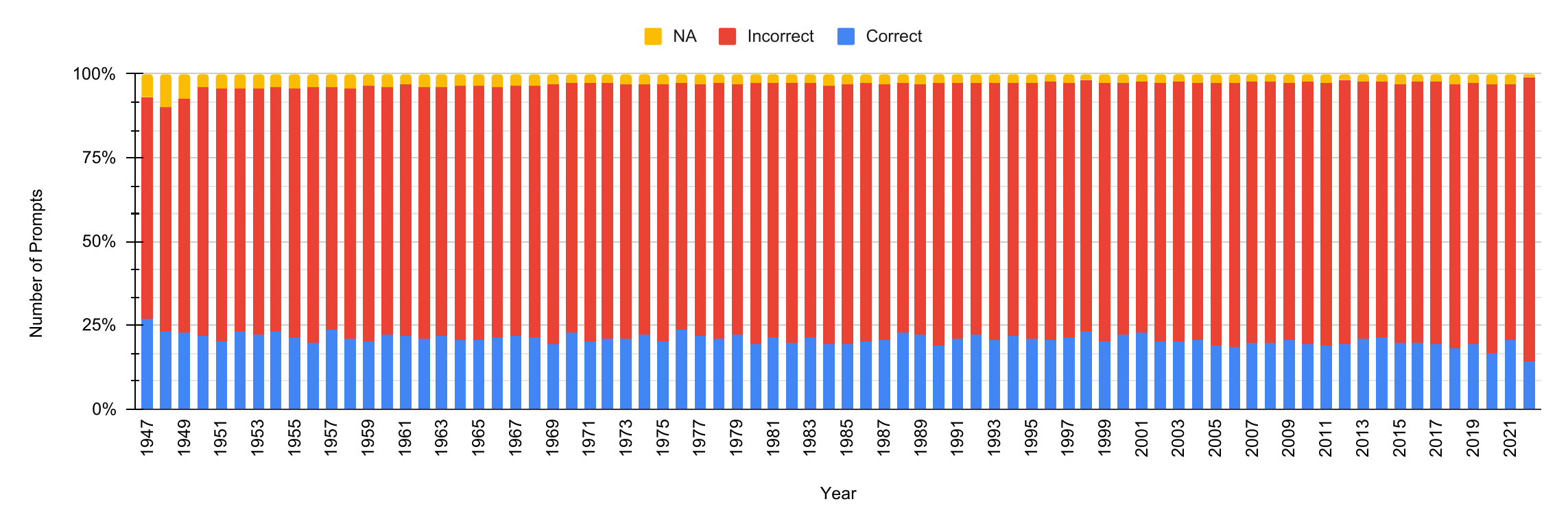}
\end{center}
\caption{Plot for the Date-based metric ($DB$) for year-wise count for \texttt{llama-2-chat} in \textbf{Zeroshot evaluation}. }
\label{fig:date-based-LLaMA}
\end{figure*}

\begin{figure*}
\begin{center}
\includegraphics[width=0.9\linewidth]{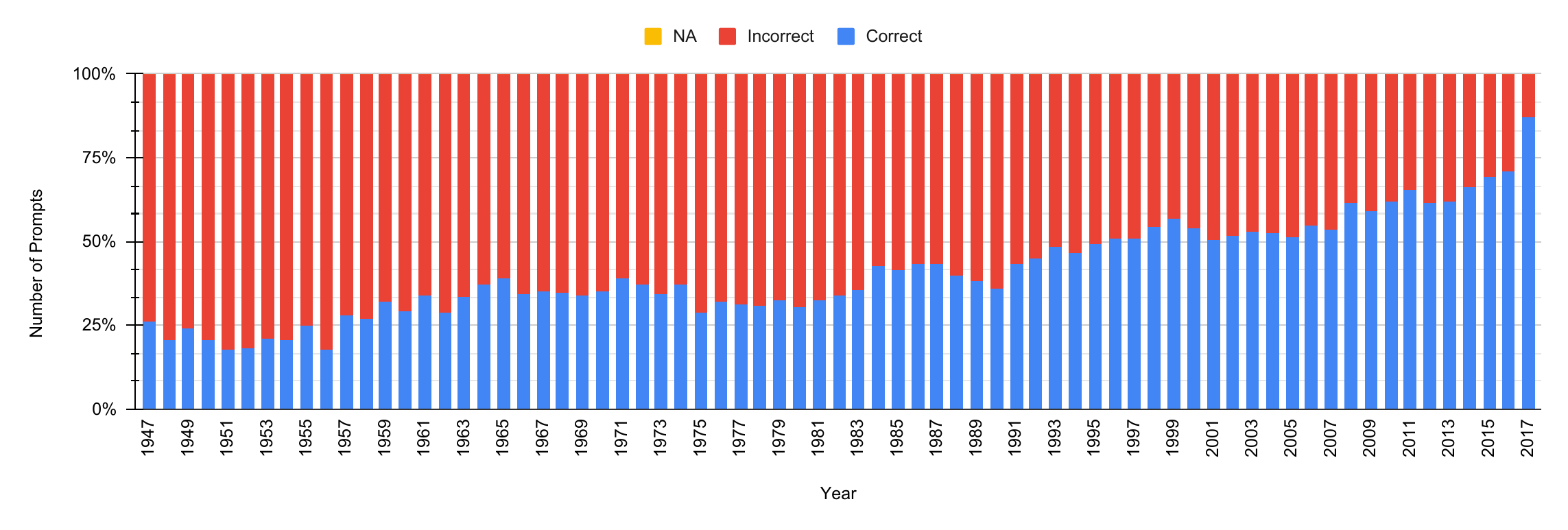}
\end{center}
\caption{Plot for the Comparative-based metric ($CP$) for year-wise count for \texttt{llama-2-chat} in \textbf{Zeroshot evaluation}. }
\label{fig:range-based-LLaMA}
\end{figure*}

\begin{figure*}
\begin{center}
\includegraphics[width=0.9\linewidth]{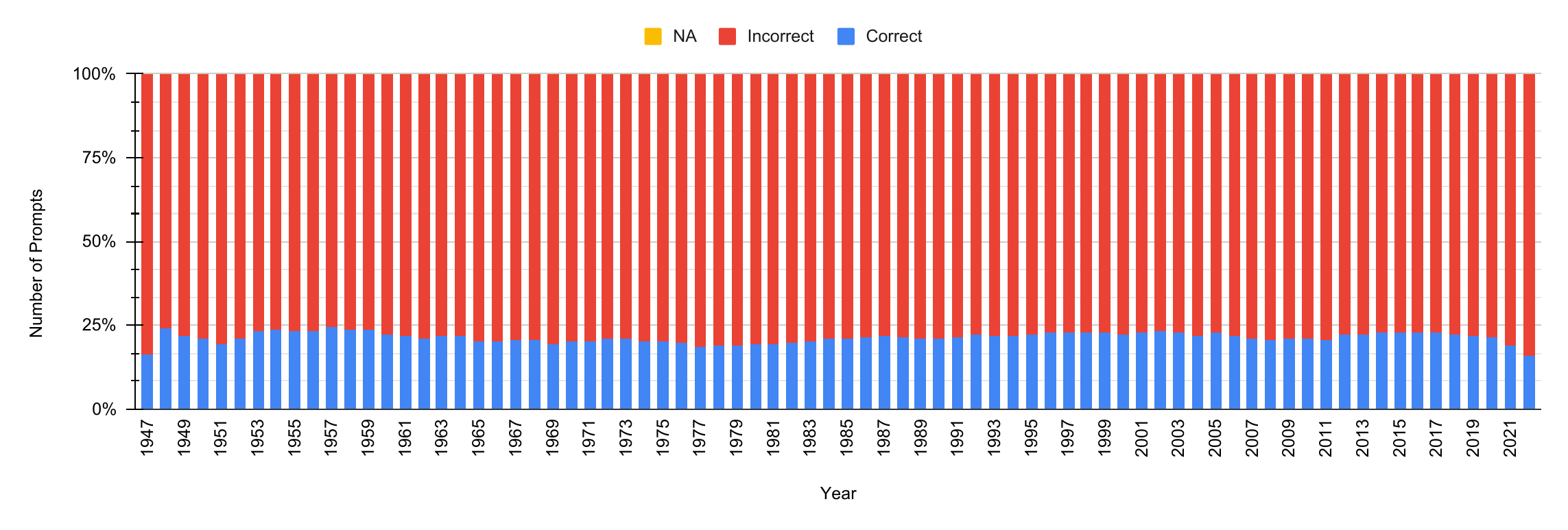}
\end{center}
\caption{Plot for the Window-based metric ($WB$) for year-wise count for \texttt{llama-2-chat} in \textbf{Zeroshot evaluation}. }
\label{fig:window-based-LLaMA}
\end{figure*}

\begin{figure*}
\begin{center}
\includegraphics[width=0.9\linewidth]{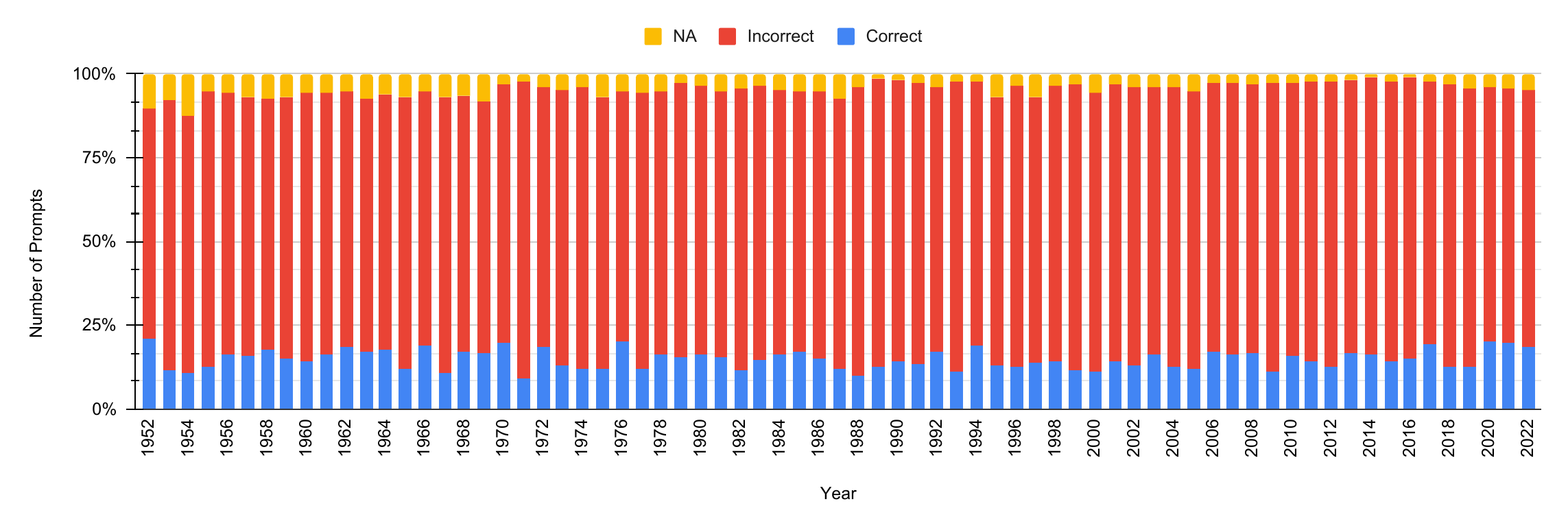}
\end{center}
\caption{Plot for the Min/Max-based metric ($MM$) for year-wise count for \texttt{llama-2-chat} in \textbf{Zeroshot evaluation}. }
\label{fig:mmb-based-LLaMA}
\end{figure*}

\begin{figure*}
\begin{center}
\includegraphics[width=0.9\linewidth]{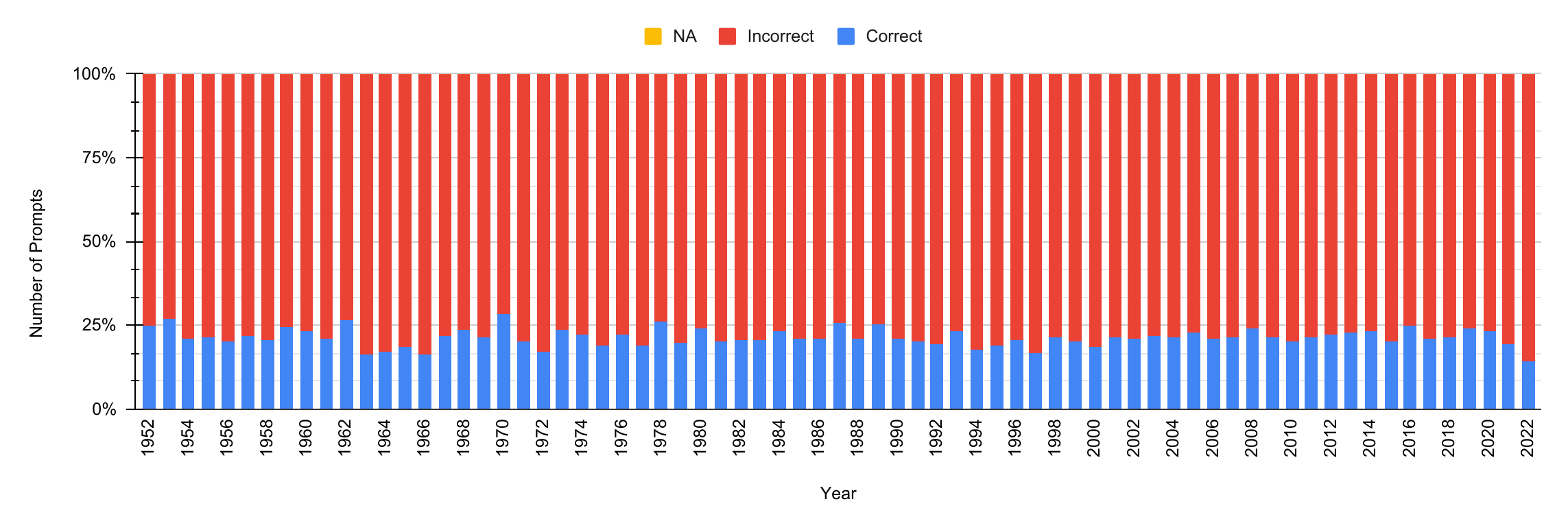}
\end{center}
\caption{Plot for the Range-based metric ($RB$) for year-wise count for \texttt{llama-2-chat} in \textbf{Zeroshot evaluation}. }
\label{fig:rab-based-LLaMA}
\end{figure*}

\begin{figure*}
\begin{center}
\includegraphics[width=0.9\linewidth]{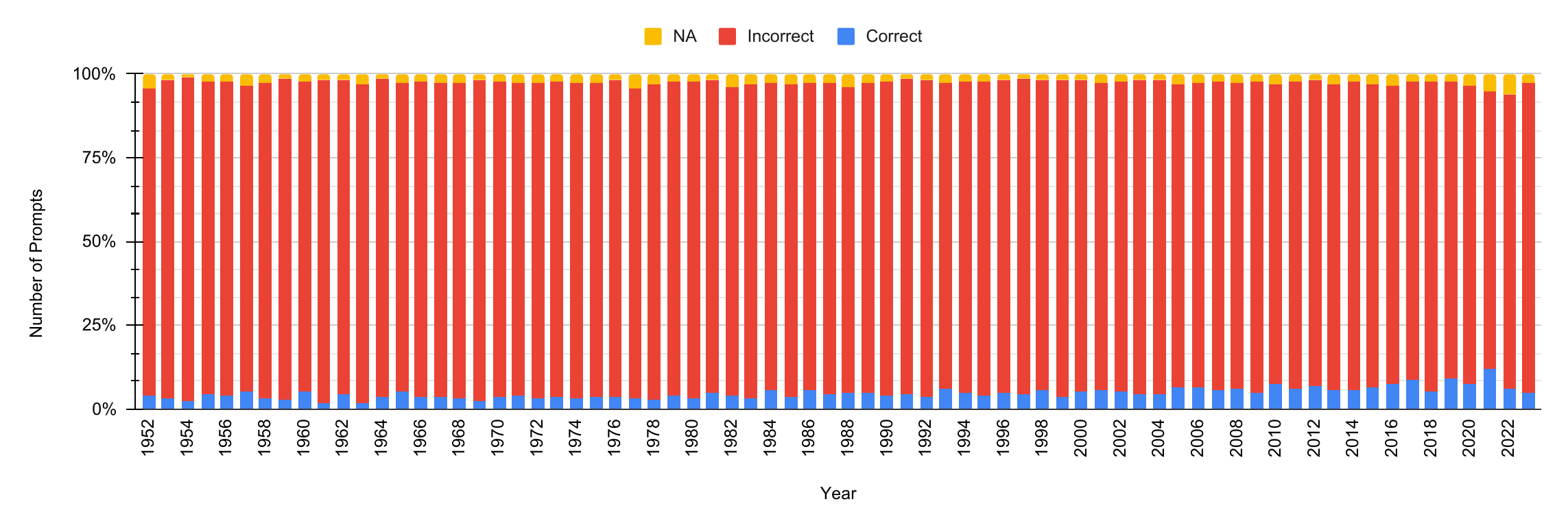}
\end{center}
\caption{Plot for the Trend-based metric ($TB$) for year-wise count for \texttt{llama-2-chat} in \textbf{Zeroshot evaluation}. }
\label{fig:trend-based-LLaMA}
\end{figure*}


\begin{figure*}
\begin{center}
\includegraphics[width=0.9\linewidth]{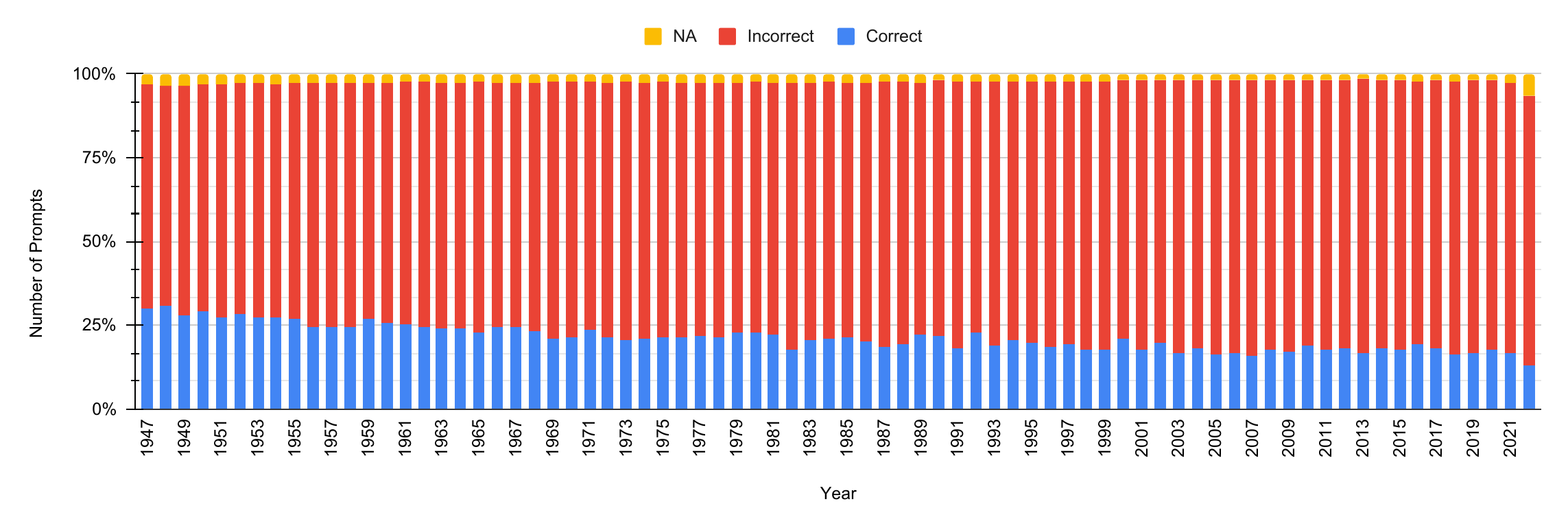}
\end{center}
\caption{Plot for the Date-based metric ($DB$) for year-wise count for \texttt{gemma-7b-it} in \textbf{Zeroshot evaluation}. }
\label{fig:date-based-gemma}
\end{figure*}
\begin{figure*}
\begin{center}
\includegraphics[width=0.9\linewidth]{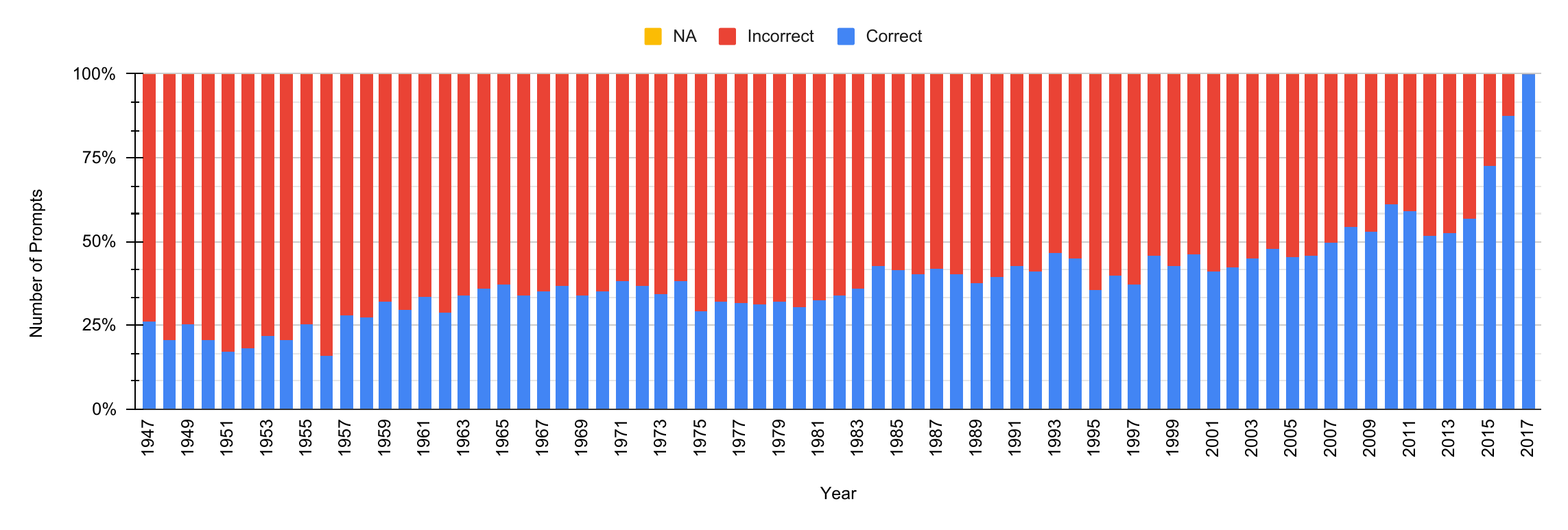}
\end{center}
\caption{Plot for the Comparative-based metric ($CP$) for year-wise count for \texttt{gemma-7b-it} in \textbf{Zeroshot evaluation}. }
\label{fig:range-based-gemma}
\end{figure*}

\begin{figure*}
\begin{center}
\includegraphics[width=0.9\linewidth]{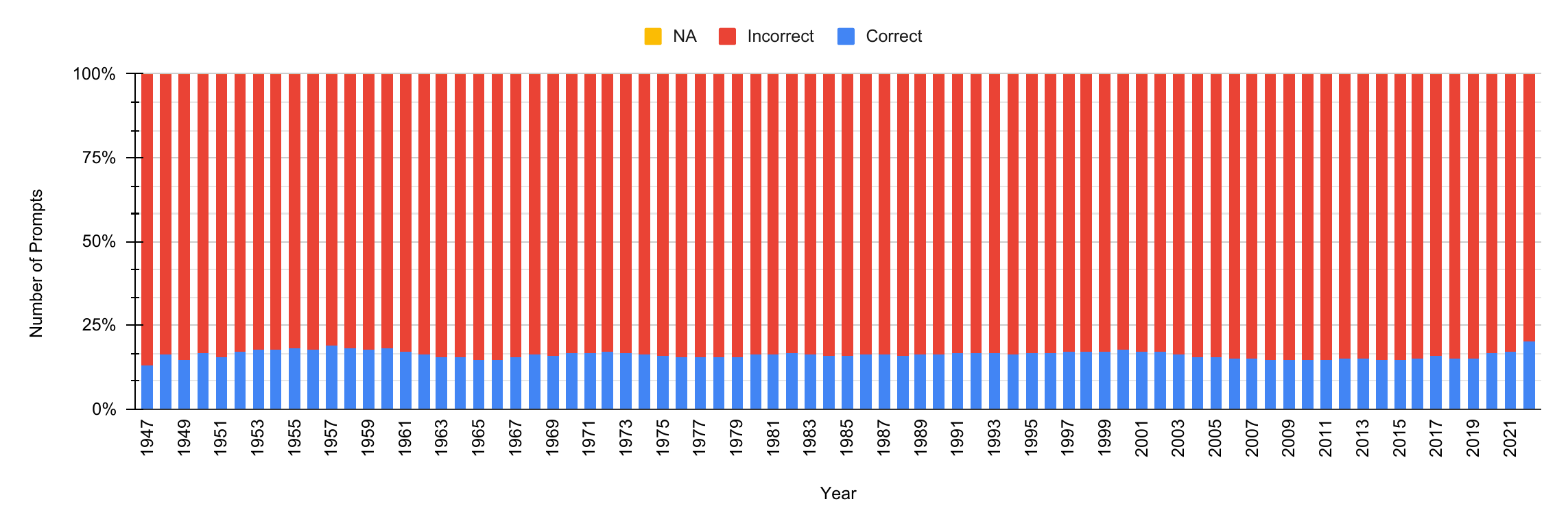}
\end{center}
\caption{Plot for the Window-based metric ($WB$) for year-wise count for \texttt{gemma-7b-it} in \textbf{Zeroshot evaluation}. }
\label{fig:window-based-gemma}
\end{figure*}

\begin{figure*}
\begin{center}
\includegraphics[width=0.9\linewidth]{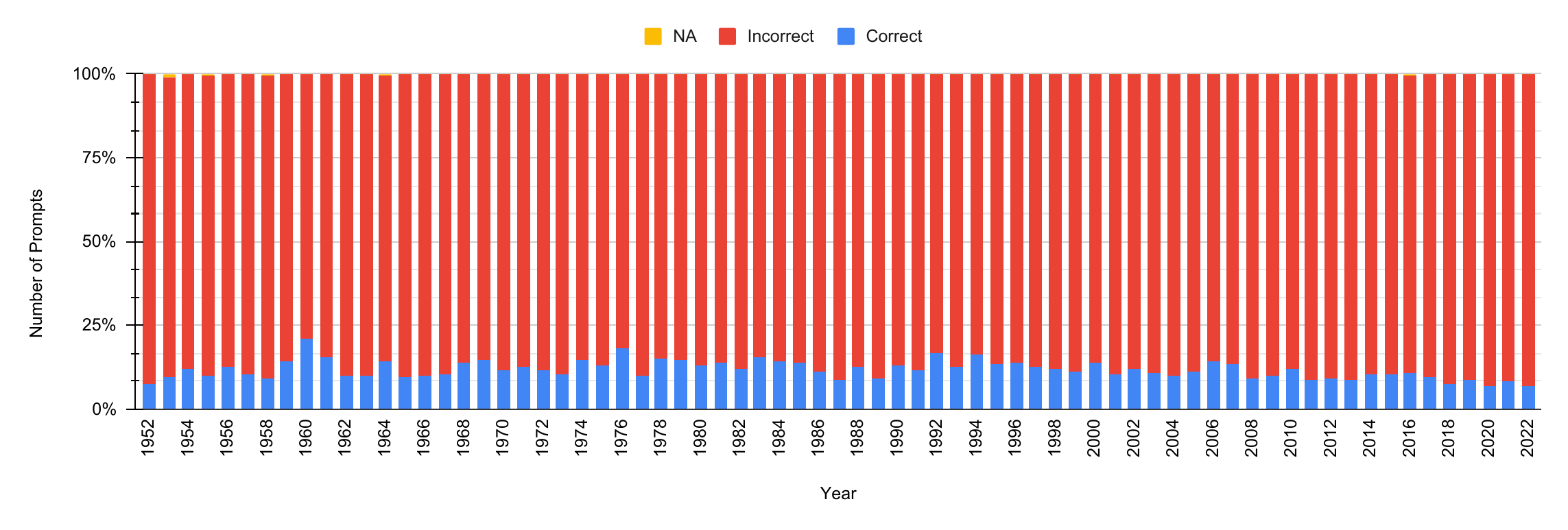}
\end{center}
\caption{Plot for the Min/Max-based metric ($MM$) for year-wise count for \texttt{gemma-7b-it} in \textbf{Zeroshot evaluation}. }
\label{fig:mmb-based-gemma}
\end{figure*}

\begin{figure*}
\begin{center}
\includegraphics[width=0.9\linewidth]{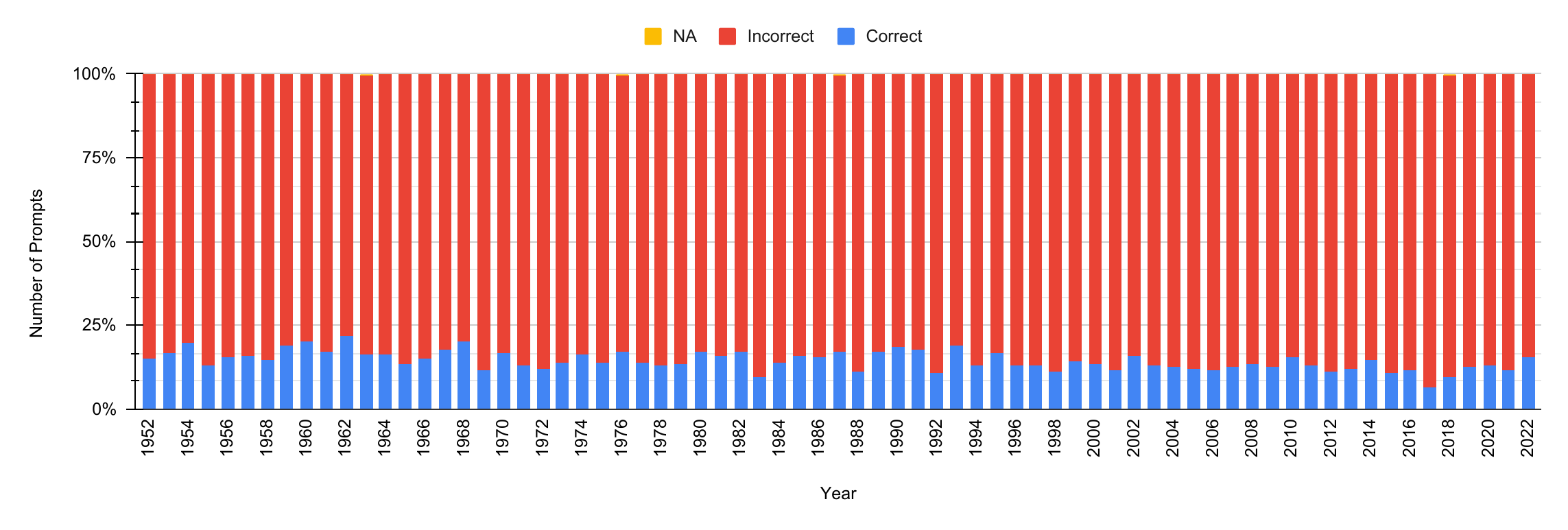}
\end{center}
\caption{Plot for the Range-based metric ($RB$) for year-wise count for \texttt{gemma-7b-it} in \textbf{Zeroshot evaluation}. }
\label{fig:rab-based-gemma}
\end{figure*}

\begin{figure*}
\begin{center}
\includegraphics[width=0.9\linewidth]{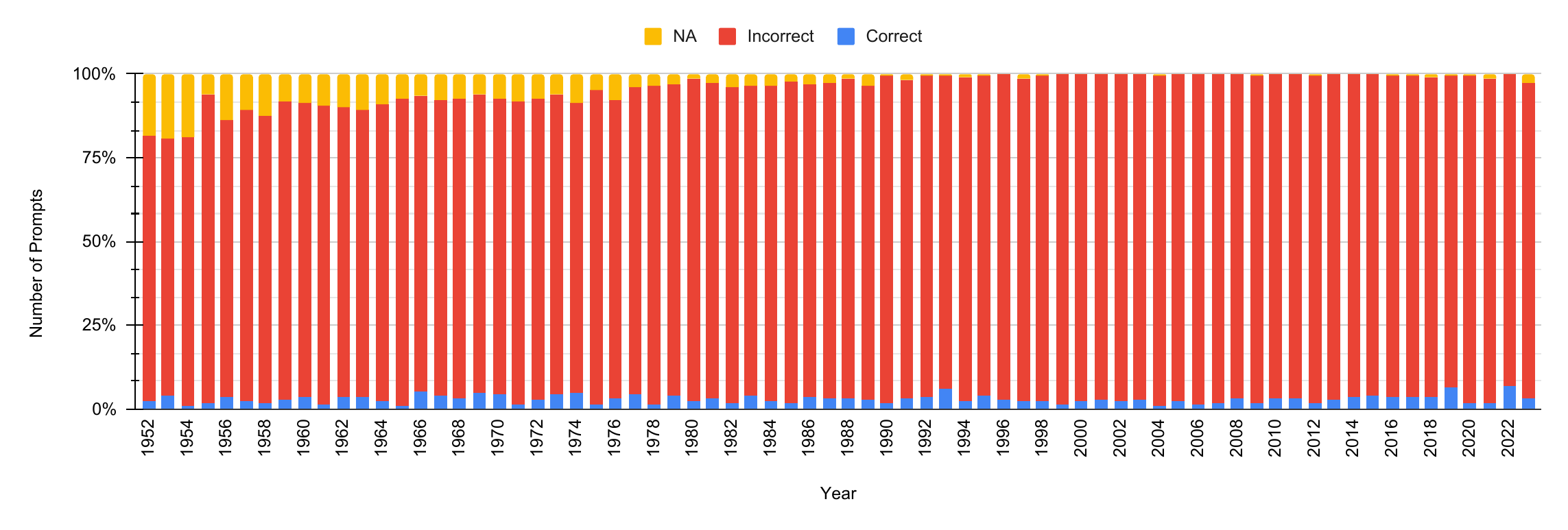}
\end{center}
\caption{Plot for the Trend-based metric ($TB$) for year-wise count for \texttt{gemma-7b-it} in \textbf{Zeroshot evaluation}. }
\label{fig:trend-based-gemma}
\end{figure*}

\begin{figure*}
\begin{center}
\includegraphics[width=0.9\linewidth]{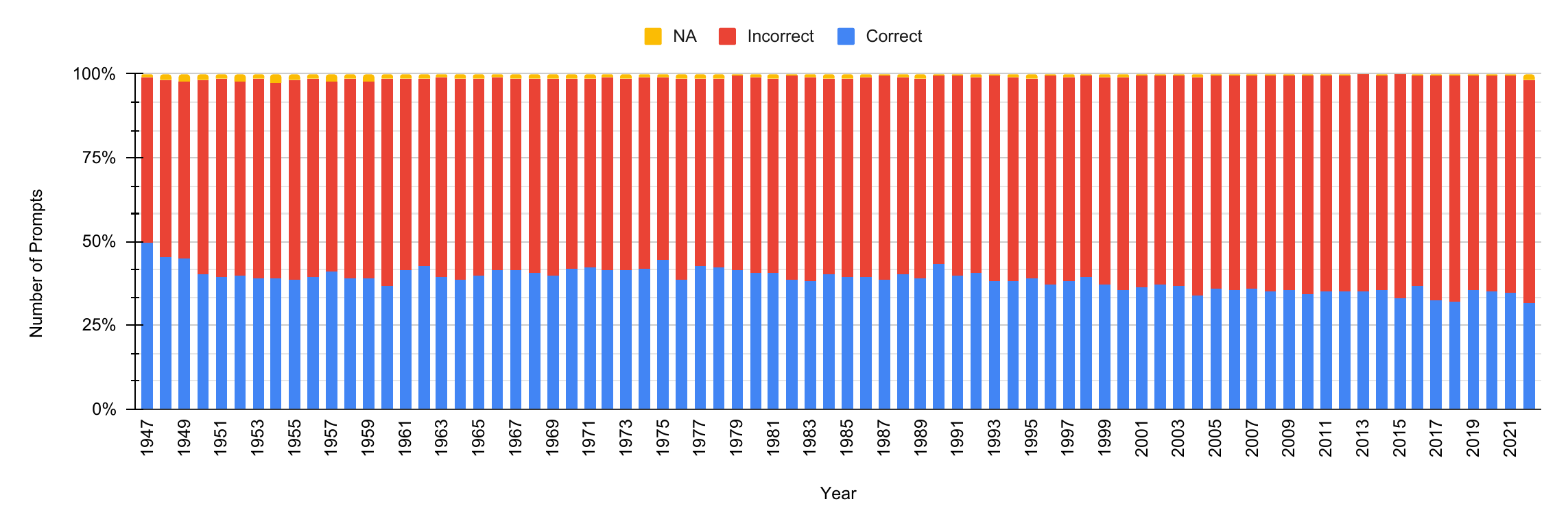}
\end{center}
\caption{Plot for the Date-based metric ($DB$) for year-wise count for \texttt{llama-3-8b} in \textbf{Zeroshot evaluation}. }
\label{fig:date-based-llama38b}
\end{figure*}

\begin{figure*}
\begin{center}
\includegraphics[width=0.9\linewidth]{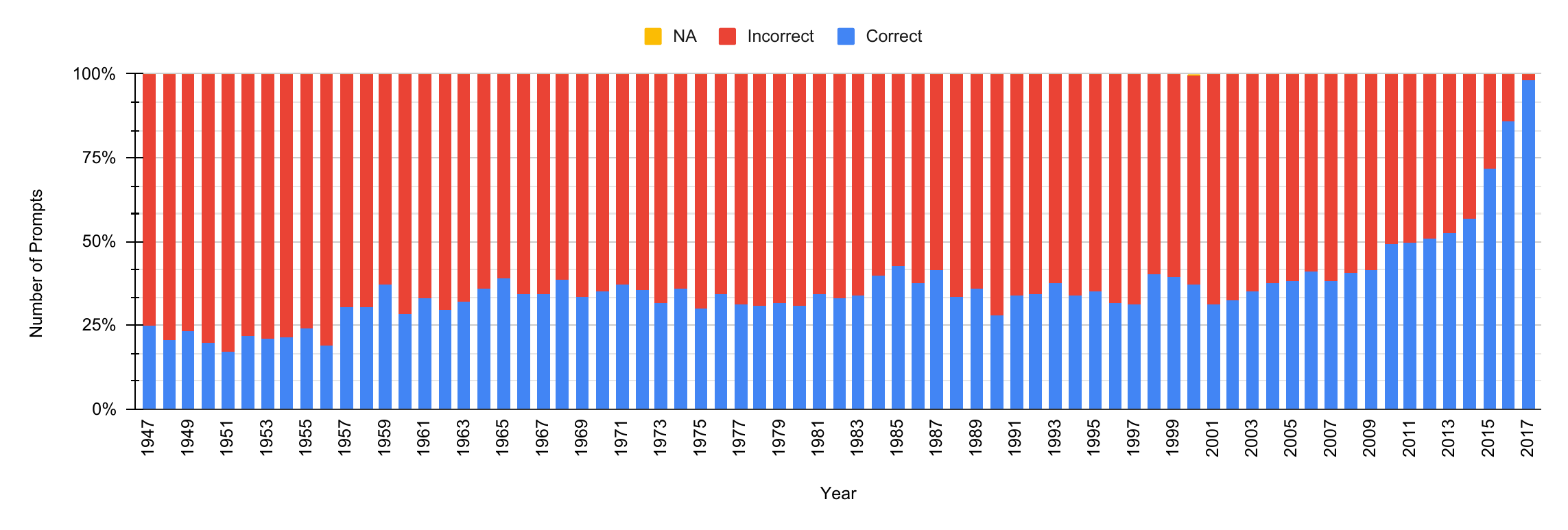}
\end{center}
\caption{Plot for the Comparative-based metric ($CP$) for year-wise count for \texttt{llama-3-8b} in \textbf{Zeroshot evaluation}. }
\label{fig:range-based-llama38b}
\end{figure*}

\begin{figure*}
\begin{center}
\includegraphics[width=0.9\linewidth]{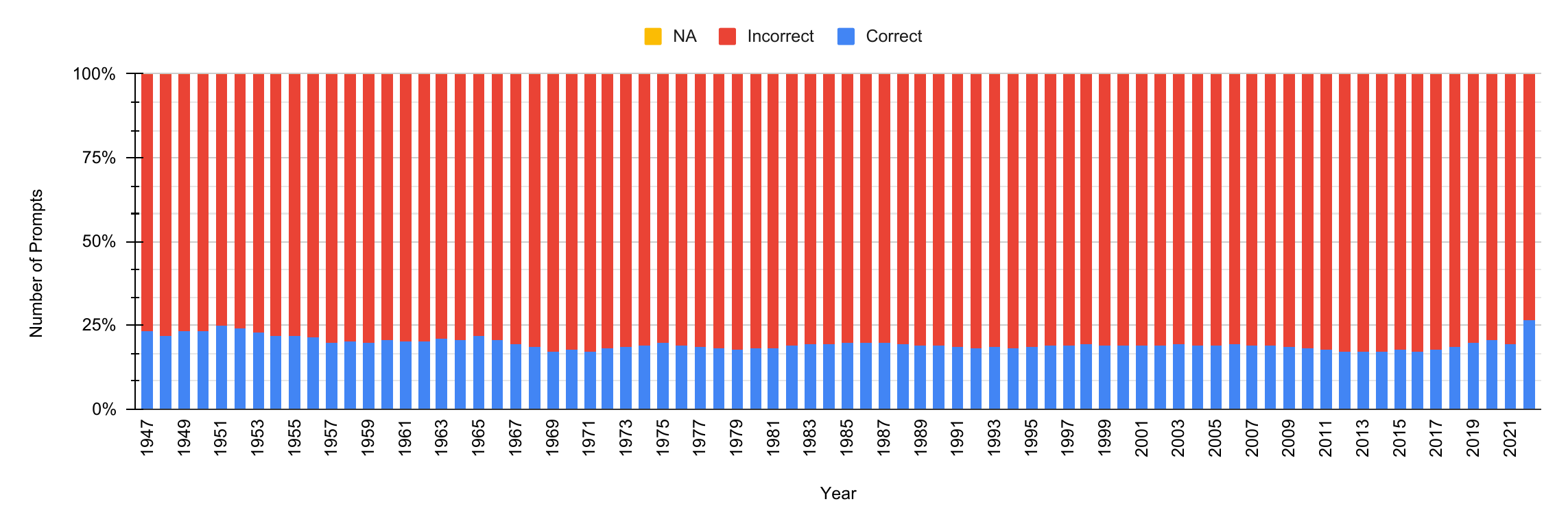}
\end{center}
\caption{Plot for the Window-based metric ($WB$) for year-wise count for \texttt{llama-3-8b} in \textbf{Zeroshot evaluation}. }
\label{fig:window-based-llama38b}
\end{figure*}

\begin{figure*}
\begin{center}
\includegraphics[width=0.9\linewidth]{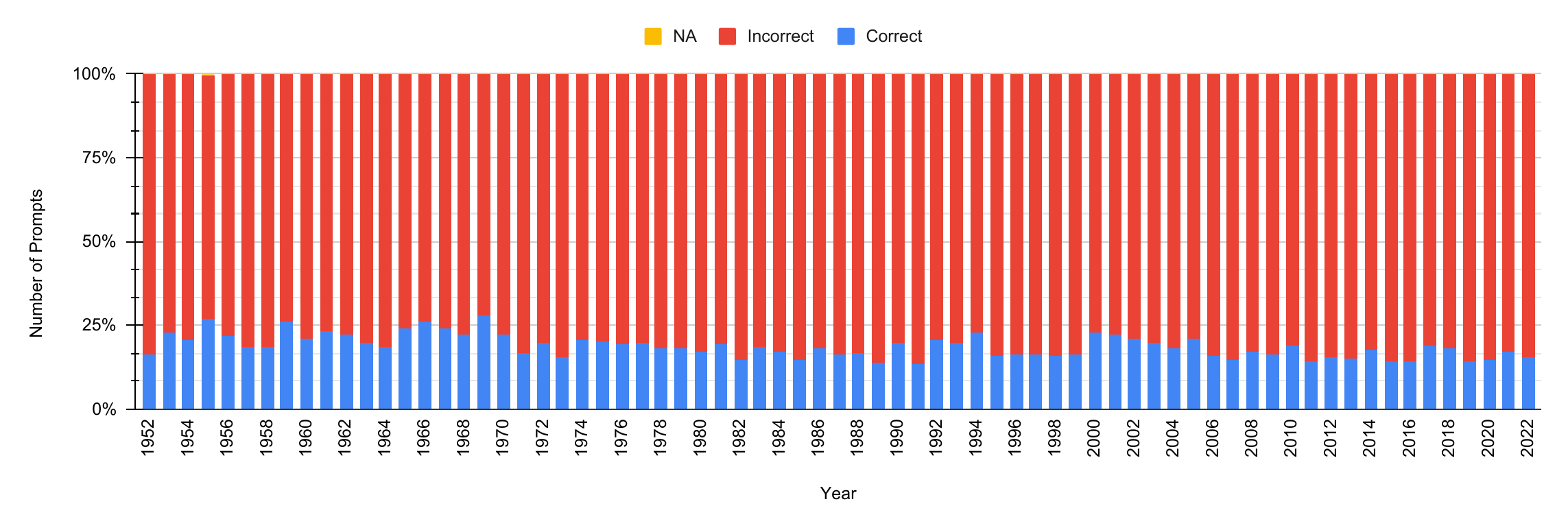}
\end{center}
\caption{Plot for the Min/Max-based metric ($MM$) for year-wise count for \texttt{llama-3-8b} in \textbf{Zeroshot evaluation}. }
\label{fig:mmb-based-llama38b}
\end{figure*}

\begin{figure*}
\begin{center}
\includegraphics[width=0.9\linewidth]{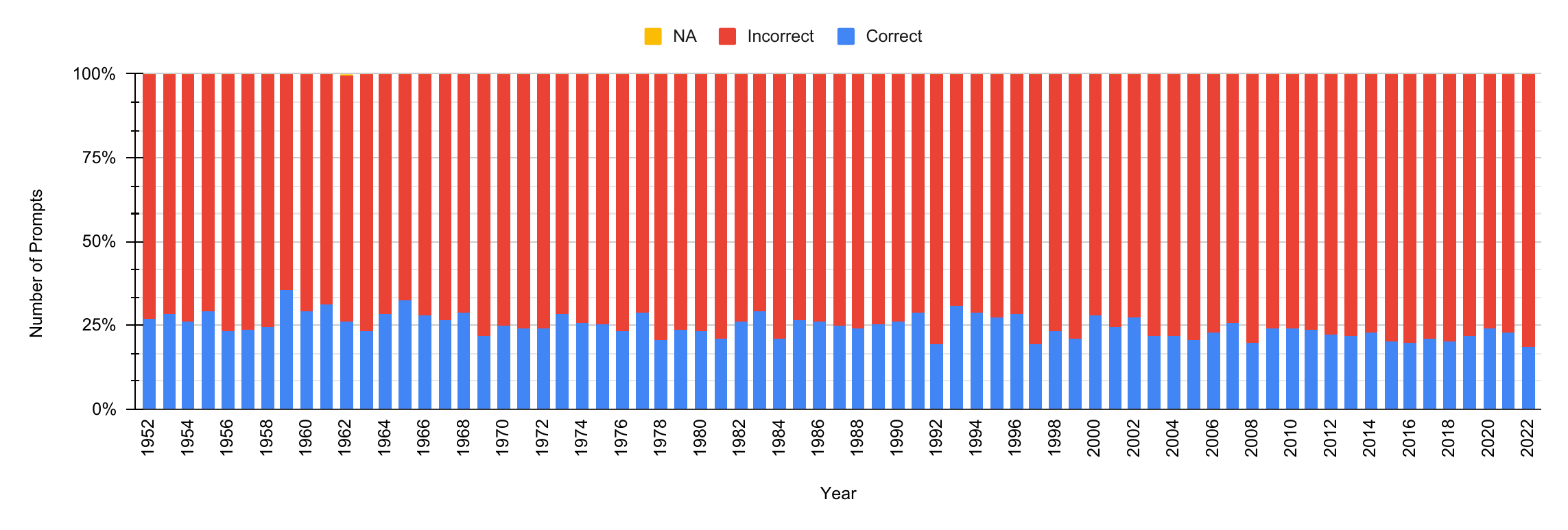}
\end{center}
\caption{Plot for the Range-based metric ($RB$) for year-wise count for \texttt{llama-3-8b} in \textbf{Zeroshot evaluation}. }
\label{fig:rab-based-llama38b}
\end{figure*}

\begin{figure*}
\begin{center}
\includegraphics[width=0.9\linewidth]{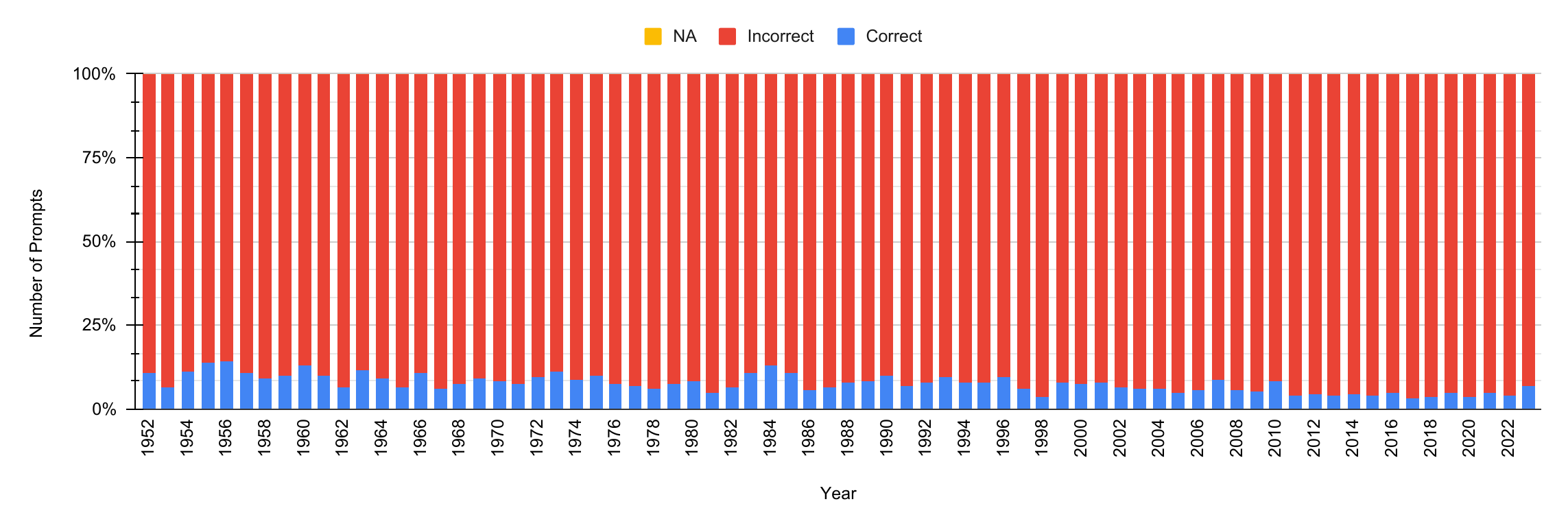} 
\end{center}
\caption{Plot for the Trend-based metric ($TB$) for year-wise count for \texttt{llama-3-8b} in \textbf{Zeroshot evaluation}. }
\label{fig:trend-based-llama38b}
\end{figure*}


\begin{figure*}
\begin{center}
\includegraphics[width=0.9\linewidth]{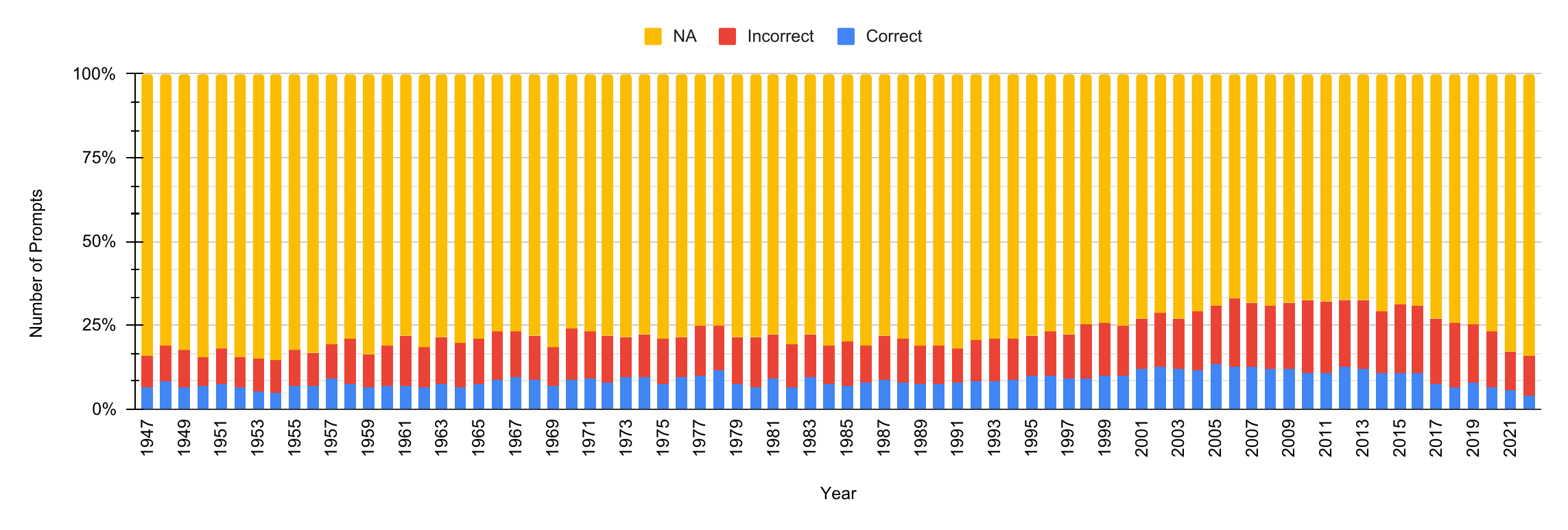}
\end{center}
\caption{Plot for the Date-based metric ($DB$) for year-wise count for \texttt{phi-3-instruct} in \textbf{Zeroshot evaluation}. }
\label{fig:date-based-phi3}
\end{figure*}

\begin{figure*}
\begin{center}
\includegraphics[width=0.9\linewidth]{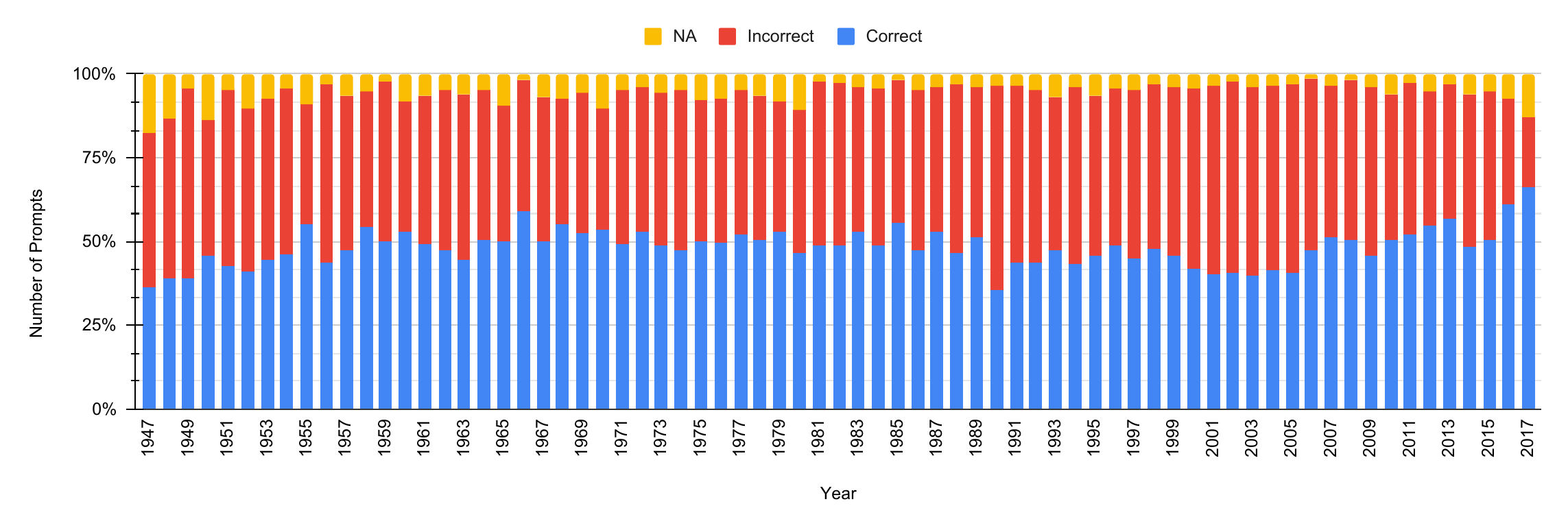}
\end{center}
\caption{Plot for the Comparative-based metric ($CP$) for year-wise count for \texttt{phi-3-instruct2} in \textbf{Zeroshot evaluation}. }
\label{fig:range-based-phi3}
\end{figure*}

\begin{figure*}
\begin{center}
\includegraphics[width=0.9\linewidth]{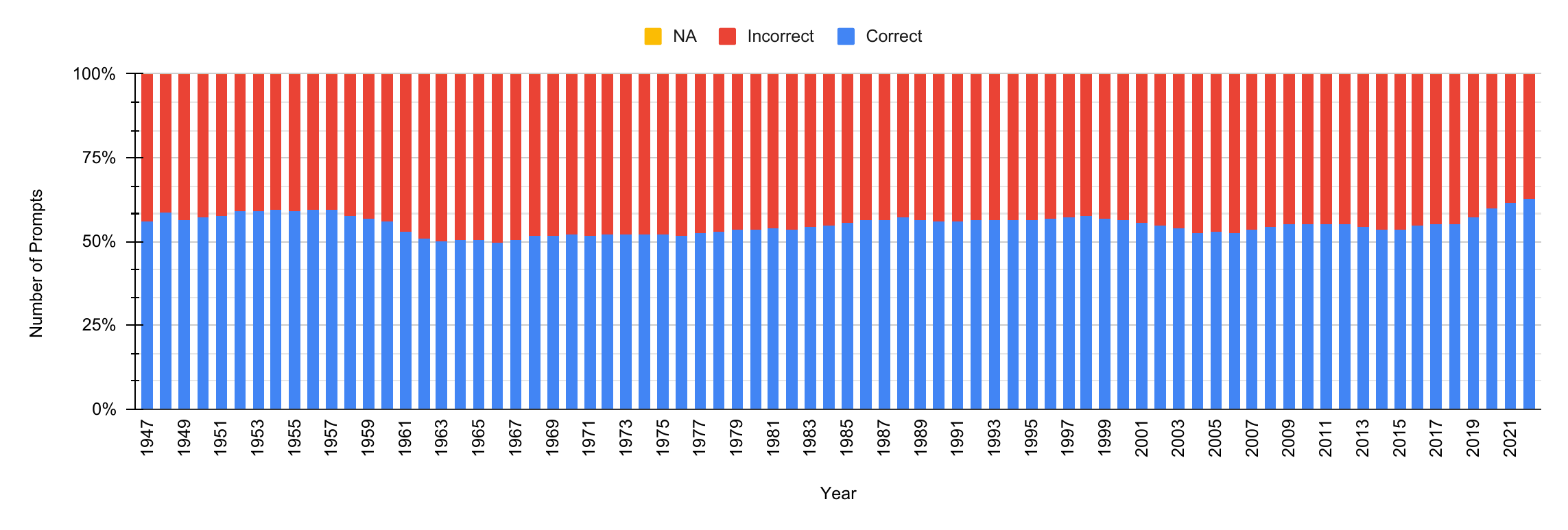}
\end{center}
\caption{Plot for the Window-based metric ($WB$) for year-wise count for \texttt{phi-3-instruct2} in \textbf{Zeroshot evaluation}. }
\label{fig:window-based-phi3}
\end{figure*}

\begin{figure*}
\begin{center}
\includegraphics[width=0.9\linewidth]{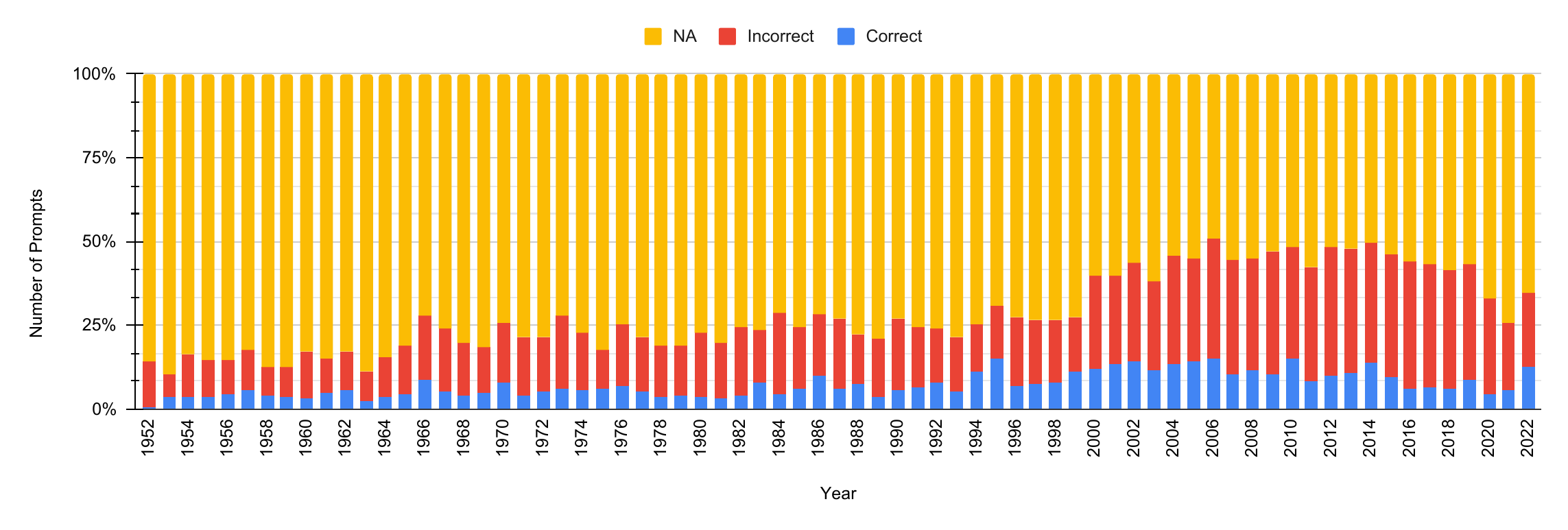}
\end{center}
\caption{Plot for the Min/Max-based metric ($MM$) for year-wise count for \texttt{phi-3-instruct2} in \textbf{Zeroshot evaluation}. }
\label{fig:mmb-based-phi3}
\end{figure*}

\begin{figure*}
\begin{center}
\includegraphics[width=0.9\linewidth]{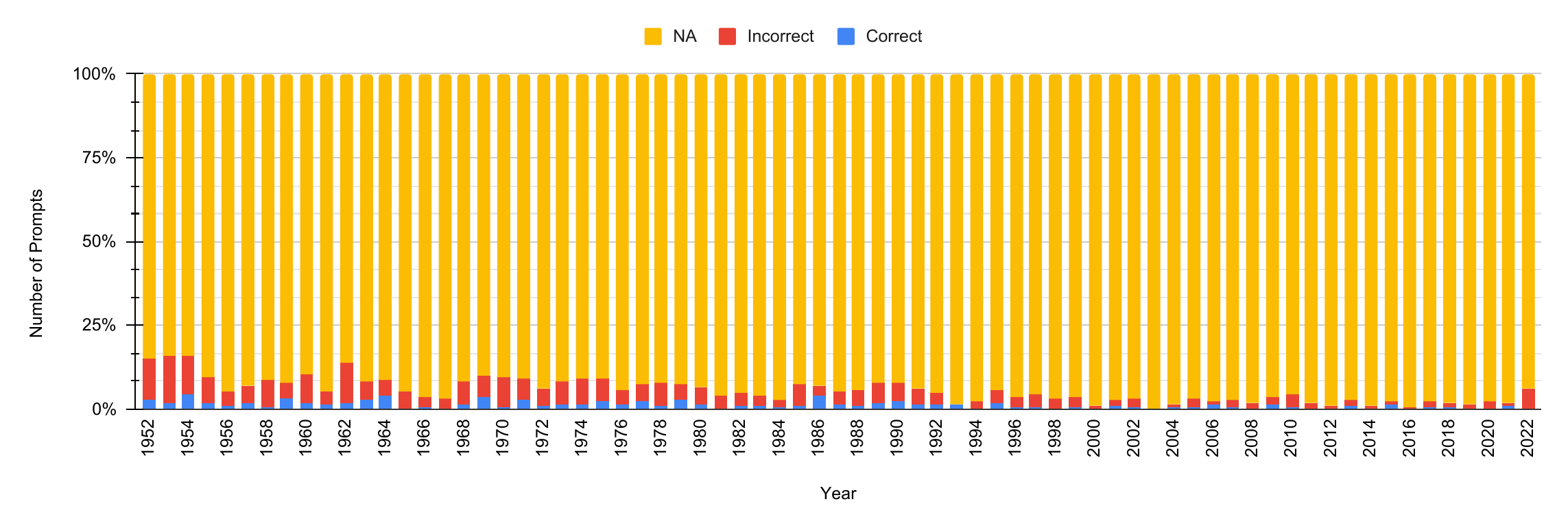}
\end{center}
\caption{Plot for the Range-based metric ($RB$) for year-wise count for \texttt{phi-3-instruct2} in \textbf{Zeroshot evaluation}. }
\label{fig:rab-based-phi3}
\end{figure*}

\begin{figure*}
\begin{center}
\includegraphics[width=0.9\linewidth]{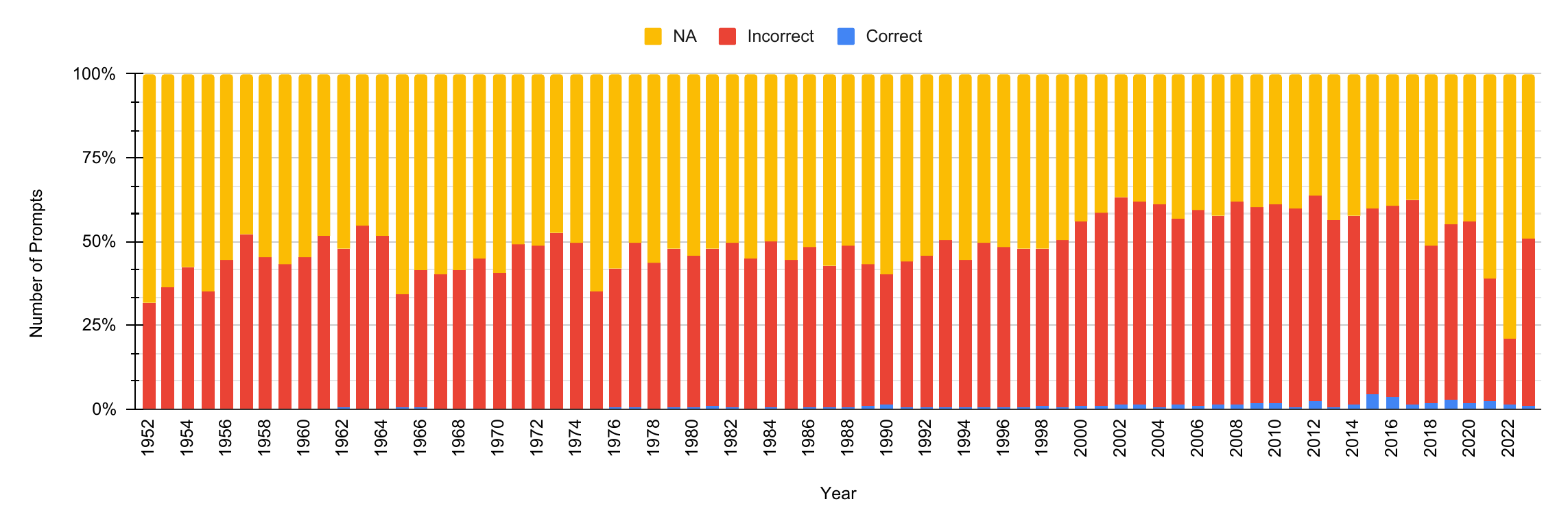} 
\end{center}
\caption{Plot for the Trend-based metric ($TB$) for year-wise count for \texttt{phi-3-instruct2} in \textbf{Zeroshot evaluation}. }
\label{fig:trend-based-phi3}
\end{figure*}


\begin{figure*}
\begin{center}
\includegraphics[width=0.9\linewidth]{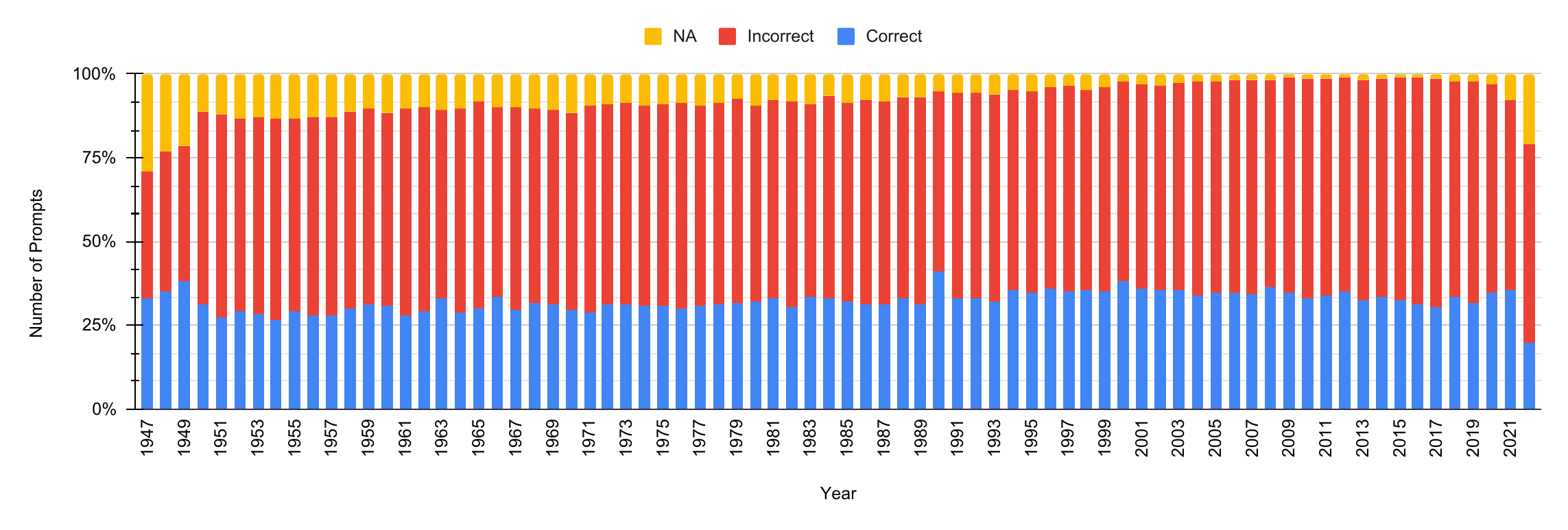}
\end{center}
\caption{Plot for the Date-based metric ($DB$) for year-wise count for \texttt{mixtral-8x7b-32768} in \textbf{Zeroshot evaluation}. }
\label{fig:date-based-mixtral}
\end{figure*}

\begin{figure*}
\begin{center}
\includegraphics[width=0.9\linewidth]{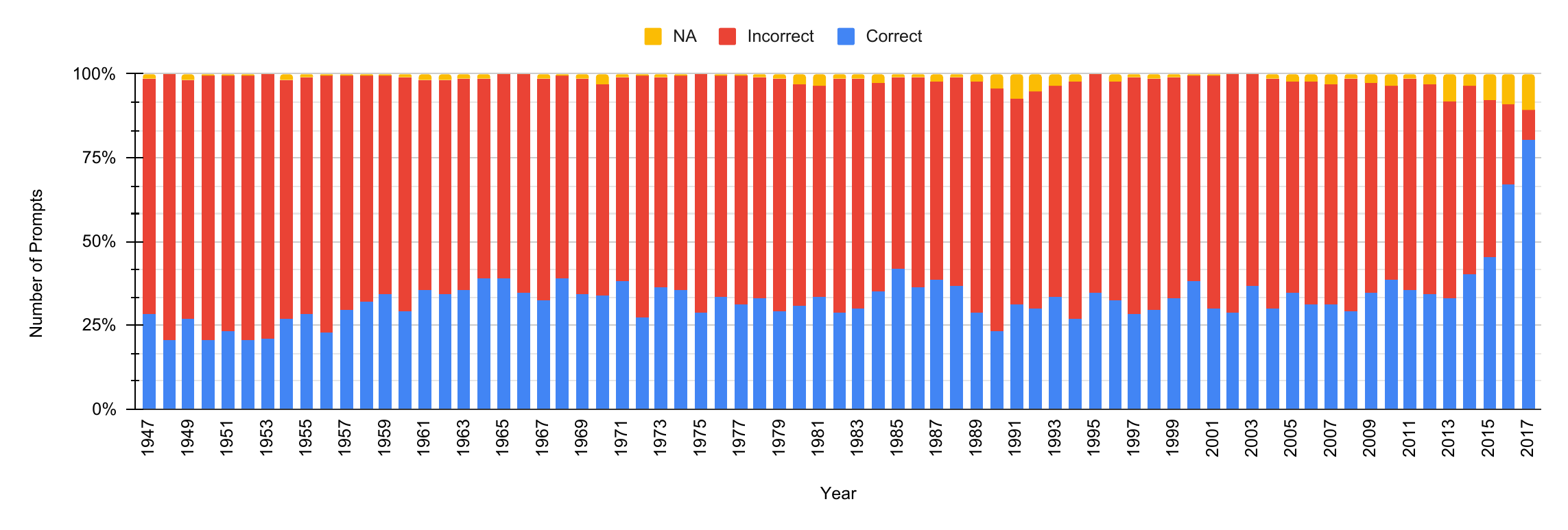}
\end{center}
\caption{Plot for the Comparative-based metric ($CP$) for year-wise count for \texttt{mixtral-8x7b-32768} in \textbf{Zeroshot evaluation}. }
\label{fig:range-based-mixtral}
\end{figure*}

\begin{figure*}
\begin{center}
\includegraphics[width=0.9\linewidth]{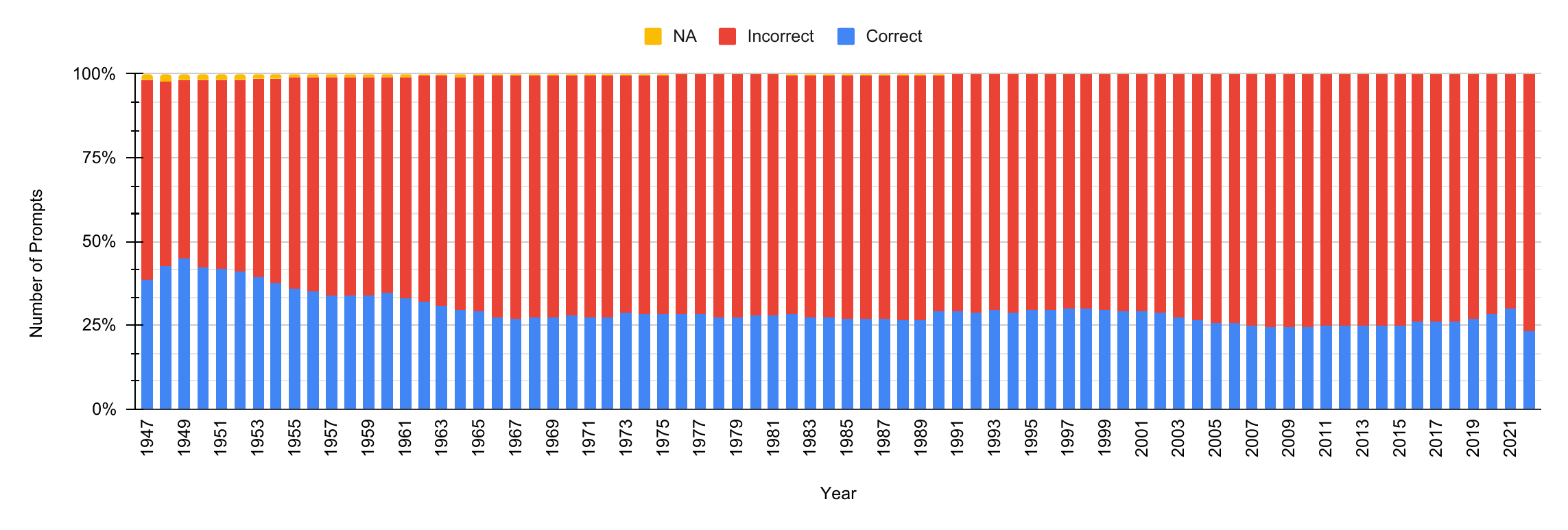}
\end{center}
\caption{Plot for the Window-based metric ($WB$) for year-wise count for \texttt{mixtral-8x7b-32768} in \textbf{Zeroshot evaluation}. }
\label{fig:window-based-mixtral}
\end{figure*}

\begin{figure*}
\begin{center}
\includegraphics[width=0.9\linewidth]{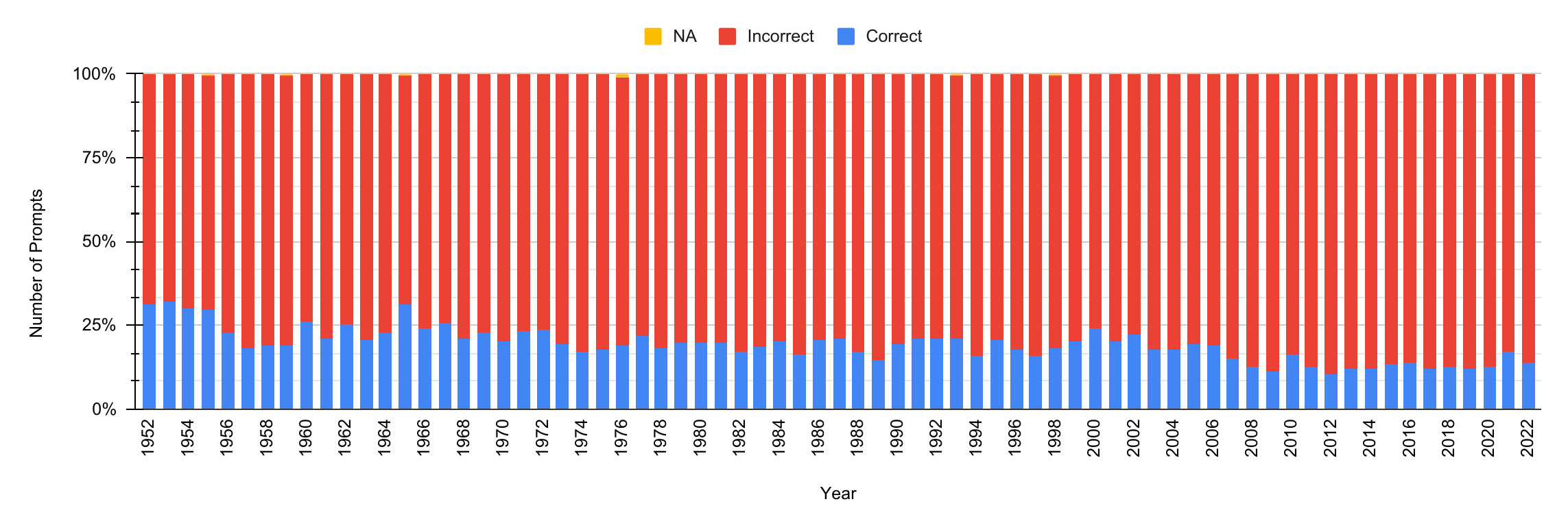}
\end{center}
\caption{Plot for the Min/Max-based metric ($MM$) for year-wise count for \texttt{mixtral-8x7b-32768} in \textbf{Zeroshot evaluation}. }
\label{fig:mmb-based-mixtral}
\end{figure*}

\begin{figure*}
\begin{center}
\includegraphics[width=0.9\linewidth]{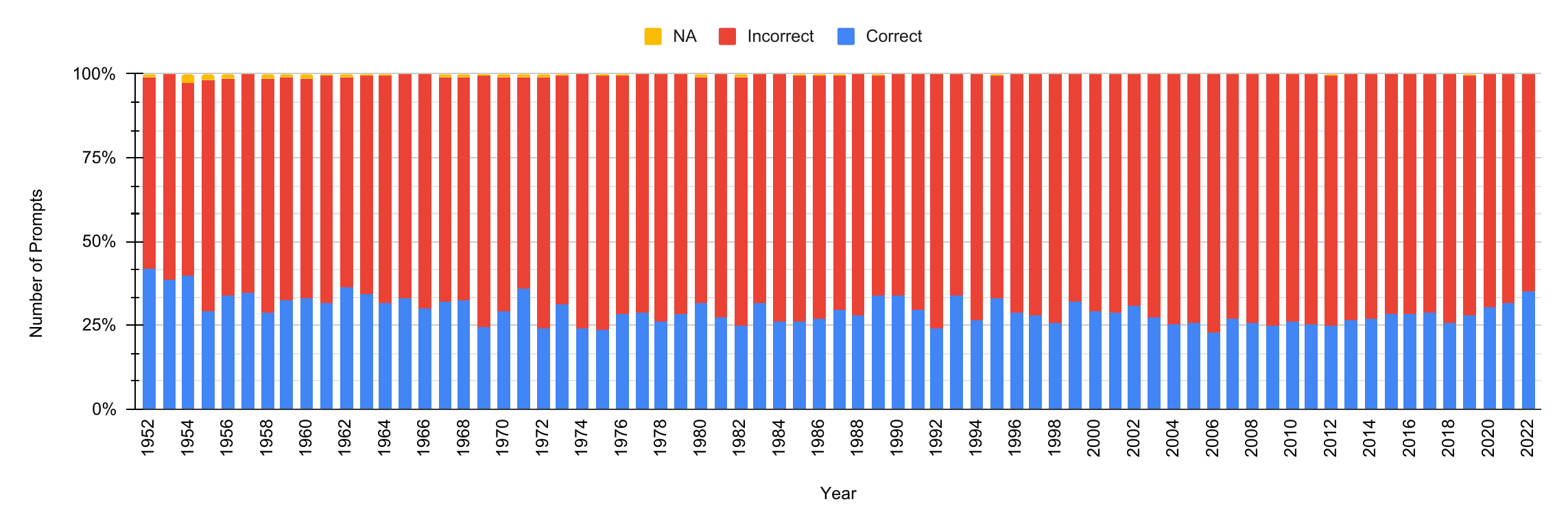}
\end{center}
\caption{Plot for the Range-based metric ($RB$) for year-wise count for \texttt{mixtral-8x7b-32768} in \textbf{Zeroshot evaluation}. }
\label{fig:rab-based-mixtral}
\end{figure*}

\begin{figure*}
\begin{center}
\includegraphics[width=0.9\linewidth]{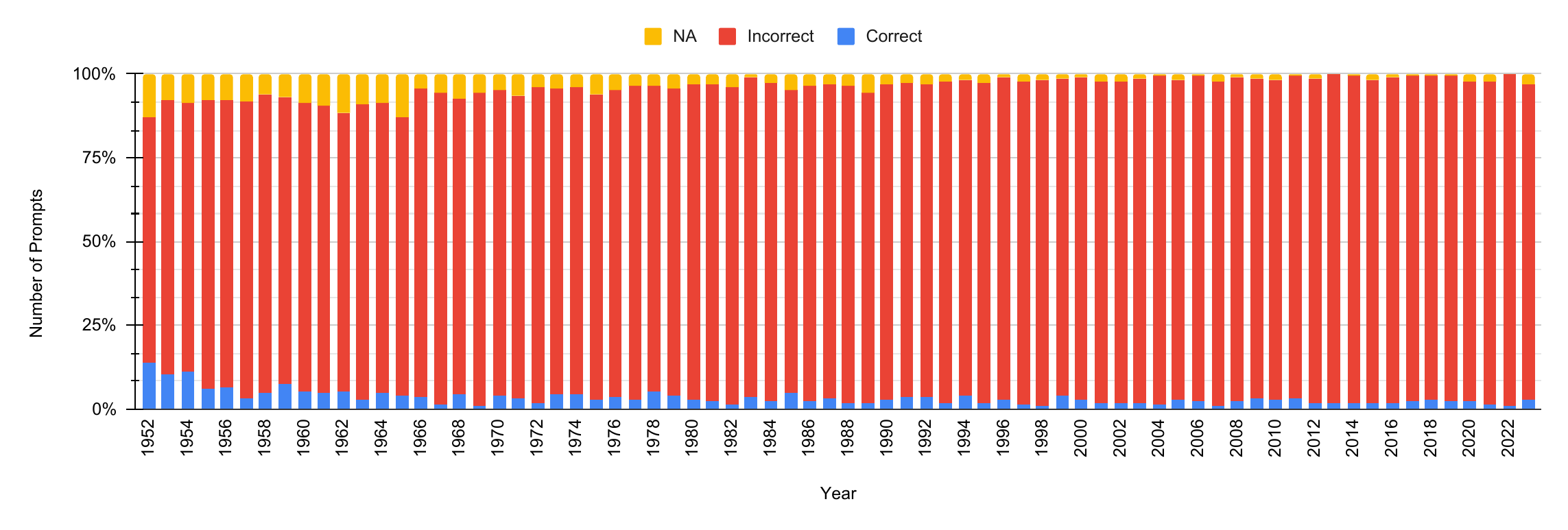} 
\end{center}
\caption{Plot for the Trend-based metric ($TB$) for year-wise count for \texttt{mixtral-8x7b-32768} in \textbf{Zeroshot evaluation}. }
\label{fig:trend-based-mixtral}
\end{figure*}


\begin{figure*}
\begin{center}
\includegraphics[width=0.9\linewidth]{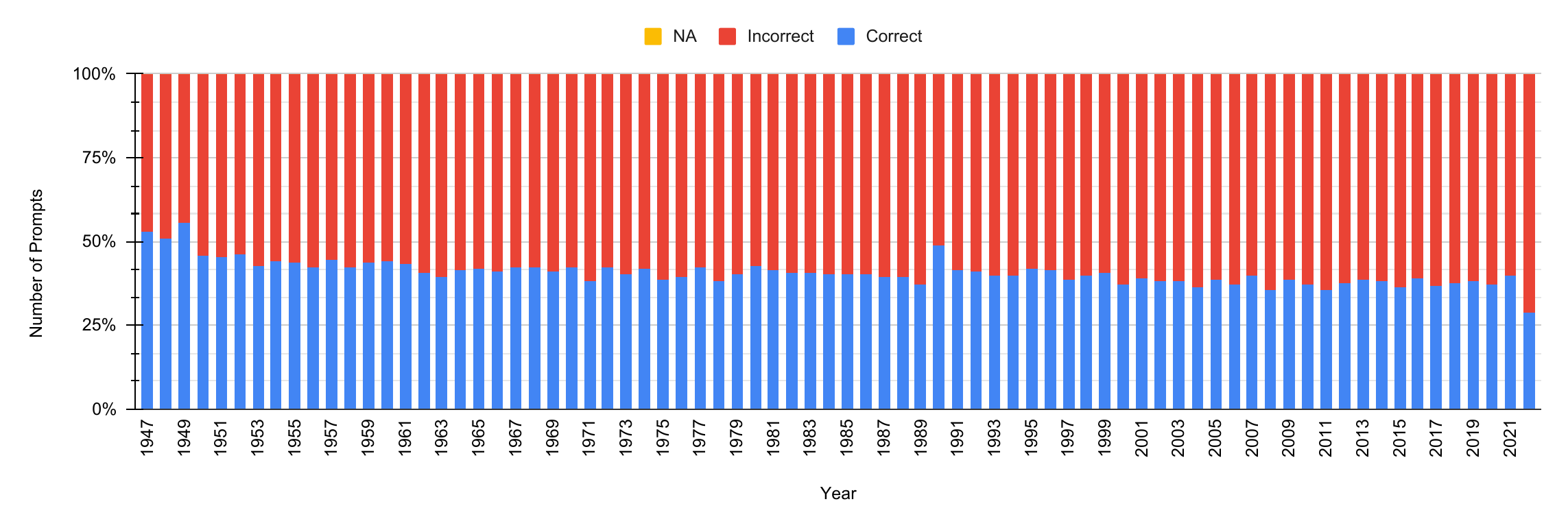}
\end{center}
\caption{Plot for the Date-based metric ($DB$) for year-wise count for \texttt{llama-3-70b-8192} in \textbf{Zeroshot evaluation}. }
\label{fig:date-based-llama70b}
\end{figure*}

\begin{figure*}
\begin{center}
\includegraphics[width=0.9\linewidth]{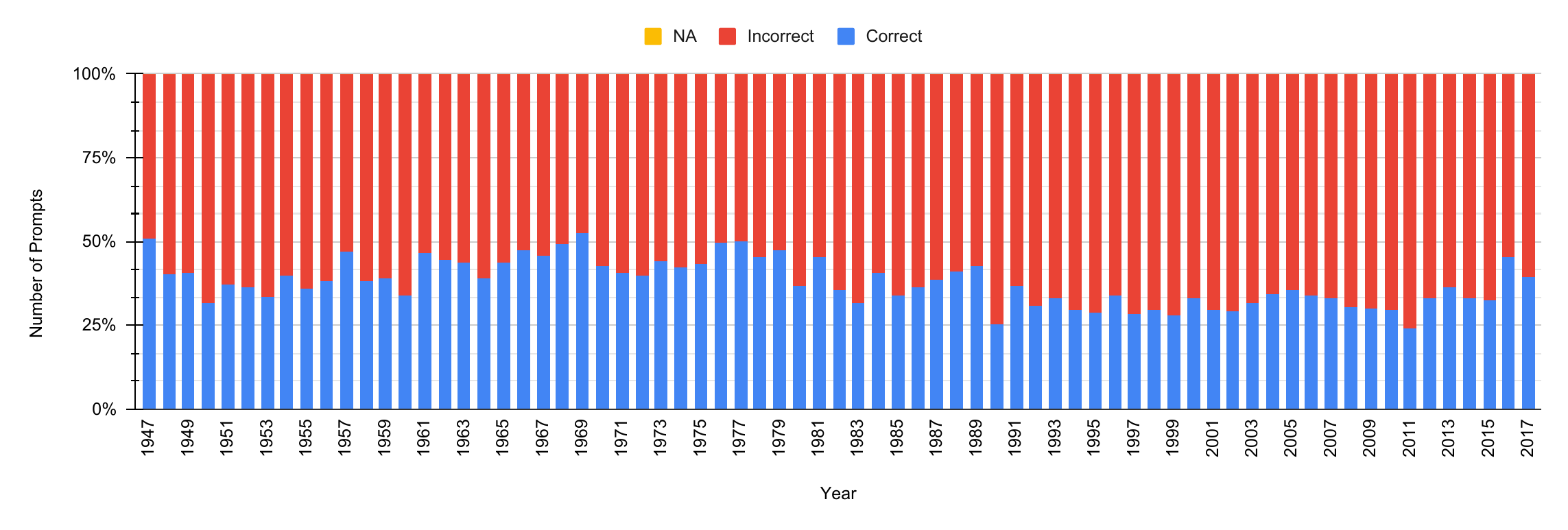}
\end{center}
\caption{Plot for the Comparative-based metric ($CP$) for year-wise count for \texttt{llama-3-70b-8192} in \textbf{Zeroshot evaluation}. }
\label{fig:range-based-llama70b}
\end{figure*}

\begin{figure*}
\begin{center}
\includegraphics[width=0.9\linewidth]{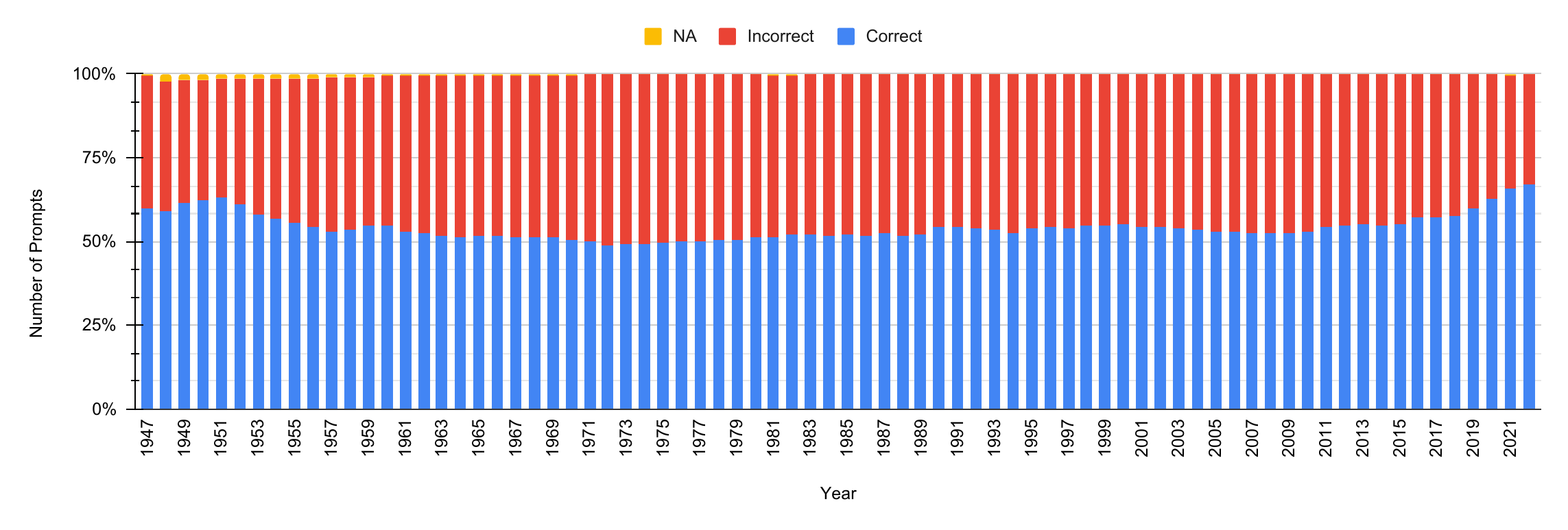}
\end{center}
\caption{Plot for the Window-based metric ($WB$) for year-wise count for \texttt{llama-3-70b-8192} in \textbf{Zeroshot evaluation}. }
\label{fig:window-based-llama70b}
\end{figure*}

\begin{figure*}
\begin{center}
\includegraphics[width=0.9\linewidth]{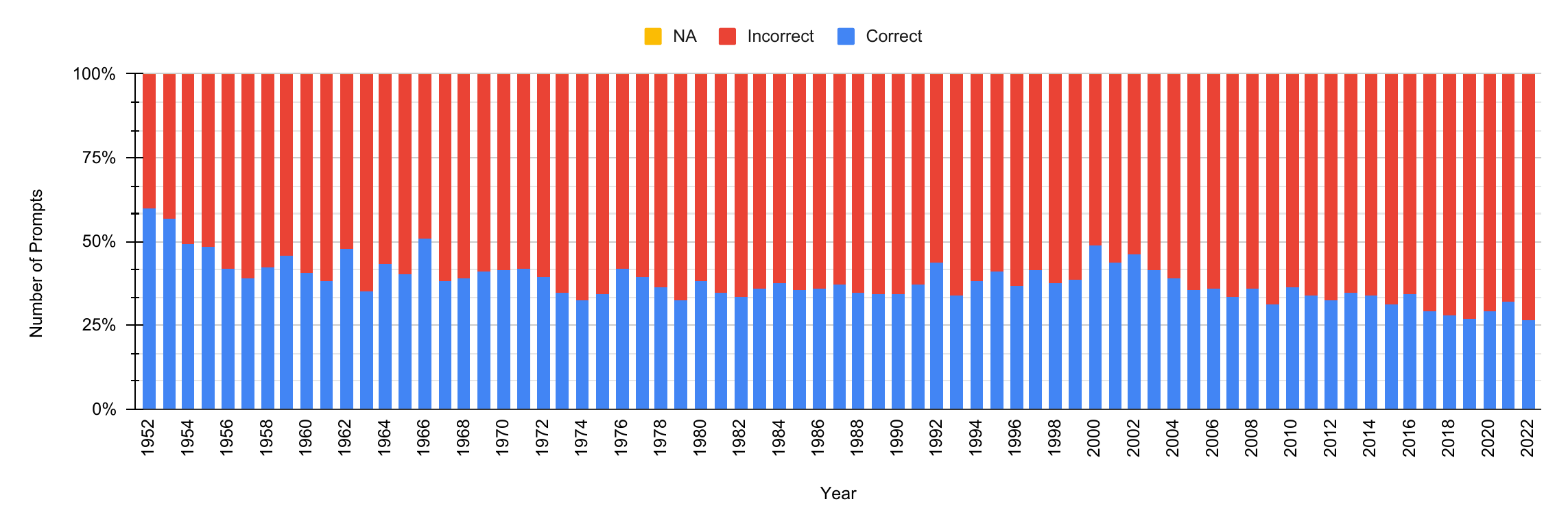}
\end{center}
\caption{Plot for the Min/Max-based metric ($MM$) for year-wise count for \texttt{llama-3-70b-8192} in \textbf{Zeroshot evaluation}. }
\label{fig:mmb-based-llama70b}
\end{figure*}

\begin{figure*}
\begin{center}
\includegraphics[width=0.9\linewidth]{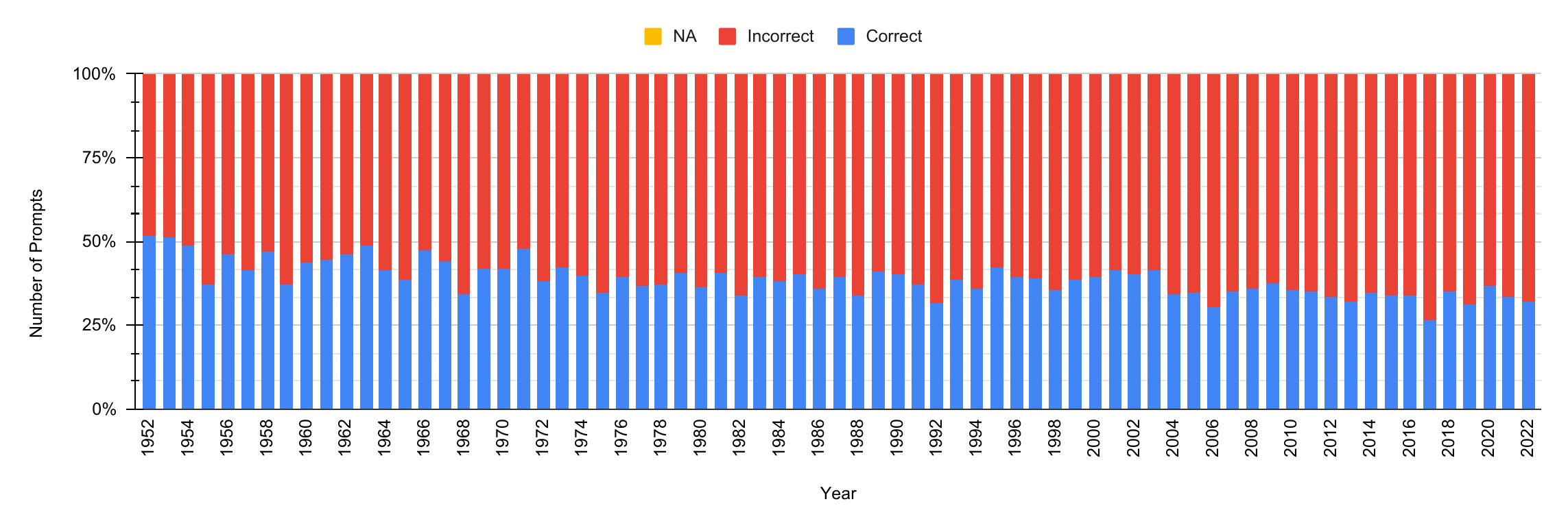}
\end{center}
\caption{Plot for the Range-based metric ($RB$) for year-wise count for \texttt{llama-3-70b-8192} in \textbf{Zeroshot evaluation}. }
\label{fig:rab-based-llama70b}
\end{figure*}

\begin{figure*}
\begin{center}
\includegraphics[width=0.9\linewidth]{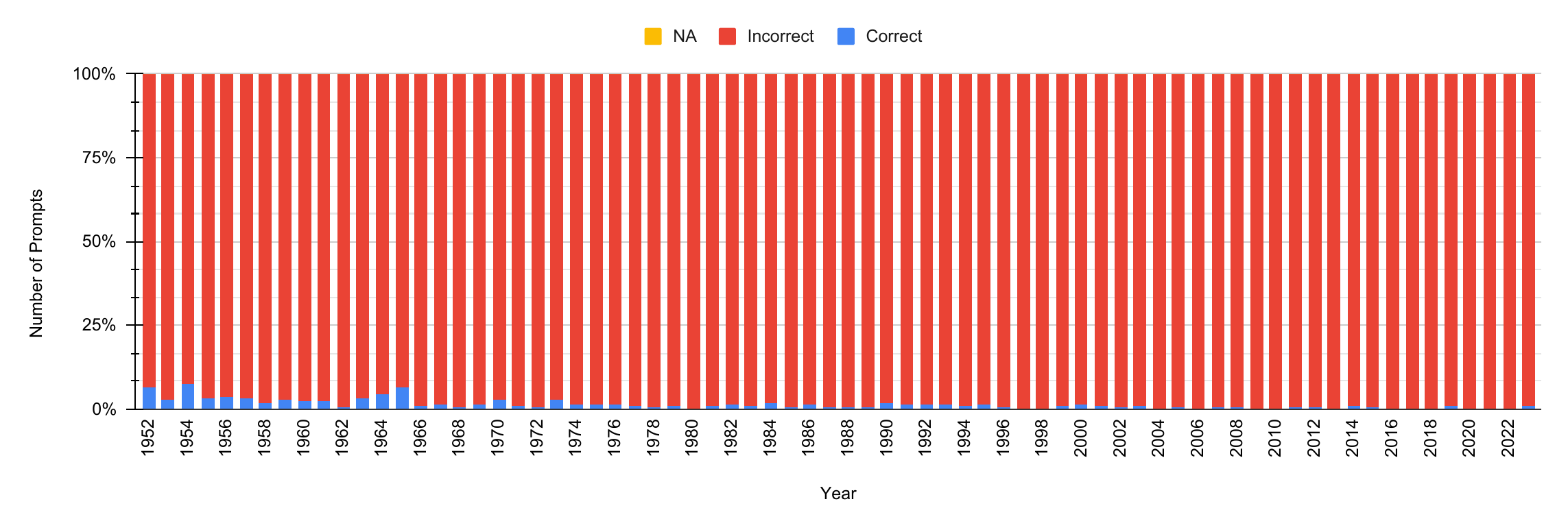} 
\end{center}
\caption{Plot for the Trend-based metric ($TB$) for year-wise count for \texttt{llama-3-70b-8192} in \textbf{Zeroshot evaluation}. }
\label{fig:trend-based-llama70b}
\end{figure*}

\begin{figure*}
\begin{center}
\includegraphics[width=0.9\linewidth]{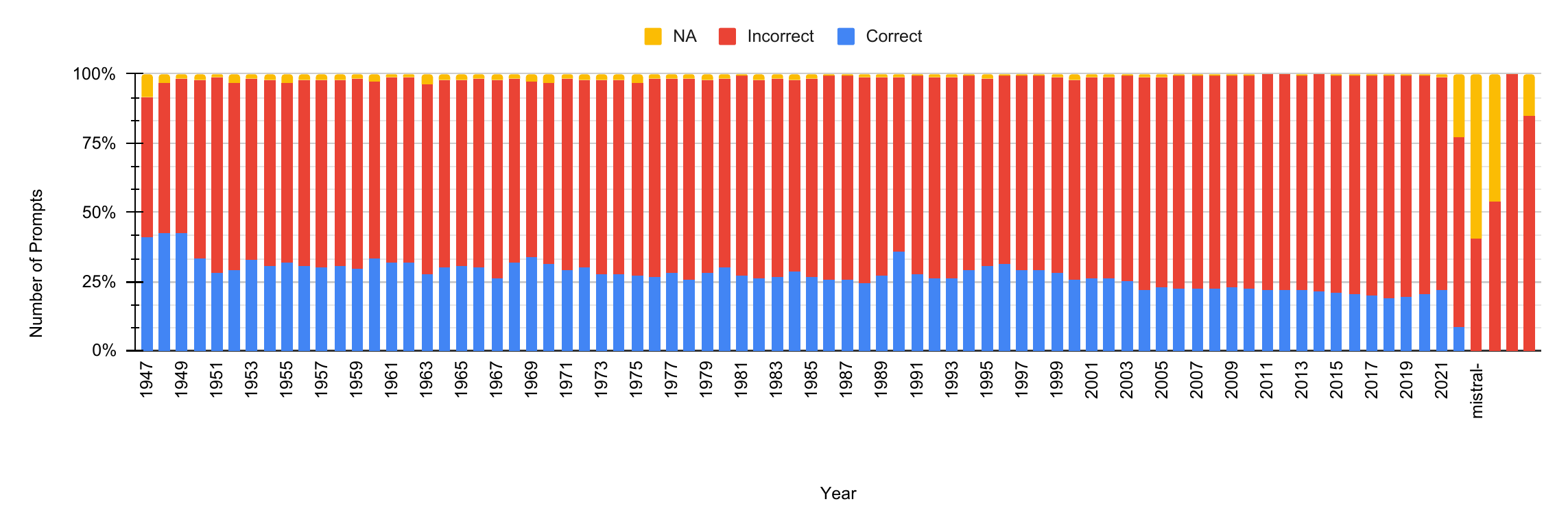}
\end{center}
\caption{Plot for the Date-based metric ($DB$) for year-wise count for \texttt{gpt-3.5} in \textbf{Zeroshot evaluation}. }
\label{fig:date-based-gpt-35}
\end{figure*}

\begin{figure*}
\begin{center}
\includegraphics[width=0.9\linewidth]{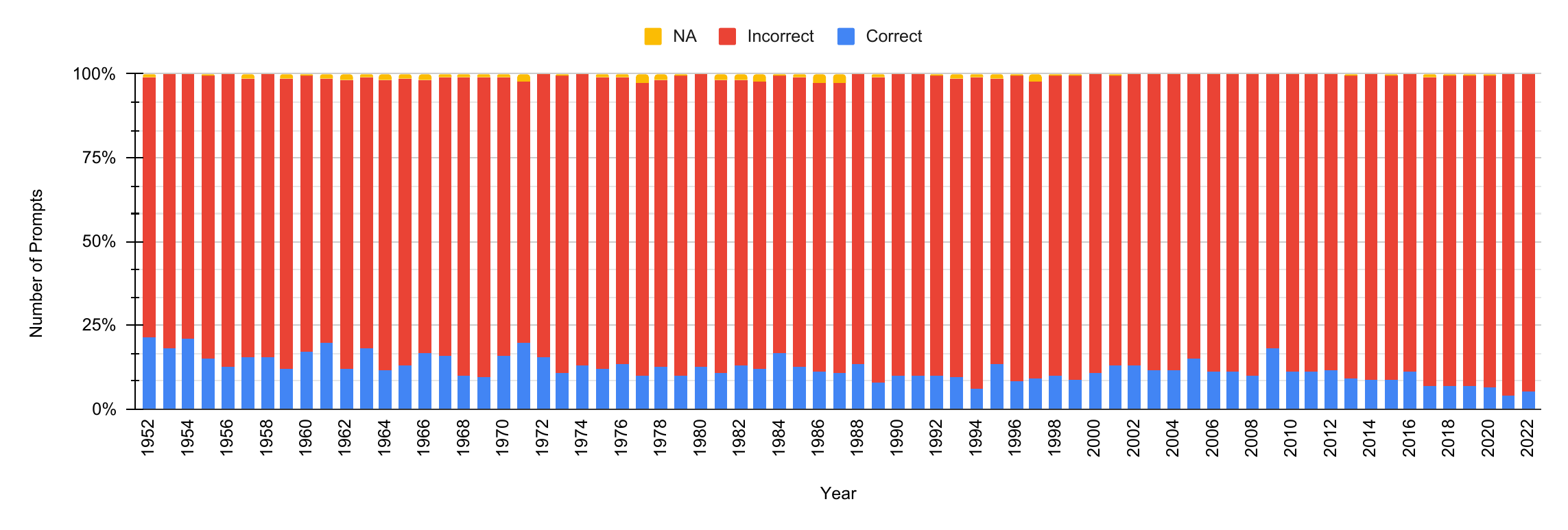}
\end{center}
\caption{Plot for the Comparative-based metric ($CP$) for year-wise count for \texttt{gpt-3.5} in \textbf{Zeroshot evaluation}. }
\label{fig:range-based-gpt-35}
\end{figure*}

\begin{figure*}
\begin{center}
\includegraphics[width=0.9\linewidth]{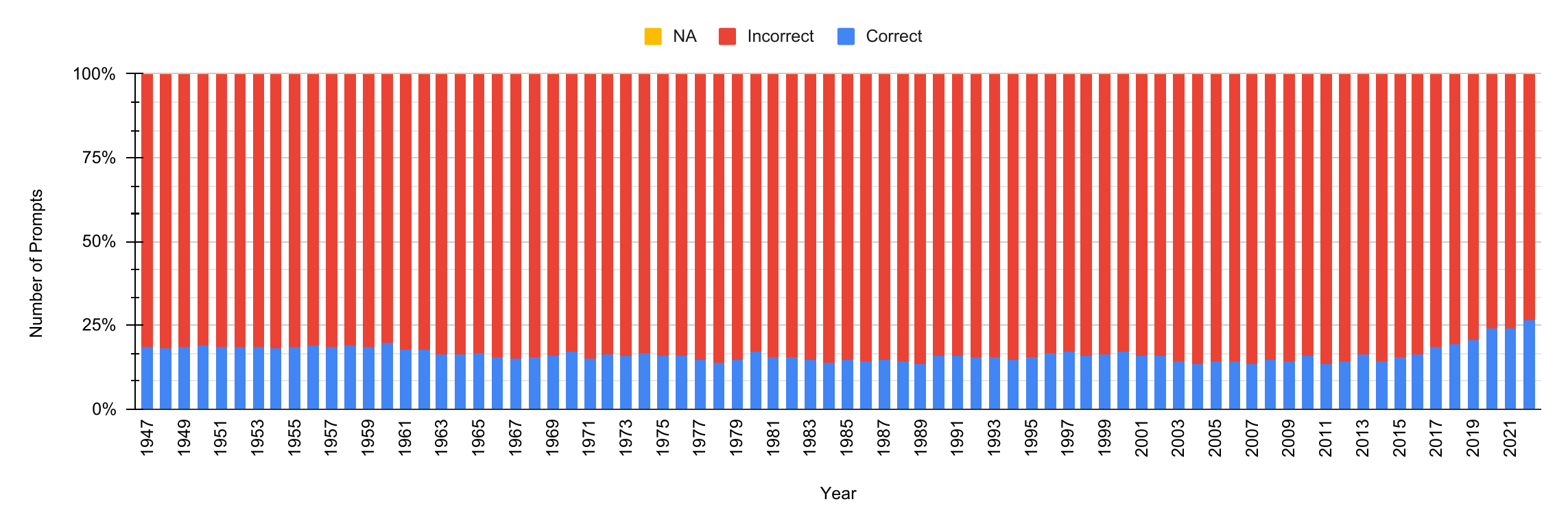}
\end{center}
\caption{Plot for the Window-based metric ($WB$) for year-wise count for \texttt{gpt-3.5} in \textbf{Zeroshot evaluation}. }
\label{fig:window-based-gpt-35}
\end{figure*}

\begin{figure*}
\begin{center}
\includegraphics[width=0.9\linewidth]{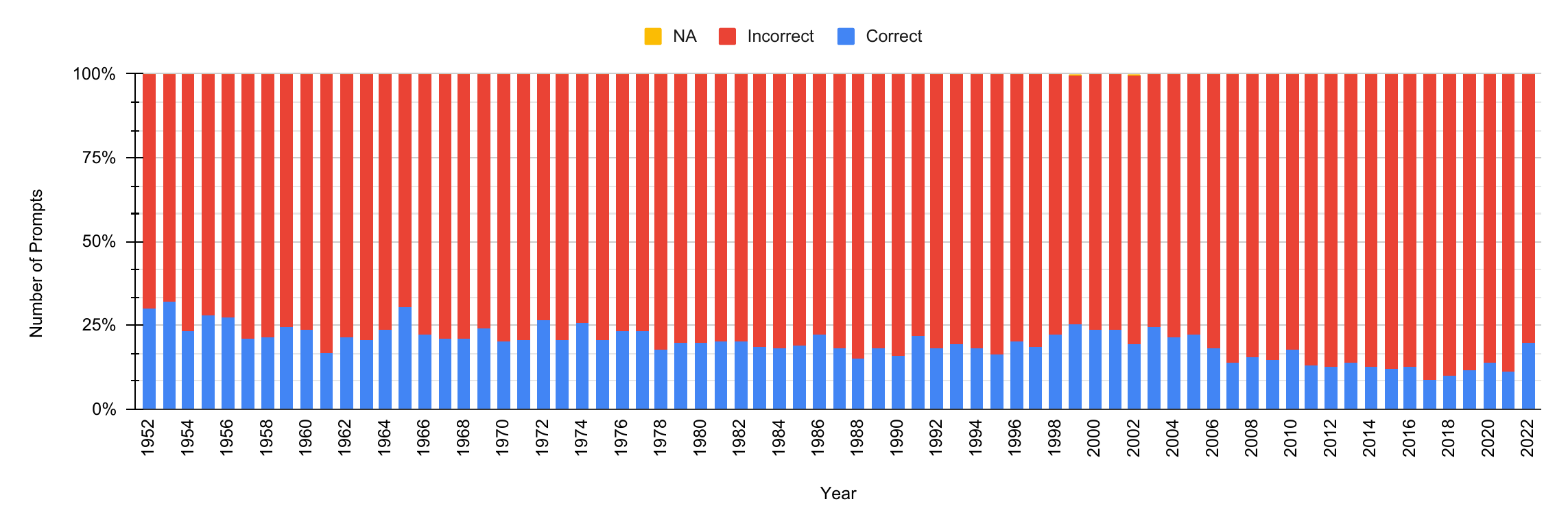}
\end{center}
\caption{Plot for the Min/Max-based metric ($MM$) for year-wise count for \texttt{gpt-3.5} in \textbf{Zeroshot evaluation}. }
\label{fig:mmb-based-gpt-35}
\end{figure*}

\begin{figure*}
\begin{center}
\includegraphics[width=0.9\linewidth]{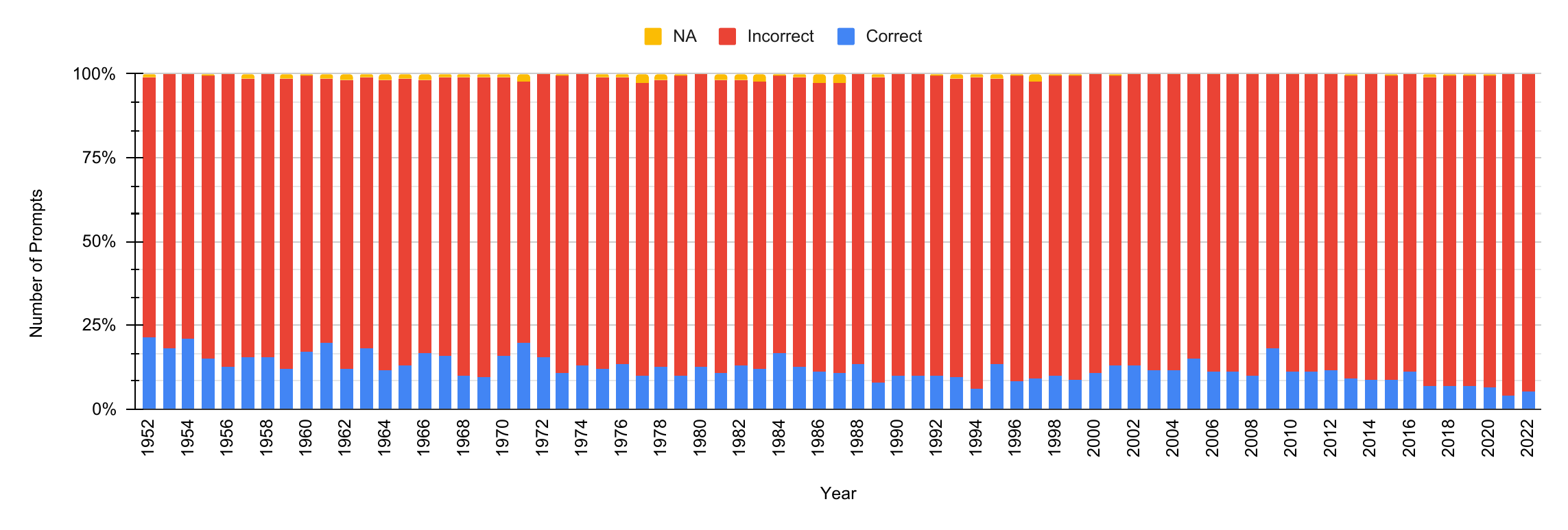}
\end{center}
\caption{Plot for the Range-based metric ($RB$) for year-wise count for \texttt{gpt-3.5} in \textbf{Zeroshot evaluation}. }
\label{fig:rab-based-gpt-35}
\end{figure*}

\begin{figure*}
\begin{center}
\includegraphics[width=0.9\linewidth]{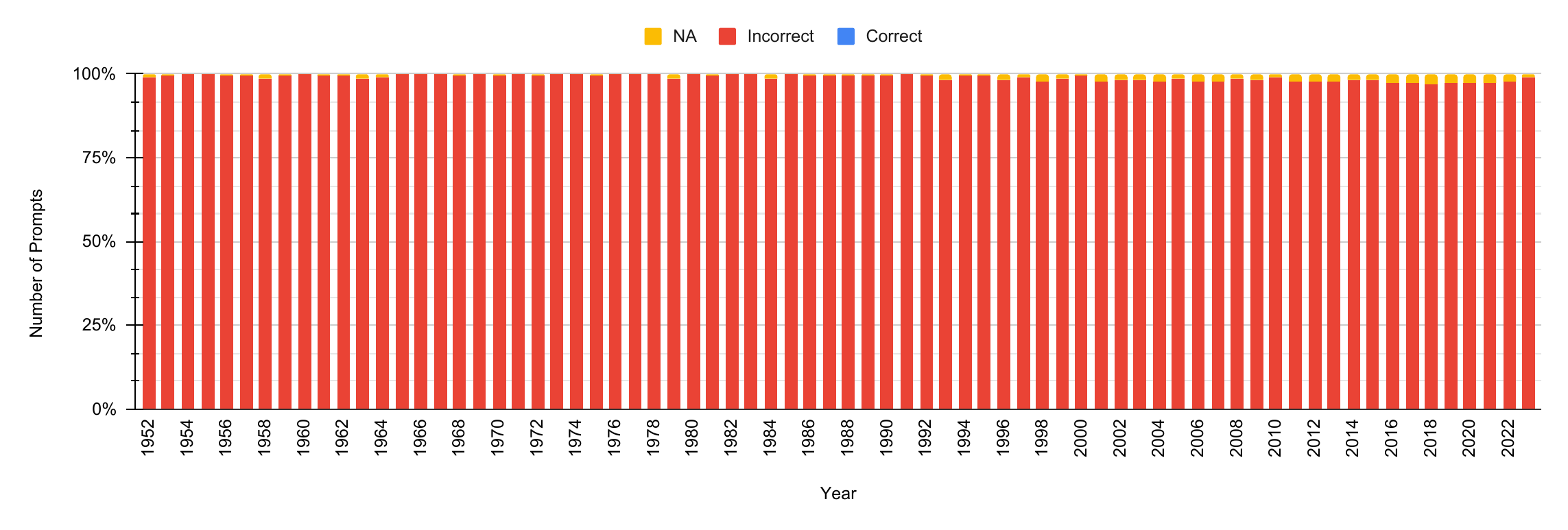}
\end{center}
\caption{Plot for the Trend-based metric ($TB$) for year-wise count for \texttt{gpt-3.5} in \textbf{Zeroshot evaluation}. }
\label{fig:trend-based-gpt-35}
\end{figure*}

\begin{figure*}
\begin{center}
\includegraphics[width=0.9\linewidth]{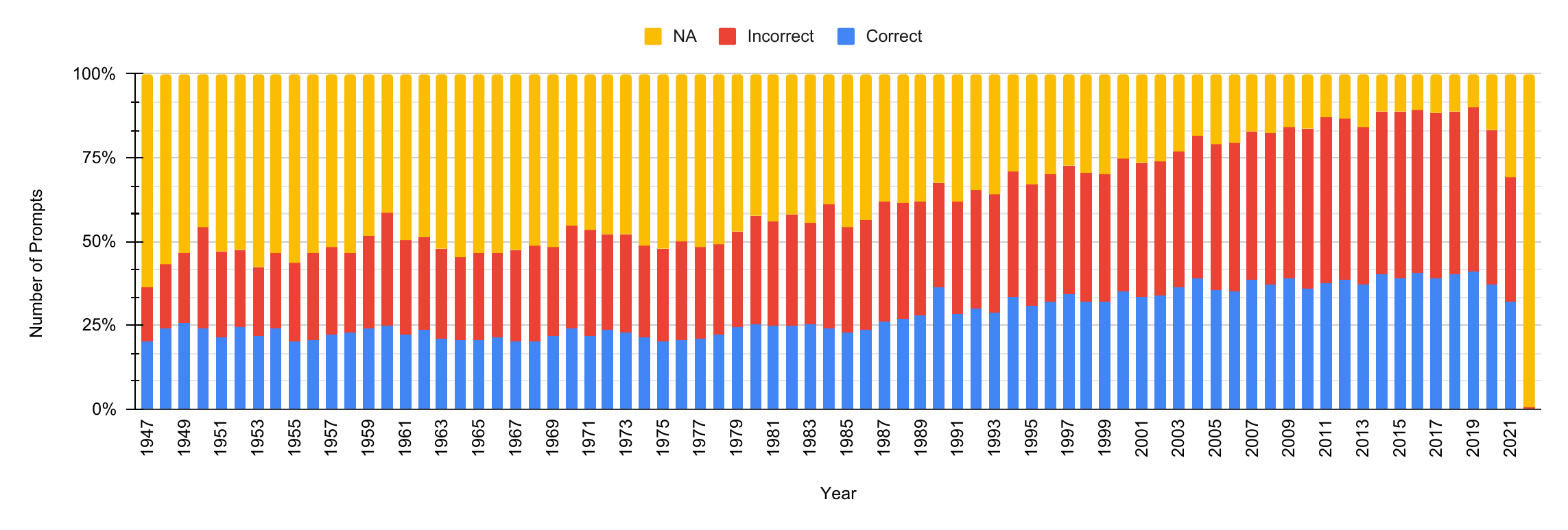}
\end{center}
\caption{Plot for the Date-based metric ($DB$) for year-wise count for \texttt{gpt-4} in \textbf{Zeroshot evaluation}. }
\label{fig:date-based-gpt-4}
\end{figure*}

\begin{figure*}
\begin{center}
\includegraphics[width=0.9\linewidth]{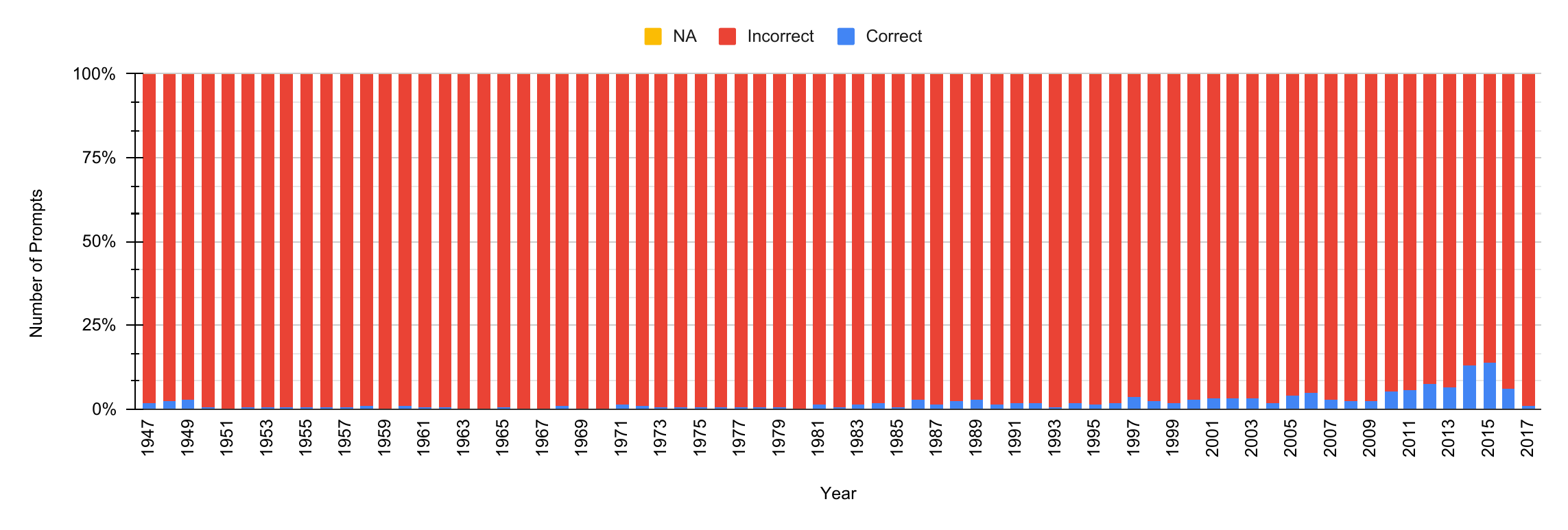}
\end{center}
\caption{Plot for the Comparative-based metric ($CP$) for year-wise count for \texttt{gpt-4} in \textbf{Zeroshot evaluation}. }
\label{fig:range-based-gpt-4}
\end{figure*}

\begin{figure*}
\begin{center}
\includegraphics[width=0.9\linewidth]{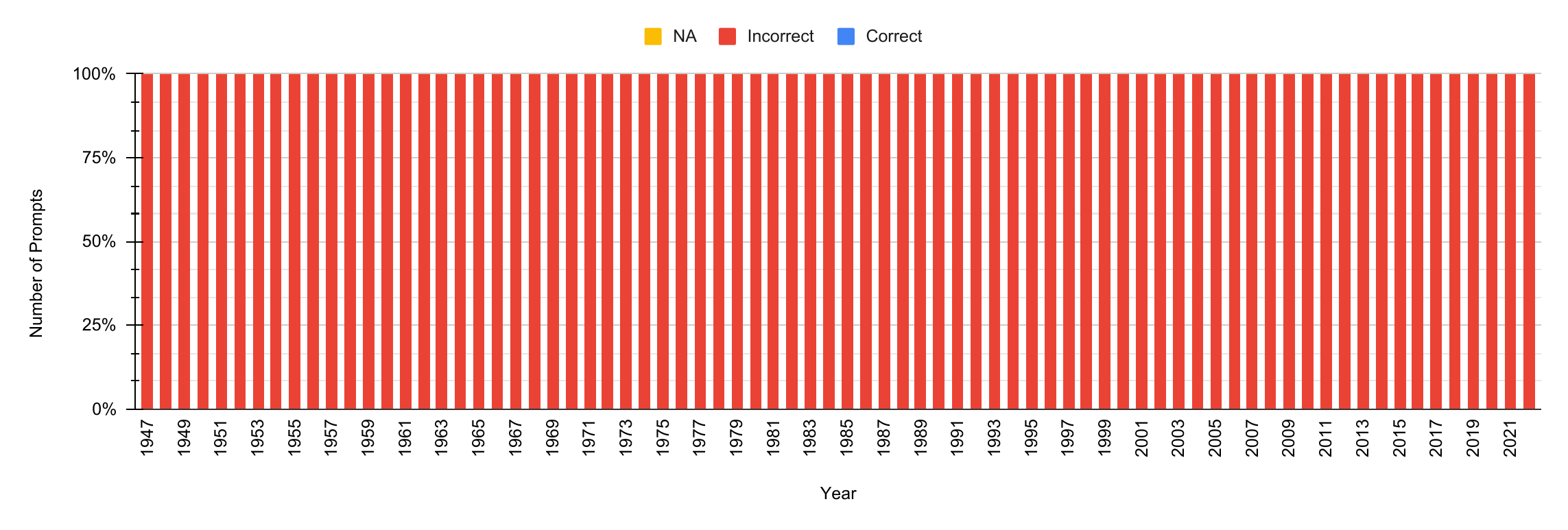}
\end{center}
\caption{Plot for the Window-based metric ($WB$) for year-wise count for \texttt{gpt-4} in \textbf{Zeroshot evaluation}. }
\label{fig:window-based-gpt-4}
\end{figure*}

\begin{figure*}
\begin{center}
\includegraphics[width=0.9\linewidth]{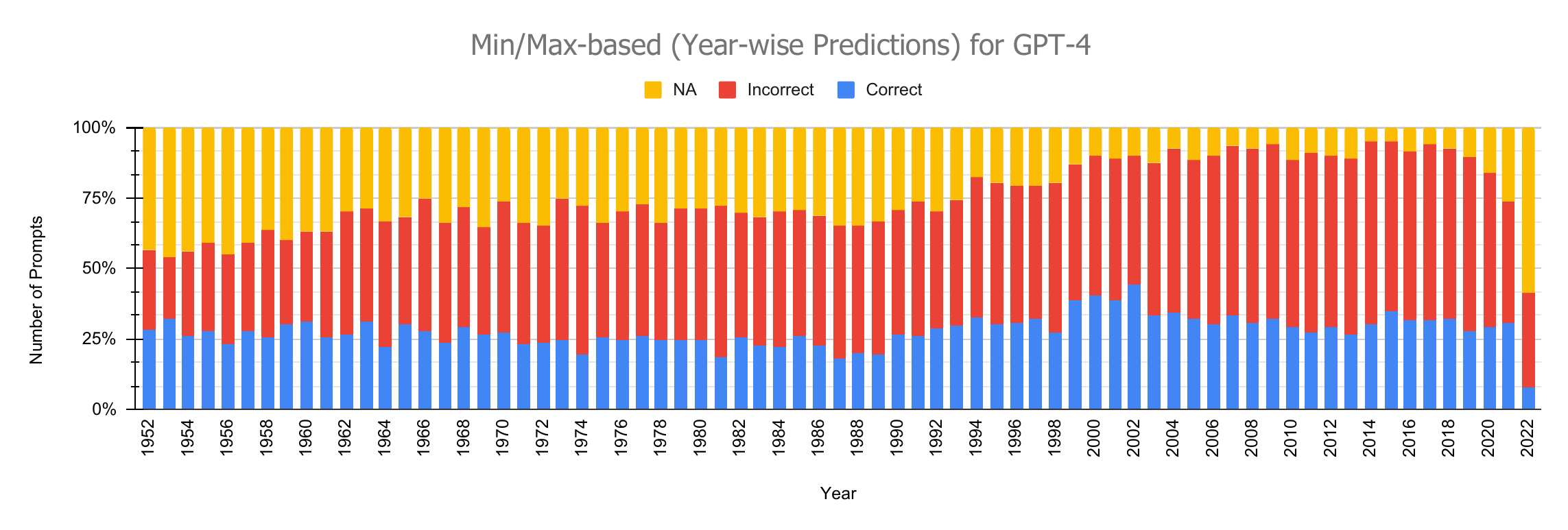}
\end{center}
\caption{Plot for the Min/Max-based metric ($MM$) for year-wise count for \texttt{gpt-4} in \textbf{Zeroshot evaluation}. }
\label{fig:mmb-based-gpt-4}
\end{figure*}

\begin{figure*}
\begin{center}
\includegraphics[width=0.9\linewidth]{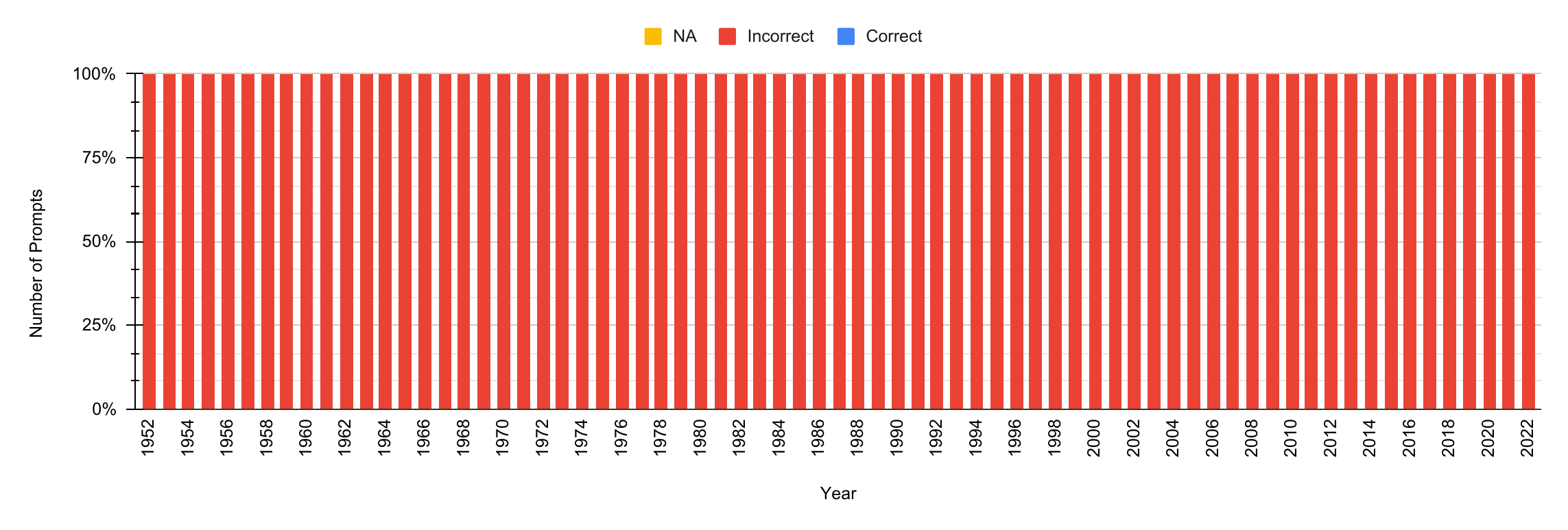}
\end{center}
\caption{Plot for the Range-based metric ($RB$) for year-wise count for \texttt{gpt-4} in \textbf{Zeroshot evaluation}. }
\label{fig:rab-based-gpt-4}
\end{figure*}

\begin{figure*}
\begin{center}
\includegraphics[width=0.9\linewidth]{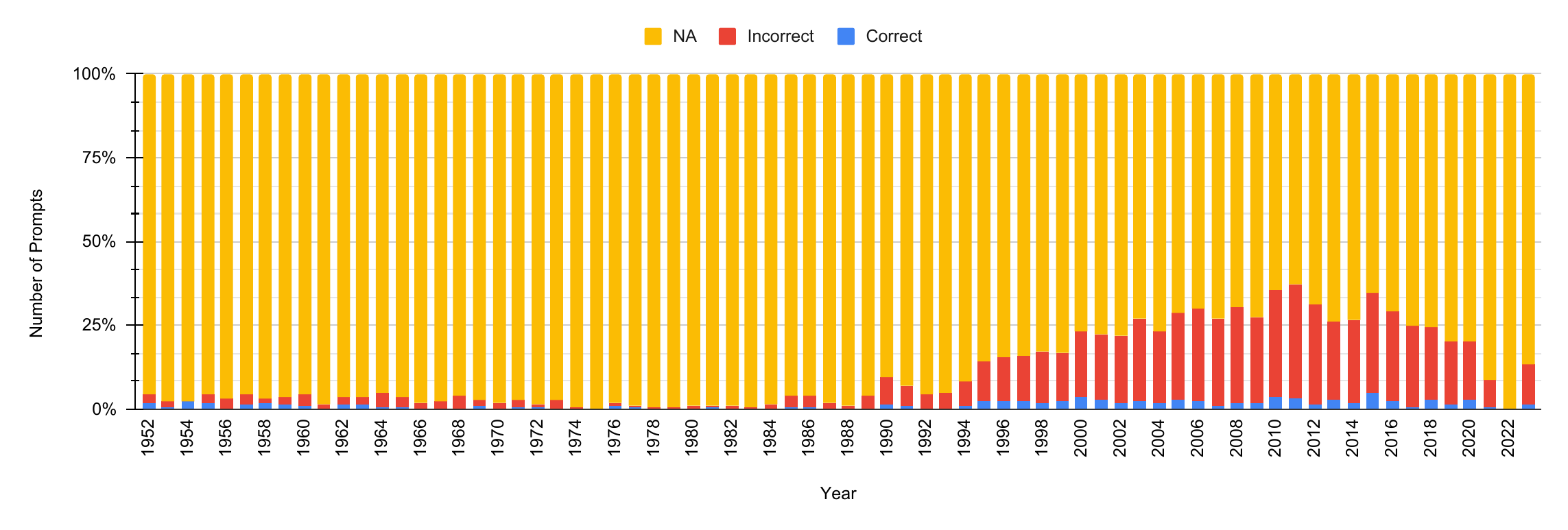}
\end{center}
\caption{Plot for the Trend-based metric ($TB$) for year-wise count for \texttt{gpt-4} in \textbf{Zeroshot evaluation}. }
\label{fig:tb-based-gpt-4}
\end{figure*}

\begin{figure*}
\begin{center}
\includegraphics[width=0.9\linewidth]{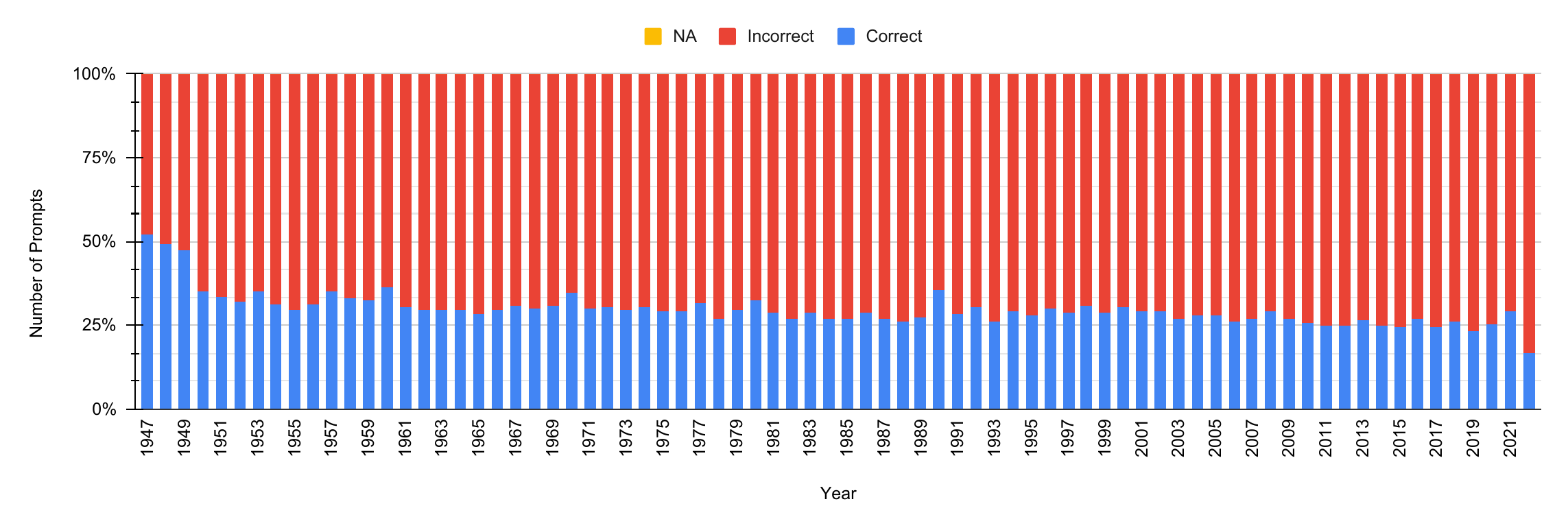}
\end{center}
\caption{Plot for the Date-based metric ($DB$) for year-wise count for \texttt{gemini-pro} in \textbf{Zeroshot evaluation}. }
\label{fig:date-based-gemini-pro}
\end{figure*}

\begin{figure*}
\begin{center}
\includegraphics[width=0.9\linewidth]{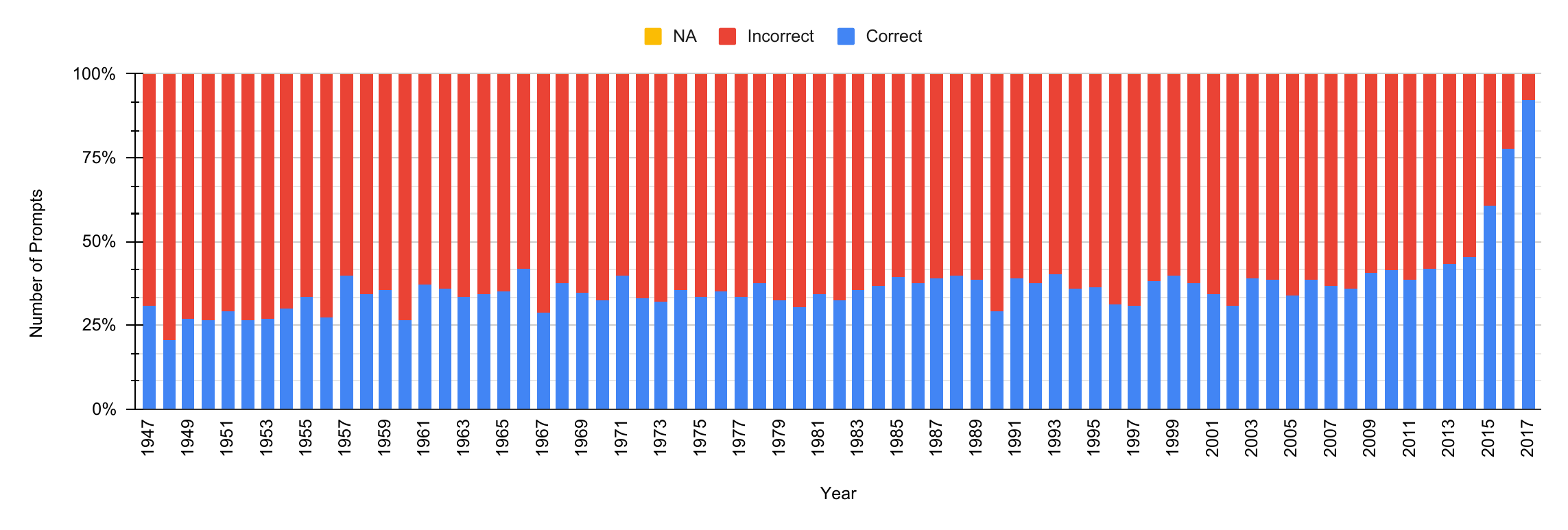}
\end{center}
\caption{Plot for the Comparative-based metric ($CP$) for year-wise count for \texttt{gemini-pro} in \textbf{Zeroshot evaluation}. }
\label{fig:range-based-gemini-pro}
\end{figure*}

\begin{figure*}
\begin{center}
\includegraphics[width=0.9\linewidth]{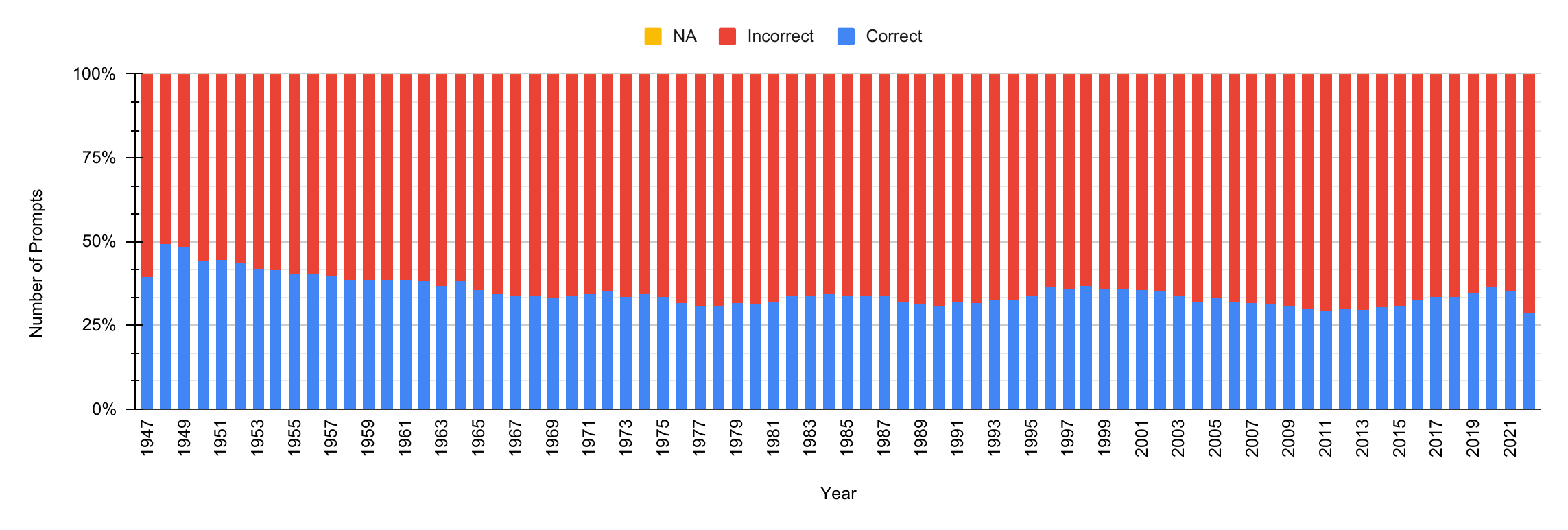}
\end{center}
\caption{Plot for the Window-based metric ($WB$) for year-wise count for \texttt{gemini-pro} in \textbf{Zeroshot evaluation}. }
\label{fig:window-based-gemini-pro}
\end{figure*}

\begin{figure*}
\begin{center}
\includegraphics[width=0.9\linewidth]{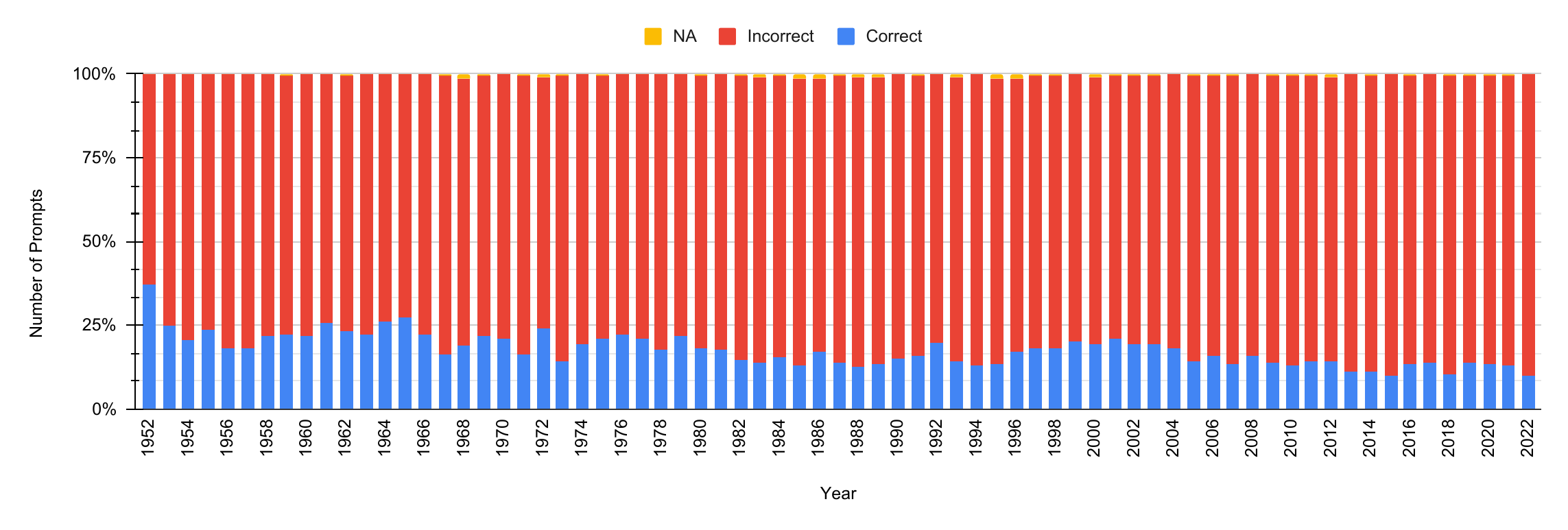}
\end{center}
\caption{Plot for the Min/Max-based metric ($MM$) for year-wise count for \texttt{gemini-pro} in \textbf{Zeroshot evaluation}. }
\label{fig:mmb-based-gemini-pro}
\end{figure*}

\begin{figure*}
\begin{center}
\includegraphics[width=0.9\linewidth]{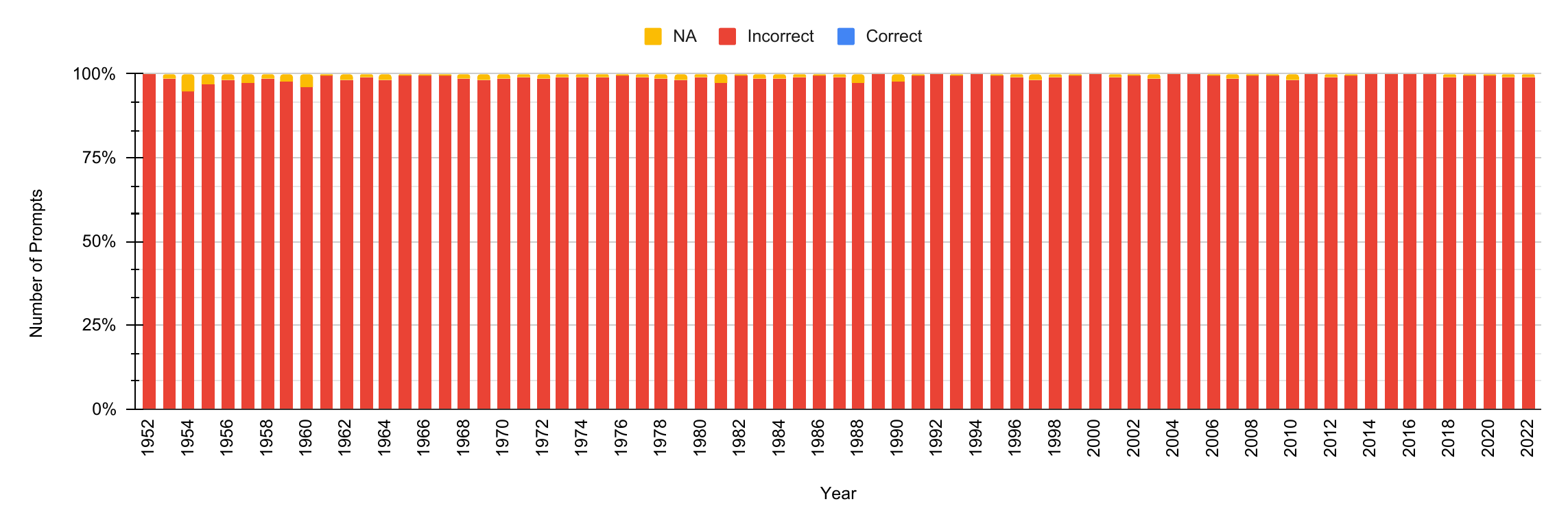}
\end{center}
\caption{Plot for the Range-based metric ($RB$) for year-wise count for \texttt{gemini-pro} in \textbf{Zeroshot evaluation}. }
\label{fig:rab-based-gemini-pro}
\end{figure*}

\begin{figure*}
\begin{center}
\includegraphics[width=0.9\linewidth]{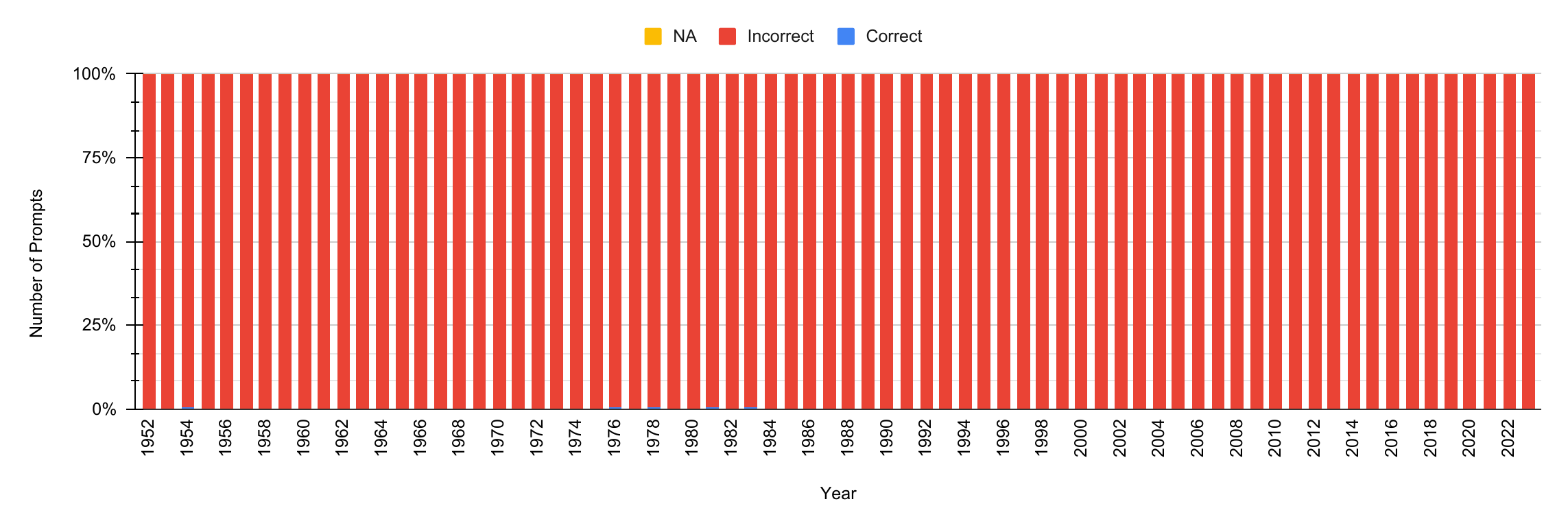}
\end{center}
\caption{Plot for the Trend-based metric ($TB$) for year-wise count for \texttt{gemini-pro} in \textbf{Zeroshot evaluation}. }
\label{fig:trend-based-gemini-pro}
\end{figure*}


\begin{figure*}
\begin{center}
\includegraphics[width=0.9\linewidth]{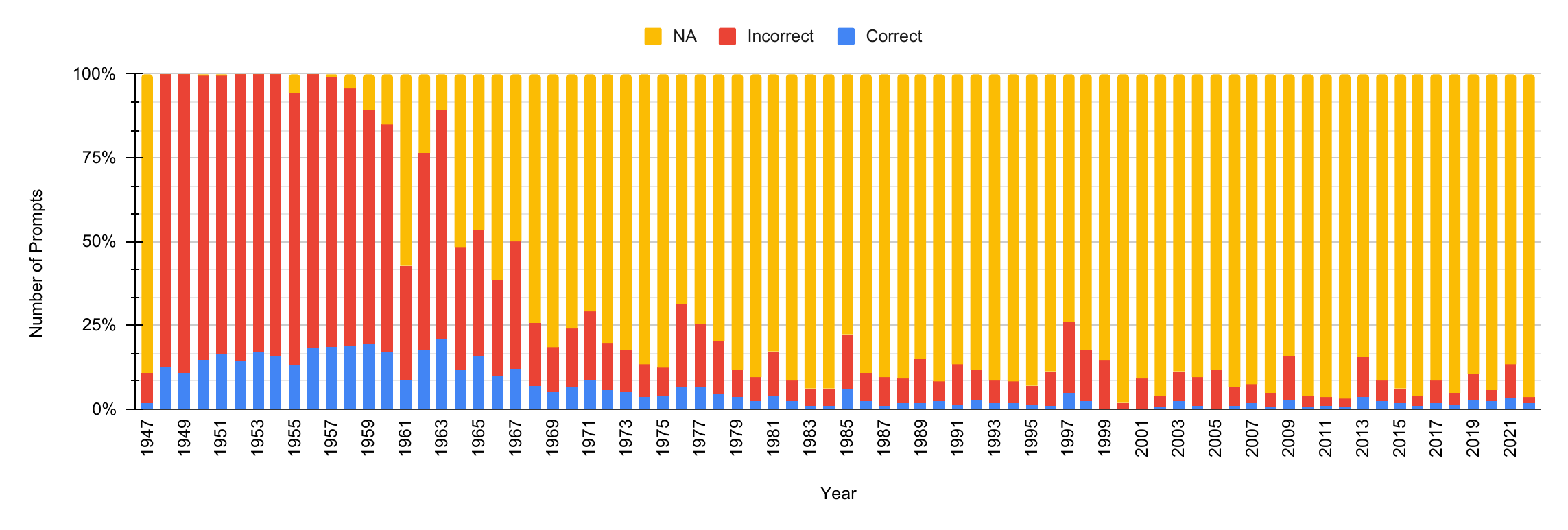}
\end{center}
\caption{Plot for the Date-based metric ($DB$) as year-wise count (In percentage) for \textbf{continual fine-tuning} for \texttt{phi-2}.}
\label{fig:date-based-ft-phi-2}
\end{figure*}

\begin{figure*}
\begin{center}
\includegraphics[width=0.9\linewidth]{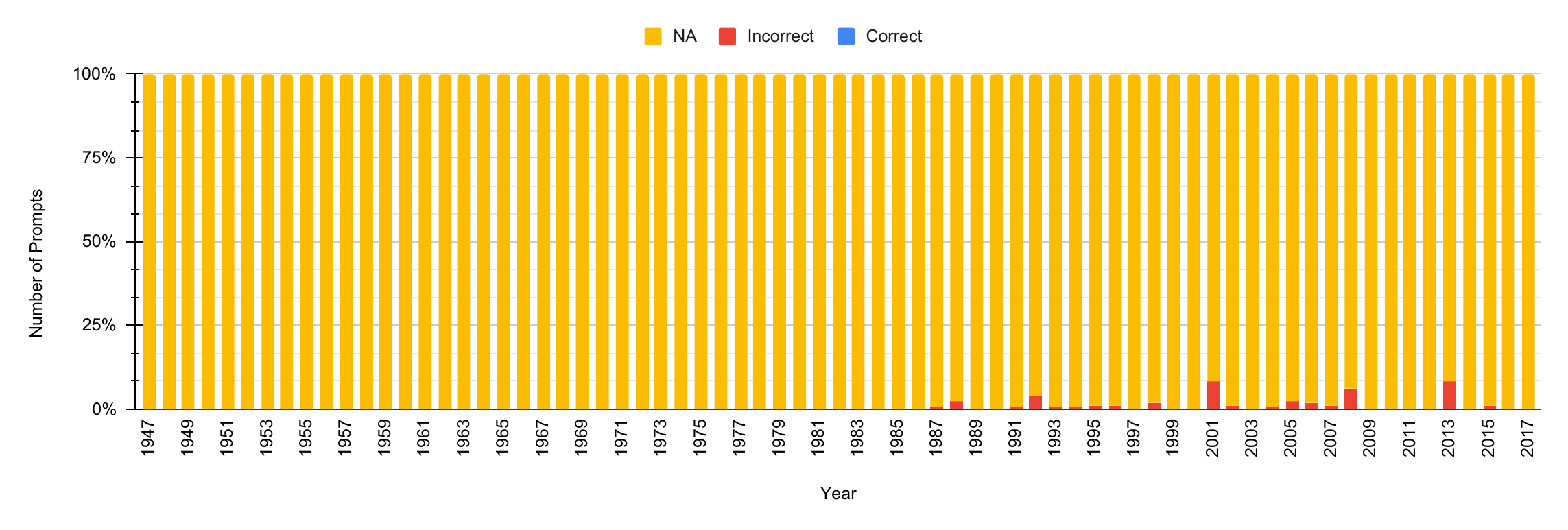}
\end{center}
\caption{Plot for the Comparative-based metric ($CP$) as year-wise count (In percentage) for \textbf{continual fine-tuning} for \texttt{phi-2}.}
\label{fig:rb-based-ft-phi-2}
\end{figure*}

\begin{figure*}
\begin{center}
\includegraphics[width=0.9\linewidth]{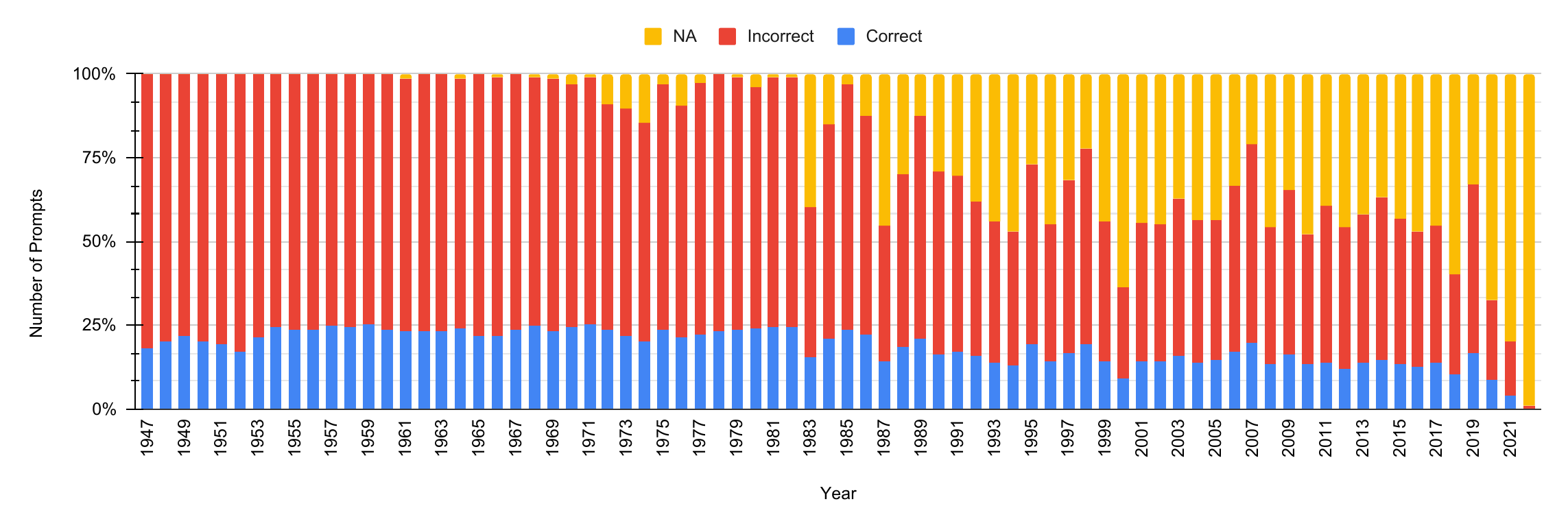}
\end{center}
\caption{Plot for the Window-based metric ($WB$) as year-wise count (In percentage) for \textbf{continual fine-tuning} for \texttt{phi-2}.}
\label{fig:window-based-ft-phi-2}
\end{figure*}

\begin{figure*}
\begin{center}
\includegraphics[width=0.9\linewidth]{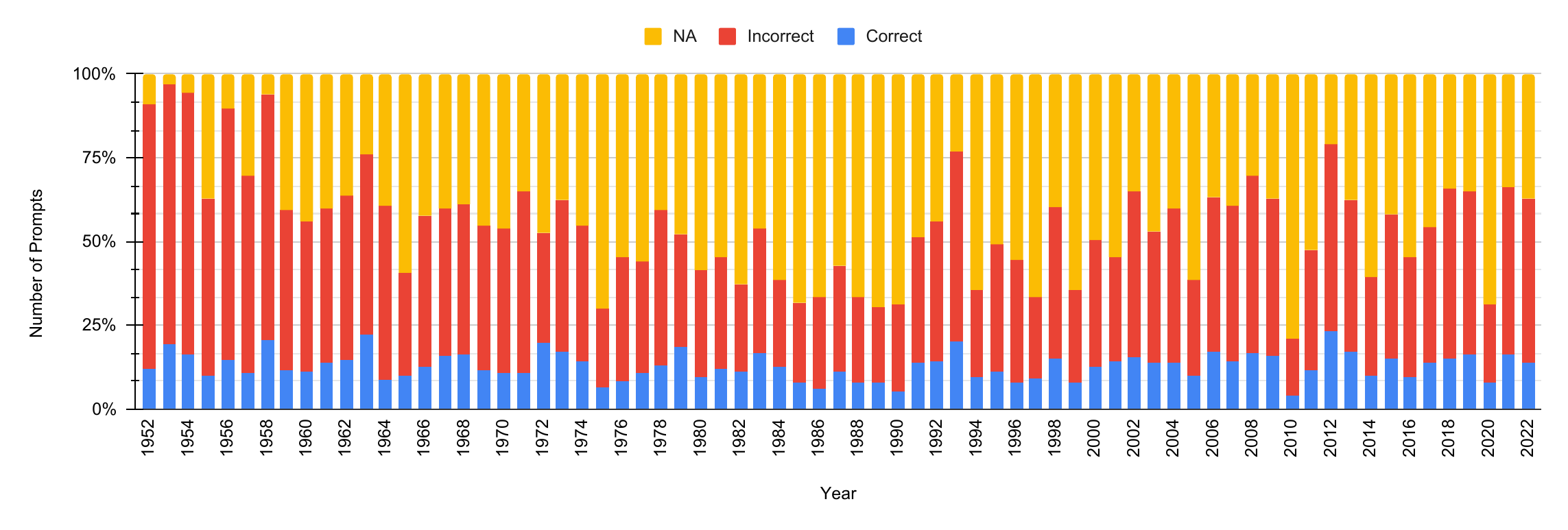}
\end{center}
\caption{Plot for the Min/Max-based metric ($MM$) as year-wise count (In percentage) for \textbf{continual fine-tuning} for \texttt{phi-2}.}
\label{fig:minmax-based-ft-phi-2}
\end{figure*}

\begin{figure*}
\begin{center}
\includegraphics[width=0.9\linewidth]{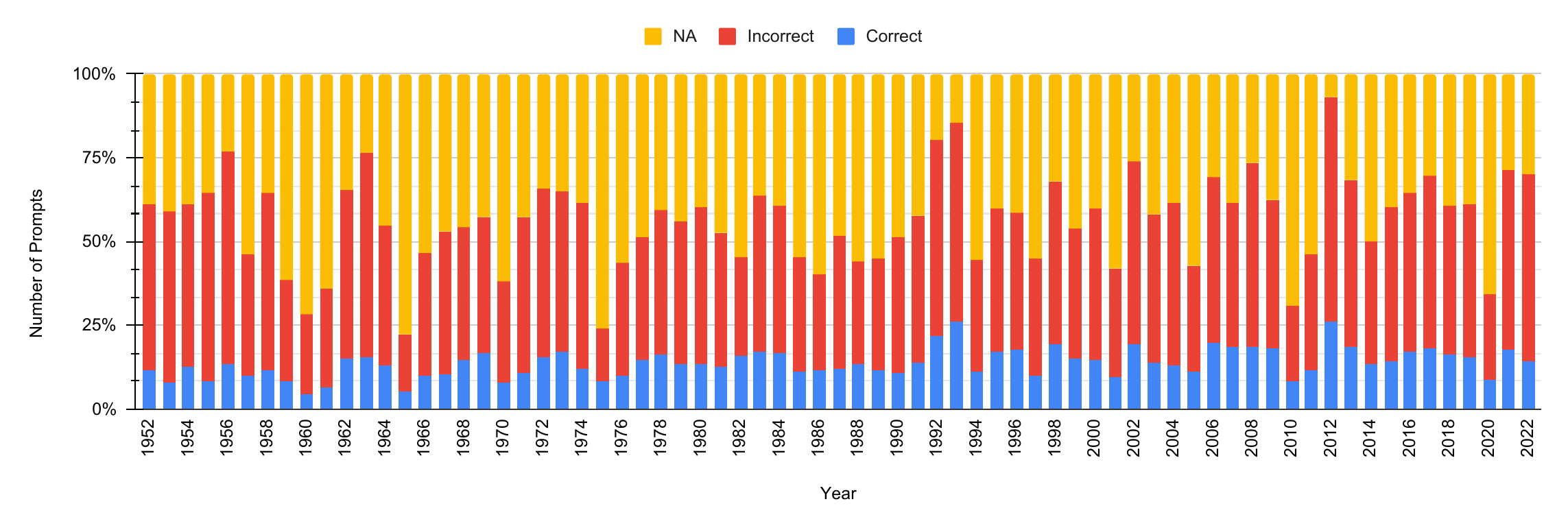}
\end{center}
\caption{Plot for the Range-based metric ($RB$) as year-wise count (In percentage) for \textbf{continual fine-tuning} for \texttt{phi-2}.}
\label{fig:rab-based-ft-phi-2}
\end{figure*}

\begin{figure*}
\begin{center}
\includegraphics[width=0.9\linewidth]{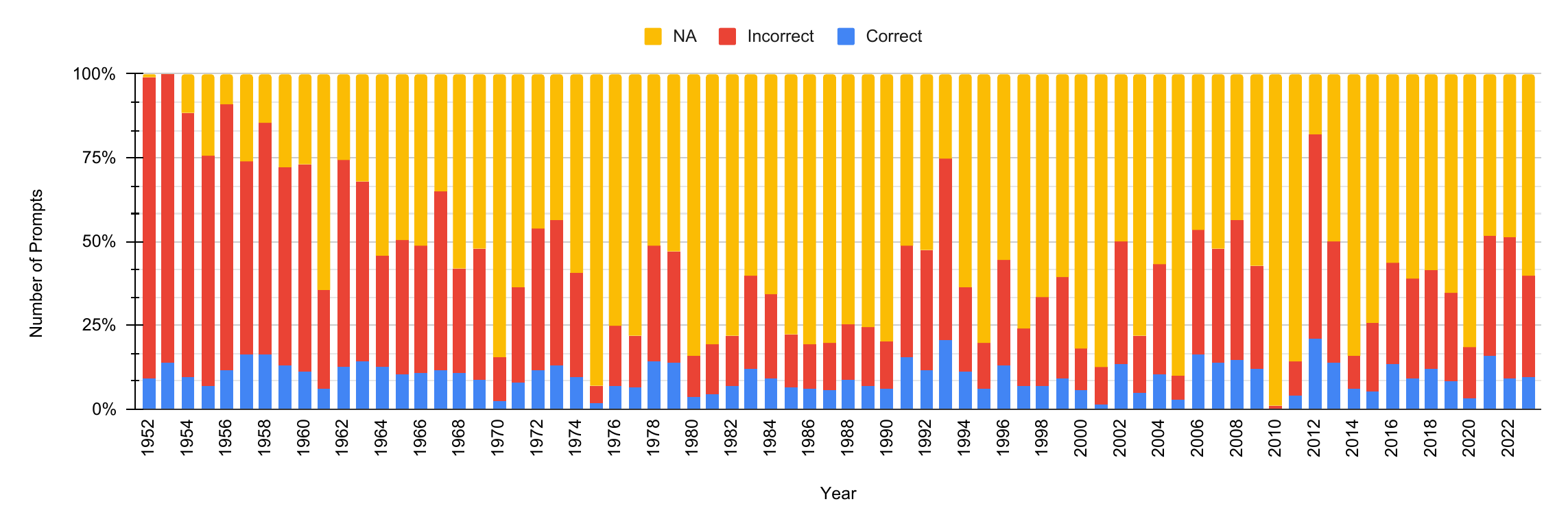}
\end{center}
\caption{Plot for the Trend-based metric ($TB$) as year-wise count (In percentage) for \textbf{continual fine-tuning} for \texttt{phi-2}.}
\label{fig:tb-based-ft-phi-2}
\end{figure*}

\begin{figure*}
\begin{center}
\includegraphics[width=0.9\linewidth]{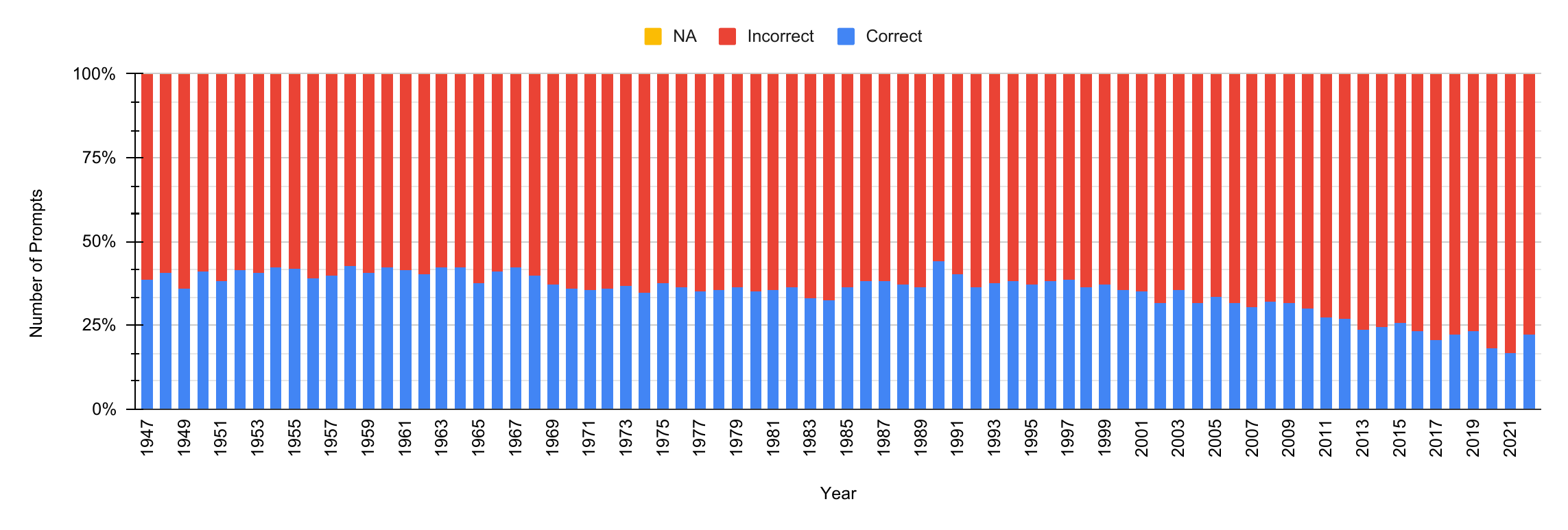}
\end{center}
\caption{Plot for the Date-based metric ($DB$) as year-wise count (In percentage) for \textbf{continual fine-tuning} for \texttt{flan-t5-xl}.}
\label{fig:date-based-ft-flan-t5-xl}
\end{figure*}

\begin{figure*}
\begin{center}
\includegraphics[width=0.9\linewidth]{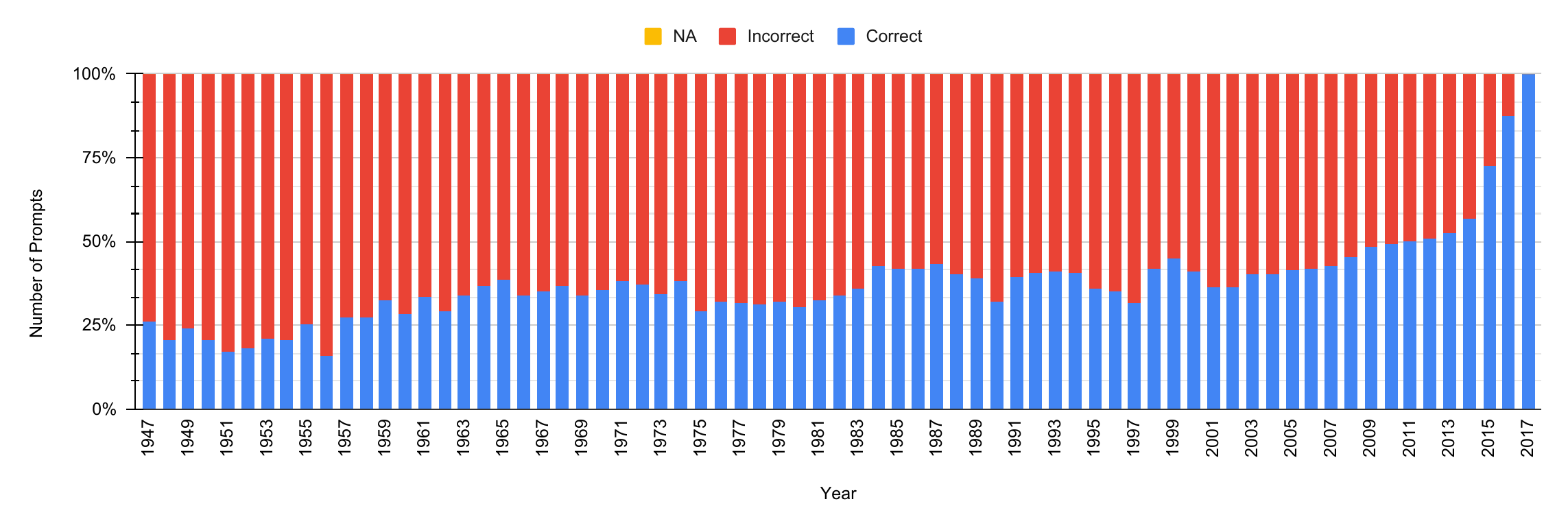}
\end{center}
\caption{Plot for the Comparative-based metric ($CP$) as year-wise count (In percentage) for \textbf{continual fine-tuning} for \texttt{flan-t5-xl}.}
\label{fig:rb-based-ft-flan-t5-xl}
\end{figure*}

\begin{figure*}
\begin{center}
\includegraphics[width=0.9\linewidth]{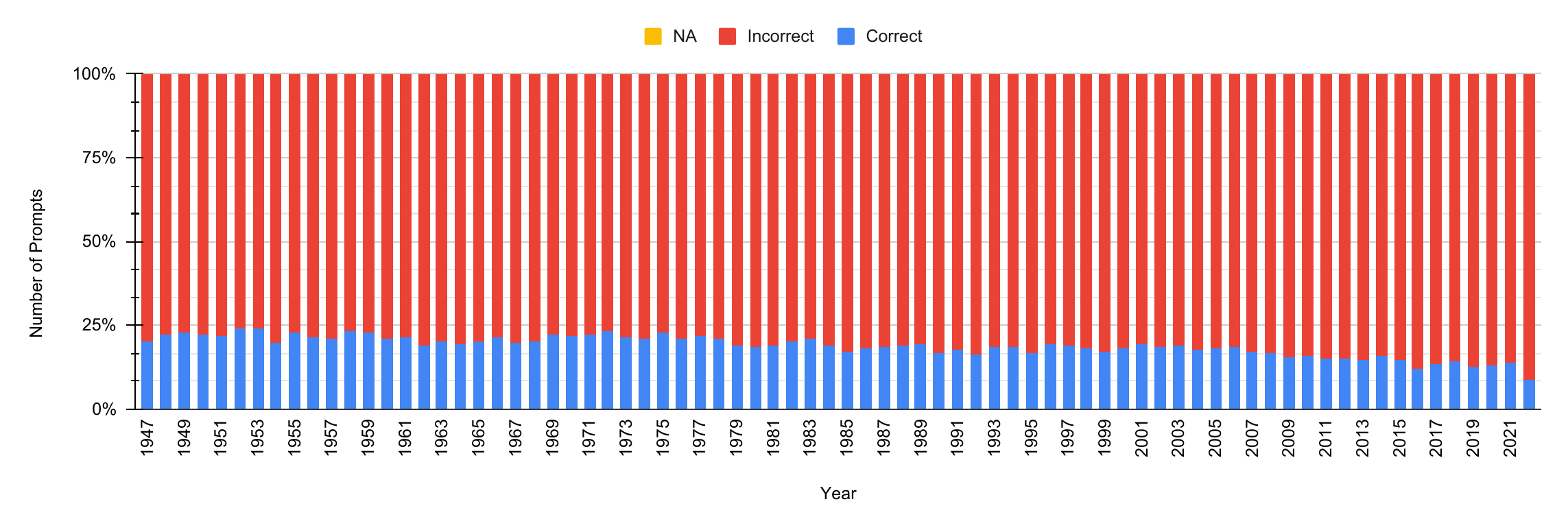}
\end{center}
\caption{Plot for the Window-based metric ($WB$) as year-wise count (In percentage) for \textbf{continual fine-tuning} for \texttt{flan-t5-xl}.}
\label{fig:window-based-ft-flan-t5-xl}
\end{figure*}

\begin{figure*}
\begin{center}
\includegraphics[width=0.9\linewidth]{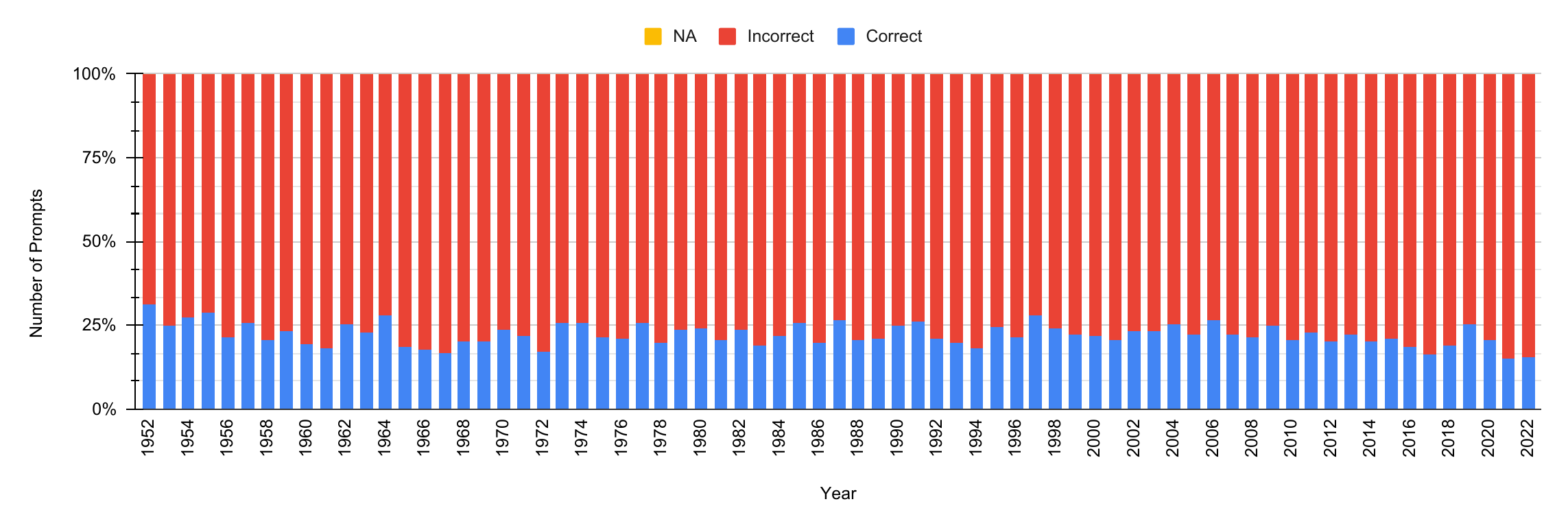}
\end{center}
\caption{Plot for the Min/Max-based metric ($MM$) as year-wise count (In percentage) for \textbf{continual fine-tuning} for \texttt{flan-t5-xl}.}
\label{fig:minmax-based-ft-flan-t5-xl}
\end{figure*}

\begin{figure*}
\begin{center}
\includegraphics[width=0.9\linewidth]{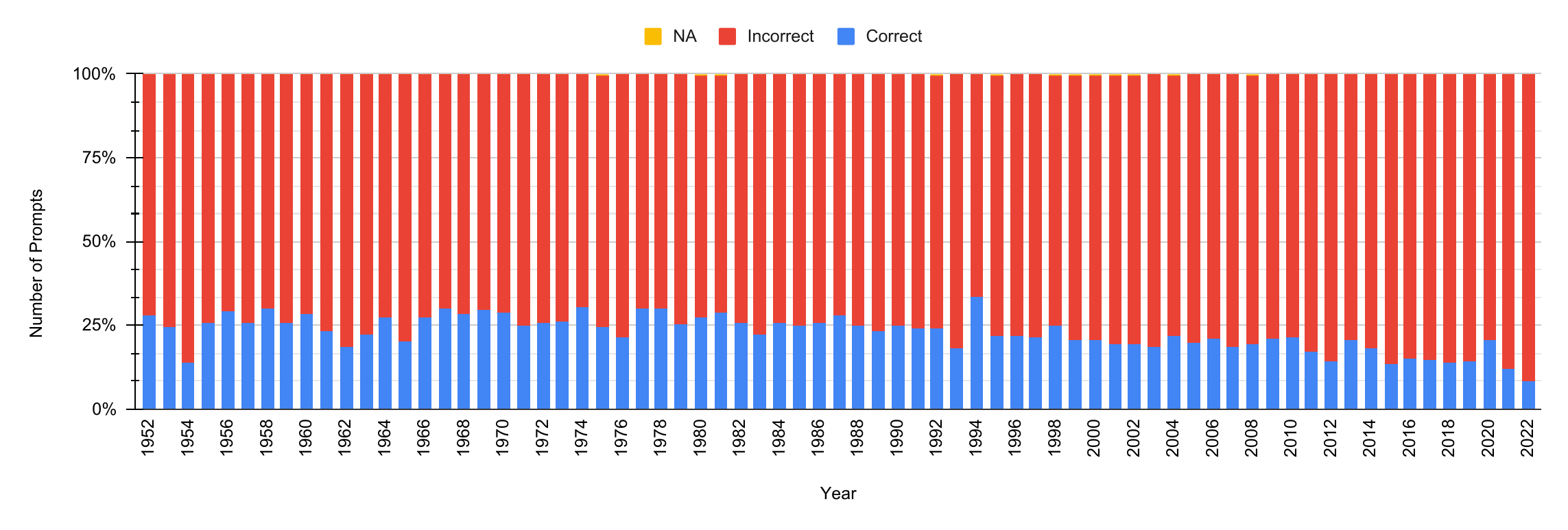}
\end{center}
\caption{Plot for the Range-based metric ($RB$) as year-wise count (In percentage) for \textbf{continual fine-tuning} for \texttt{flan-t5-xl}.}
\label{fig:rab-based-ft-flan-t5-xl}
\end{figure*}

\begin{figure*}
\begin{center}
\includegraphics[width=0.9\linewidth]{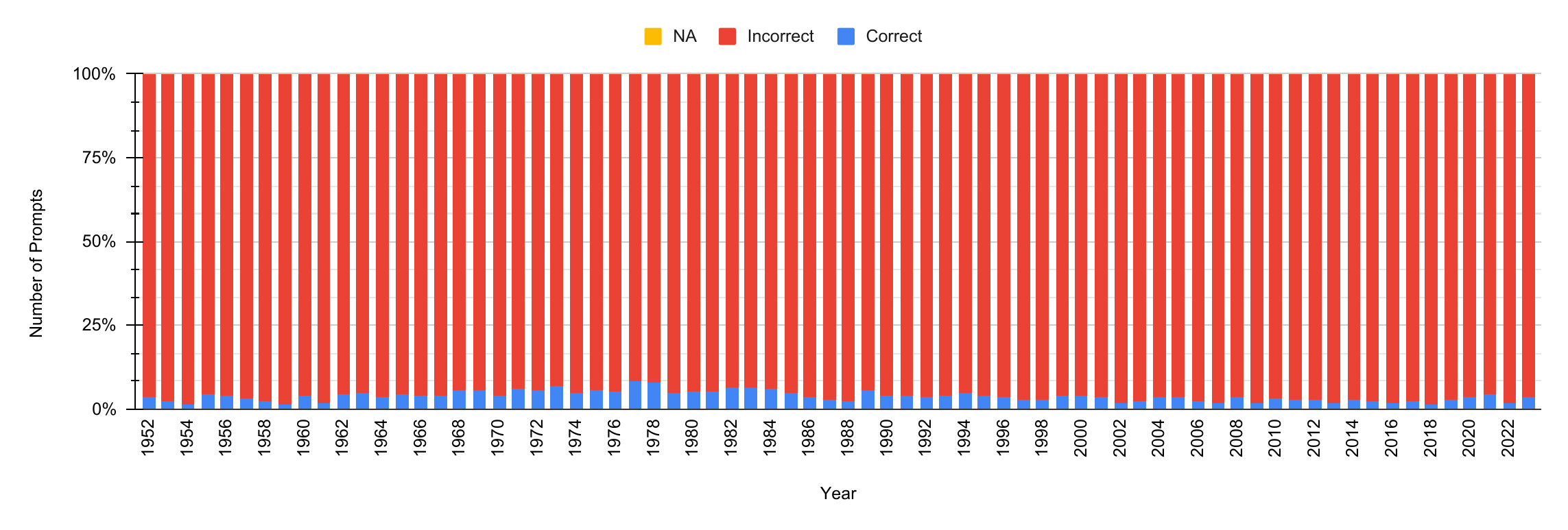}
\end{center}
\caption{Plot for the Trend-based metric ($TB$) as year-wise count (In percentage) for \textbf{continual fine-tuning} for \texttt{flan-t5-xl}.}
\label{fig:tb-based-ft-flan-t5-xl}
\end{figure*}


\begin{figure*}
\begin{center}
\includegraphics[width=0.9\linewidth]{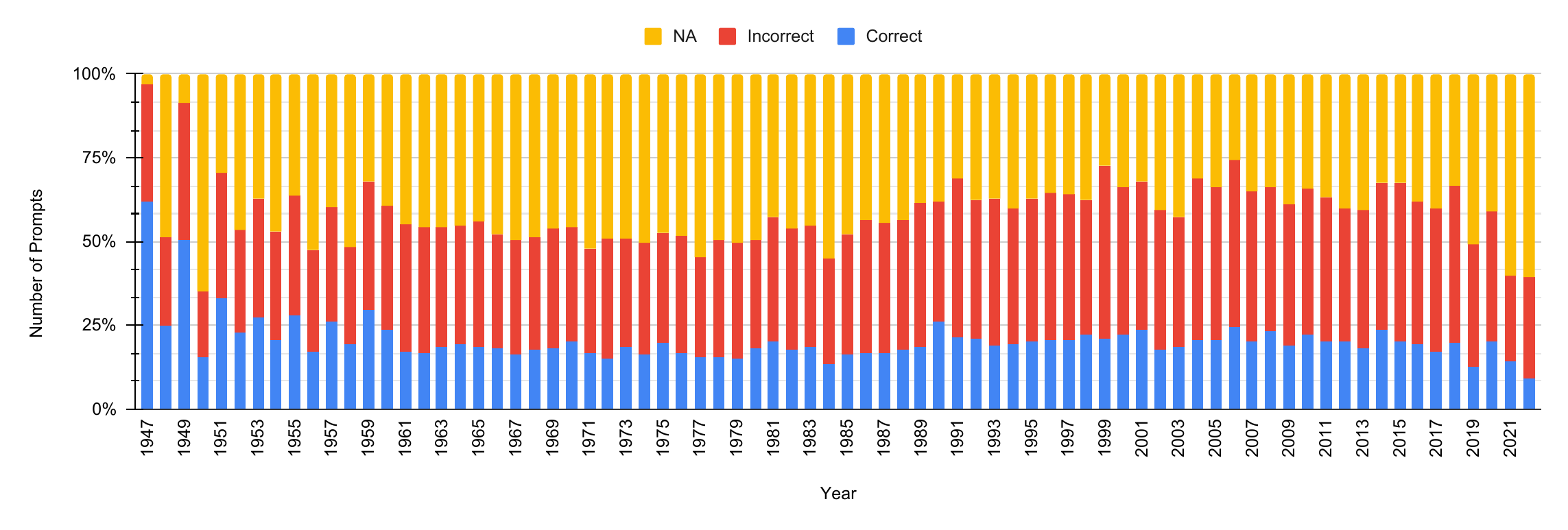}
\end{center}
\caption{Plot for the Date-based metric ($DB$) as year-wise count (In percentage) for \textbf{continual fine-tuning} for \texttt{mistral-instruct}.}
\label{fig:date-based-ft-mistral}
\end{figure*}

\begin{figure*}
\begin{center}
\includegraphics[width=0.9\linewidth]{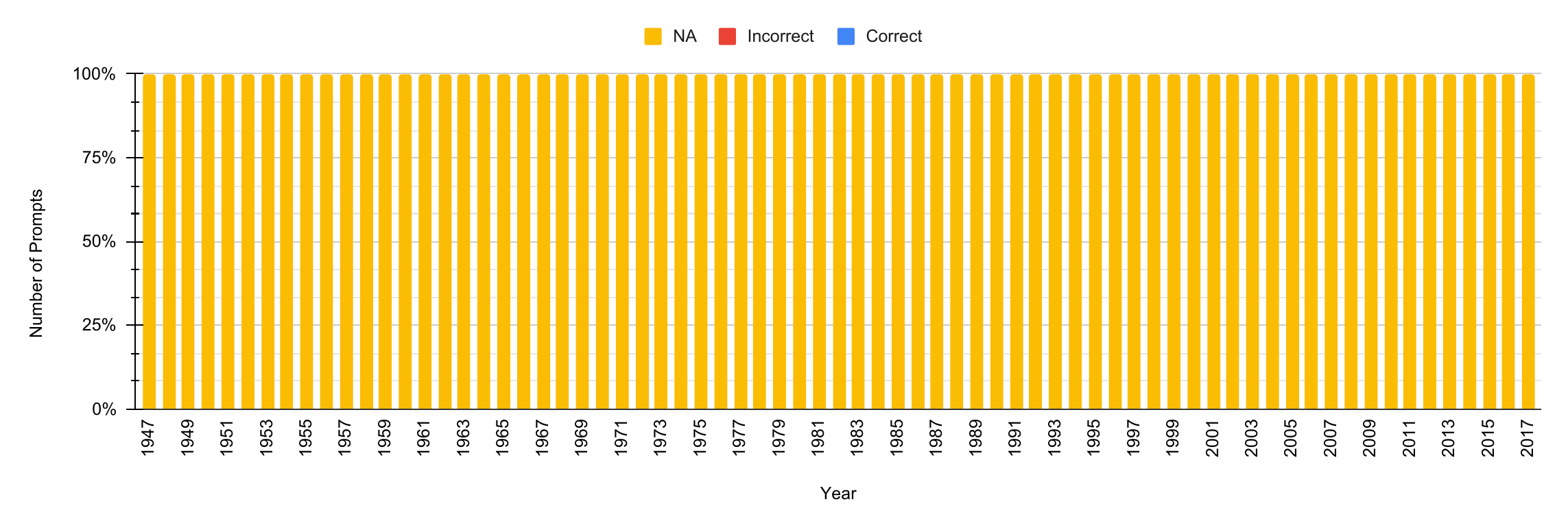}
\end{center}
\caption{Plot for the Comparative-based metric ($CP$) as year-wise count (In percentage) for \textbf{continual fine-tuning} for \texttt{mistral-instruct}.}
\label{fig:range-based-ft-mistral}
\end{figure*}

\begin{figure*}
\begin{center}
\includegraphics[width=0.9\linewidth]{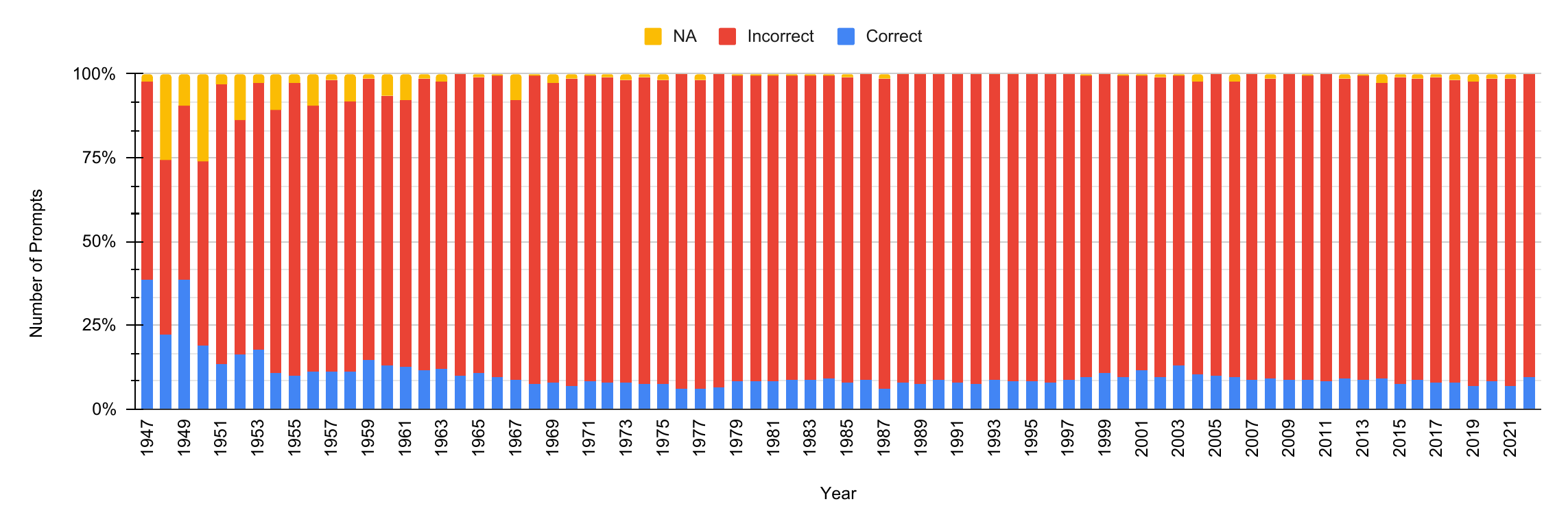}
\end{center}
\caption{Plot for the Window-based metric ($WB$) as year-wise count (In percentage) for \textbf{continual fine-tuning} for \texttt{mistral-instruct}.}
\label{fig:window-based-ft-mistral}
\end{figure*}
\begin{figure*}
\begin{center}
\includegraphics[width=0.9\linewidth]{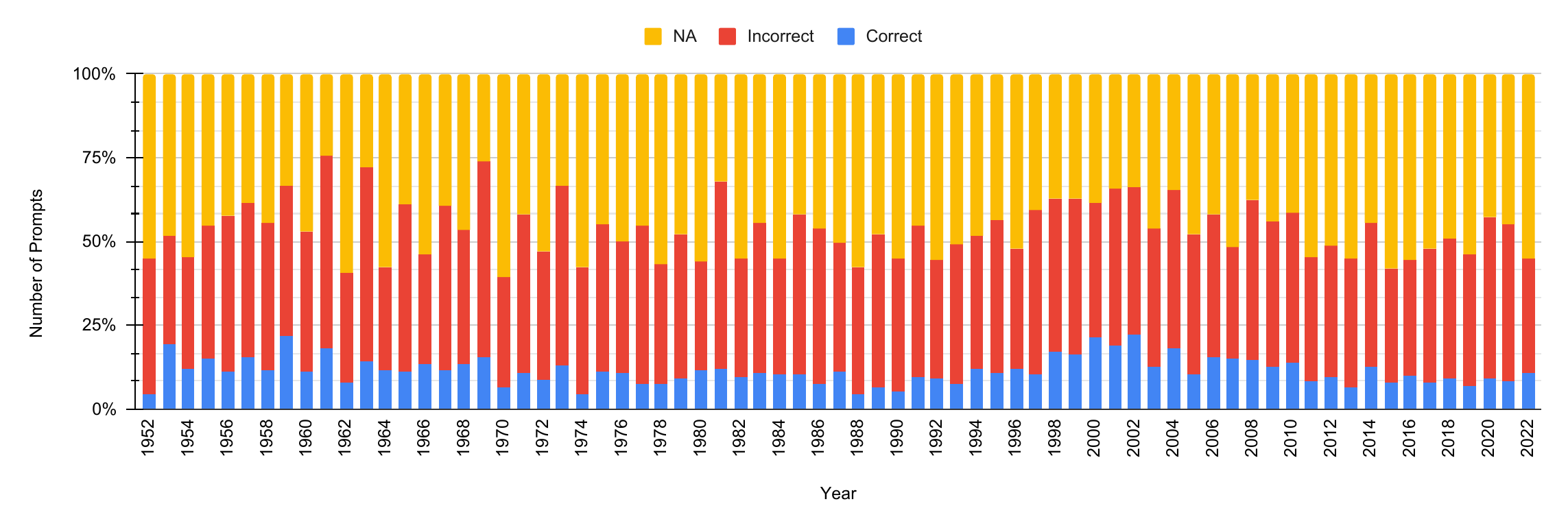}
\end{center}
\caption{Plot for the Min/Max-based metric ($MM$) as year-wise count (In percentage) for \textbf{continual fine-tuning} for \texttt{mistral-instruct}.}
\label{fig:minmax-based-ft-mistral}
\end{figure*}

\begin{figure*}
\begin{center}
\includegraphics[width=0.9\linewidth]{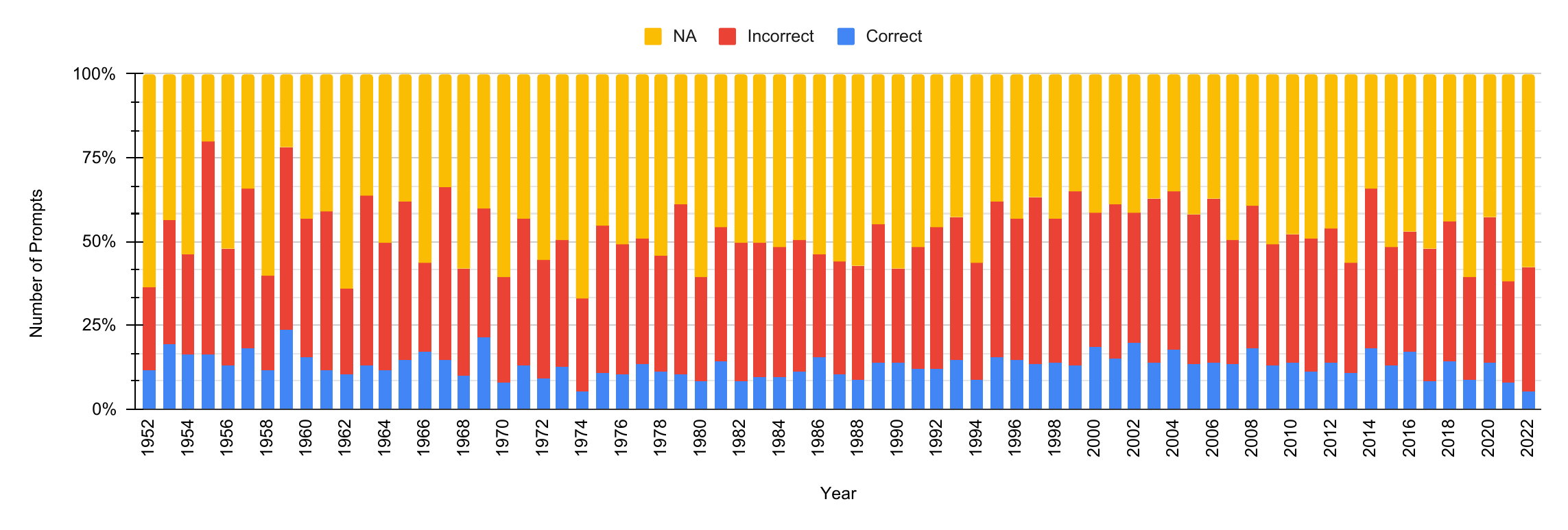}
\end{center}
\caption{Plot for the Range-based metric ($RB$) as year-wise count (In percentage) for \textbf{continual fine-tuning} for \texttt{mistral-instruct}.}
\label{fig:rab-based-ft-mistral}
\end{figure*}

\begin{figure*}
\begin{center}
\includegraphics[width=0.9\linewidth]{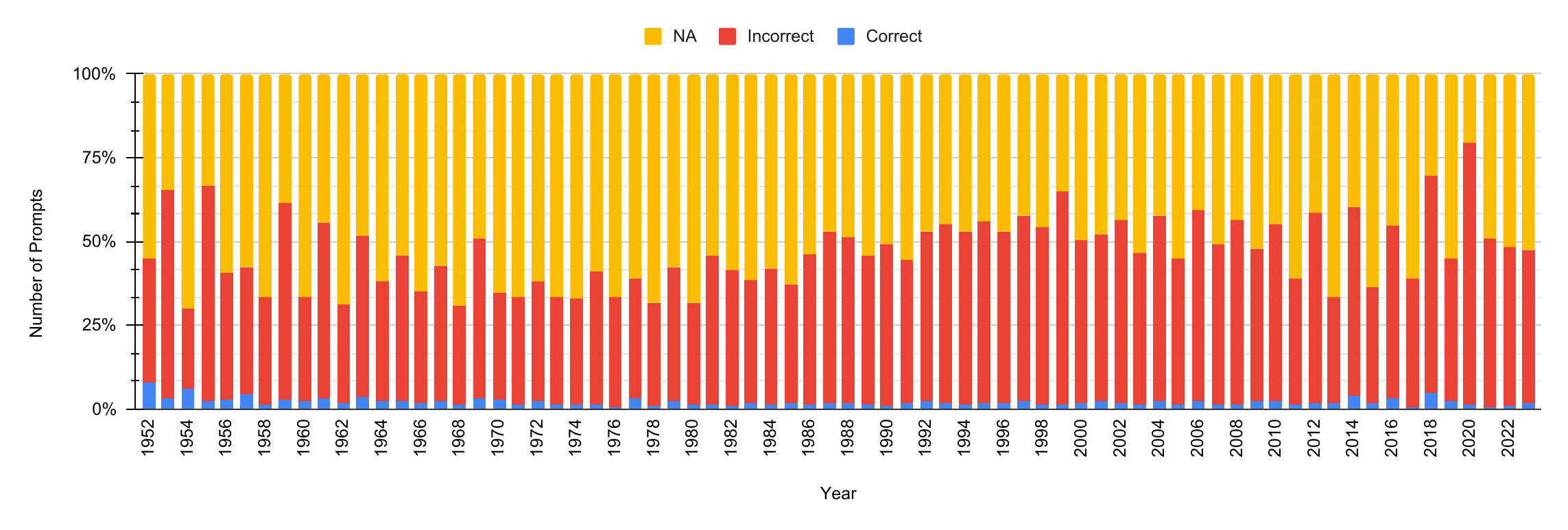}
\end{center}
\caption{Plot for the Trend-based metric ($TB$) as year-wise count (In percentage) for \textbf{continual fine-tuning} for \texttt{mistral-instruct}.}
\label{fig:tb-based-ft-mistral}
\end{figure*}


\begin{figure*}
\begin{center}
\includegraphics[width=0.9\linewidth]{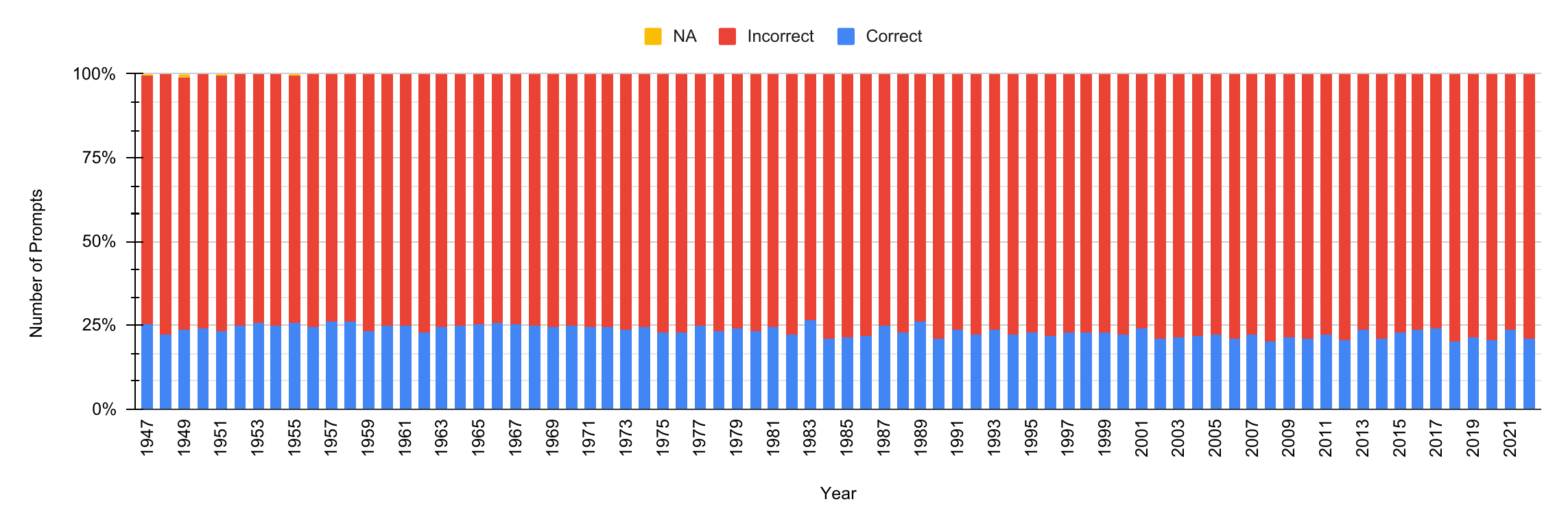}
\end{center}
\caption{Plot for the Date-based metric ($DB$) as year-wise count (In percentage) for \textbf{continual fine-tuning} for \texttt{llama-2}.}
\label{fig:date-based-ft-llama-2}
\end{figure*}

\begin{figure*}
\begin{center}
\includegraphics[width=0.9\linewidth]{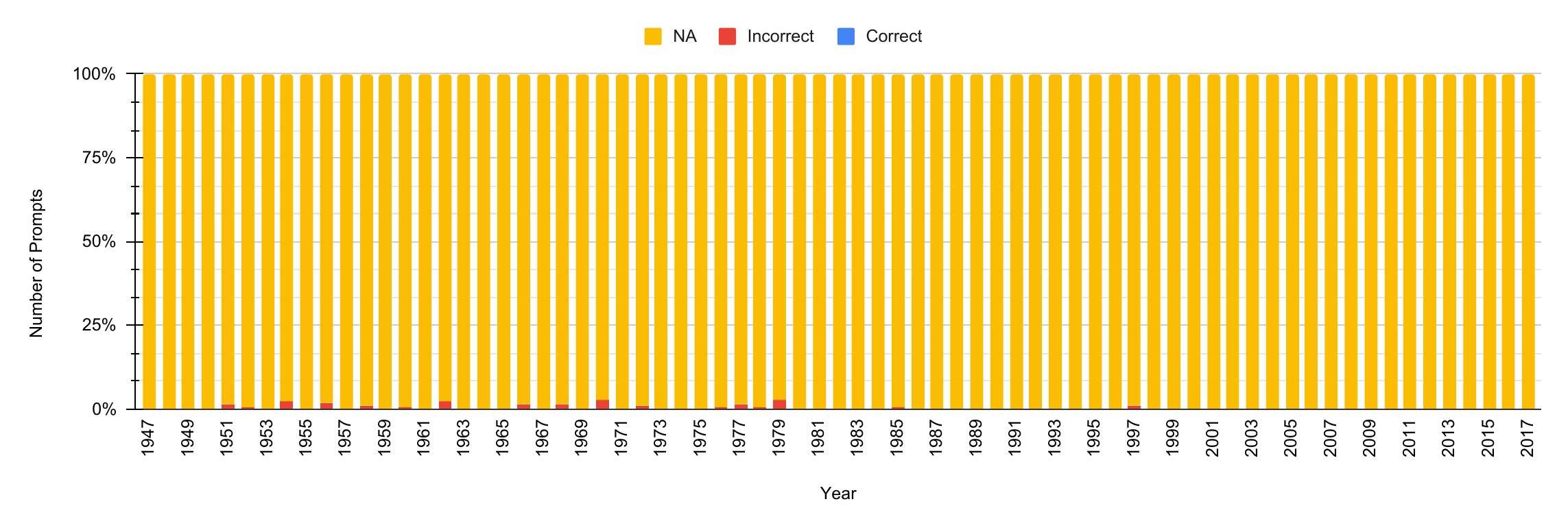}
\end{center}
\caption{Plot for the Comparative-based metric ($CP$) as year-wise count (In percentage) for \textbf{continual fine-tuning} for \texttt{llama-2}.}
\label{fig:rb-based-ft-llama-2}
\end{figure*}

\begin{figure*}
\begin{center}
\includegraphics[width=0.9\linewidth]{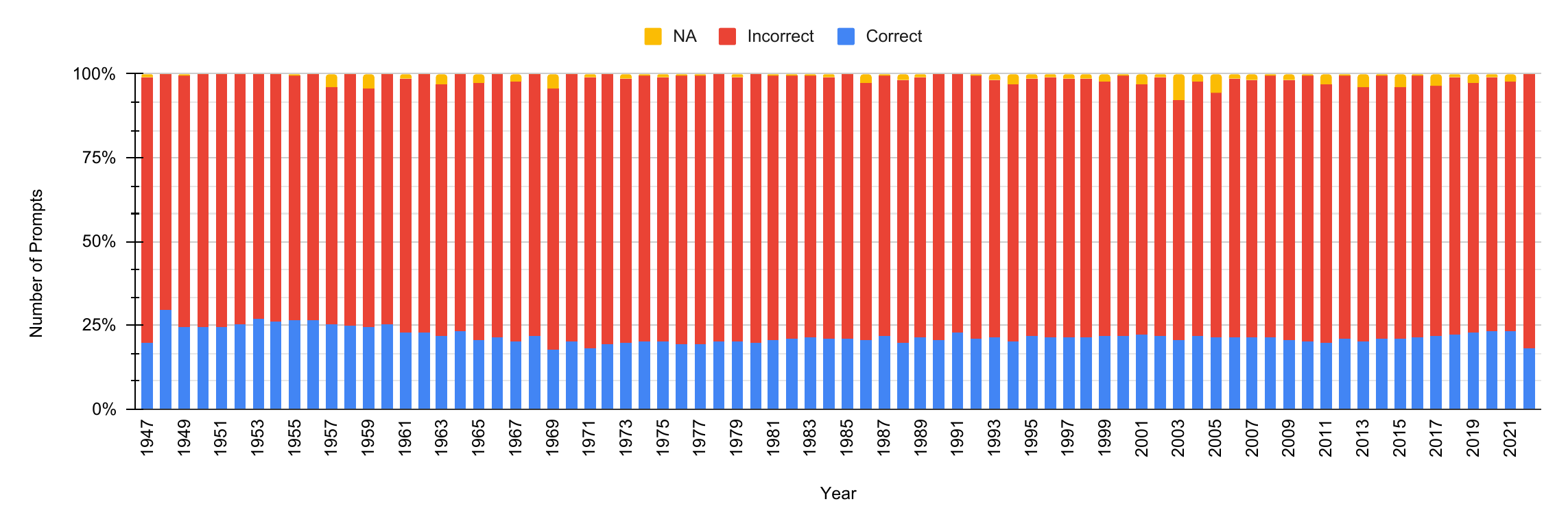}
\end{center}
\caption{Plot for the Window-based metric ($WB$) as year-wise count (In percentage) for \textbf{continual fine-tuning} for \texttt{llama-2}.}
\label{fig:window-based-ft-llama-2}
\end{figure*}

\begin{figure*}
\begin{center}
\includegraphics[width=0.9\linewidth]{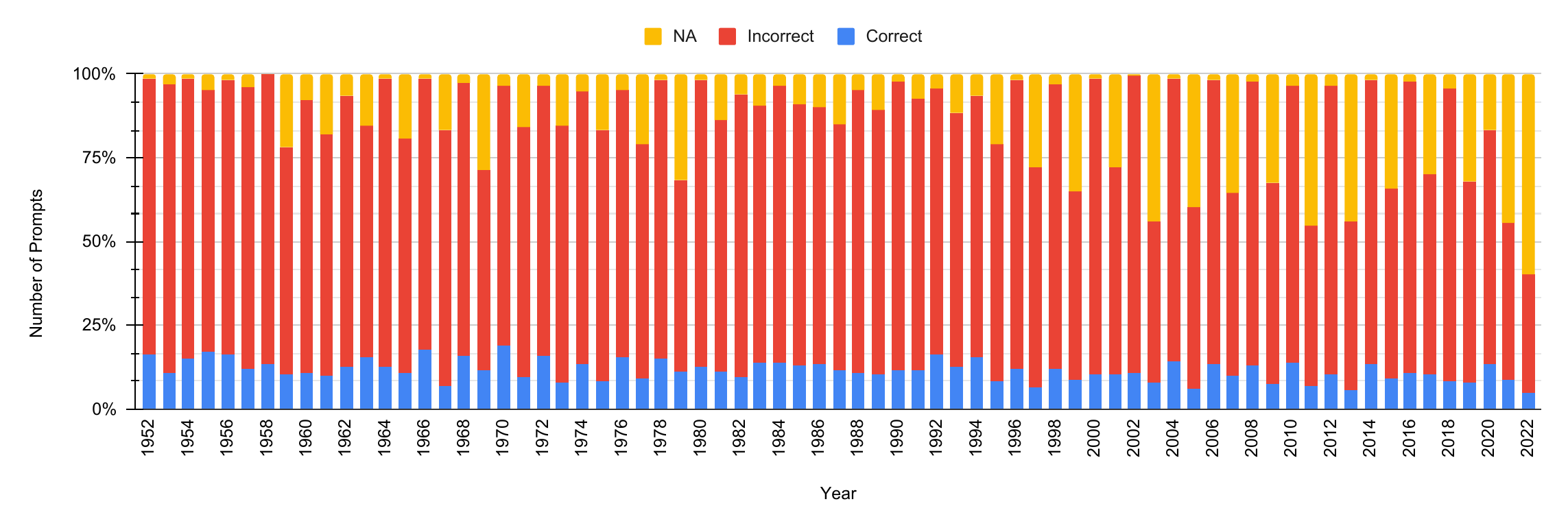}
\end{center}
\caption{Plot for the Min/Max-based metric ($MM$) as year-wise count (In percentage) for \textbf{continual fine-tuning} for \texttt{llama-2}.}
\label{fig:minmax-based-ft-llama-2}
\end{figure*}

\begin{figure*}
\begin{center}
\includegraphics[width=0.9\linewidth]{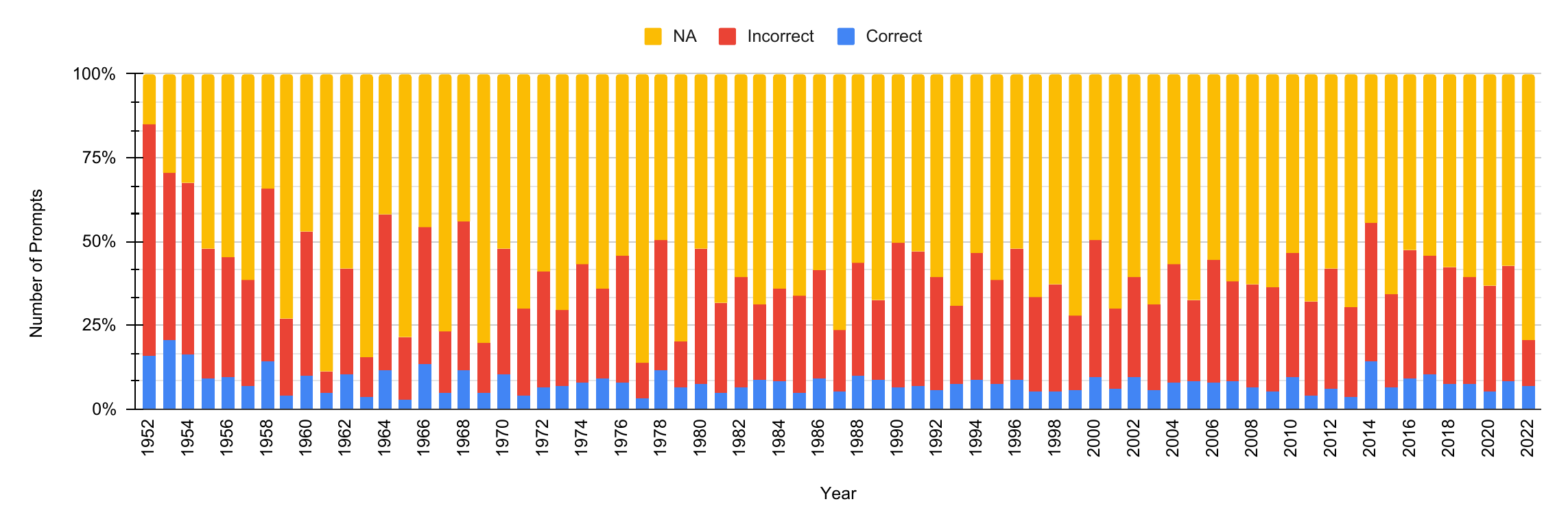}
\end{center}
\caption{Plot for the Range-based metric ($RB$) as year-wise count (In percentage) for \textbf{continual fine-tuning} for \texttt{llama-2}.}
\label{fig:rab-based-ft-llama-2}
\end{figure*}

\begin{figure*}
\begin{center}
\includegraphics[width=0.9\linewidth]{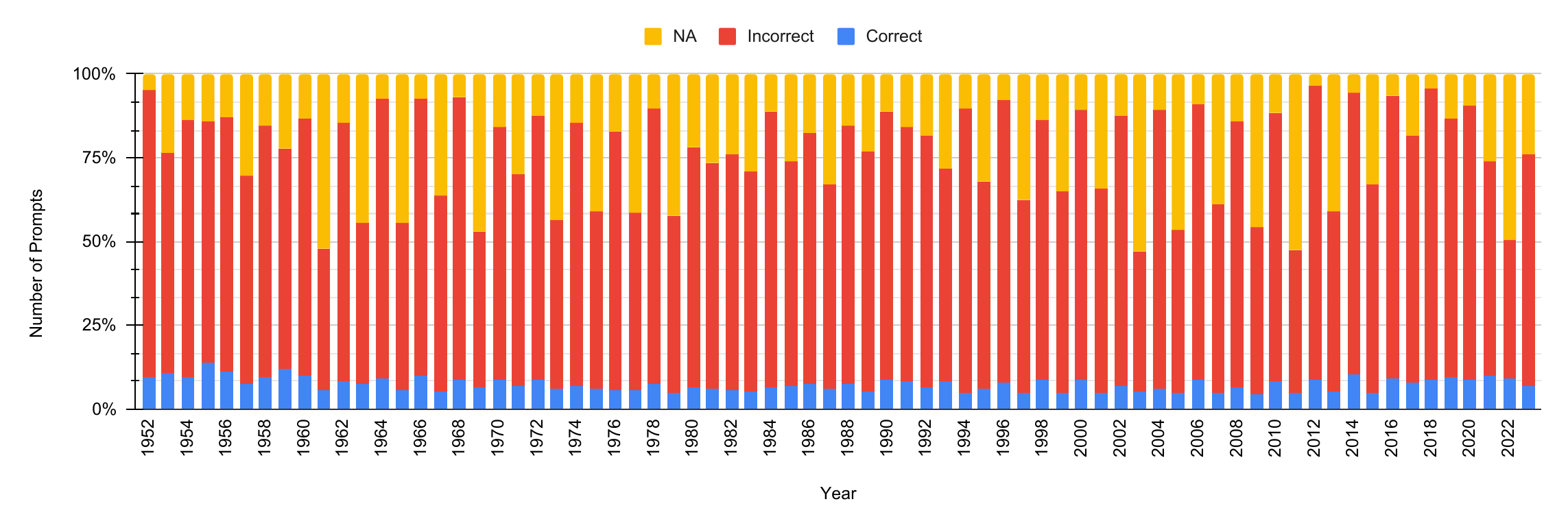}
\end{center}
\caption{Plot for the Trend-based metric ($TB$) as year-wise count (In percentage) for \textbf{continual fine-tuning} for \texttt{llama-2}.}
\label{fig:tb-based-ft-llama-2}
\end{figure*}


\begin{figure*}
\begin{center}
\includegraphics[width=0.9\linewidth]{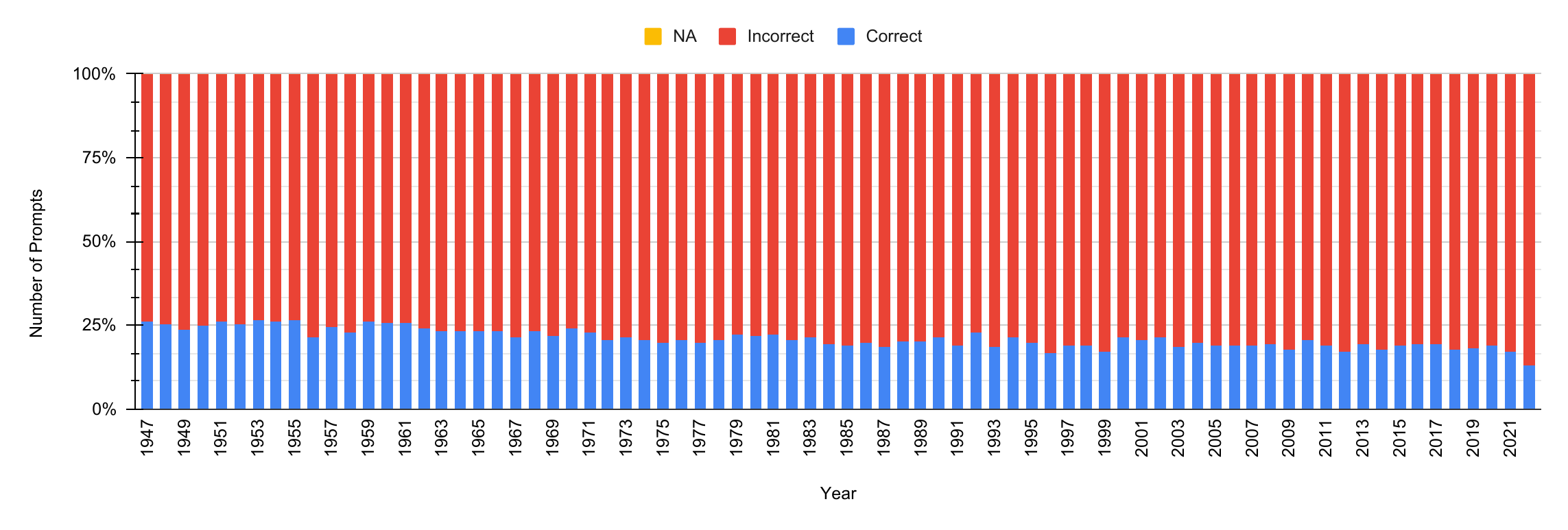}
\end{center}
\caption{Plot for the Date-based metric ($DB$) as year-wise count (In percentage) for \textbf{continual fine-tuning} for \texttt{gemma-7b-it}.}
\label{fig:date-based-ft-gemma-7b-it}
\end{figure*}

\begin{figure*}
\begin{center}
\includegraphics[width=0.9\linewidth]{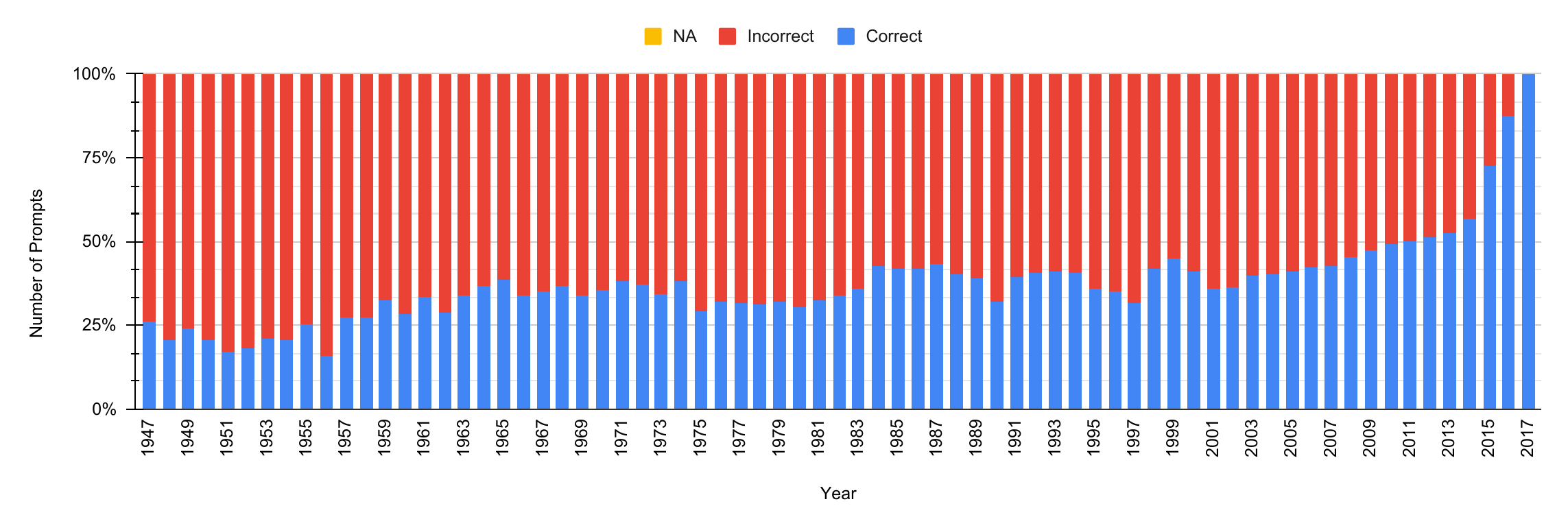}
\end{center}
\caption{Plot for the Comparative-based metric ($CP$) as year-wise count (In percentage) for \textbf{continual fine-tuning} for \texttt{gemma-7b-it}.}
\label{fig:rb-based-ft-gemma-7b-it}
\end{figure*}

\begin{figure*}
\begin{center}
\includegraphics[width=0.9\linewidth]{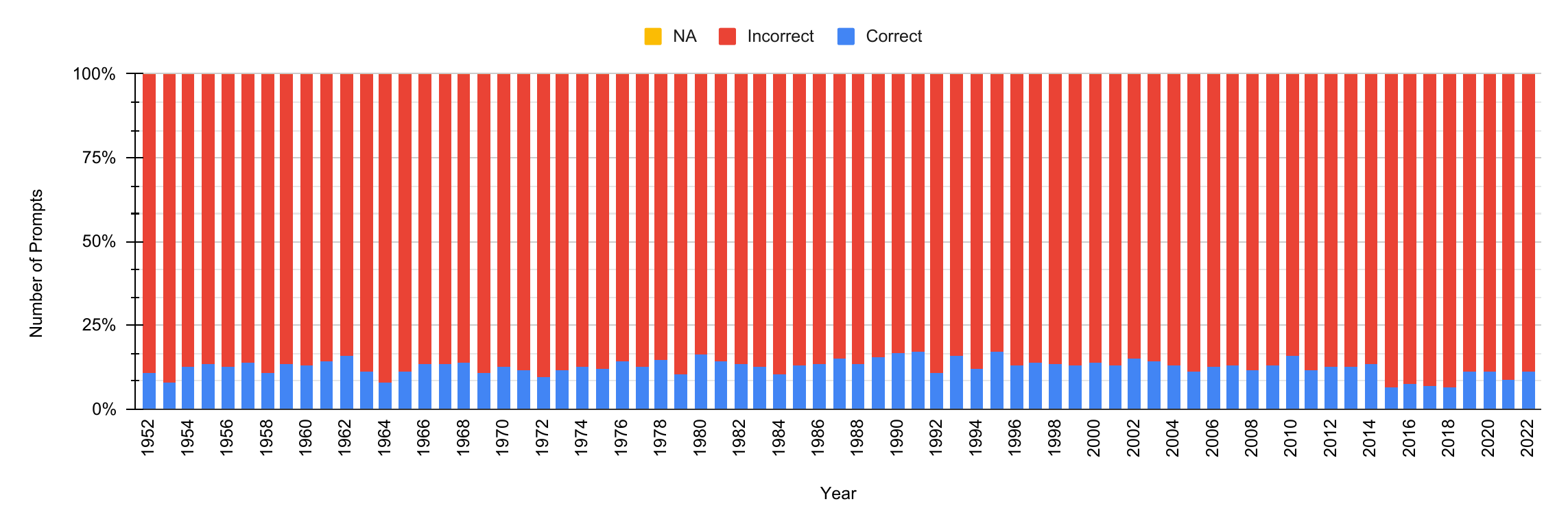}
\end{center}
\caption{Plot for the Window-based metric ($WB$) as year-wise count (In percentage) for \textbf{continual fine-tuning} for \texttt{gemma-7b-it}.}
\label{fig:window-based-ft-gemma-7b-it}
\end{figure*}

\begin{figure*}
\begin{center}
\includegraphics[width=0.9\linewidth]{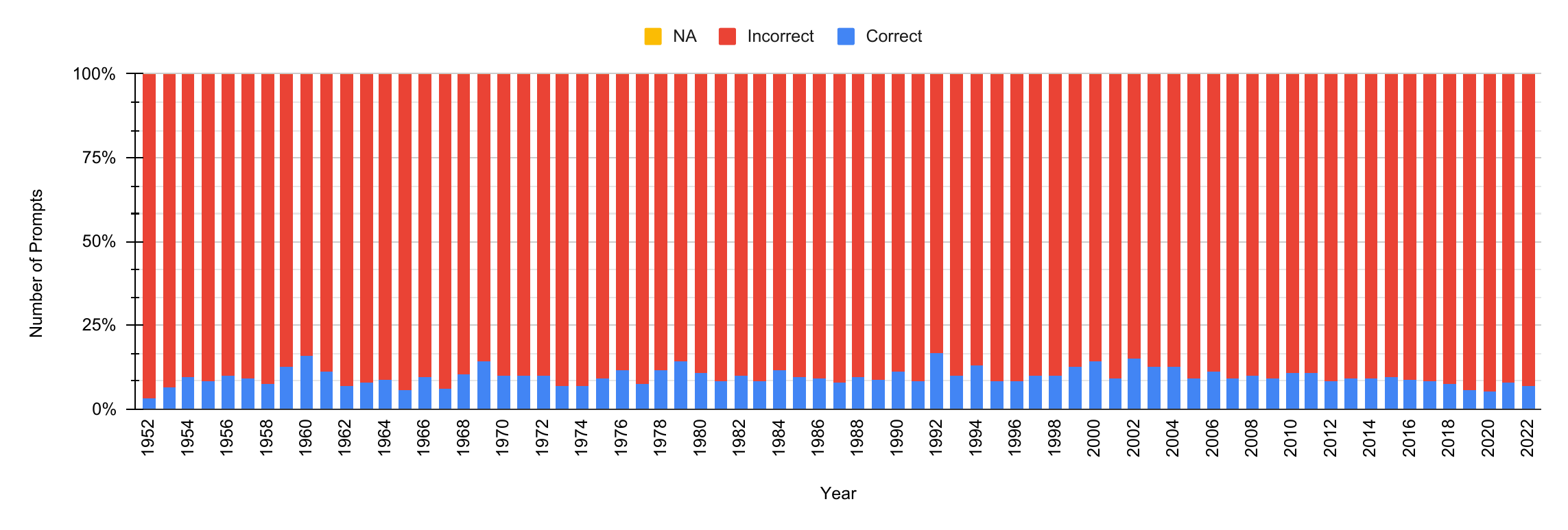}
\end{center}
\caption{Plot for the Min/Max-based metric ($MM$) as year-wise count (In percentage) for \textbf{continual fine-tuning} for \texttt{gemma-7b-it}.}
\label{fig:minmax-based-ft-gemma-7b-it}
\end{figure*}

\begin{figure*}
\begin{center}
\includegraphics[width=0.9\linewidth]{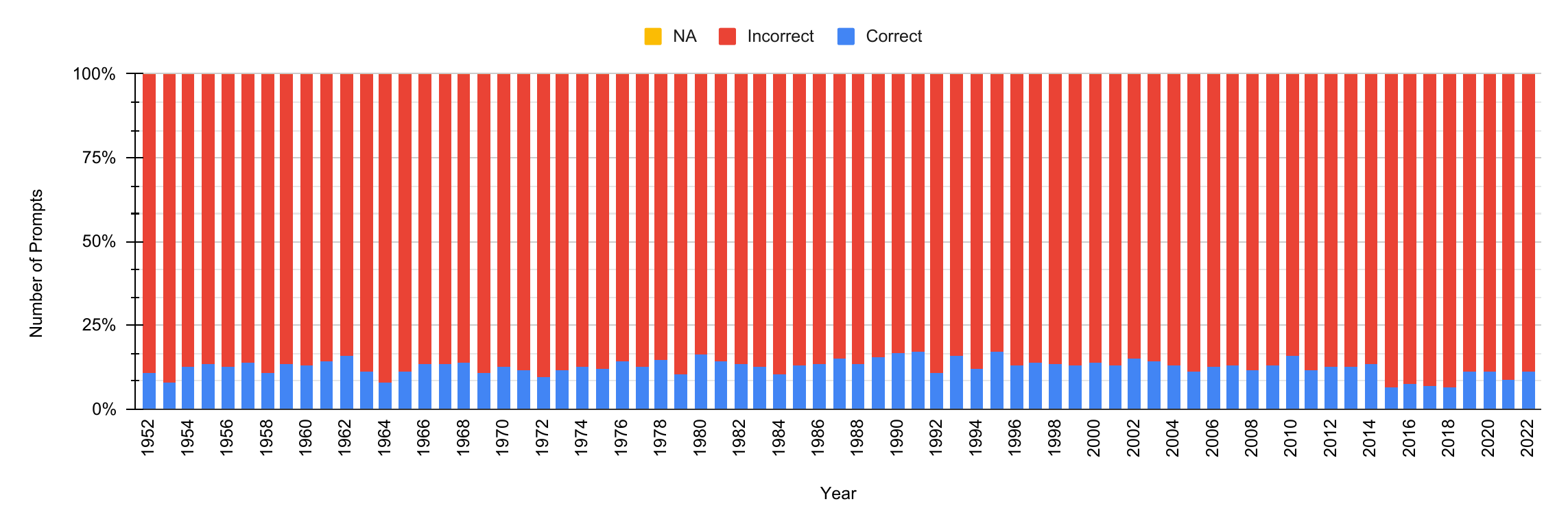}
\end{center}
\caption{Plot for the Range-based metric ($RB$) as year-wise count (In percentage) for \textbf{continual fine-tuning} for \texttt{gemma-7b-it}.}
\label{fig:rab-based-ft-gemma-7b-it}
\end{figure*}

\begin{figure*}
\begin{center}
\includegraphics[width=0.9\linewidth]{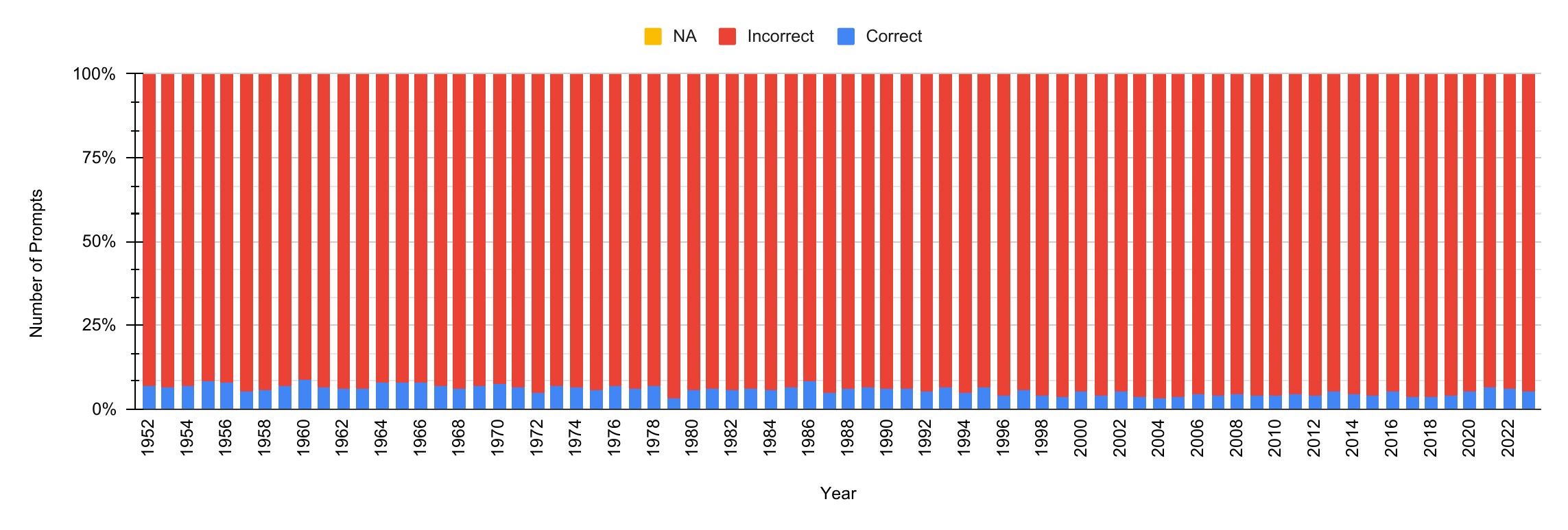}
\end{center}
\caption{Plot for the Trend-based metric ($TB$) as year-wise count (In percentage) for \textbf{continual fine-tuning} for \texttt{gemma-7b-it}.}
\label{fig:tb-based-ft-gemma-7b-it}
\end{figure*}


\begin{figure*}
\begin{center}
\includegraphics[width=0.9\linewidth]{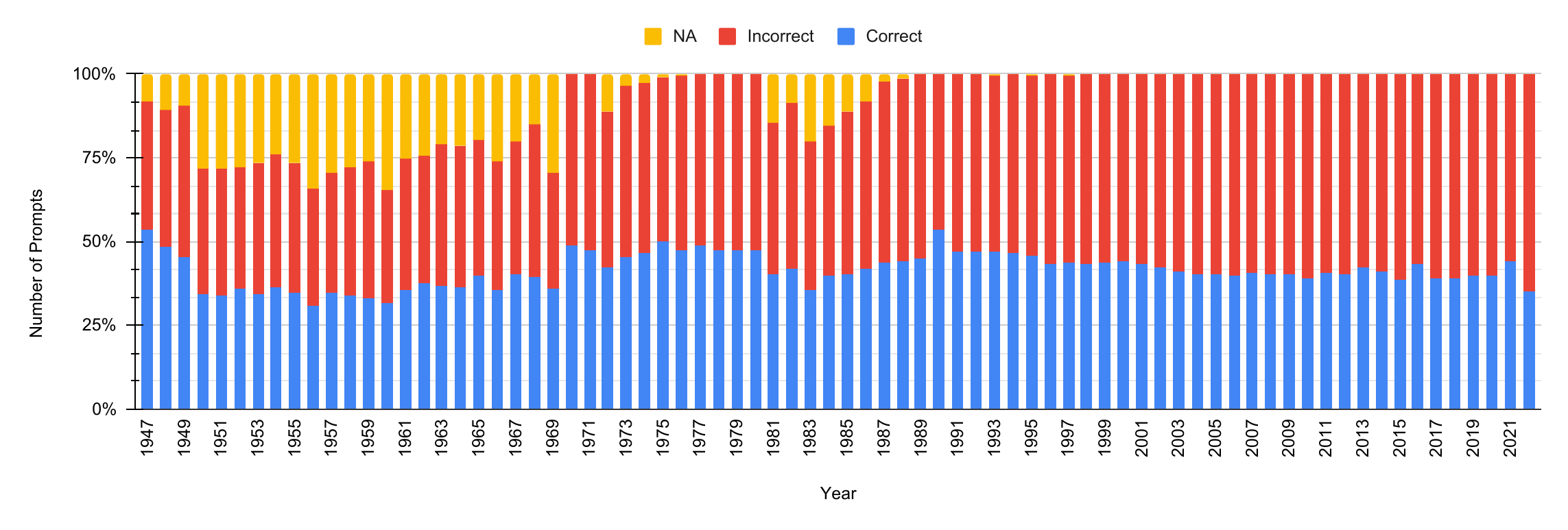}
\end{center}
\caption{Plot for the Date-based metric ($DB$) as year-wise count (In percentage) for \textbf{continual fine-tuning} for \texttt{llama-3-8b}.}
\label{fig:date-based-ft-llama-3-8b}
\end{figure*}

\begin{figure*}
\begin{center}
\includegraphics[width=0.9\linewidth]{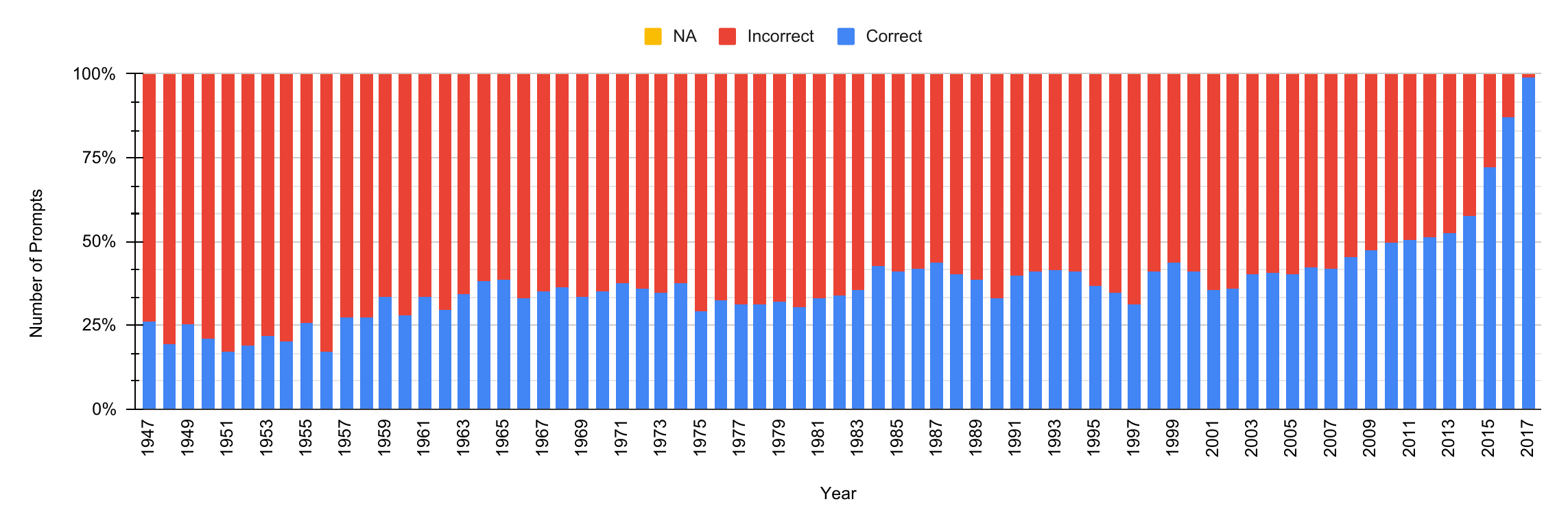}
\end{center}
\caption{Plot for the Comparative-based metric ($CP$) as year-wise count (In percentage) for \textbf{continual fine-tuning} for \texttt{llama-3-8b}.}
\label{fig:rb-based-ft-llama-3-8b}
\end{figure*}

\begin{figure*}
\begin{center}
\includegraphics[width=0.9\linewidth]{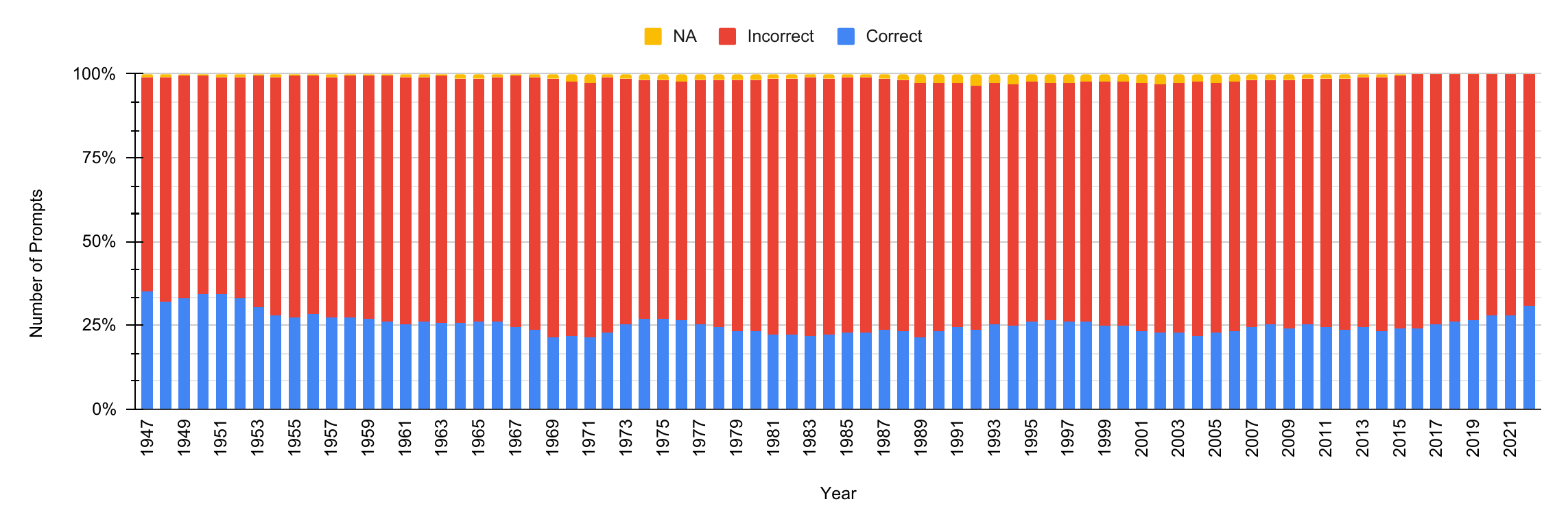}
\end{center}
\caption{Plot for the Window-based metric ($WB$) as year-wise count (In percentage) for \textbf{continual fine-tuning} for \texttt{llama-3-8b}.}
\label{fig:window-based-ft-llama-3-8b}
\end{figure*}

\begin{figure*}
\begin{center}
\includegraphics[width=0.9\linewidth]{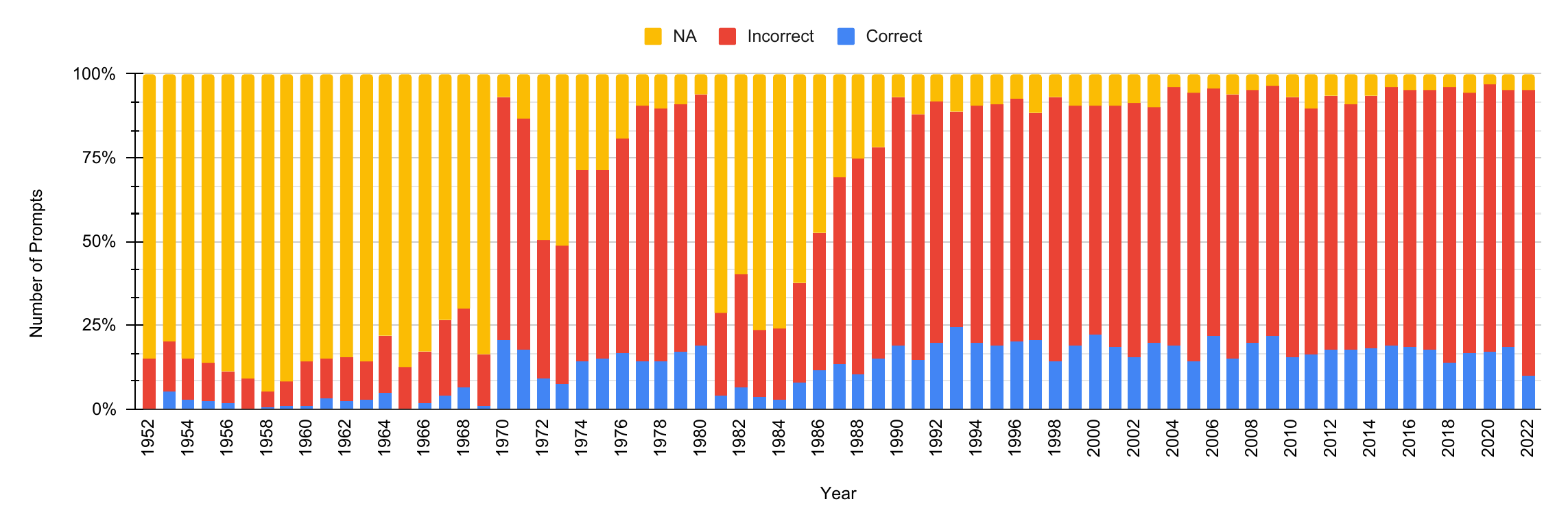}
\end{center}
\caption{Plot for the Min/Max-based metric ($MM$) as year-wise count (In percentage) for \textbf{continual fine-tuning} for \texttt{llama-3-8b}.}
\label{fig:minmax-based-ft-llama-3-8b}
\end{figure*}

\begin{figure*}
\begin{center}
\includegraphics[width=0.9\linewidth]{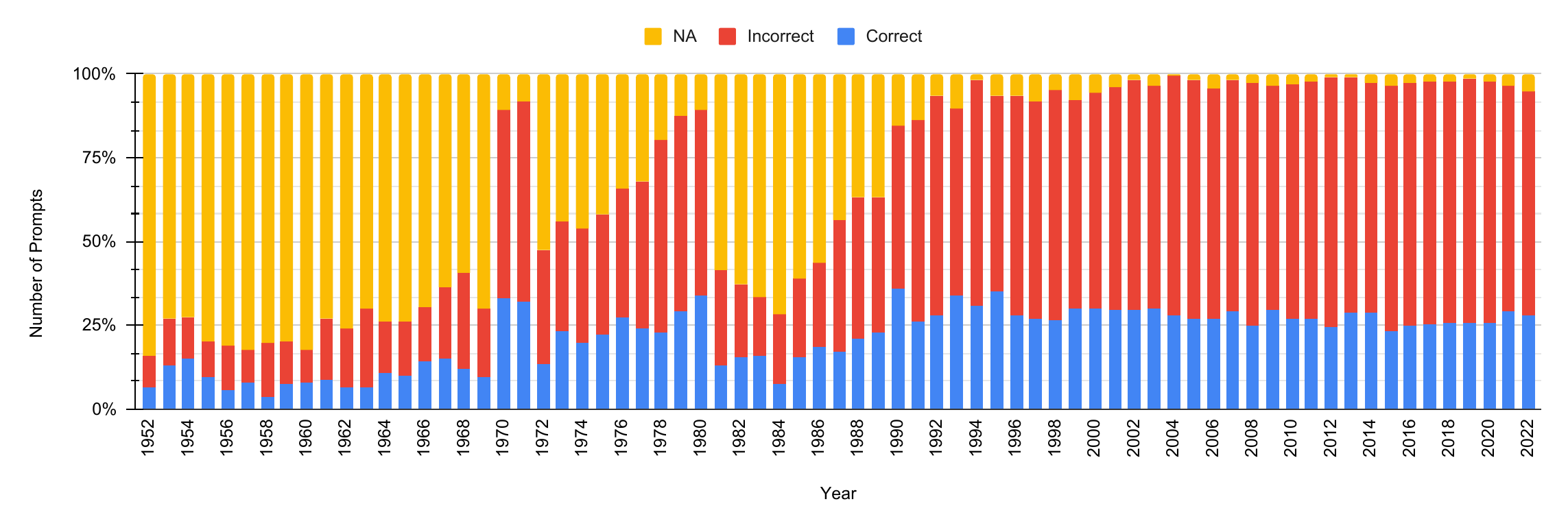}
\end{center}
\caption{Plot for the Range-based metric ($RB$) as year-wise count (In percentage) for \textbf{continual fine-tuning} for \texttt{llama-3-8b}.}
\label{fig:rab-based-ft-llama-3-8b}
\end{figure*}

\begin{figure*}
\begin{center}
\includegraphics[width=0.9\linewidth]{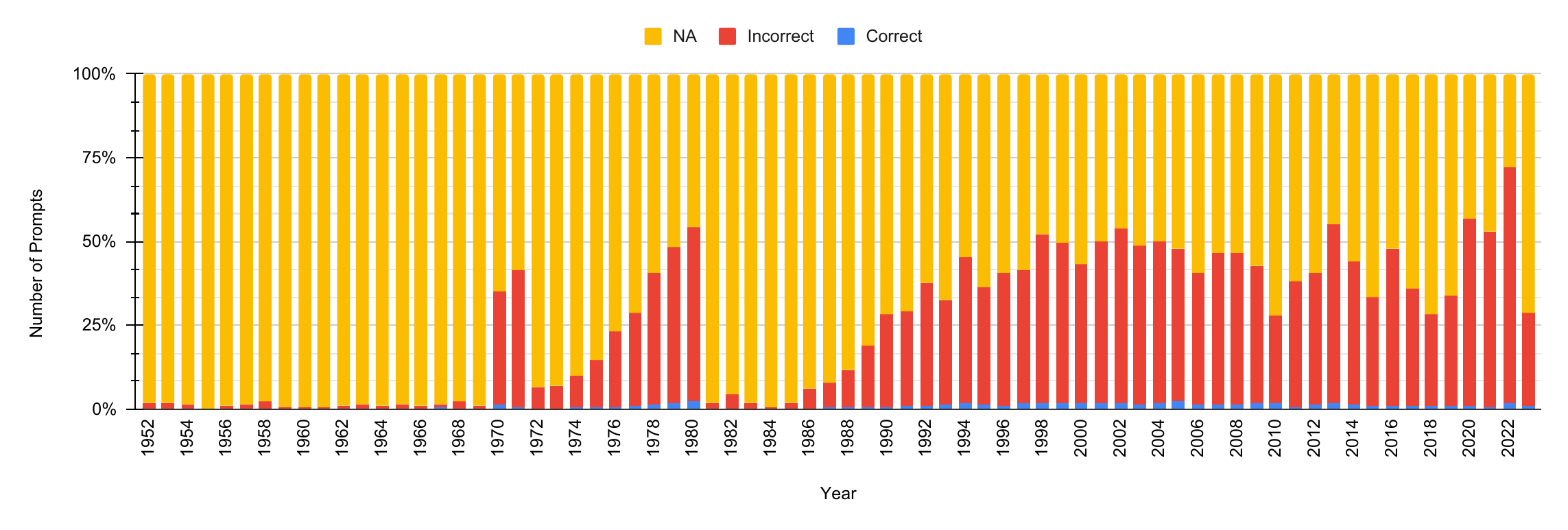}
\end{center}
\caption{Plot for the Trend-based metric ($TB$) as year-wise count (In percentage) for \textbf{continual fine-tuning} for \texttt{llama-3-8b}.}
\label{fig:tb-based-ft-llama-3-8b}
\end{figure*}


\begin{figure*}
\begin{center}
\includegraphics[width=0.9\linewidth]{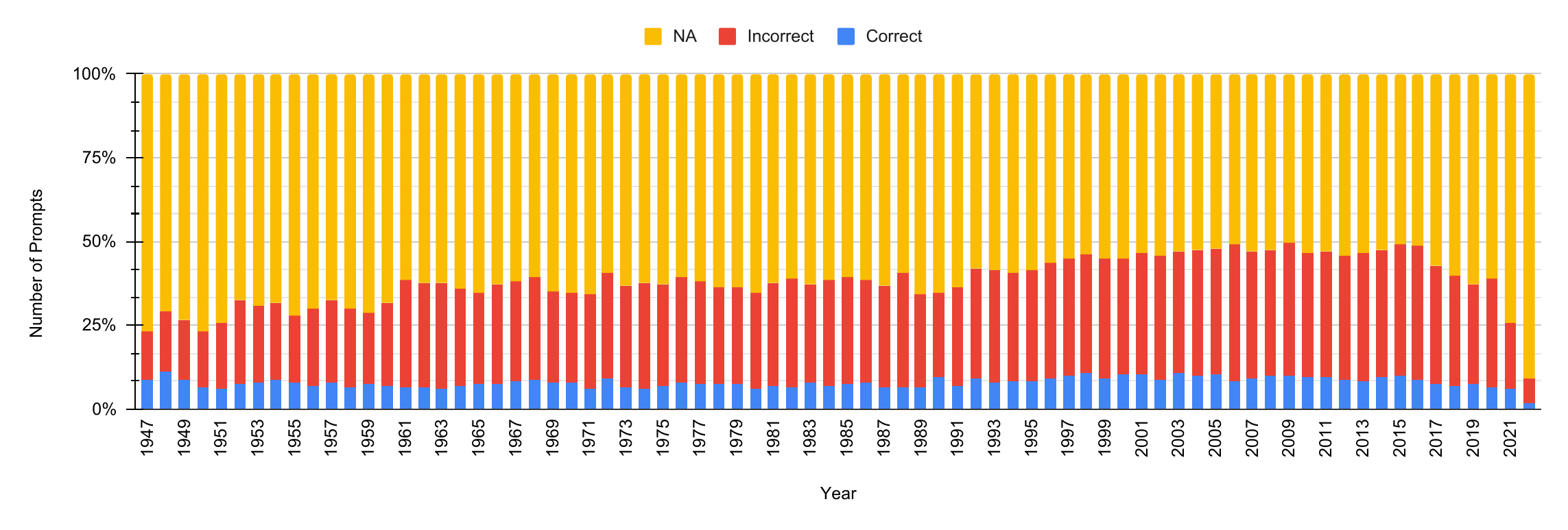}
\end{center}
\caption{Plot for the Date-based metric ($DB$) as year-wise count (In percentage) for \textbf{continual fine-tuning} for \texttt{phi-3-medium}.}
\label{fig:date-based-ft-phi-3-medium}
\end{figure*}

\begin{figure*}
\begin{center}
\includegraphics[width=0.9\linewidth]{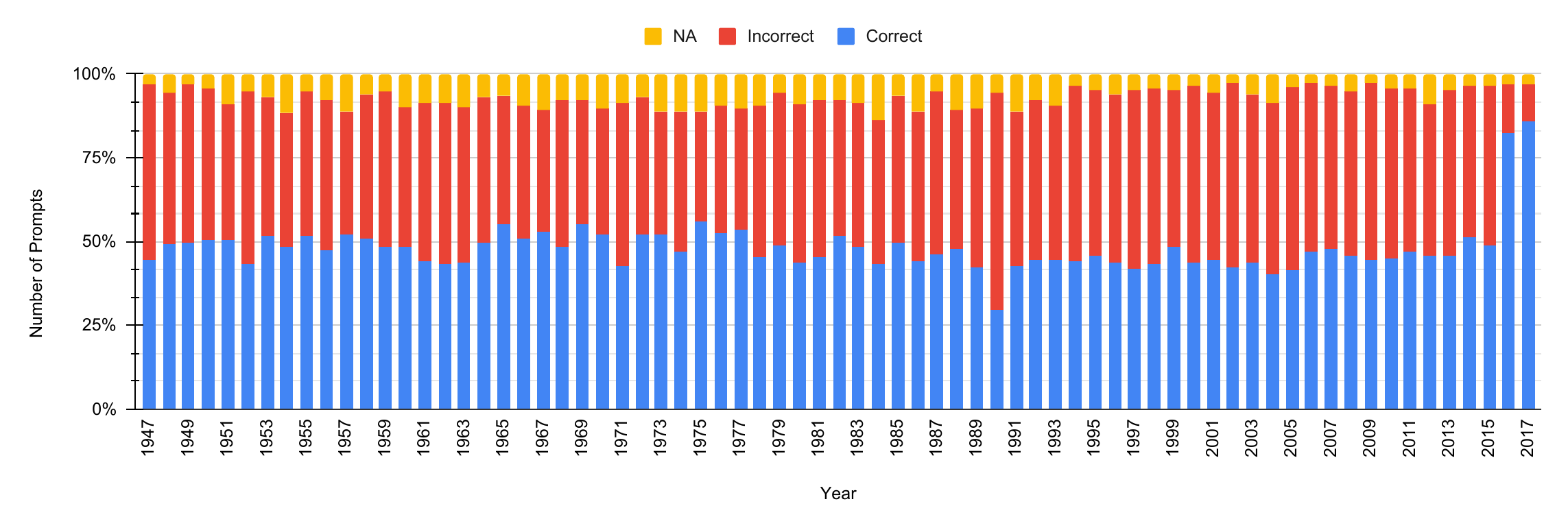}
\end{center}
\caption{Plot for the Comparative-based metric ($CP$) as year-wise count (In percentage) for \textbf{continual fine-tuning} for \texttt{phi-3-medium}.}
\label{fig:rb-based-ft-phi-3-medium}
\end{figure*}

\begin{figure*}
\begin{center}
\includegraphics[width=0.9\linewidth]{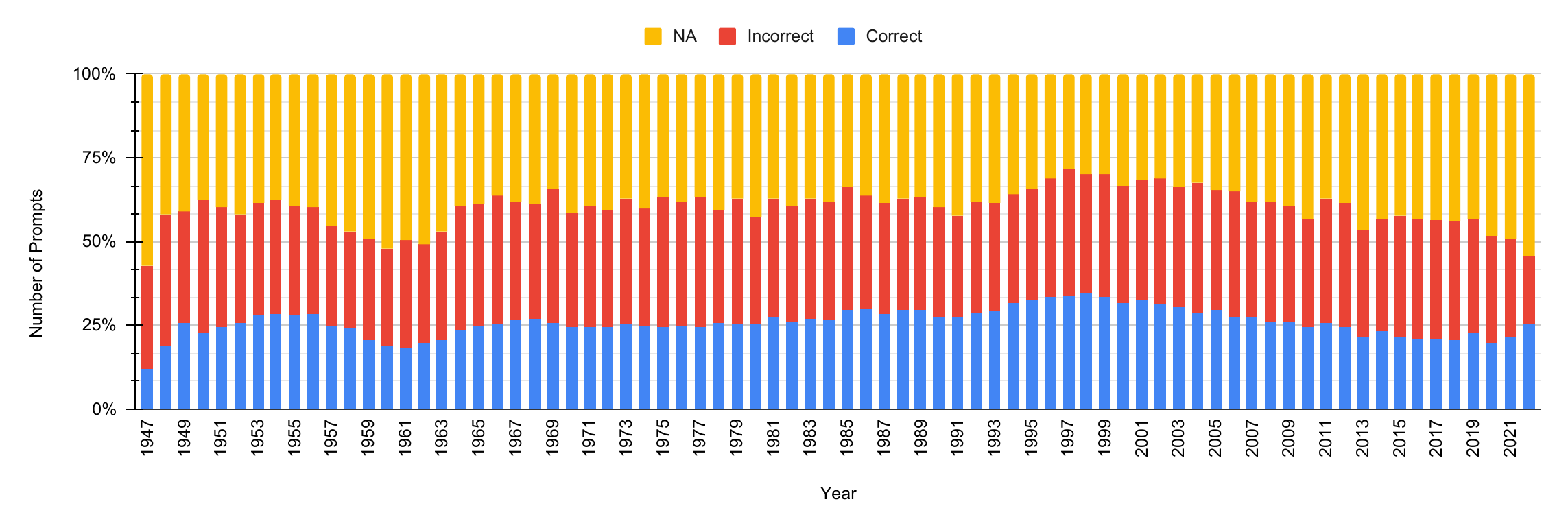}
\end{center}
\caption{Plot for the Window-based metric ($WB$) as year-wise count (In percentage) for \textbf{continual fine-tuning} for \texttt{phi-3-medium}.}
\label{fig:window-based-ft-phi-3-medium}
\end{figure*}

\begin{figure*}
\begin{center}
\includegraphics[width=0.9\linewidth]{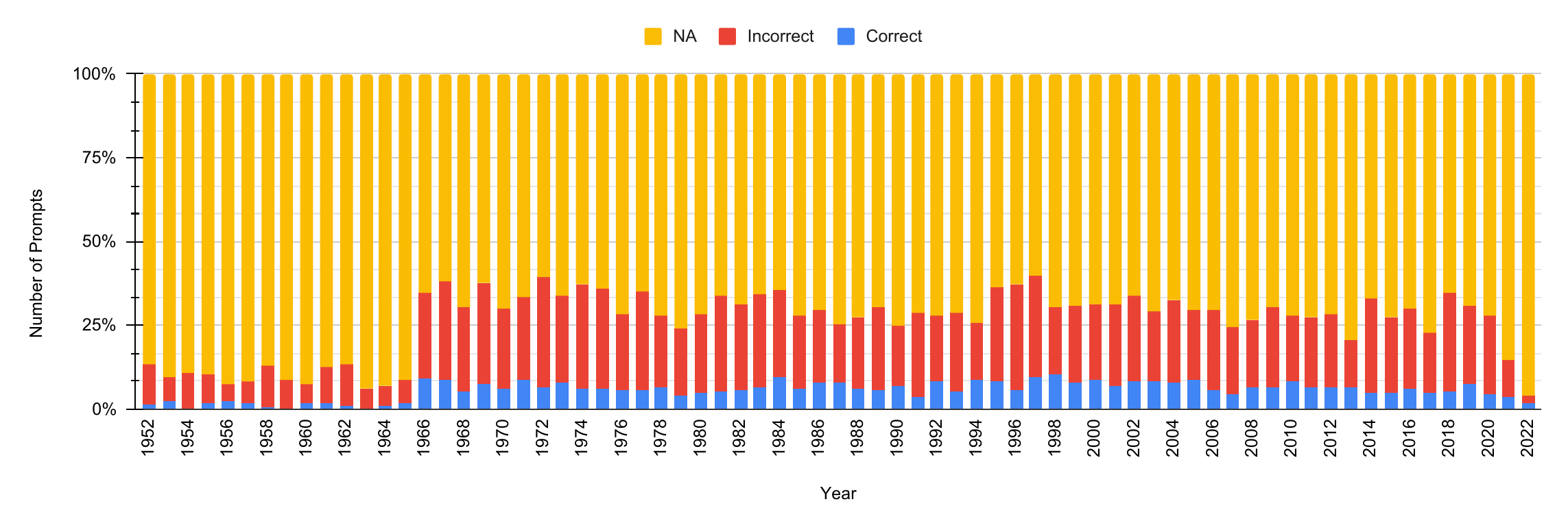}
\end{center}
\caption{Plot for the Min/Max-based metric ($MM$) as year-wise count (In percentage) for \textbf{continual fine-tuning} for \texttt{phi-3-medium}.}
\label{fig:minmax-based-ft-phi-3-medium}
\end{figure*}

\begin{figure*}
\begin{center}
\includegraphics[width=0.9\linewidth]{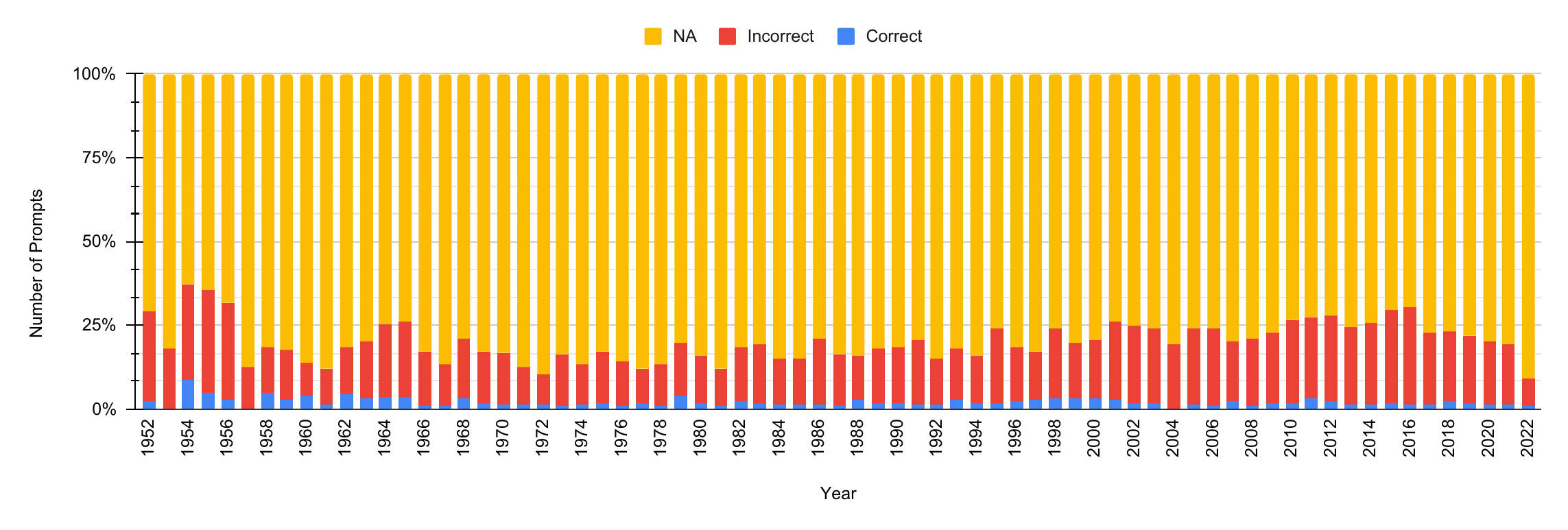}
\end{center}
\caption{Plot for the Range-based metric ($RB$) as year-wise count (In percentage) for \textbf{continual fine-tuning} for \texttt{phi-3-medium}.}
\label{fig:rab-based-ft-phi-3-medium}
\end{figure*}

\begin{figure*}
\begin{center}
\includegraphics[width=0.9\linewidth]{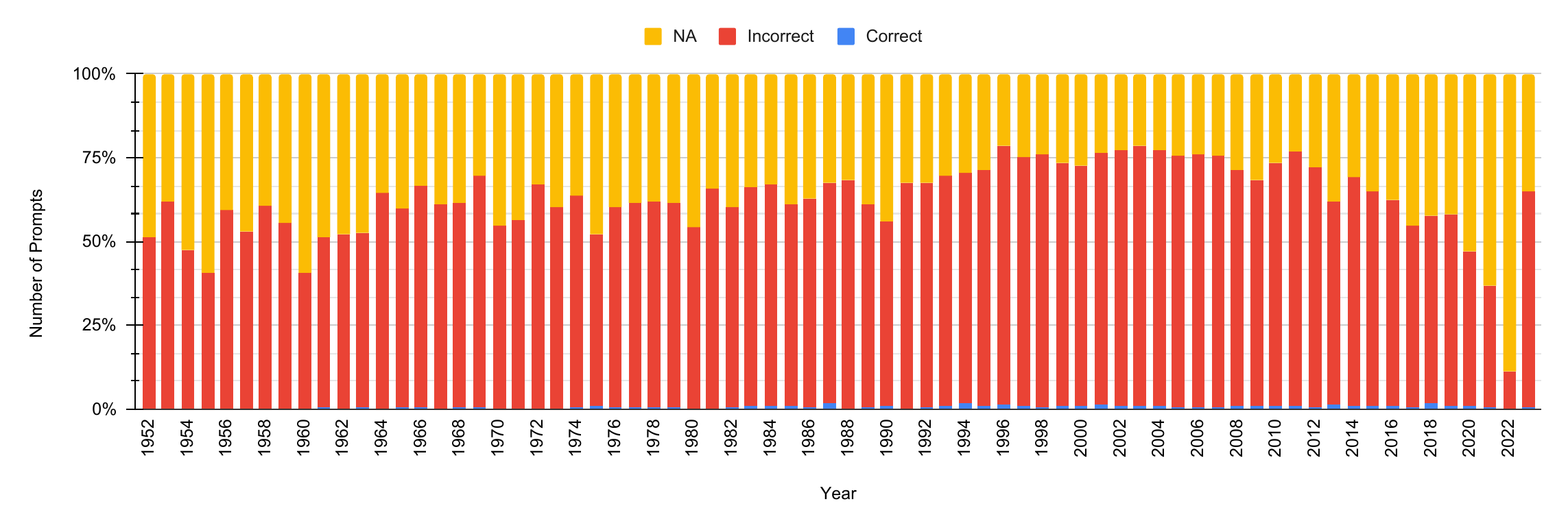}
\end{center}
\caption{Plot for the Trend-based metric ($TB$) as year-wise count (In percentage) for \textbf{continual fine-tuning} for \texttt{phi-3-medium}.}
\label{fig:tb-based-ft-phi-3-medium}
\end{figure*}


\begin{figure*}
\begin{center}
\includegraphics[width=0.9\linewidth]{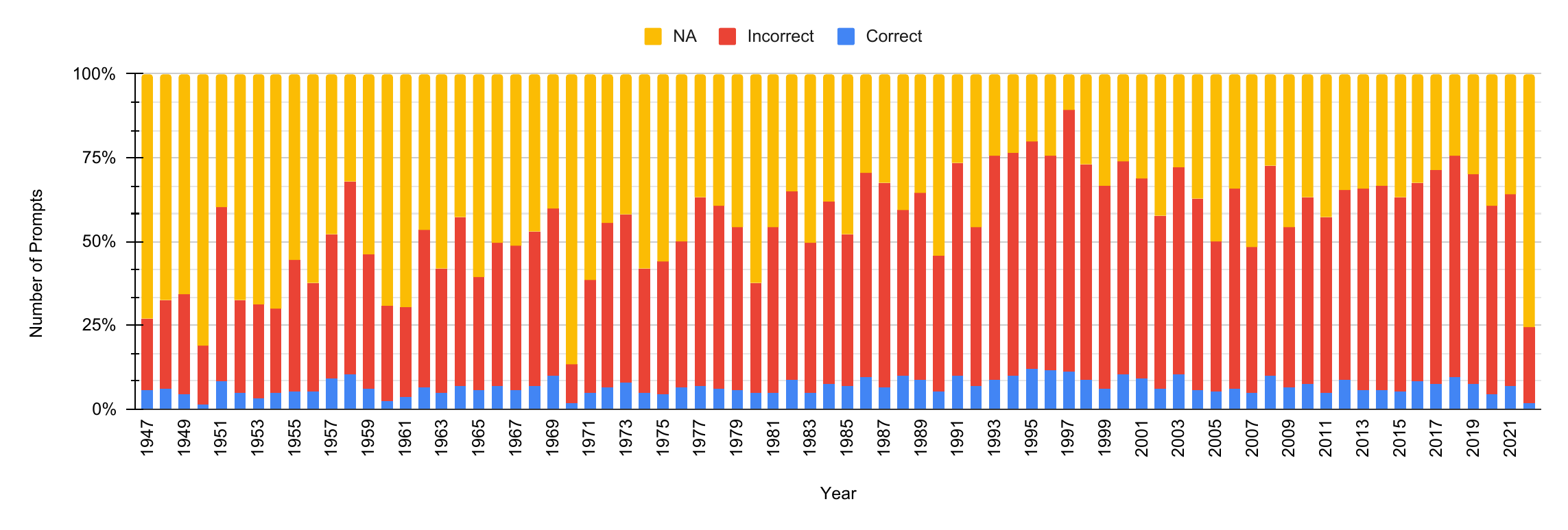}
\end{center}
\caption{Plot for the Date-based metric ($DB$) as year-wise count (In percentage) for \textbf{yearwise fine-tuning} for \texttt{phi-2}.}
\label{fig:date-based-yl-phi2}
\end{figure*}

\begin{figure*}
\begin{center}
\includegraphics[width=0.9\linewidth]{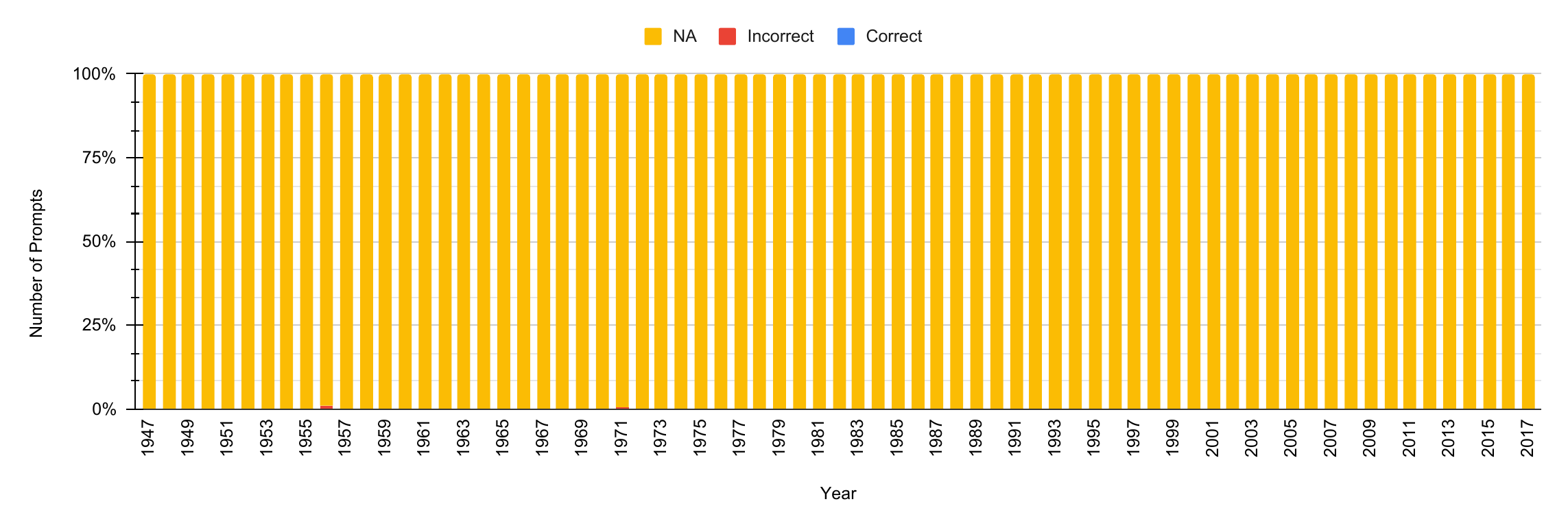}
\end{center}
\caption{Plot for the Comparative-based metric ($CP$) as year-wise count (In percentage) for \textbf{yearwise fine-tuning} for \texttt{phi-2}.}
\label{fig:cp-yl-phi2}
\end{figure*}

\begin{figure*}
\begin{center}
\includegraphics[width=0.9\linewidth]{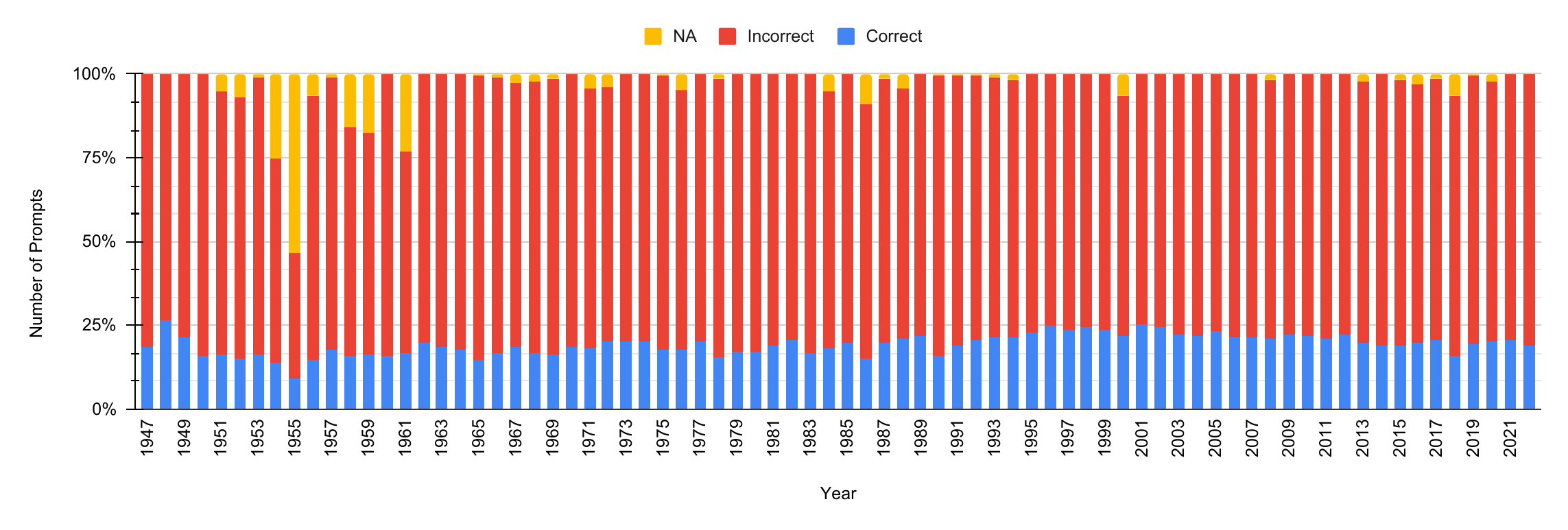}
\end{center}
\caption{Plot for the Window-based metric ($WB$) as year-wise count (In percentage) for \textbf{yearwise fine-tuning} for \texttt{phi-2}.}
\label{fig:window-based-yl-phi2}
\end{figure*}
\begin{figure*}
\begin{center}
\includegraphics[width=0.9\linewidth]{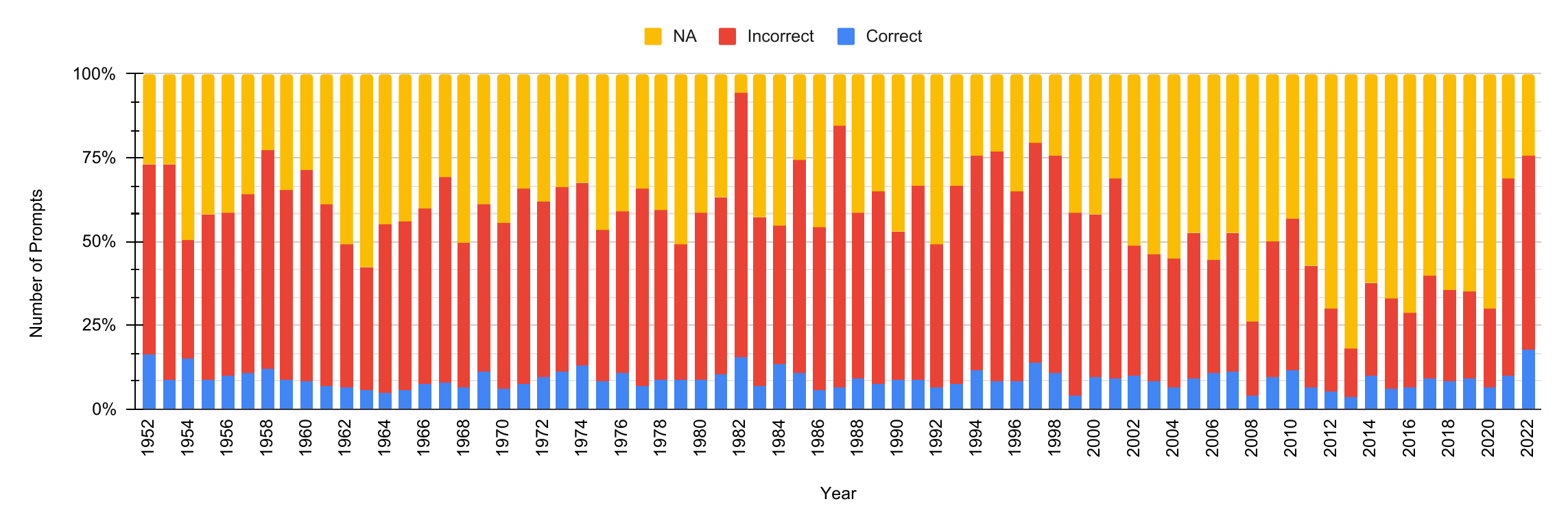}
\end{center}
\caption{Plot for the Min/Max-based metric ($MM$) as year-wise count (In percentage) for \textbf{yearwise fine-tuning} for \texttt{phi-2}.}
\label{fig:minmax-based-yl-phi2}
\end{figure*}

\begin{figure*}
\begin{center}
\includegraphics[width=0.9\linewidth]{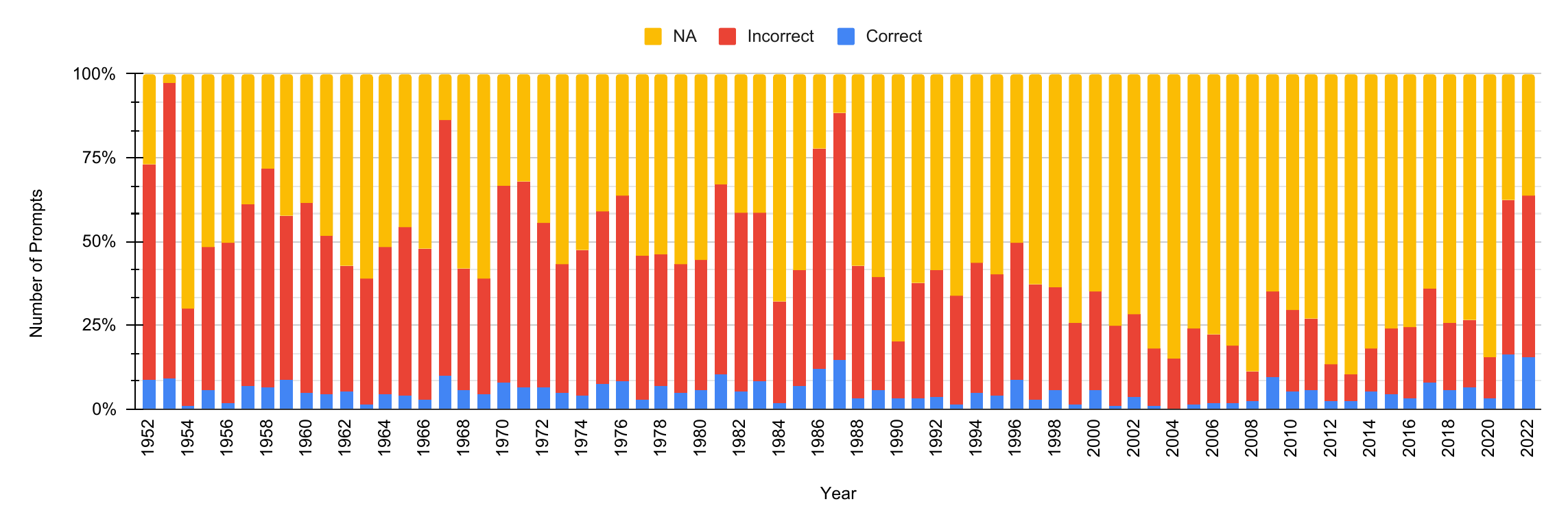}
\end{center}
\caption{Plot for the Range-based metric ($RB$) as year-wise count (In percentage) for \textbf{yearwise fine-tuning} for \texttt{phi-2}.}
\label{fig:rab-based-yl-phi2}
\end{figure*}

\begin{figure*}
\begin{center}
\includegraphics[width=0.9\linewidth]{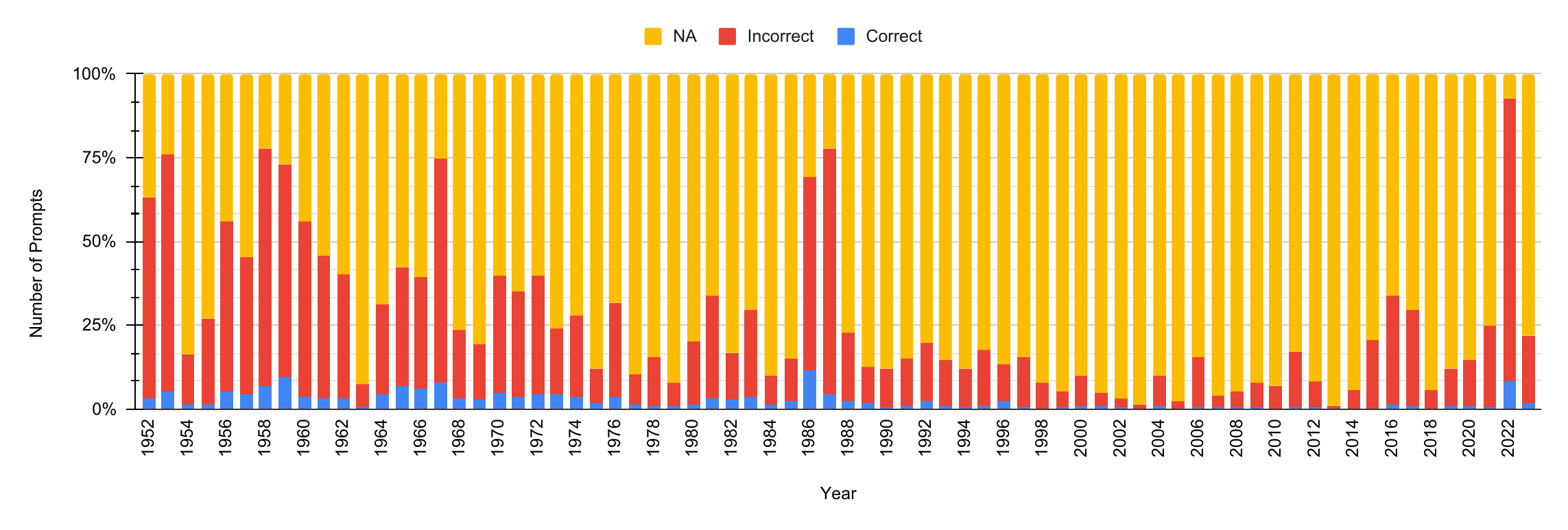}
\end{center}
\caption{Plot for the Trend-based metric ($TB$) as year-wise count (In percentage) for \textbf{yearwise fine-tuning} for \texttt{phi-2}.}
\label{fig:tb-based-yl-phi2}
\end{figure*}


\begin{figure*}
\begin{center}
\includegraphics[width=0.9\linewidth]{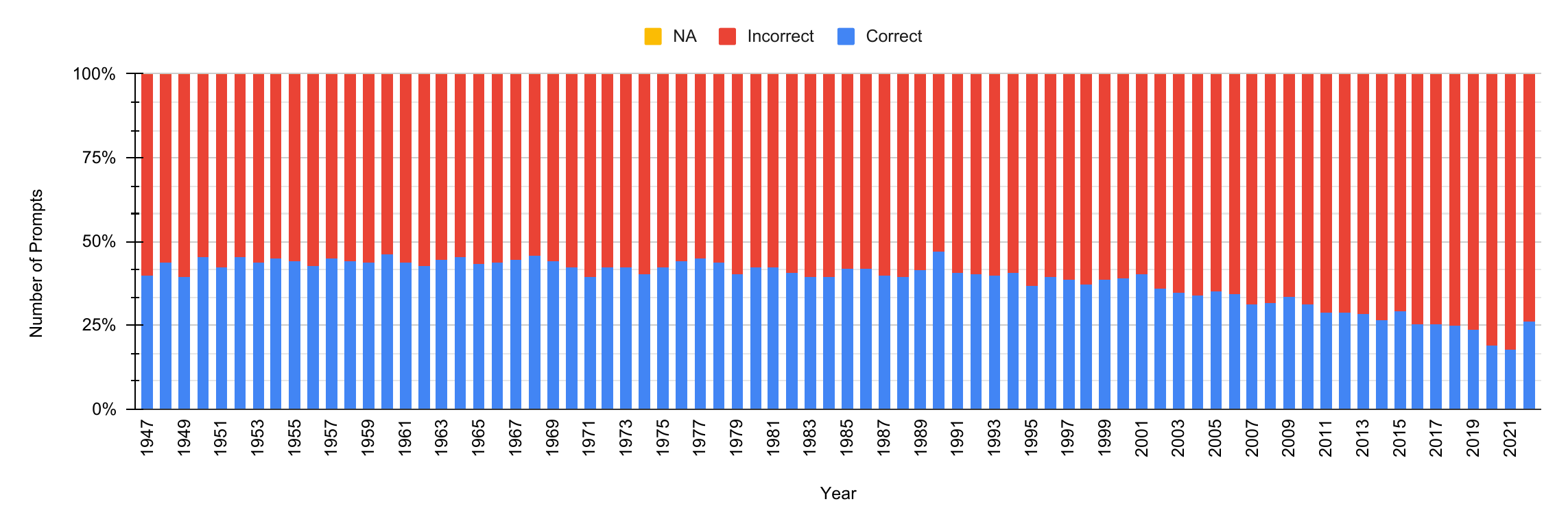}
\end{center}
\caption{Plot for the Date-based metric ($DB$) as year-wise count (In percentage) for \textbf{yearwise fine-tuning} for \texttt{flan-t5-xl}.}
\label{fig:date-based-yl-flan-t5-xl}
\end{figure*}

\begin{figure*}
\begin{center}
\includegraphics[width=0.9\linewidth]{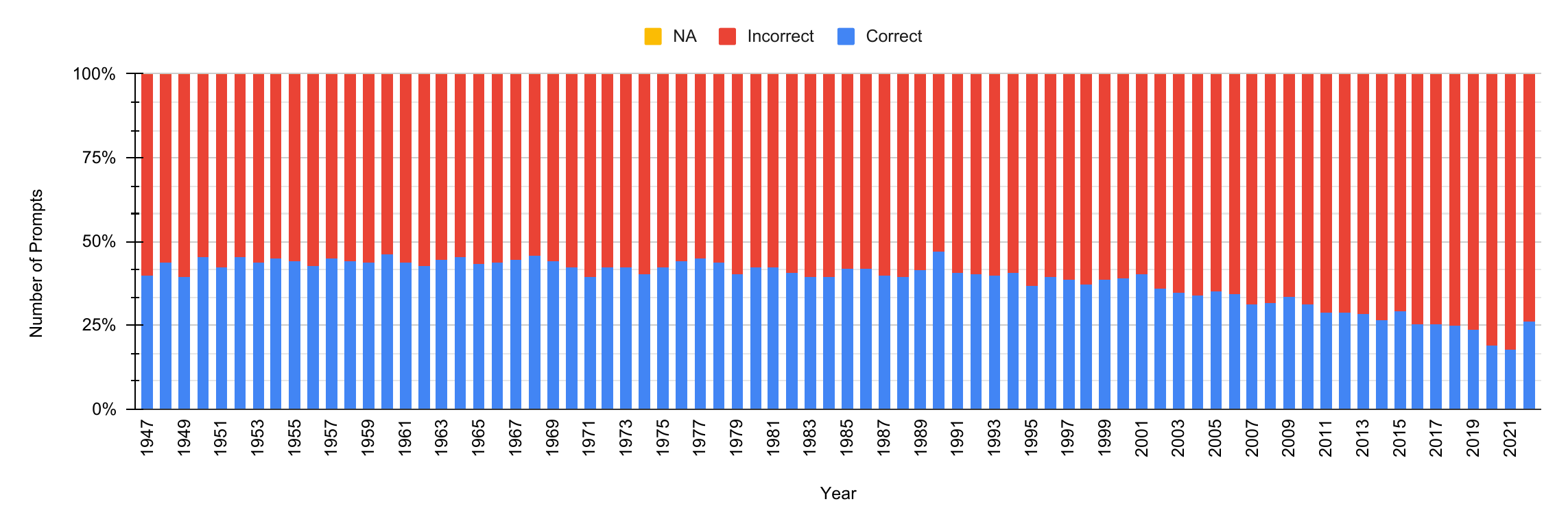}
\end{center}
\caption{Plot for the Comparative-based metric ($CP$) as year-wise count (In percentage) for \textbf{yearwise fine-tuning} for \texttt{flan-t5-xl}.}
\label{fig:cate-yl-flan-t5-xl}
\end{figure*}

\begin{figure*}
\begin{center}
\includegraphics[width=0.9\linewidth]{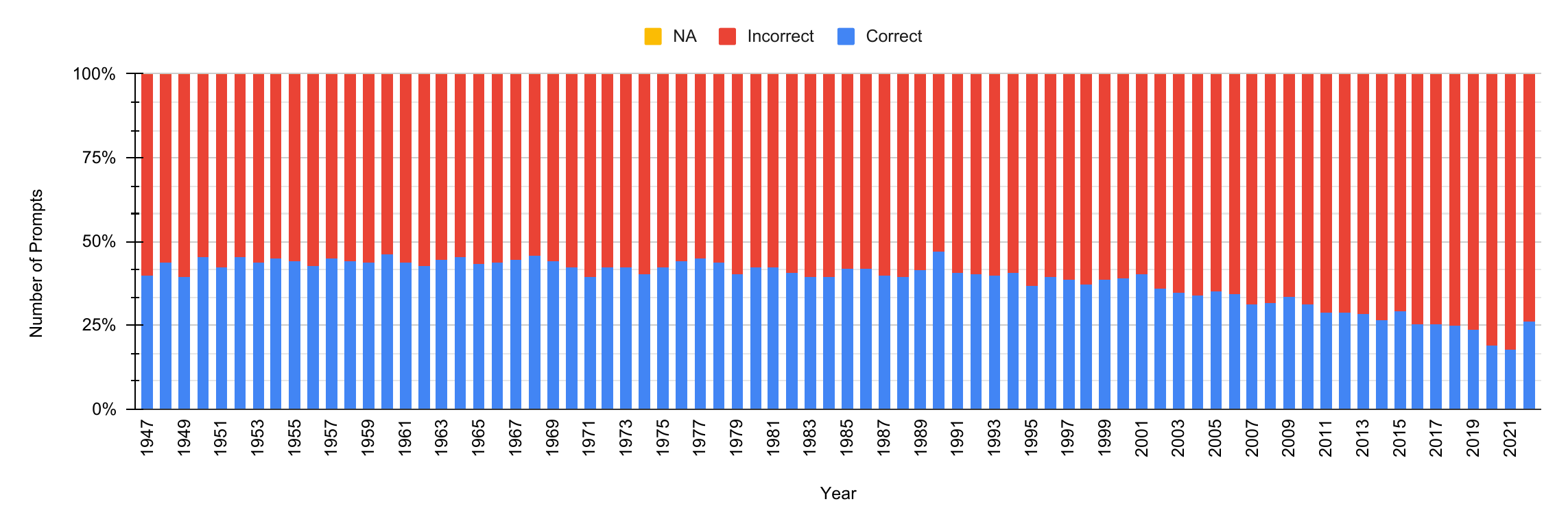}
\end{center}
\caption{Plot for the Window-based metric ($WB$) as year-wise count (In percentage) for \textbf{yearwise fine-tuning} for \texttt{flan-t5-xl}.}
\label{fig:window-based-yl-flan-t5-xl}
\end{figure*}
\begin{figure*}
\begin{center}
\includegraphics[width=0.9\linewidth]{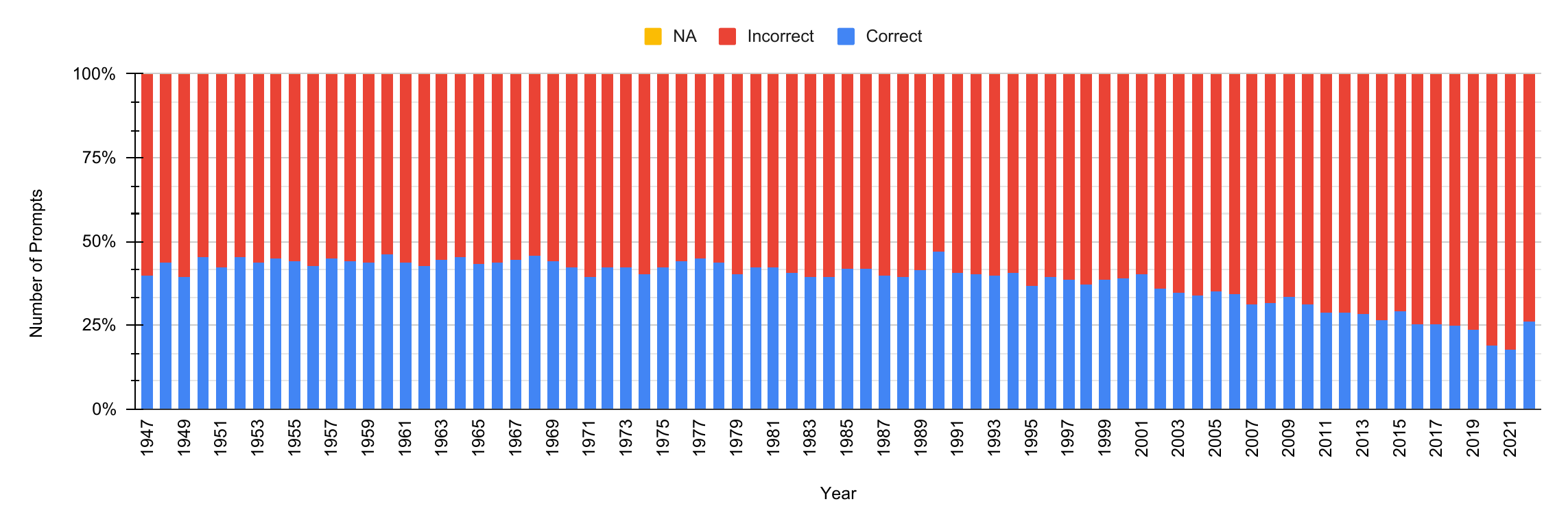}
\end{center}
\caption{Plot for the Min/Max-based metric ($MM$) as year-wise count (In percentage) for \textbf{yearwise fine-tuning} for \texttt{flan-t5-xl}.}
\label{fig:minmax-based-yl-flan-t5-xl}
\end{figure*}

\begin{figure*}
\begin{center}
\includegraphics[width=0.9\linewidth]{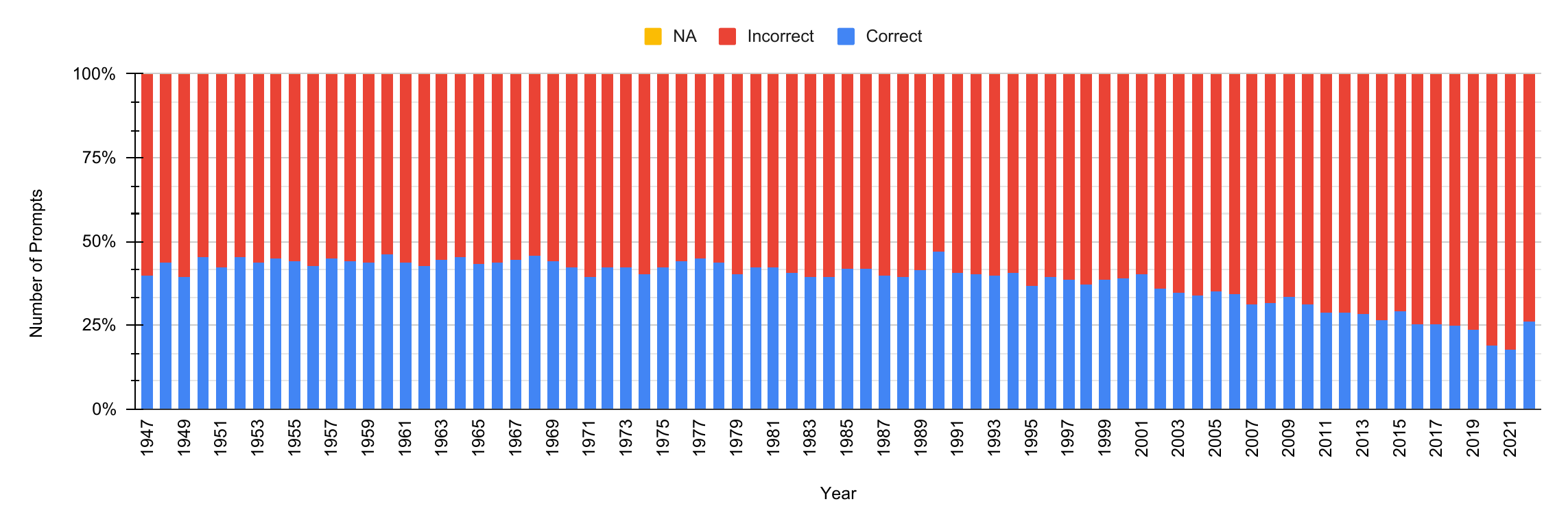}
\end{center}
\caption{Plot for the Range-based metric ($RB$) as year-wise count (In percentage) for \textbf{yearwise fine-tuning} for \texttt{flan-t5-xl}.}
\label{fig:rab-based-yl-flan-t5-xl}
\end{figure*}

\begin{figure*}
\begin{center}
\includegraphics[width=0.9\linewidth]{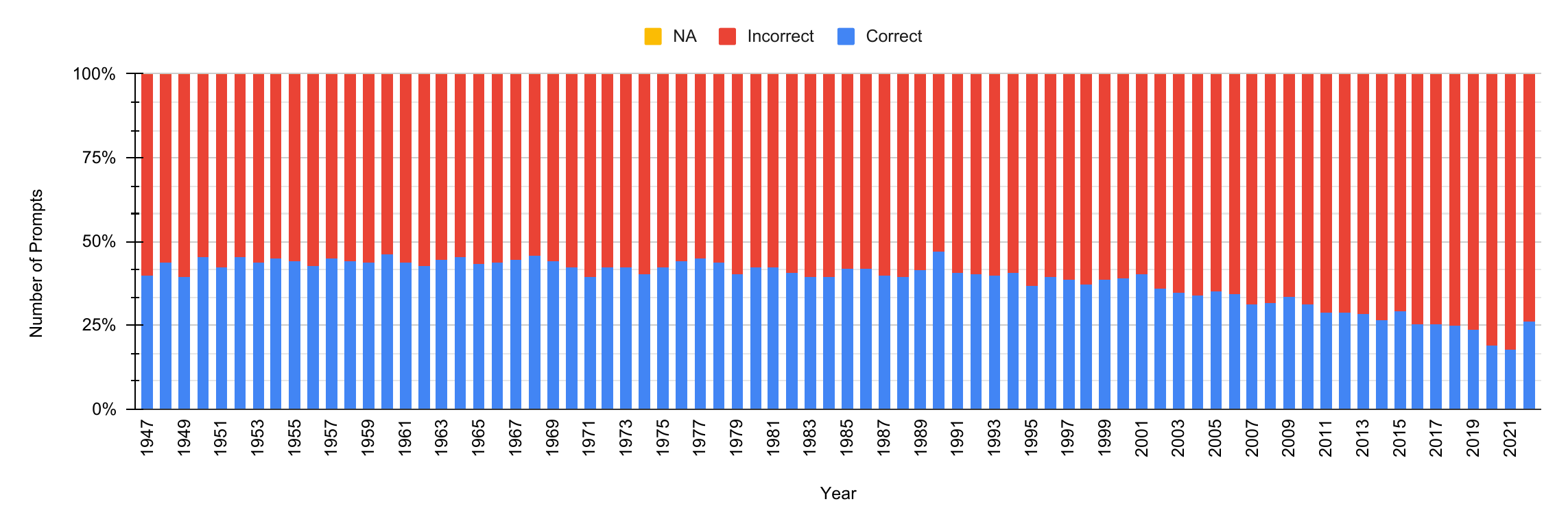}
\end{center}
\caption{Plot for the Trend-based metric ($TB$) as year-wise count (In percentage) for \textbf{yearwise fine-tuning} for \texttt{flan-t5-xl}.}
\label{fig:tb-based-yl-flan-t5-xl}
\end{figure*}

\begin{figure*}
\begin{center}
\includegraphics[width=0.9\linewidth]{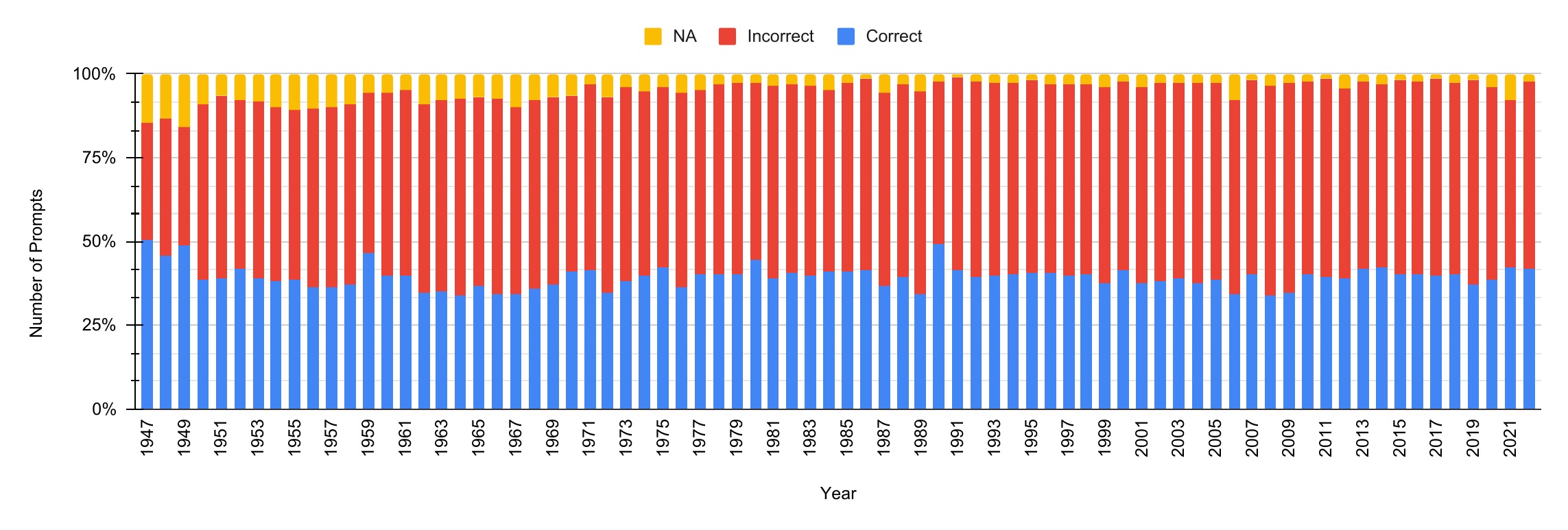}
\end{center}
\caption{Plot for the Date-based metric ($DB$) as year-wise count (In percentage) for \textbf{yearwise fine-tuning} for for \texttt{mistral-instruct}.}
\label{fig:date-based-yl-mistral}
\end{figure*}

\begin{figure*}
\begin{center}
\includegraphics[width=0.9\linewidth]{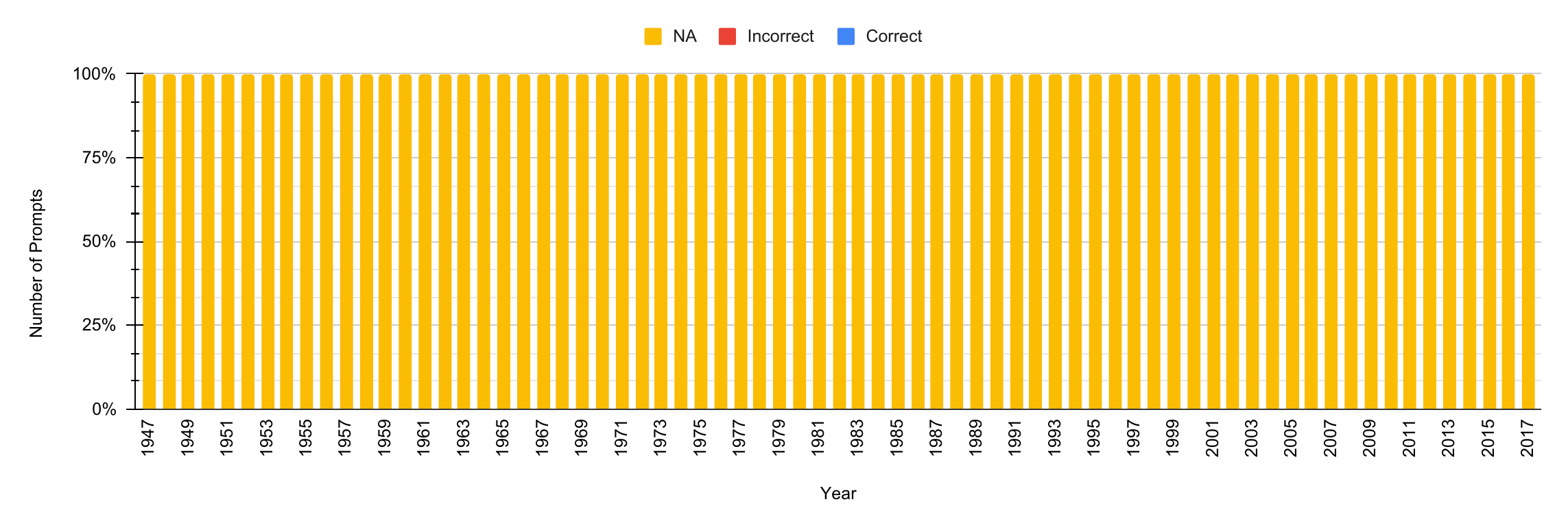}
\end{center}
\caption{Plot for the Comparative-based metric ($CP$) as year-wise count (In percentage) for \textbf{yearwise fine-tuning} for \texttt{mistral-instruct}.}
\label{fig:cp-yl-mistral}
\end{figure*}

\begin{figure*}
\begin{center}
\includegraphics[width=0.9\linewidth]{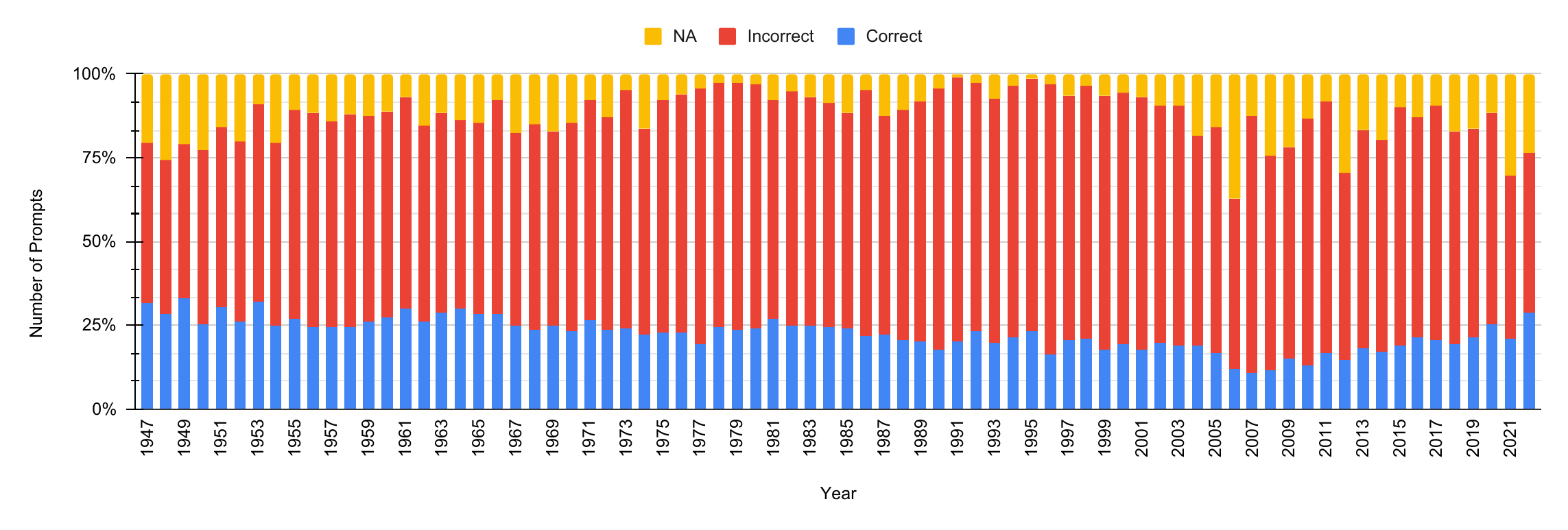}
\end{center}
\caption{Plot for the Window-based metric ($WB$) as year-wise count (In percentage) for \textbf{yearwise fine-tuning} for \texttt{mistral-instruct}.}
\label{fig:window-based-yl-mistral}
\end{figure*}
\begin{figure*}
\begin{center}
\includegraphics[width=0.9\linewidth]{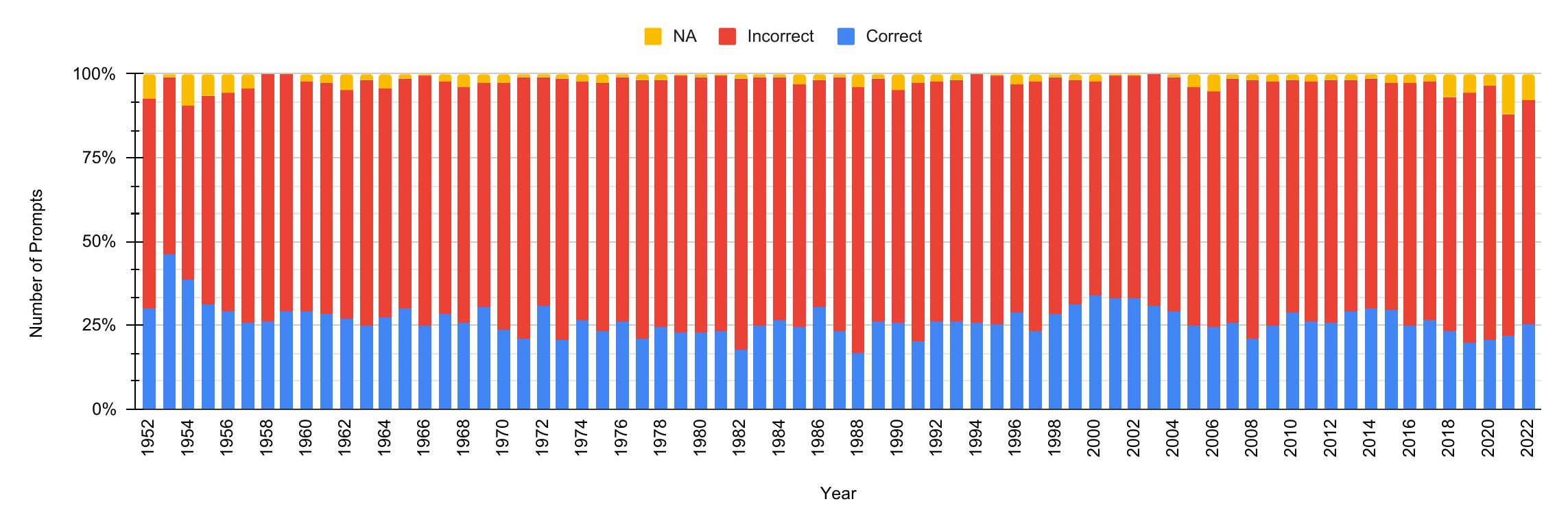}
\end{center}
\caption{Plot for the Min/Max-based metric ($MM$) as year-wise count (In percentage) for \textbf{yearwise fine-tuning} for \texttt{mistral-instruct}.}
\label{fig:minmax-based-yl-mistral}
\end{figure*}

\begin{figure*}
\begin{center}
\includegraphics[width=0.9\linewidth]{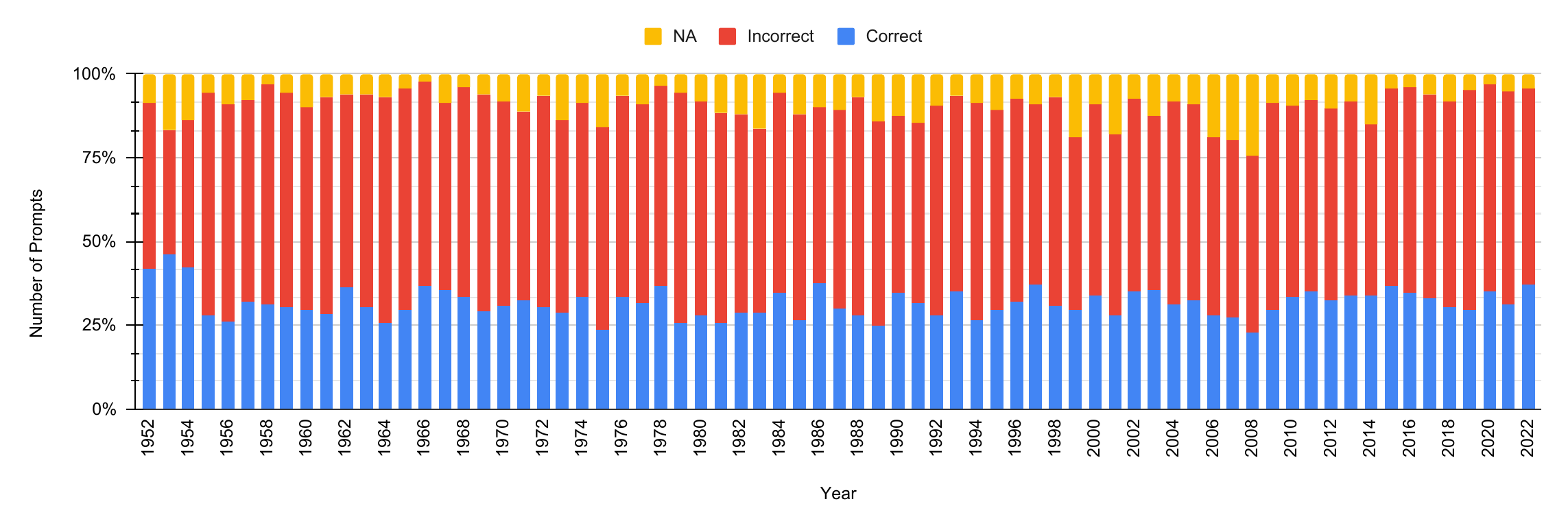}
\end{center}
\caption{Plot for the Range-based metric ($RB$) as year-wise count (In percentage) for \textbf{yearwise fine-tuning} for \texttt{mistral-instruct}.}
\label{fig:rab-based-yl-mistral}
\end{figure*}

\begin{figure*}
\begin{center}
\includegraphics[width=0.9\linewidth]{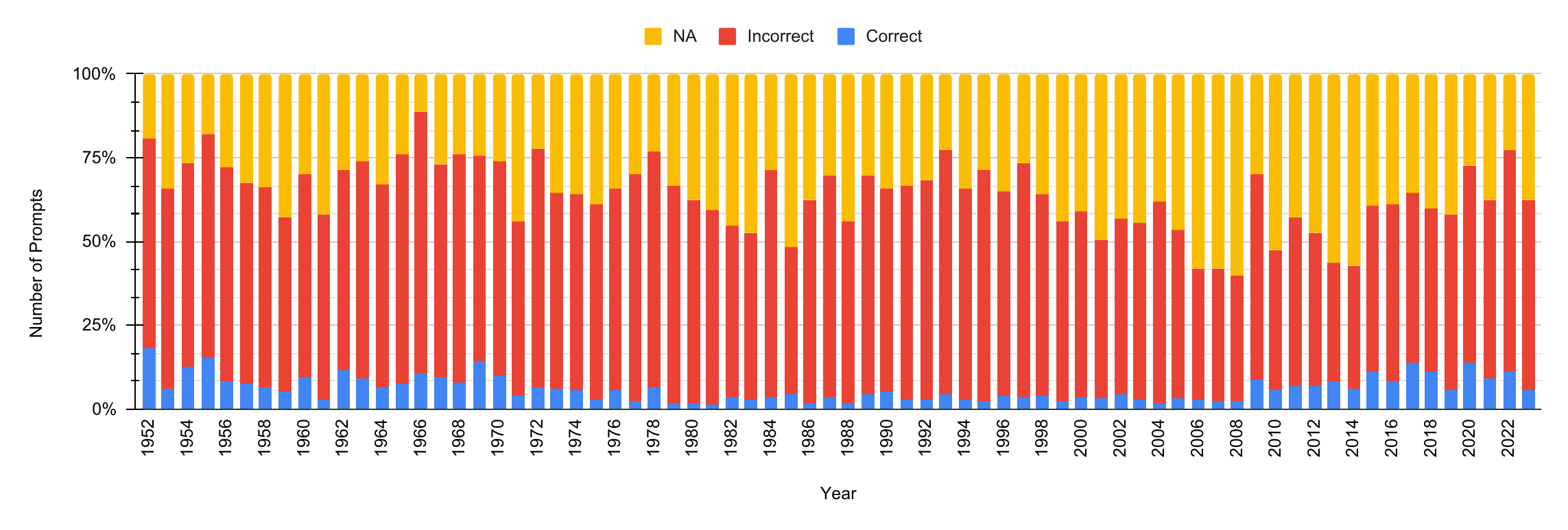}
\end{center}
\caption{Plot for the Trend-based metric ($TB$) as year-wise count (In percentage) for \textbf{yearwise fine-tuning} for \texttt{mistral-instruct}.}
\label{fig:tb-based-yl-mistral}
\end{figure*}

\begin{figure*}
\begin{center}
\includegraphics[width=0.9\linewidth]{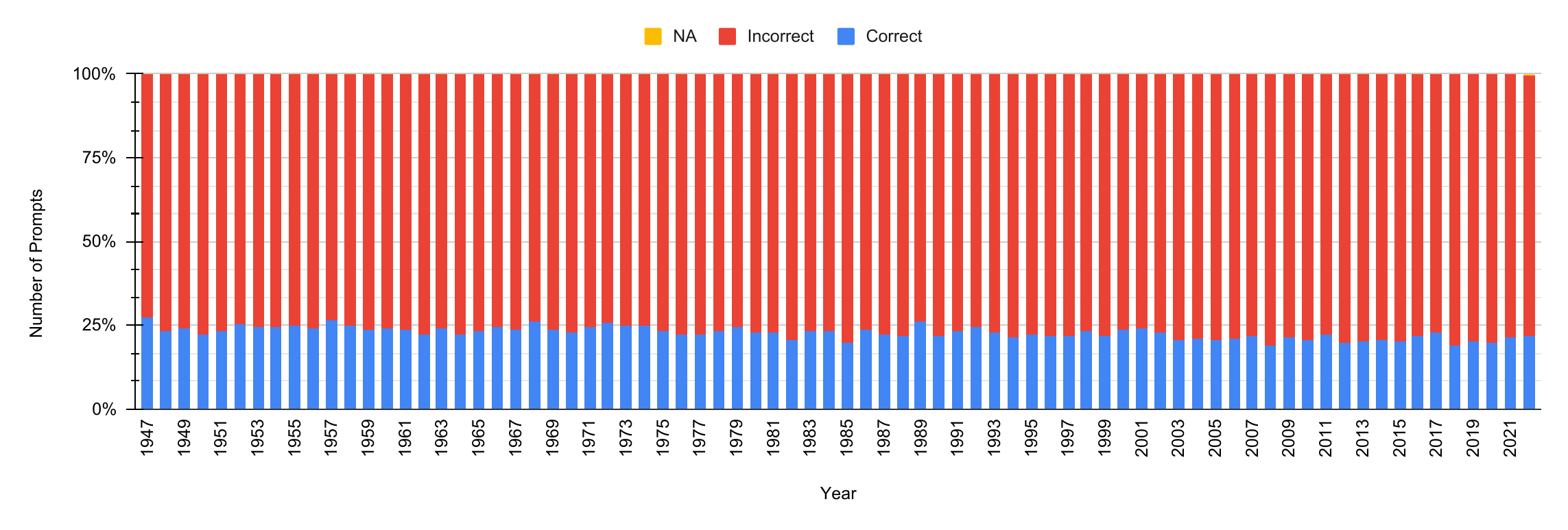}
\end{center}
\caption{Plot for the Date-based metric ($DB$) as year-wise count (In percentage) for \textbf{yearwise fine-tuning} for \texttt{llama-2}.}
\label{fig:date-based-yl-llama}
\end{figure*}

\begin{figure*}
\begin{center}
\includegraphics[width=0.9\linewidth]{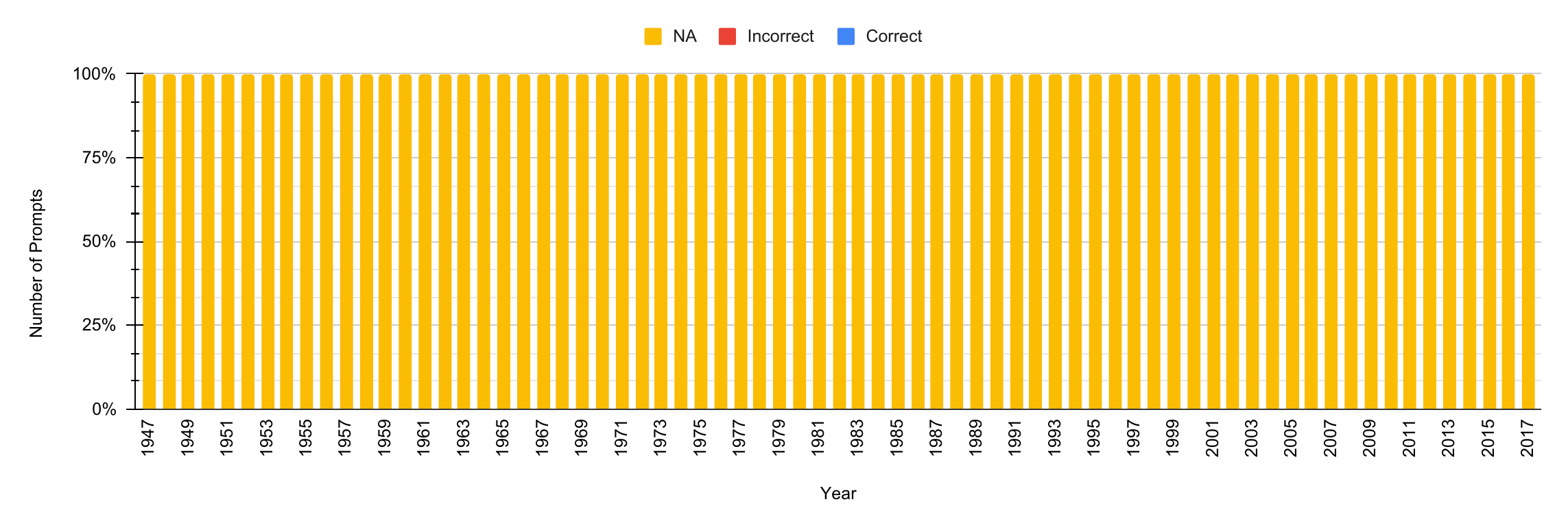}
\end{center}
\caption{Plot for the Comparative-based metric ($CP$) as year-wise count (In percentage) for \textbf{yearwise fine-tuning} for \texttt{llama-2}.}
\label{fig:cp-yl-llama}
\end{figure*}

\begin{figure*}
\begin{center}
\includegraphics[width=0.9\linewidth]{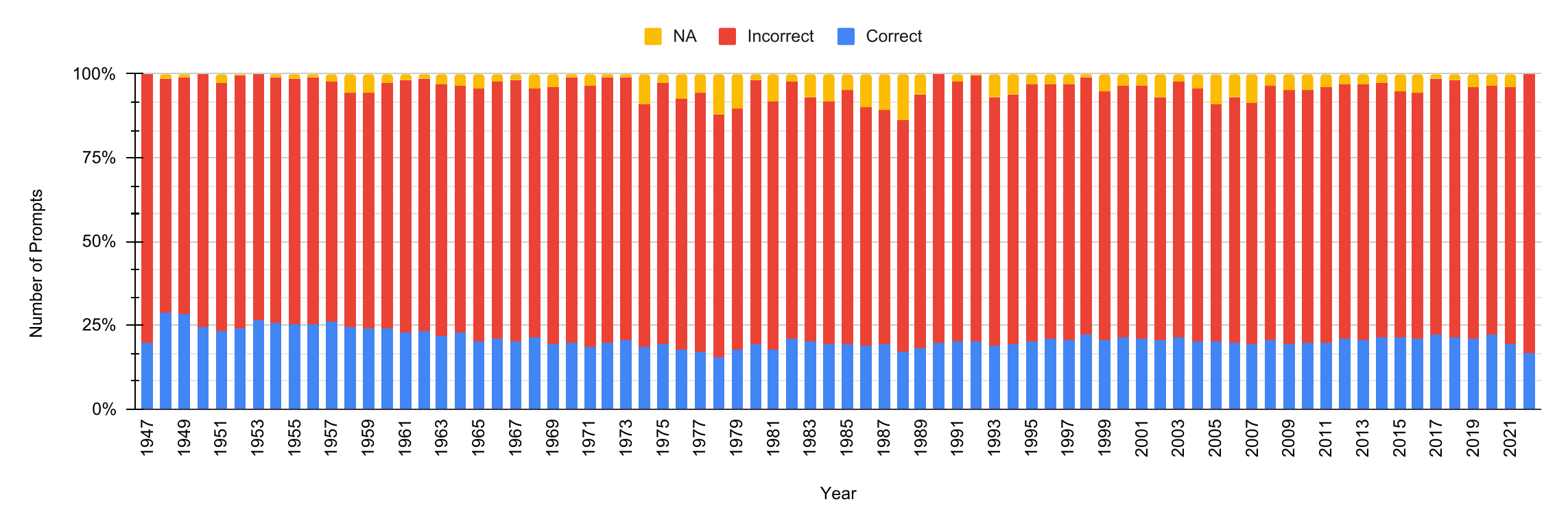}
\end{center}
\caption{Plot for the Window-based metric ($WB$) as year-wise count (In percentage) for \textbf{yearwise fine-tuning} for \texttt{llama-2}.}
\label{fig:window-based-yl-llama}
\end{figure*}
\begin{figure*}
\begin{center}
\includegraphics[width=0.9\linewidth]{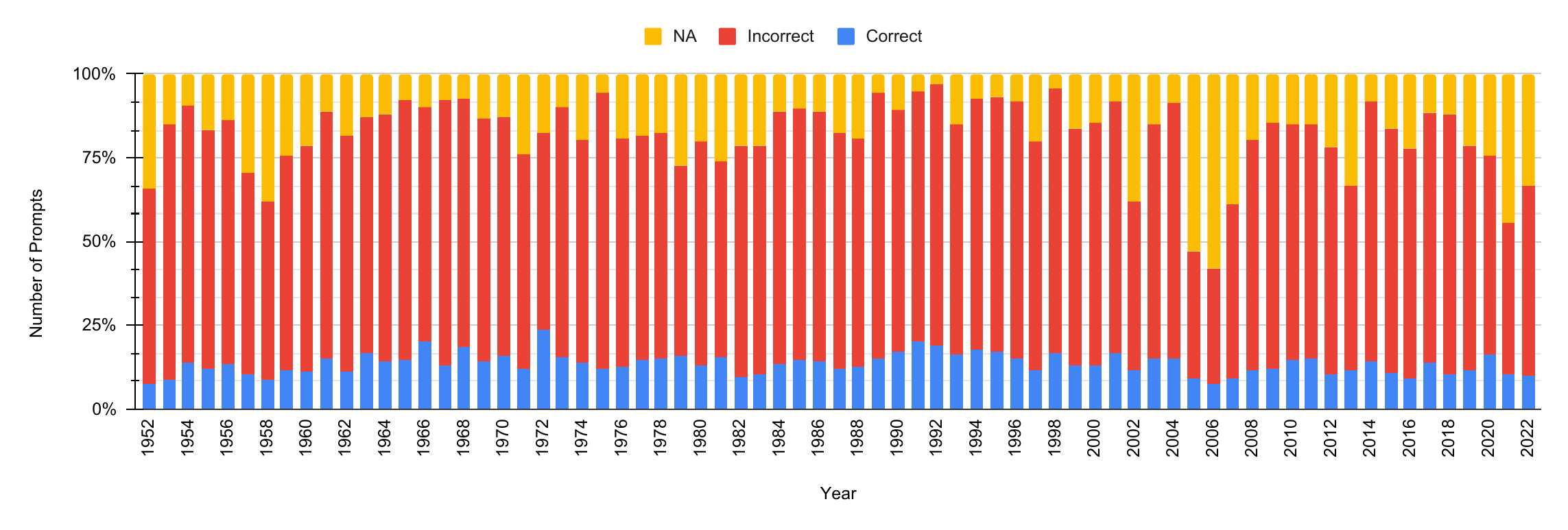}
\end{center}
\caption{Plot for the Min/Max-based metric ($MM$) as year-wise count (In percentage) for \textbf{yearwise fine-tuning} for \texttt{llama-2}.}
\label{fig:minmax-based-yl-llama}
\end{figure*}

\begin{figure*}
\begin{center}
\includegraphics[width=0.9\linewidth]{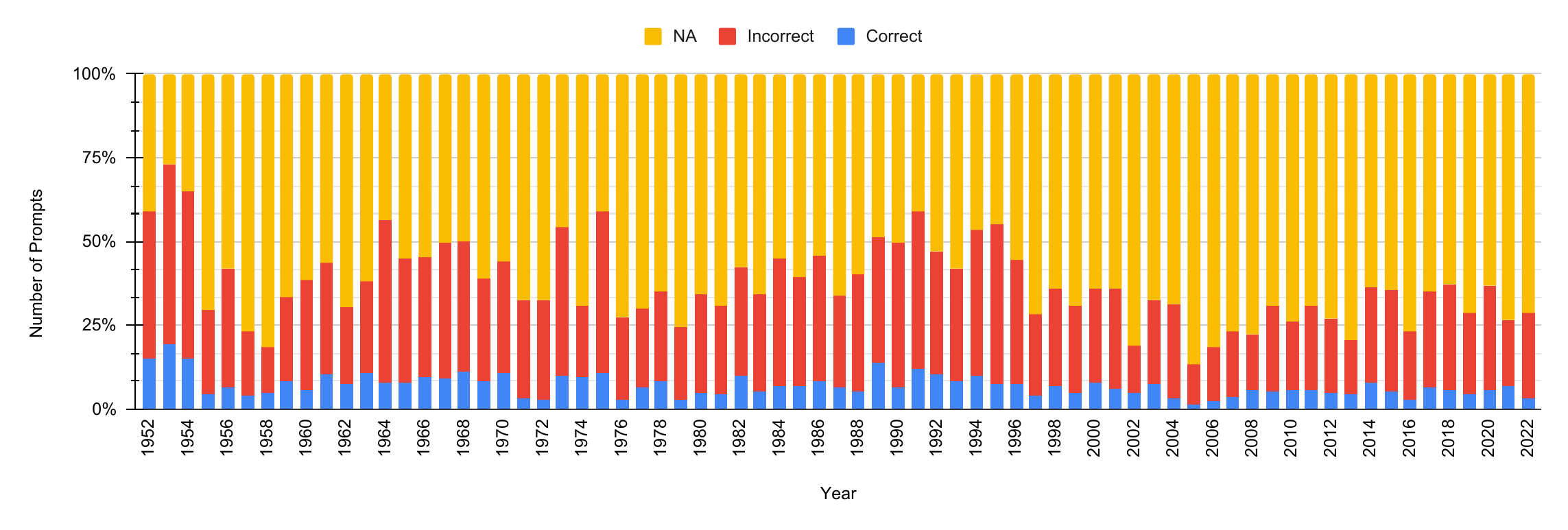}
\end{center}
\caption{Plot for the Range-based metric ($RB$) as year-wise count (In percentage) for \textbf{yearwise fine-tuning} for \texttt{llama-2}.}
\label{fig:rab-based-yl-llama}
\end{figure*}

\begin{figure*}
\begin{center}
\includegraphics[width=0.9\linewidth]{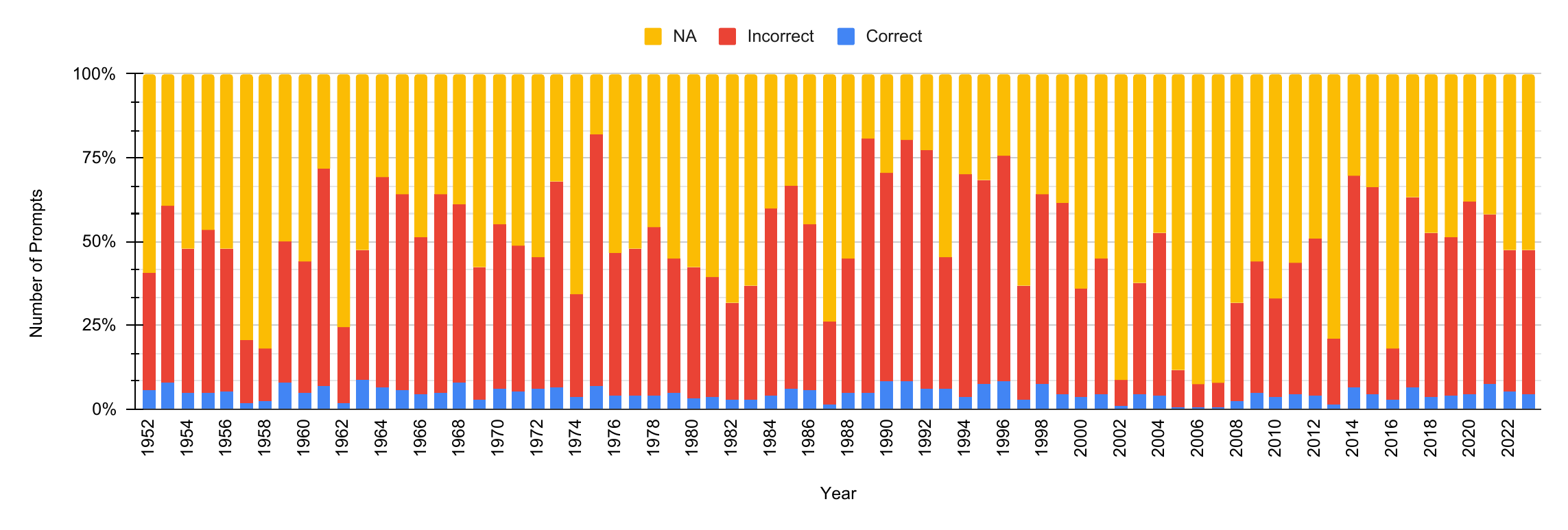}
\end{center}
\caption{Plot for the Trend-based metric ($TB$) as year-wise count (In percentage) for \textbf{yearwise fine-tuning} for \texttt{llama-2}.}
\label{fig:tb-based-yl-llama}
\end{figure*}


\begin{figure*}
\begin{center}
\includegraphics[width=0.9\linewidth]{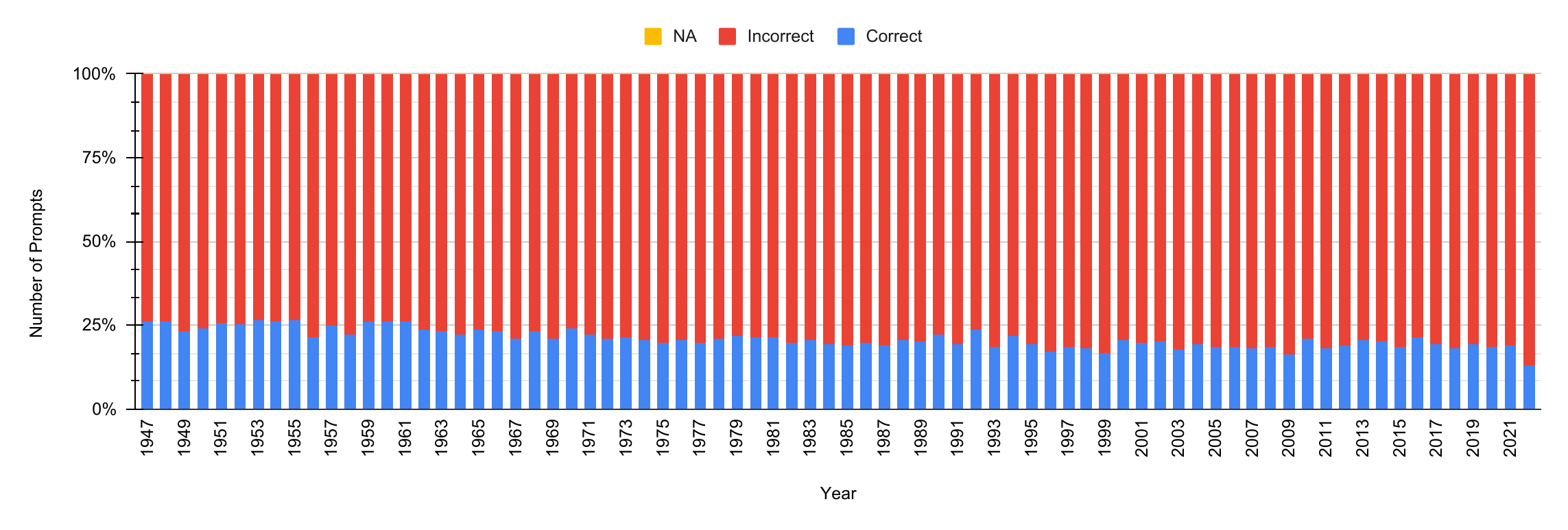}
\end{center}
\caption{Plot for the Date-based metric ($DB$) as year-wise count (In percentage) for \textbf{yearwise fine-tuning} for \texttt{gemma-7b-it}.}
\label{fig:date-based-yl-gemma-7b-it}
\end{figure*}

\begin{figure*}
\begin{center}
\includegraphics[width=0.9\linewidth]{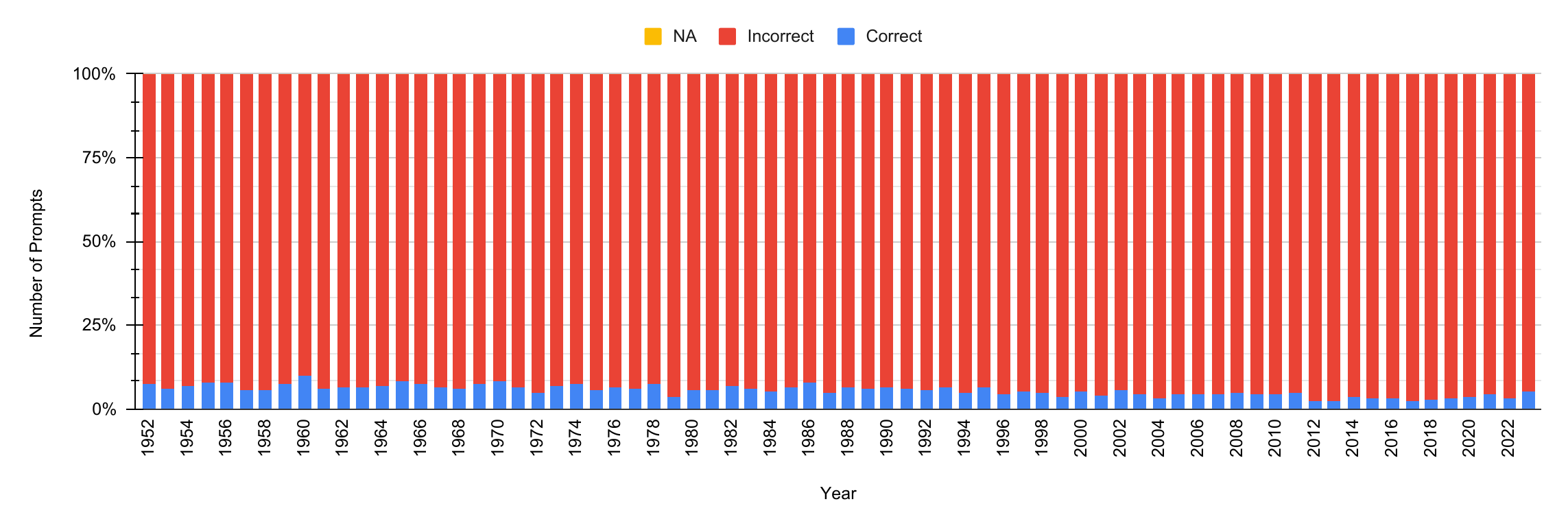}
\end{center}
\caption{Plot for the Comparative-based metric ($CP$) as year-wise count (In percentage) for \textbf{yearwise fine-tuning} for \texttt{gemma-7b-it}.}
\label{fig:cp-yl-gemma-7b-it}
\end{figure*}

\begin{figure*}
\begin{center}
\includegraphics[width=0.9\linewidth]{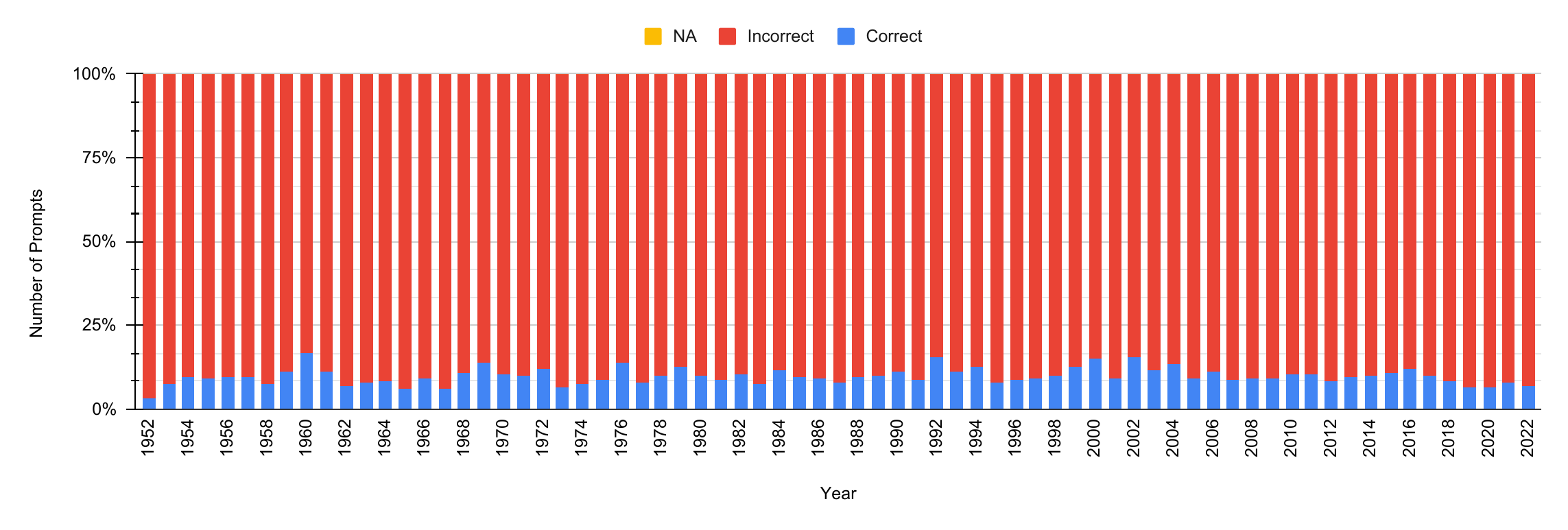}
\end{center}
\caption{Plot for the Window-based metric ($WB$) as year-wise count (In percentage) for \textbf{yearwise fine-tuning} for \texttt{gemma-7b-it}.}
\label{fig:window-based-yl-gemma-7b-it}
\end{figure*}
\begin{figure*}
\begin{center}
\includegraphics[width=0.9\linewidth]{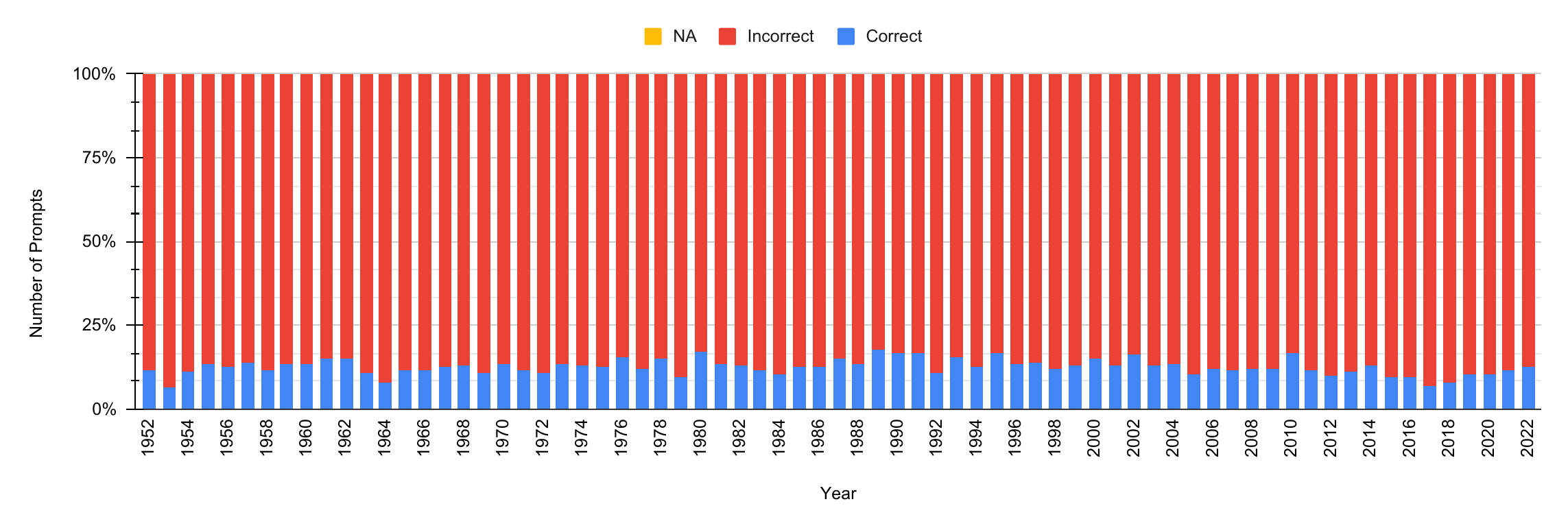}
\end{center}
\caption{Plot for the Min/Max-based metric ($MM$) as year-wise count (In percentage) for \textbf{yearwise fine-tuning} for \texttt{gemma-7b-it}.}
\label{fig:minmax-based-yl-gemma-7b-it}
\end{figure*}

\begin{figure*}
\begin{center}
\includegraphics[width=0.9\linewidth]{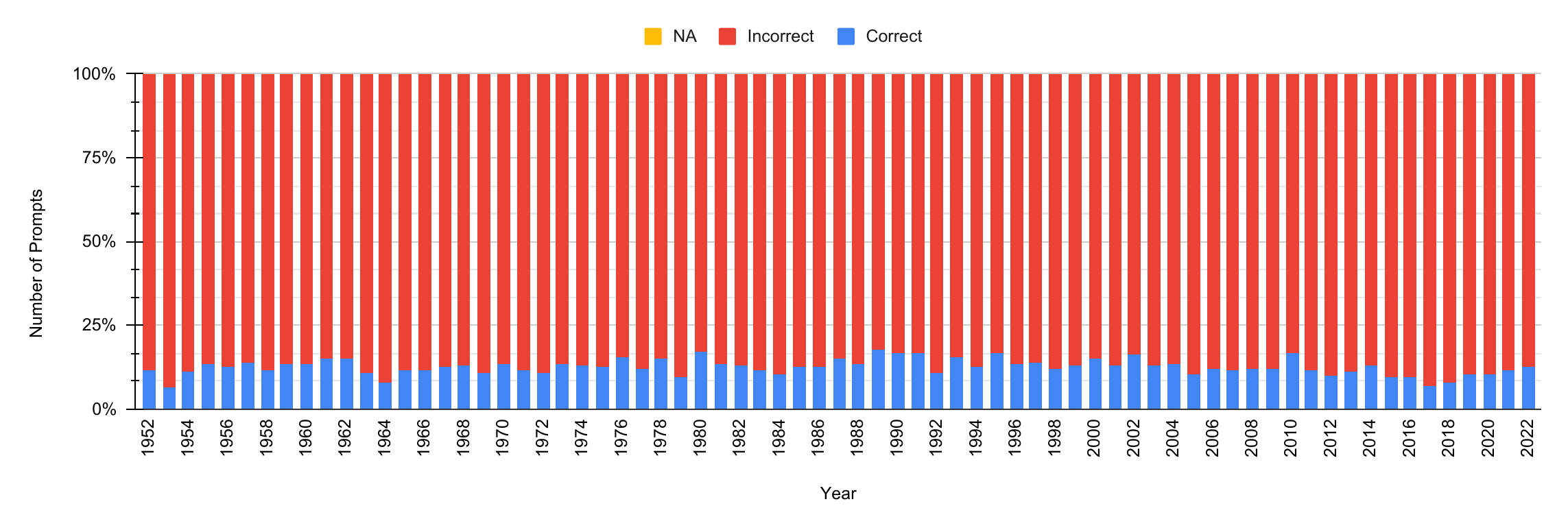}
\end{center}
\caption{Plot for the Range-based metric ($RB$) as year-wise count (In percentage) for \textbf{yearwise fine-tuning} for \texttt{gemma-7b-it}.}
\label{fig:rab-based-yl-gemma-7b-it}
\end{figure*}

\begin{figure*}
\begin{center}
\includegraphics[width=0.9\linewidth]{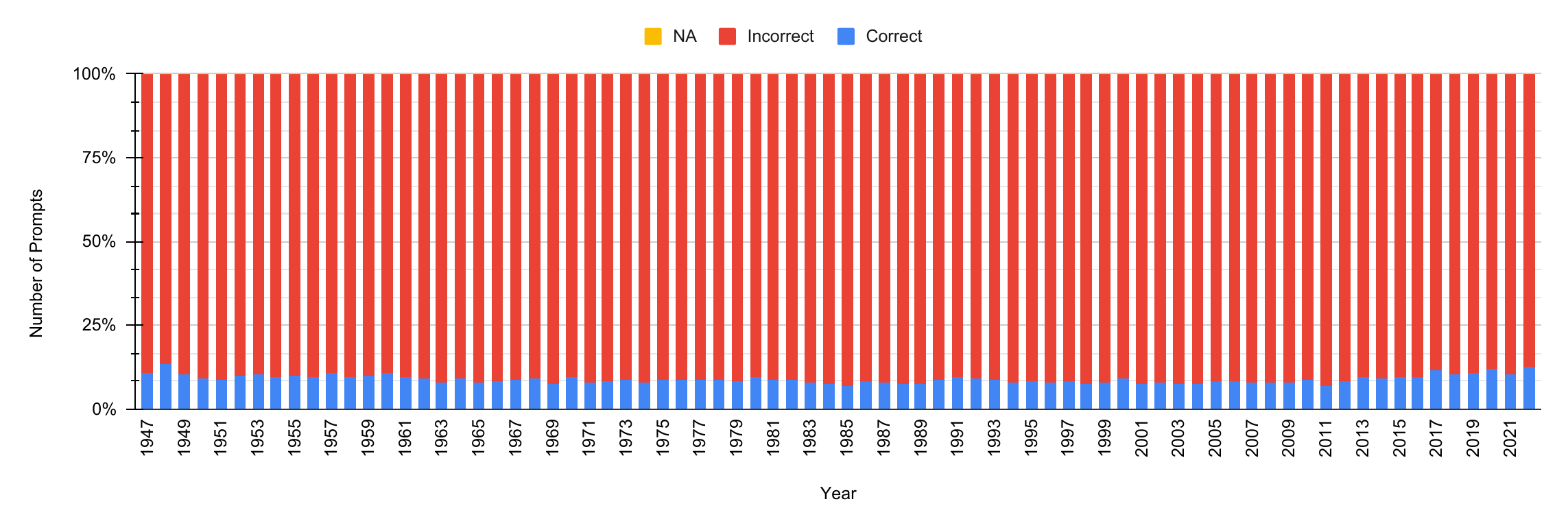}
\end{center}
\caption{Plot for the Trend-based metric ($TB$) as year-wise count (In percentage) for \textbf{yearwise fine-tuning} for \texttt{gemma-7b-it}.}
\label{fig:tb-based-yl-gemma-7b-it}
\end{figure*}


\begin{figure*}
\begin{center}
\includegraphics[width=0.9\linewidth]{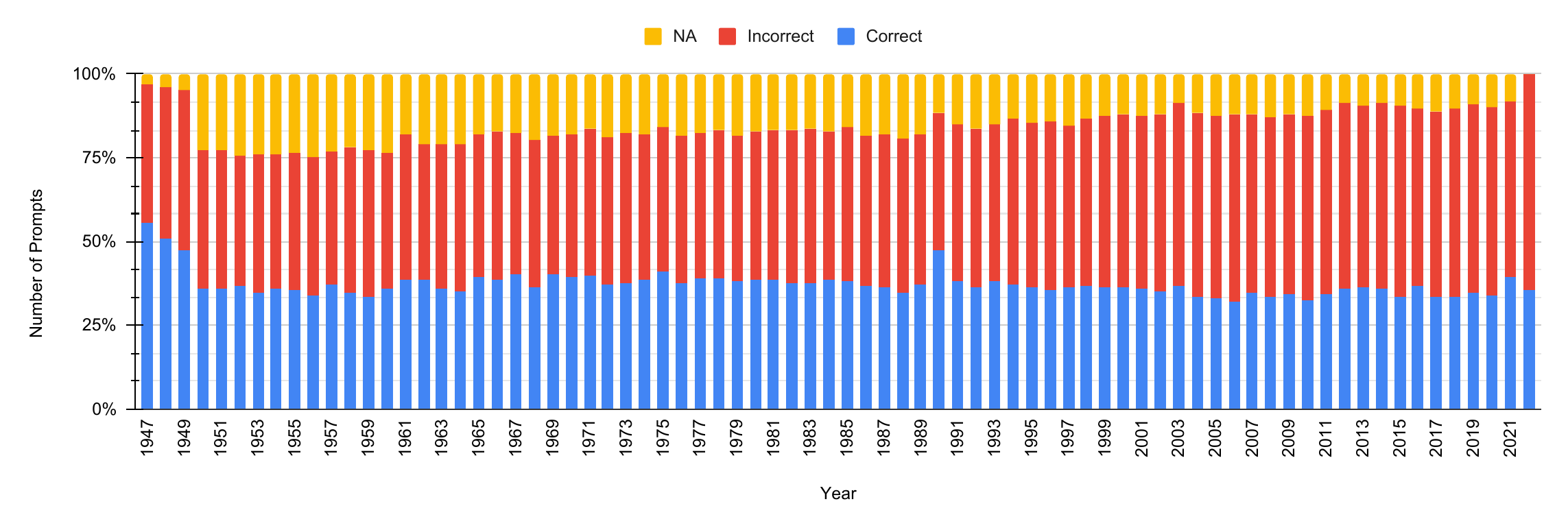}
\end{center}
\caption{Plot for the Date-based metric ($DB$) as year-wise count (In percentage) for \textbf{yearwise fine-tuning} for \texttt{llama-3-8b}.}
\label{fig:date-based-yl-llama-3-8b}
\end{figure*}

\begin{figure*}
\begin{center}
\includegraphics[width=0.9\linewidth]{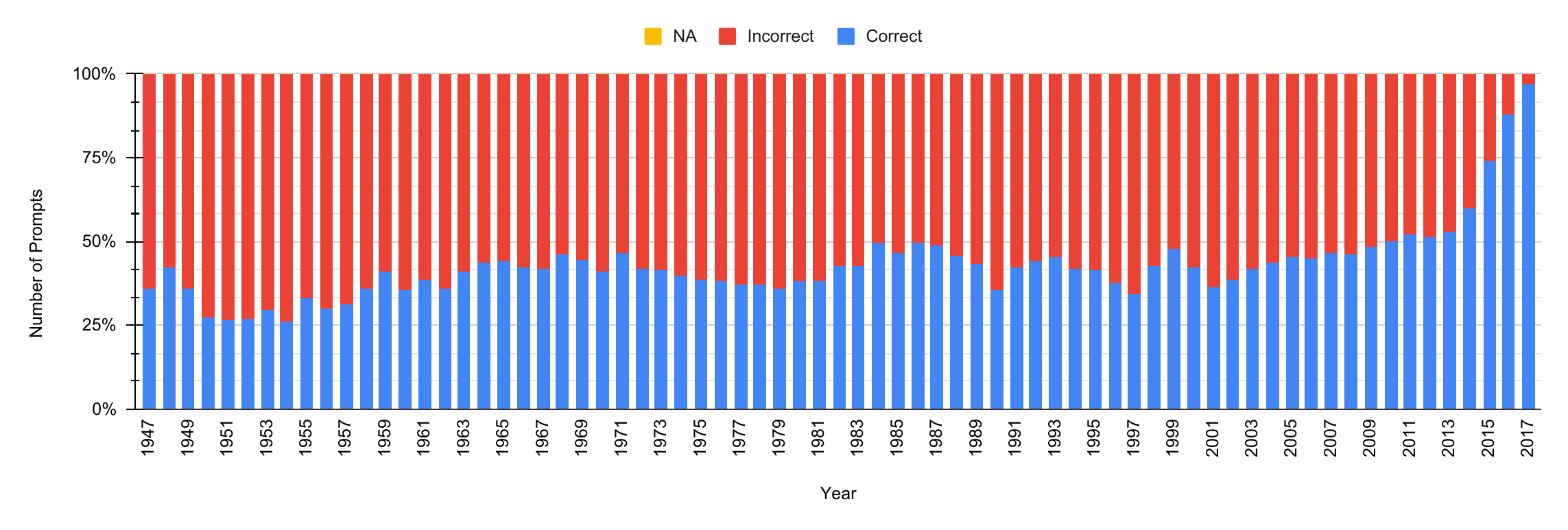}
\end{center}
\caption{Plot for the Comparative-based metric ($CP$) as year-wise count (In percentage) for \textbf{yearwise fine-tuning} for \texttt{llama-3-8b}.}
\label{fig:cp-yl-llama-3-8b}
\end{figure*}

\begin{figure*}
\begin{center}
\includegraphics[width=0.9\linewidth]{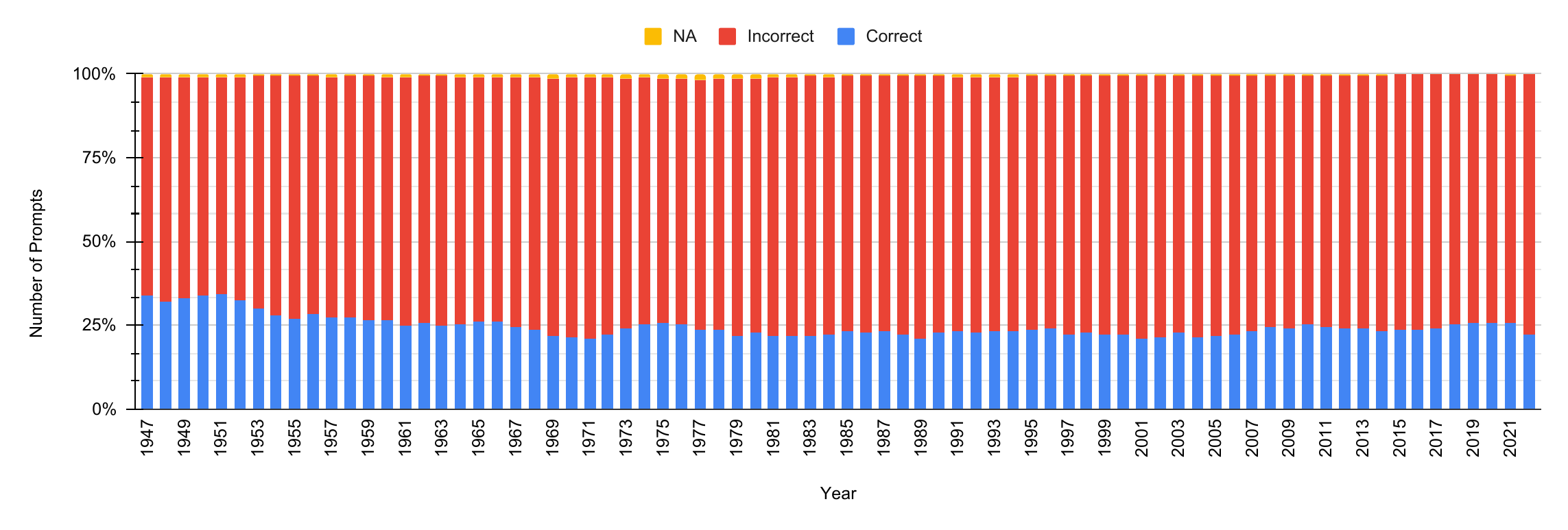}
\end{center}
\caption{Plot for the Window-based metric ($WB$) as year-wise count (In percentage) for \textbf{yearwise fine-tuning} for \texttt{llama-3-8b}.}
\label{fig:window-based-yl-llama-3-8b}
\end{figure*}
\begin{figure*}
\begin{center}
\includegraphics[width=0.9\linewidth]{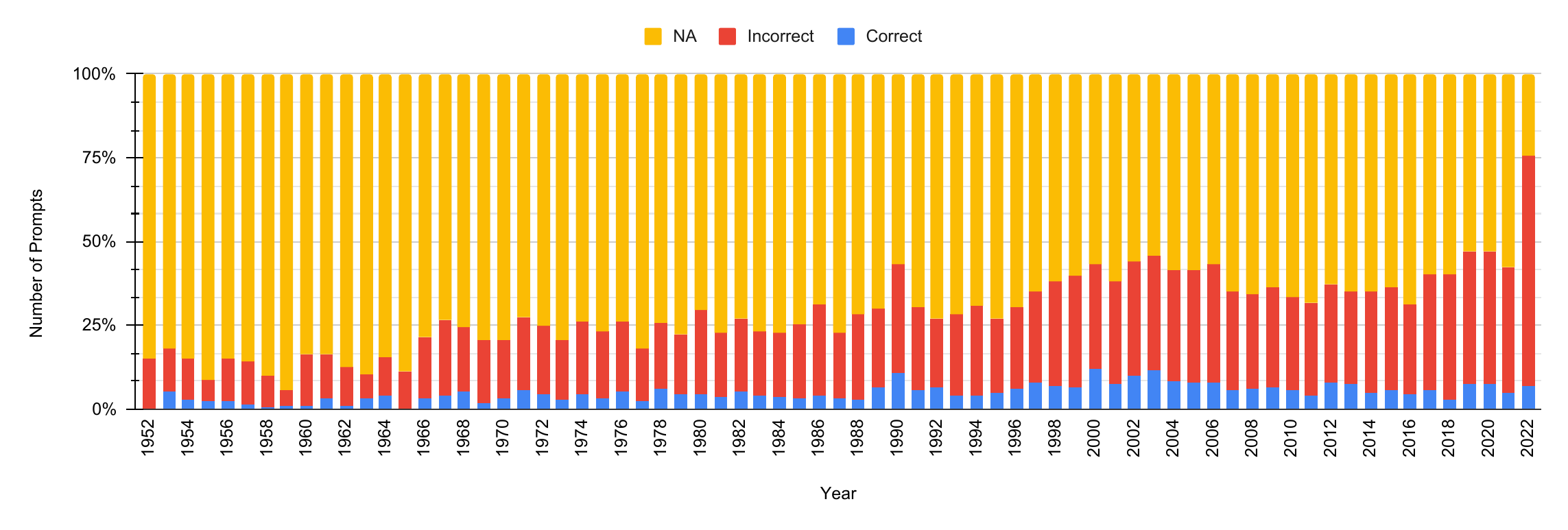}
\end{center}
\caption{Plot for the Min/Max-based metric ($MM$) as year-wise count (In percentage) for \textbf{yearwise fine-tuning} for \texttt{llama-3-8b}.}
\label{fig:minmax-based-yl-llama-3-8b}
\end{figure*}

\begin{figure*}
\begin{center}
\includegraphics[width=0.9\linewidth]{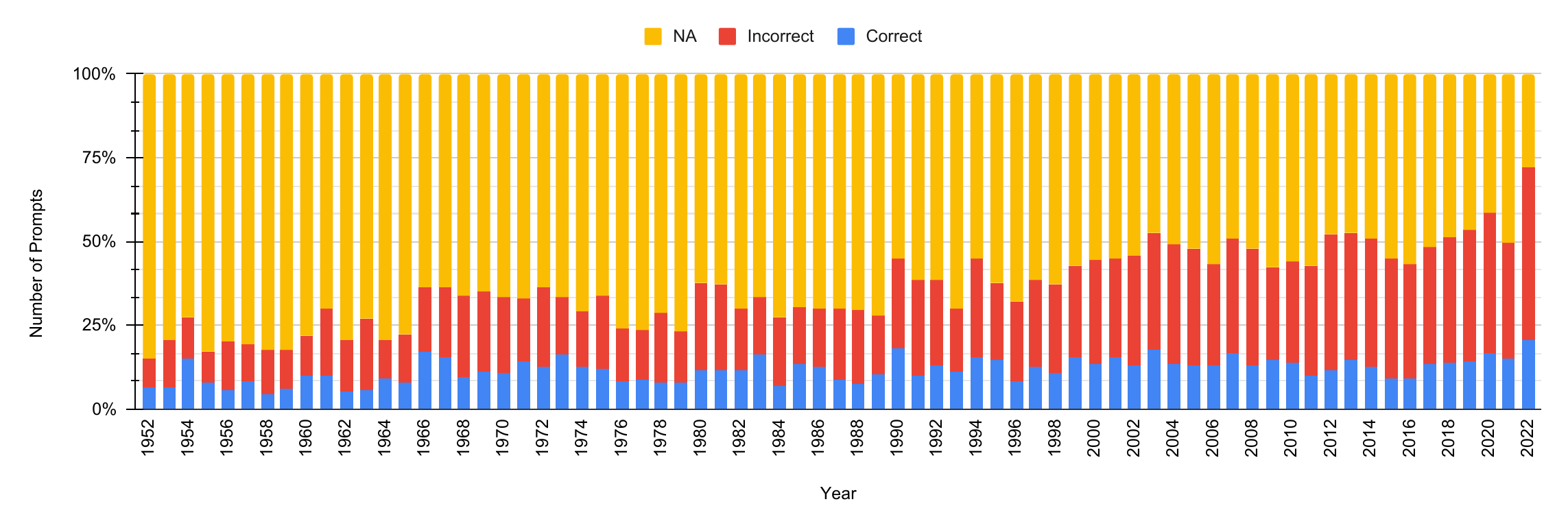}
\end{center}
\caption{Plot for the Range-based metric ($RB$) as year-wise count (In percentage) for \textbf{yearwise fine-tuning} for \texttt{llama-3-8b}.}
\label{fig:rab-based-yl-llama-3-8b}
\end{figure*}

\begin{figure*}
\begin{center}
\includegraphics[width=0.9\linewidth]{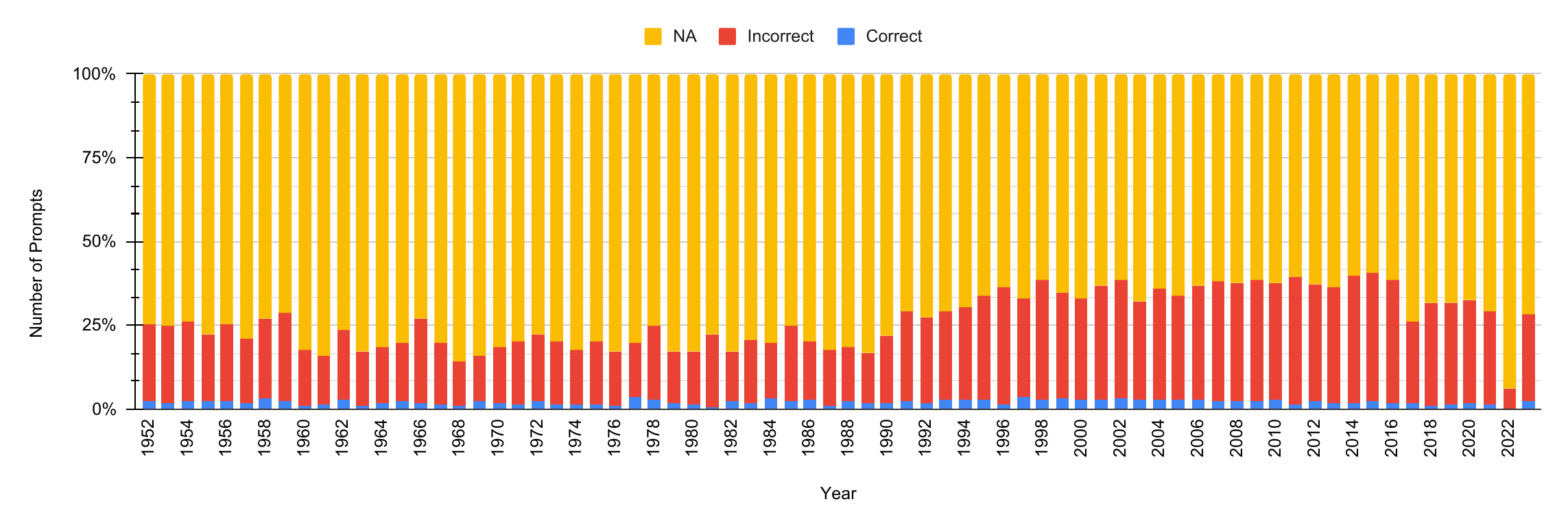}
\end{center}
\caption{Plot for the Trend-based metric ($TB$) as year-wise count (In percentage) for \textbf{yearwise fine-tuning} for \texttt{llama-3-8b}.}
\label{fig:tb-based-yl-llama-3-8b}
\end{figure*}


\begin{figure*}
\begin{center}
\includegraphics[width=0.9\linewidth]{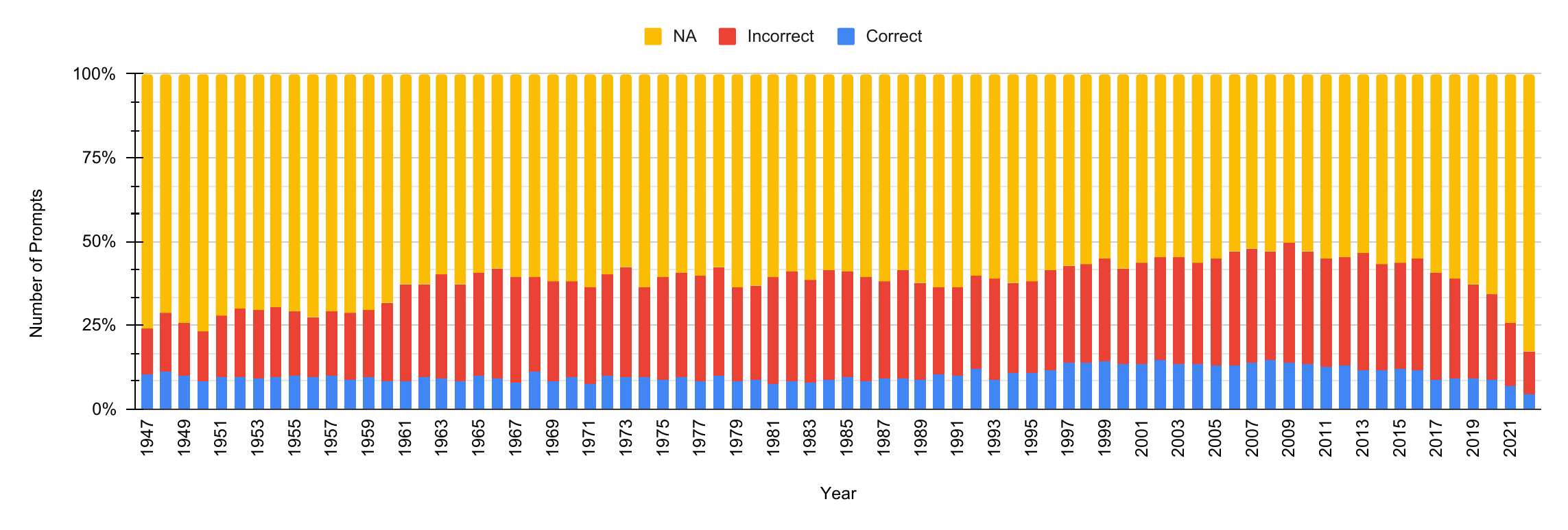}
\end{center}
\caption{Plot for the Date-based metric ($DB$) as year-wise count (In percentage) for \textbf{yearwise fine-tuning} for \texttt{phi-3-medium}.}
\label{fig:date-based-yl-phi-3-medium}
\end{figure*}

\begin{figure*}
\begin{center}
\includegraphics[width=0.9\linewidth]{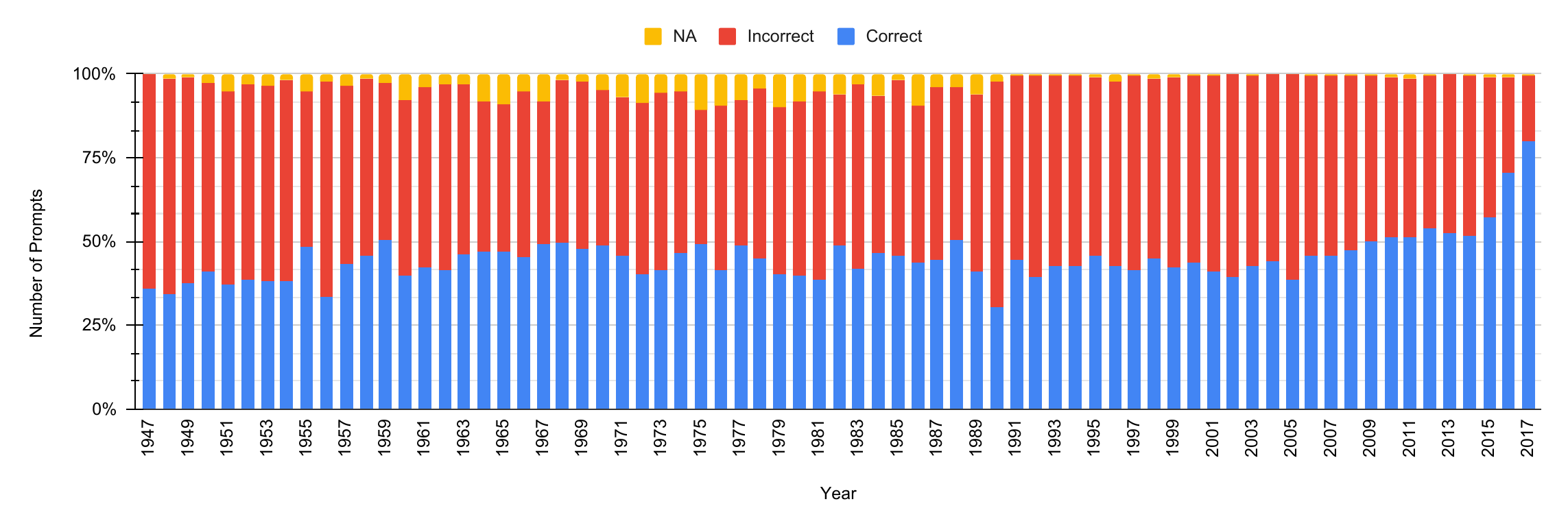}
\end{center}
\caption{Plot for the Comparative-based metric ($CP$) as year-wise count (In percentage) for \textbf{yearwise fine-tuning} for \texttt{phi-3-medium}.}
\label{fig:cp-yl-phi-3-medium}
\end{figure*}

\begin{figure*}
\begin{center}
\includegraphics[width=0.9\linewidth]{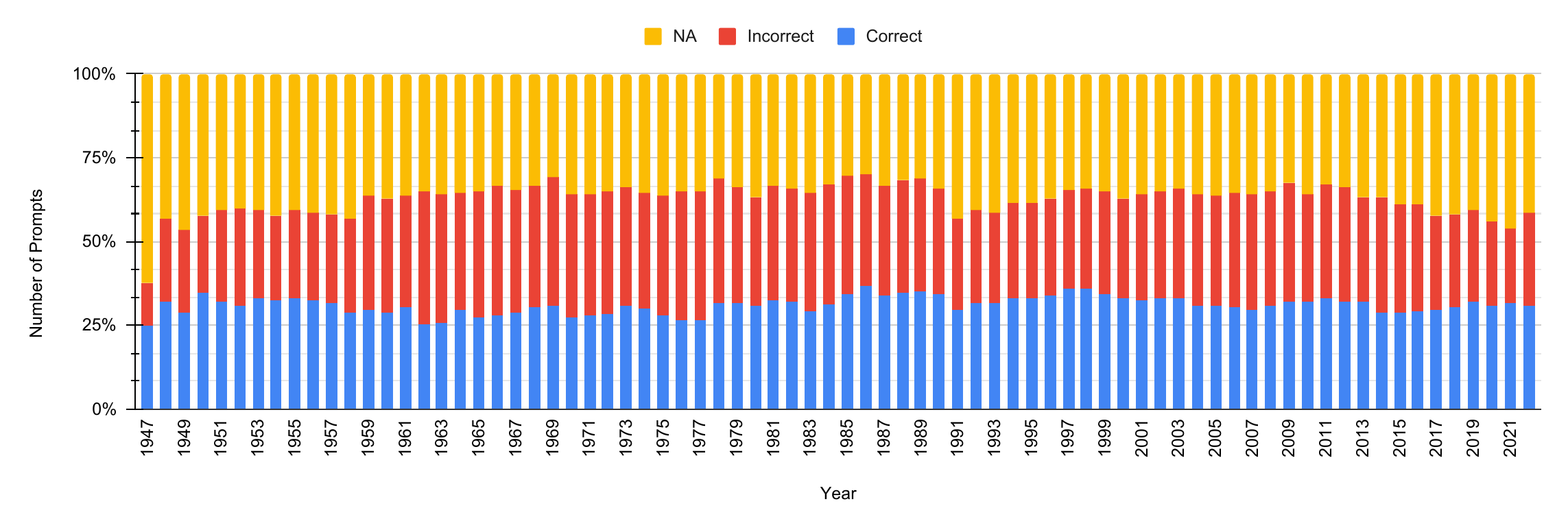}
\end{center}
\caption{Plot for the Window-based metric ($WB$) as year-wise count (In percentage) for \textbf{yearwise fine-tuning} for \texttt{phi-3-medium}.}
\label{fig:window-based-yl-phi-3-medium}
\end{figure*}
\begin{figure*}
\begin{center}
\includegraphics[width=0.9\linewidth]{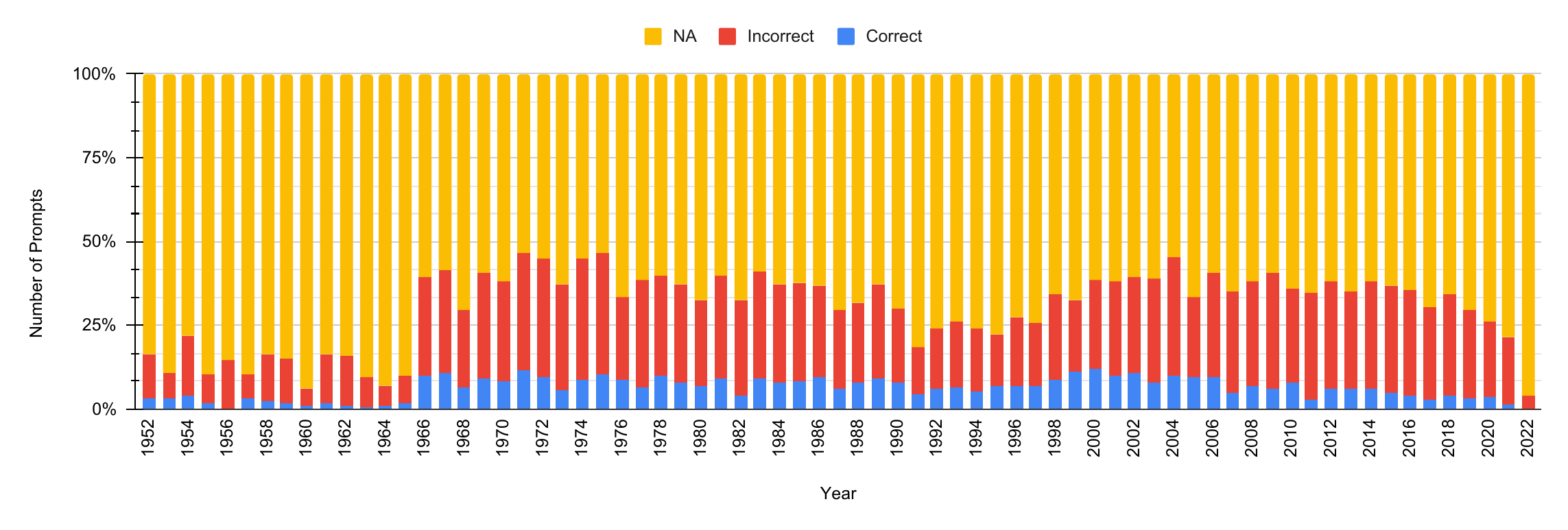}
\end{center}
\caption{Plot for the Min/Max-based metric ($MM$) as year-wise count (In percentage) for \textbf{yearwise fine-tuning} for \texttt{phi-3-medium}.}
\label{fig:minmax-based-yl-phi-3-medium}
\end{figure*}

\begin{figure*}
\begin{center}
\includegraphics[width=0.9\linewidth]{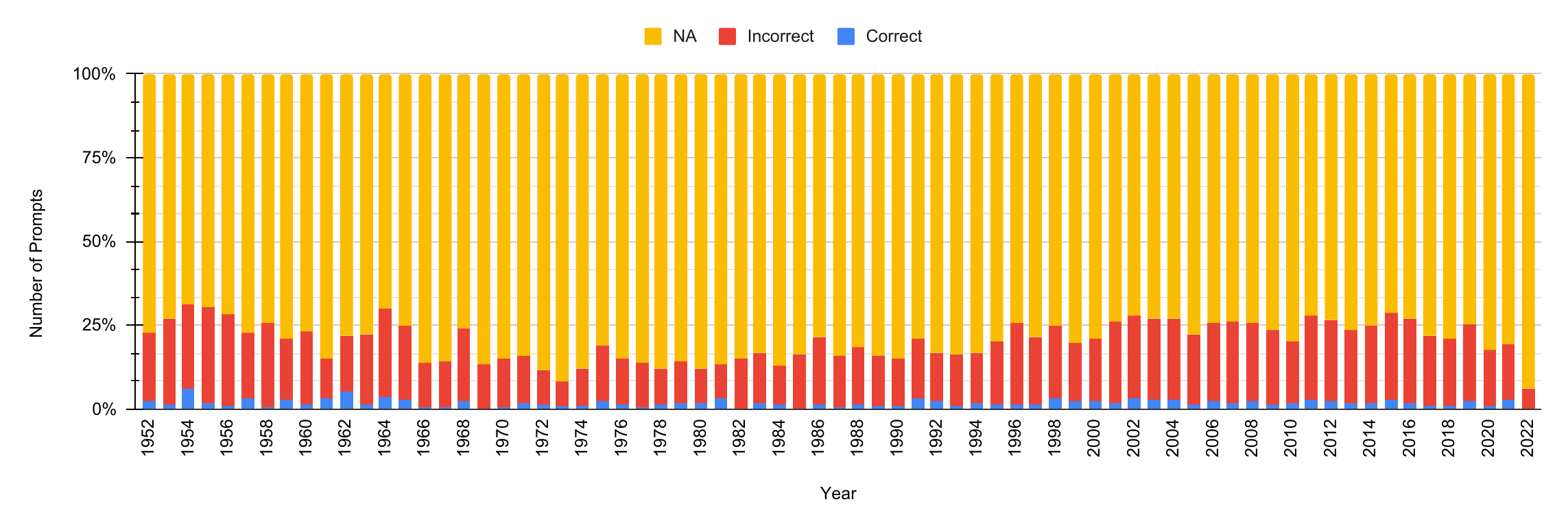}
\end{center}
\caption{Plot for the Range-based metric ($RB$) as year-wise count (In percentage) for \textbf{yearwise fine-tuning} for \texttt{phi-3-medium}.}
\label{fig:rab-based-yl-phi-3-medium}
\end{figure*}

\begin{figure*}
\begin{center}
\includegraphics[width=0.9\linewidth]{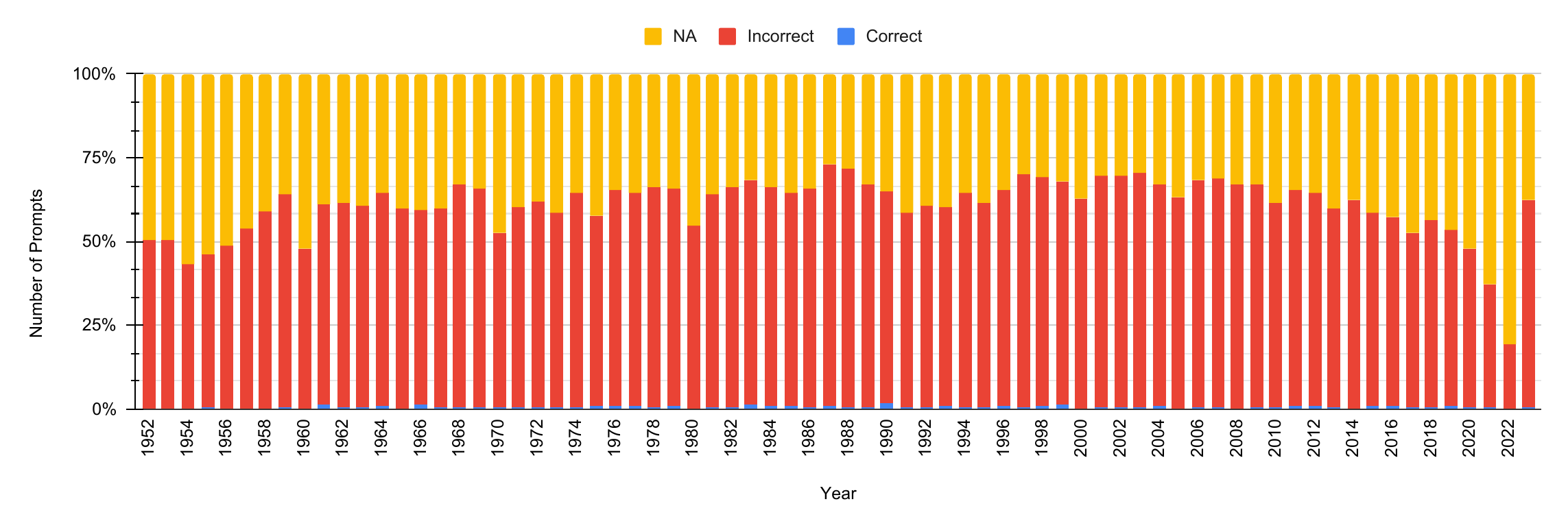}
\end{center}
\caption{Plot for the Trend-based metric ($TB$) as year-wise count (In percentage) for \textbf{yearwise fine-tuning} for \texttt{phi-3-medium}.}
\label{fig:tb-based-yl-phi-3-medium}
\end{figure*}


\begin{figure*}
\begin{center}
\includegraphics[width=0.9\linewidth]{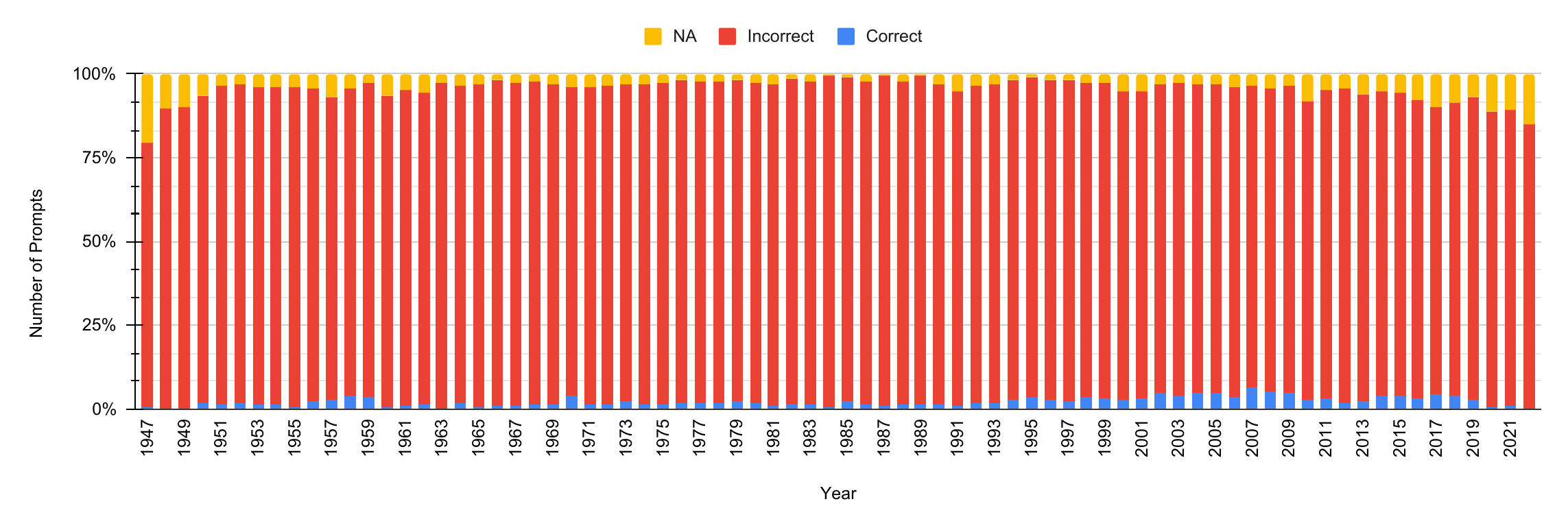}
\end{center}
\caption{Plot for the Date-based metric ($DB$) as year-wise count (In percentage) for \textbf{random fine-tuning} for \texttt{phi-2}.}
\label{fig:date-based-rn-phi2}
\end{figure*}

\begin{figure*}
\begin{center}
\includegraphics[width=0.9\linewidth]{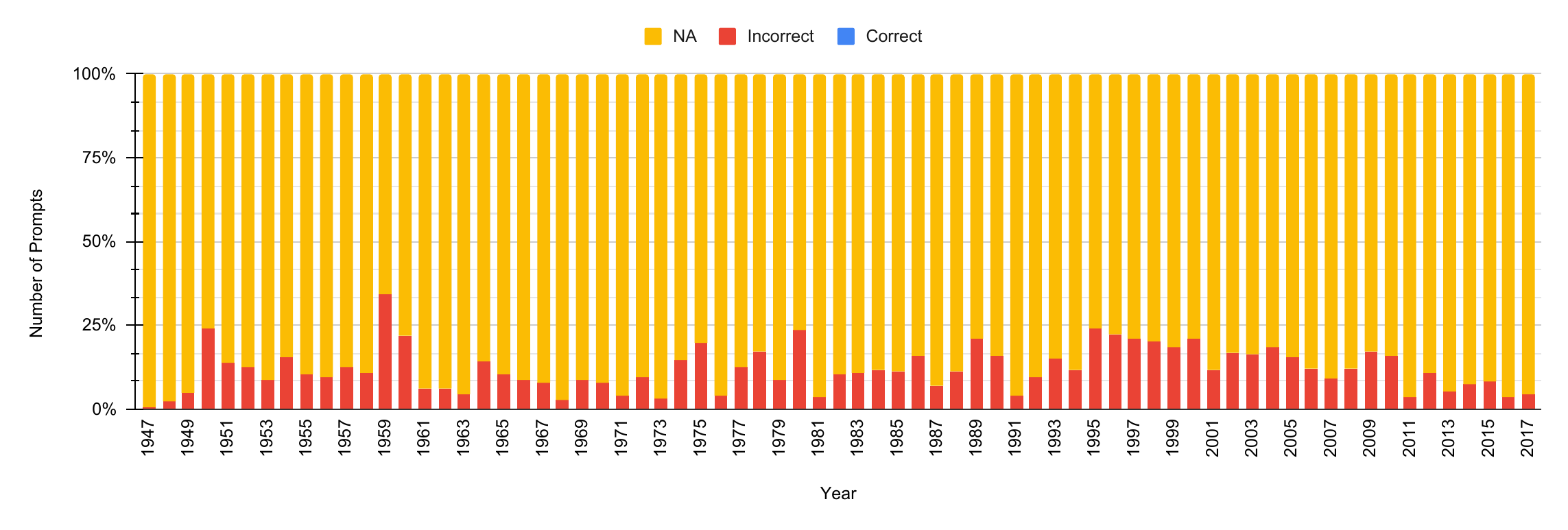}
\end{center}
\caption{Plot for the Comparative-based metric ($CP$) as year-wise count (In percentage) for \textbf{random fine-tuning} for \texttt{phi-2}.}
\label{fig:cp-rn-phi2}
\end{figure*}

\begin{figure*}
\begin{center}
\includegraphics[width=0.9\linewidth]{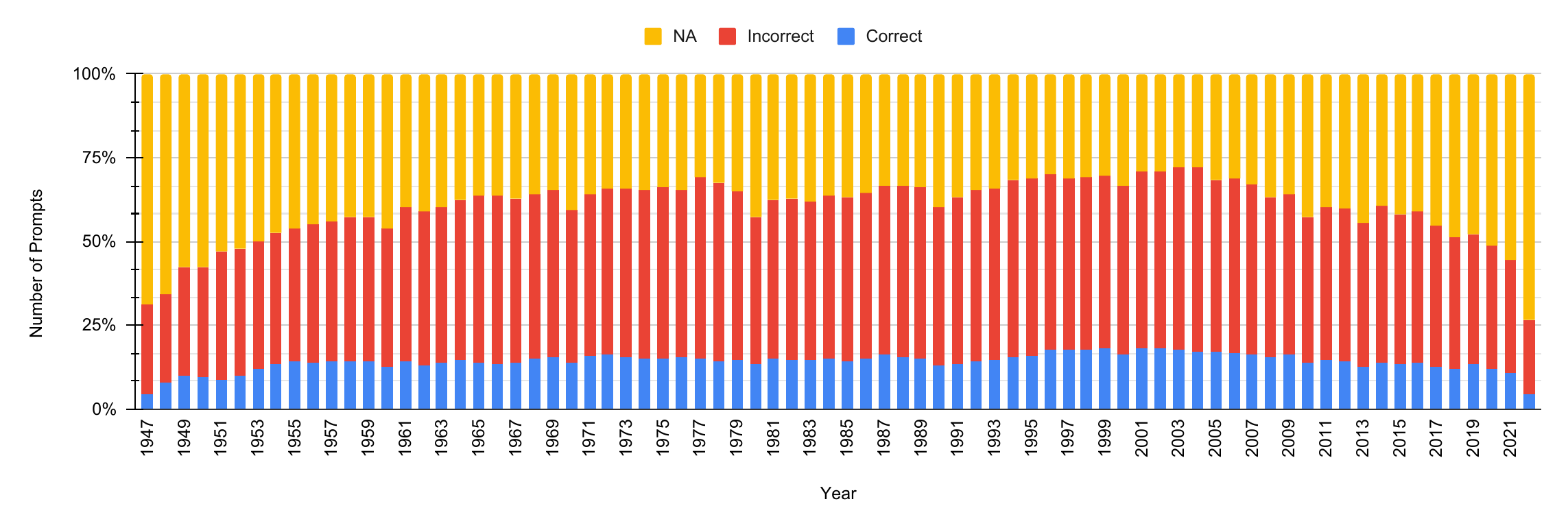}
\end{center}
\caption{Plot for the Window-based metric ($WB$) as year-wise count (In percentage) for \textbf{random fine-tuning} for \texttt{phi-2}.}
\label{fig:window-based-rn-phi2}
\end{figure*}
\begin{figure*}
\begin{center}
\includegraphics[width=0.9\linewidth]{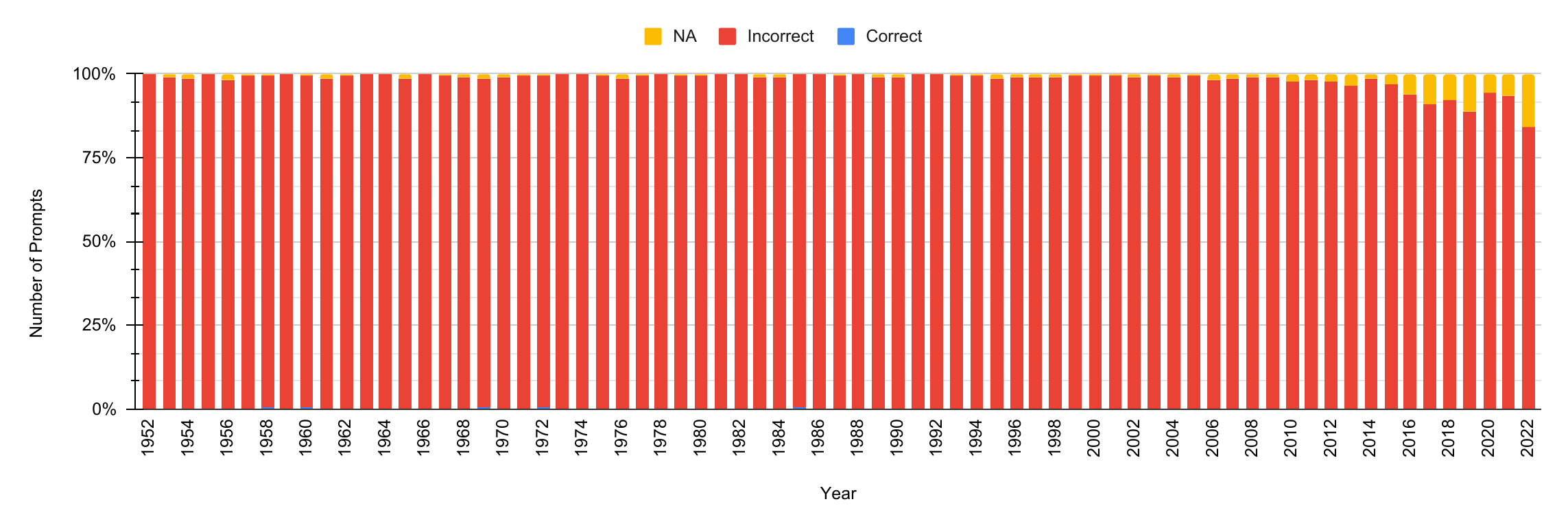}
\end{center}
\caption{Plot for the Min/Max-based metric ($MM$) as year-wise count (In percentage) for \textbf{random fine-tuning} for \texttt{phi-2}.}
\label{fig:minmax-based-rn-phi2}
\end{figure*}

\begin{figure*}
\begin{center}
\includegraphics[width=0.9\linewidth]{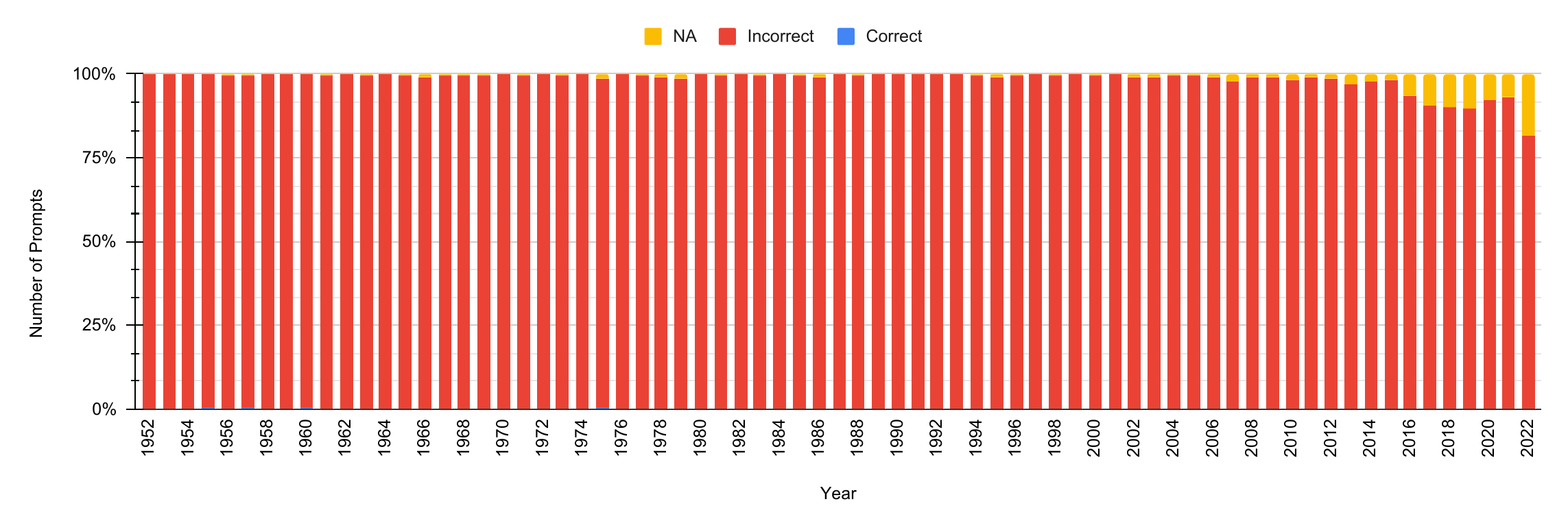}
\end{center}
\caption{Plot for the Range-based metric ($RB$) as year-wise count (In percentage) for \textbf{random fine-tuning} for \texttt{phi-2}.}
\label{fig:rab-based-rn-phi2}
\end{figure*}

\begin{figure*}
\begin{center}
\includegraphics[width=0.9\linewidth]{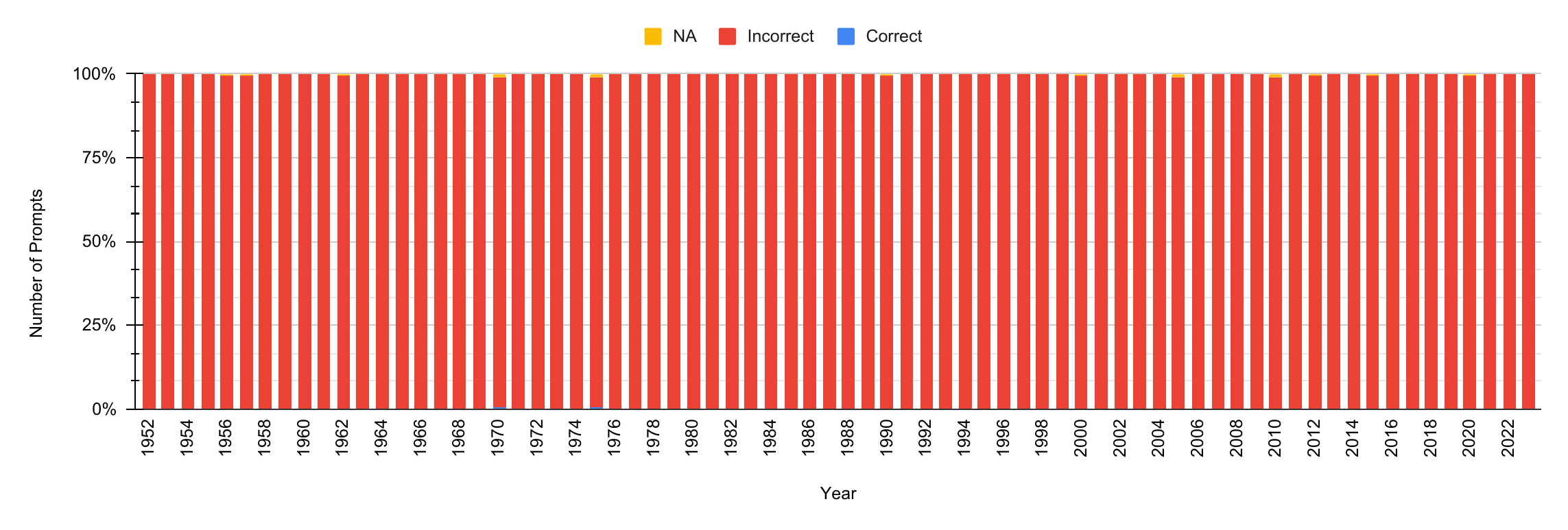}
\end{center}
\caption{Plot for the Trend-based metric ($TB$) as year-wise count (In percentage) for \textbf{random fine-tuning} for \texttt{phi-2}.}
\label{fig:tb-based-rn-phi2}
\end{figure*}

\begin{figure*}
\begin{center}
\includegraphics[width=0.9\linewidth]{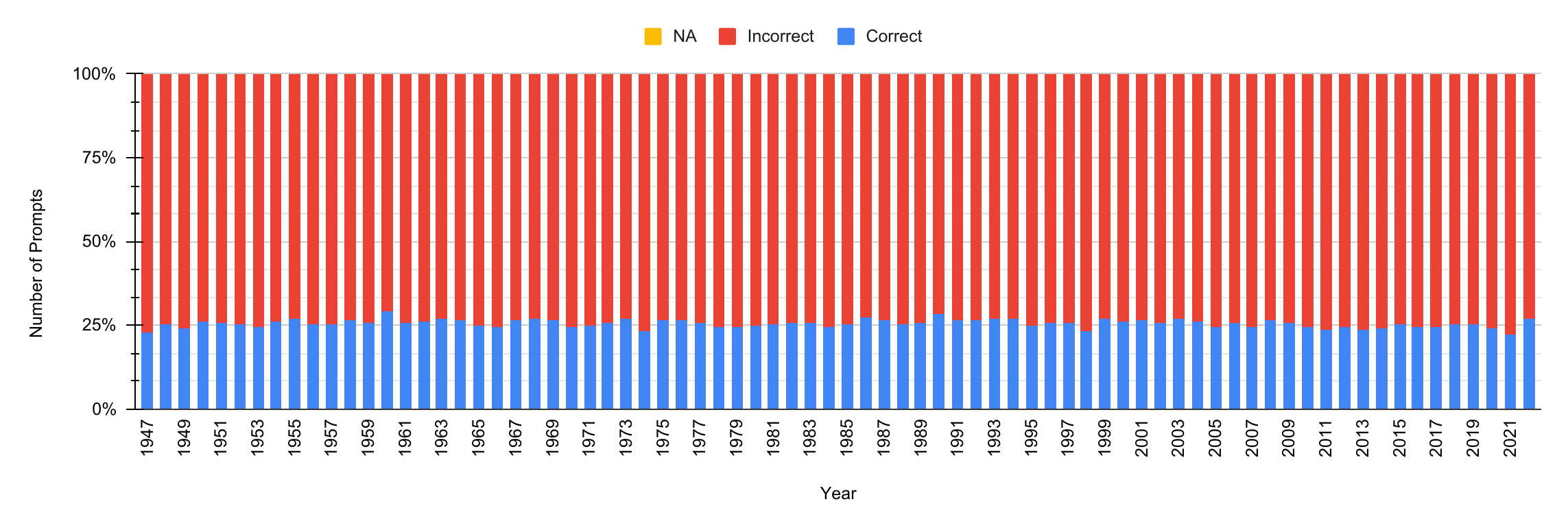}
\end{center}
\caption{Plot for the Date-based metric ($DB$) as year-wise count (In percentage) for \textbf{random fine-tuning} for \texttt{flan-t5-xl}.}
\label{fig:date-based-rn-flan-t5-xl}
\end{figure*}

\begin{figure*}
\begin{center}
\includegraphics[width=0.9\linewidth]{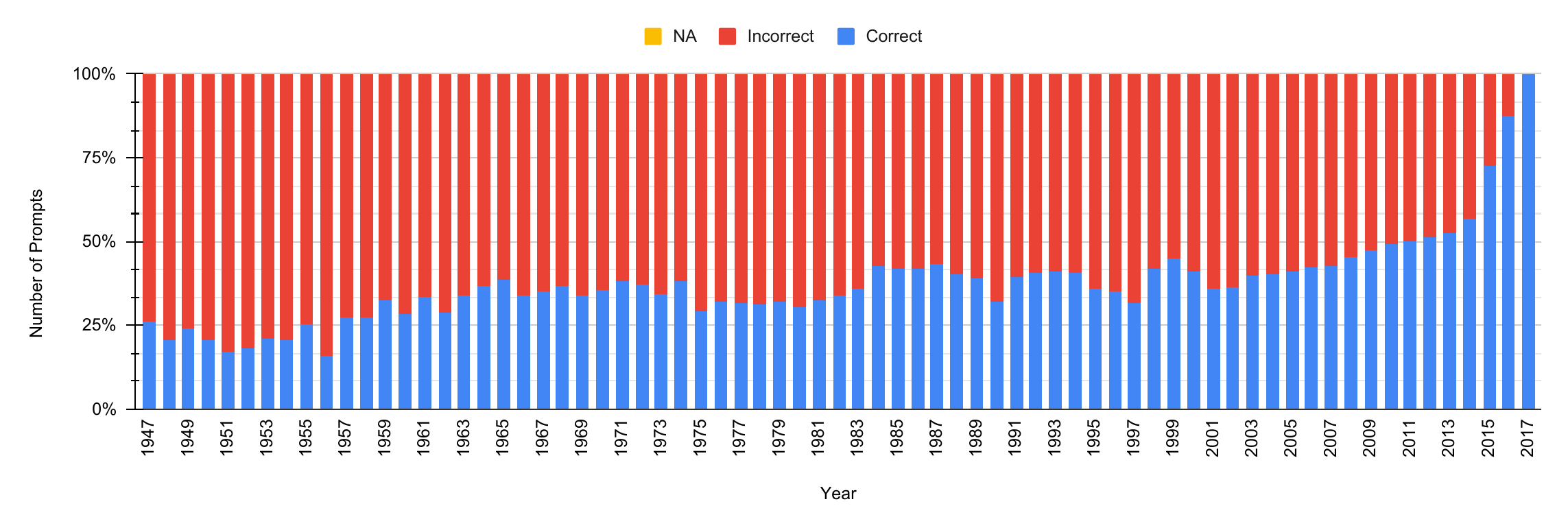}
\end{center}
\caption{Plot for the Comparative-based metric ($CP$) as year-wise count (In percentage) for \textbf{random fine-tuning} for \texttt{flan-t5-xl}.}
\label{fig:cp-rn-flan-t5-xl}
\end{figure*}

\begin{figure*}
\begin{center}
\includegraphics[width=0.9\linewidth]{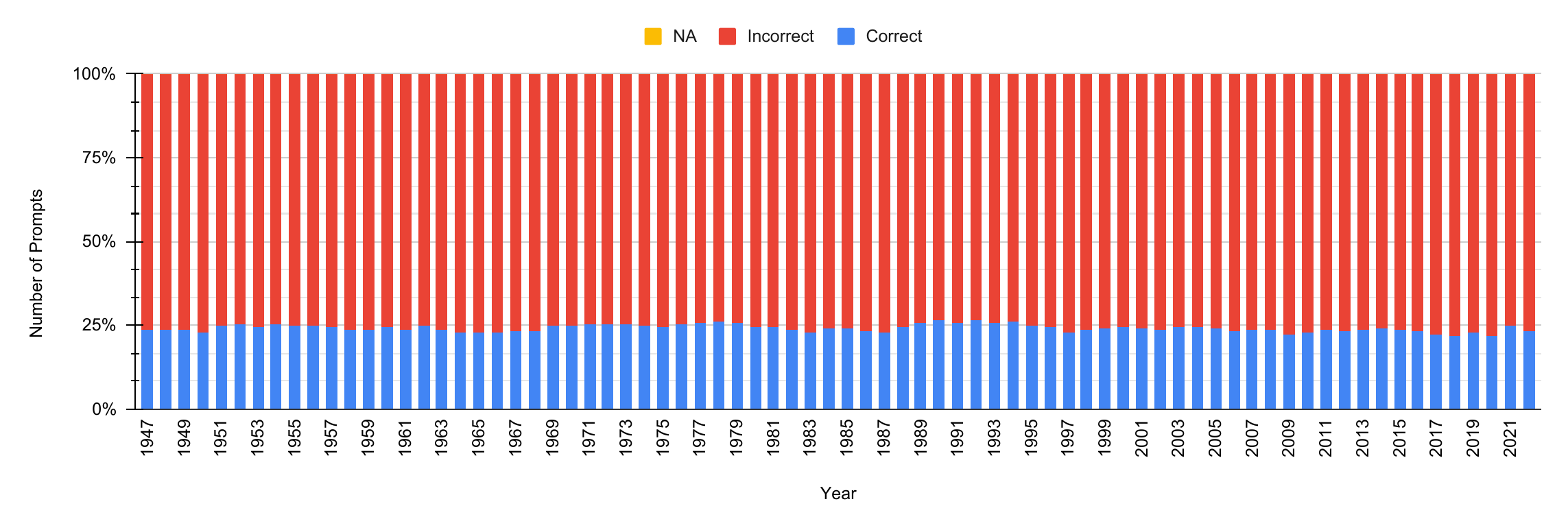}
\end{center}
\caption{Plot for the Window-based metric ($WB$) as year-wise count (In percentage) for \textbf{random fine-tuning} for \texttt{flan-t5-xl}.}
\label{fig:window-based-rn-flan-t5-xl}
\end{figure*}
\begin{figure*}
\begin{center}
\includegraphics[width=0.9\linewidth]{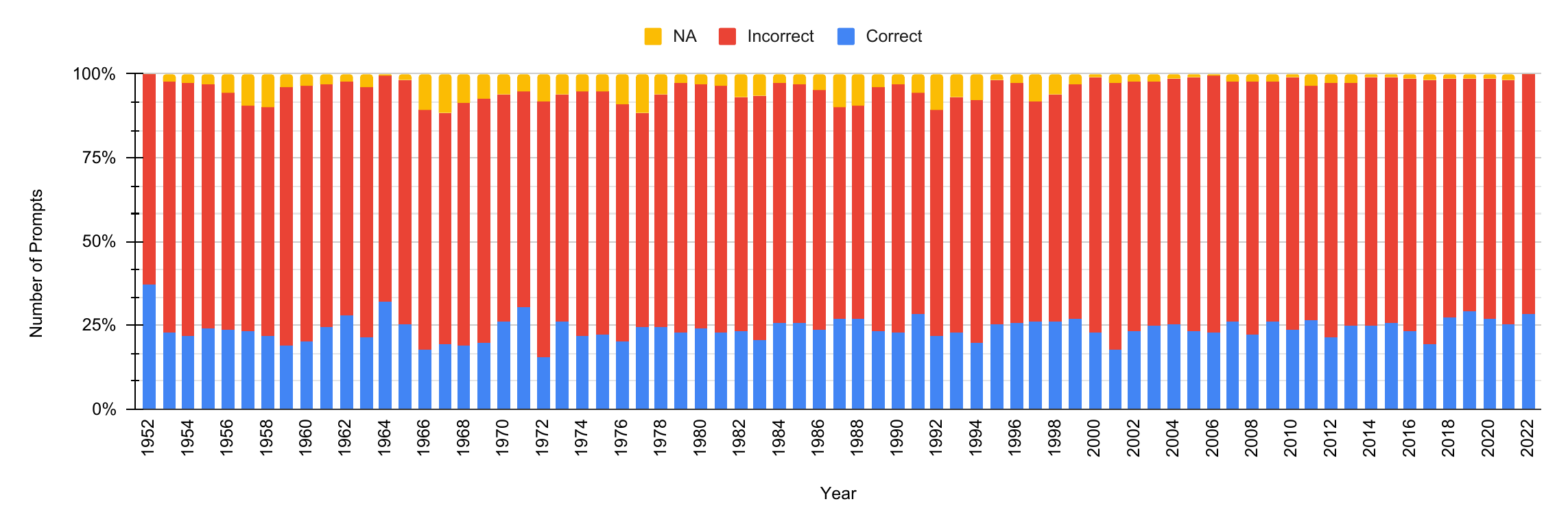}
\end{center}
\caption{Plot for the Min/Max-based metric ($MM$) as year-wise count (In percentage) for \textbf{random fine-tuning} for \texttt{flan-t5-xl}.}
\label{fig:minmax-based-rn-flan-t5-xl}
\end{figure*}

\begin{figure*}
\begin{center}
\includegraphics[width=0.9\linewidth]{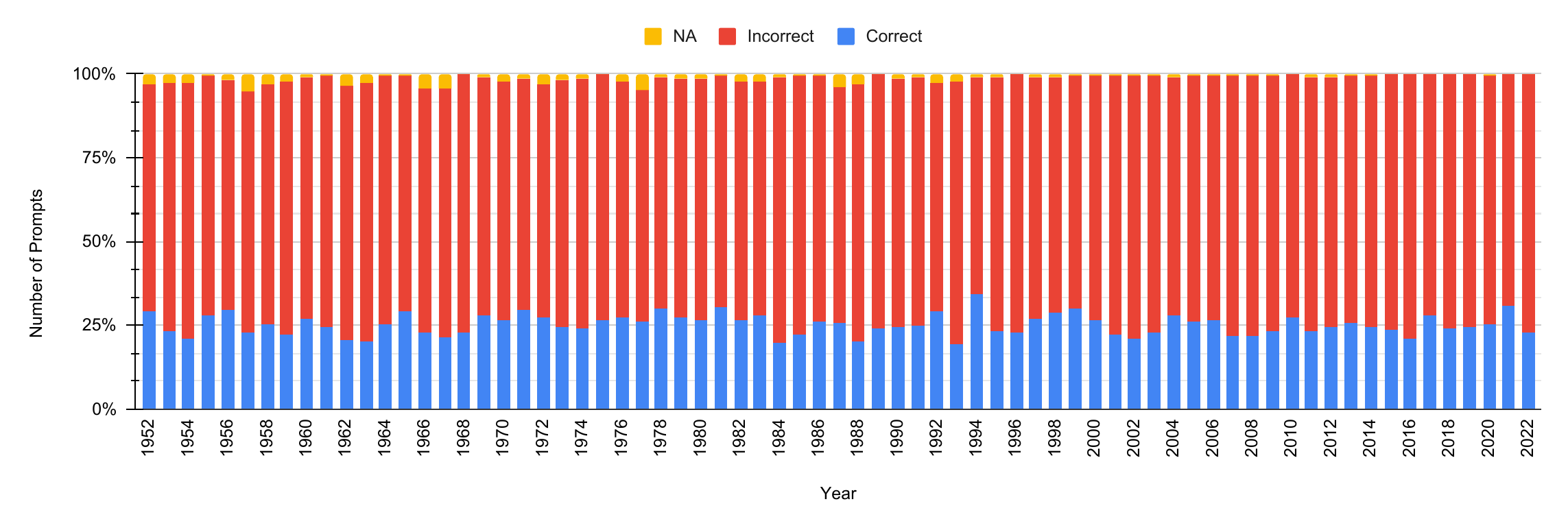}
\end{center}
\caption{Plot for the Range-based metric ($RB$) as year-wise count (In percentage) for \textbf{random fine-tuning} for \texttt{flan-t5-xl}.}
\label{fig:rab-based-rn-flan-t5-xl}
\end{figure*}

\begin{figure*}
\begin{center}
\includegraphics[width=0.9\linewidth]{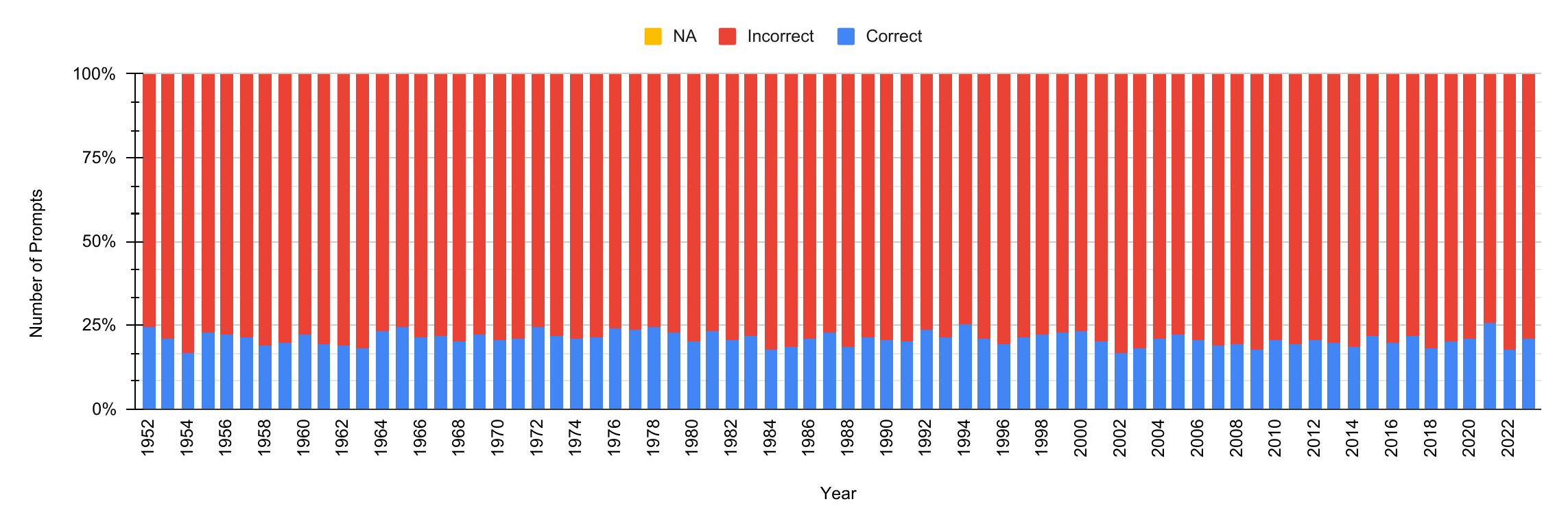}
\end{center}
\caption{Plot for the Trend-based metric ($TB$) as year-wise count (In percentage) for \textbf{random fine-tuning} for \texttt{flan-t5-xl}.}
\label{fig:tb-based-rn-flan-t5-xl}
\end{figure*}

\begin{figure*}
\begin{center}
\includegraphics[width=0.9\linewidth]{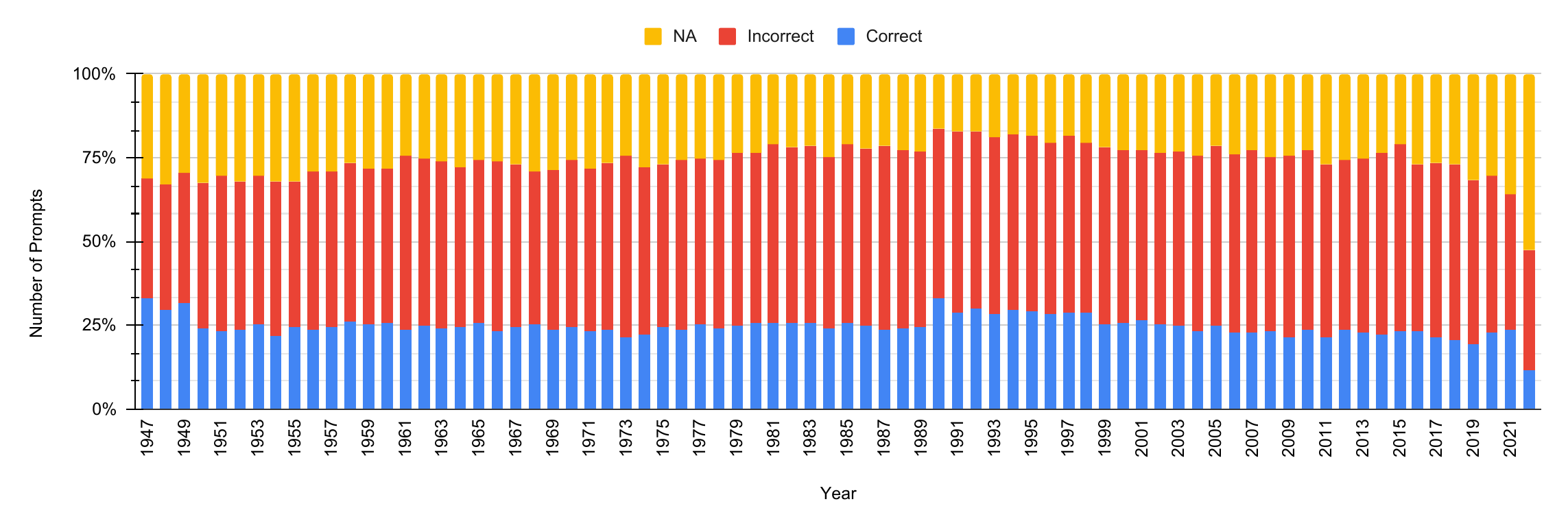}
\end{center}
\caption{Plot for the Date-based metric ($DB$) as year-wise count (In percentage) for \textbf{random fine-tuning} for \texttt{mistral-instruct}.}
\label{fig:date-based-rn-mistral}
\end{figure*}

\begin{figure*}
\begin{center}
\includegraphics[width=0.9\linewidth]{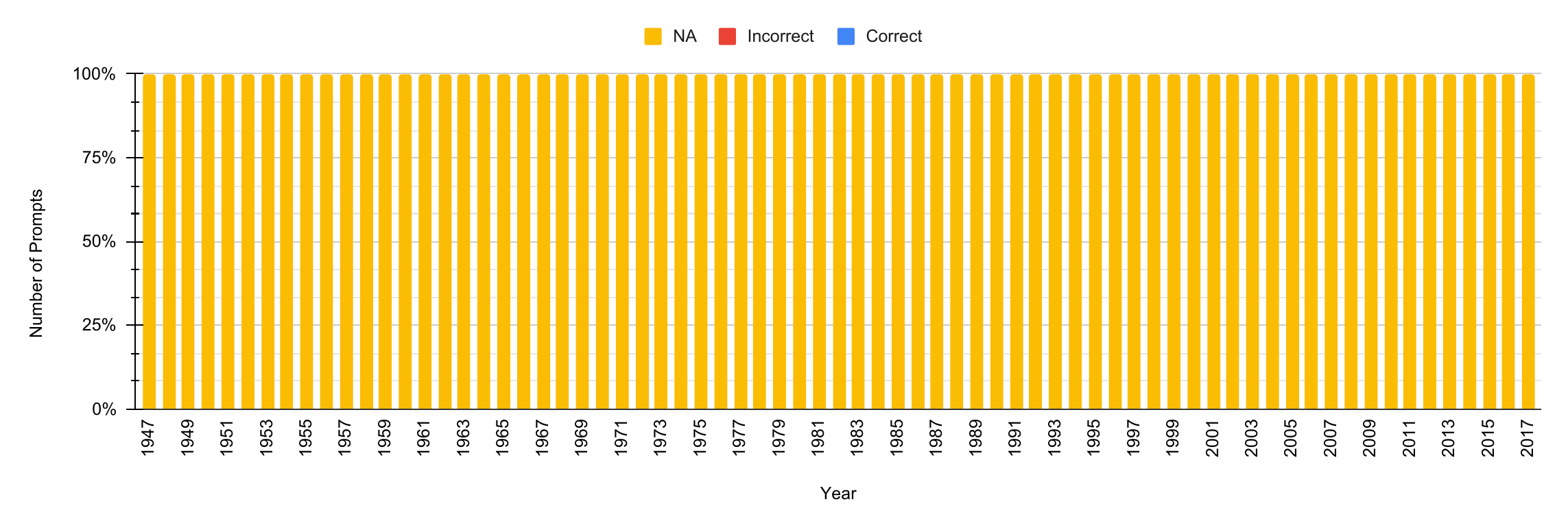}
\end{center}
\caption{Plot for the Comparative-based metric ($CP$) as year-wise count (In percentage) for \textbf{random fine-tuning} for \texttt{mistral-instruct}.}
\label{fig:cp-rn-mistral}
\end{figure*}

\begin{figure*}
\begin{center}
\includegraphics[width=0.9\linewidth]{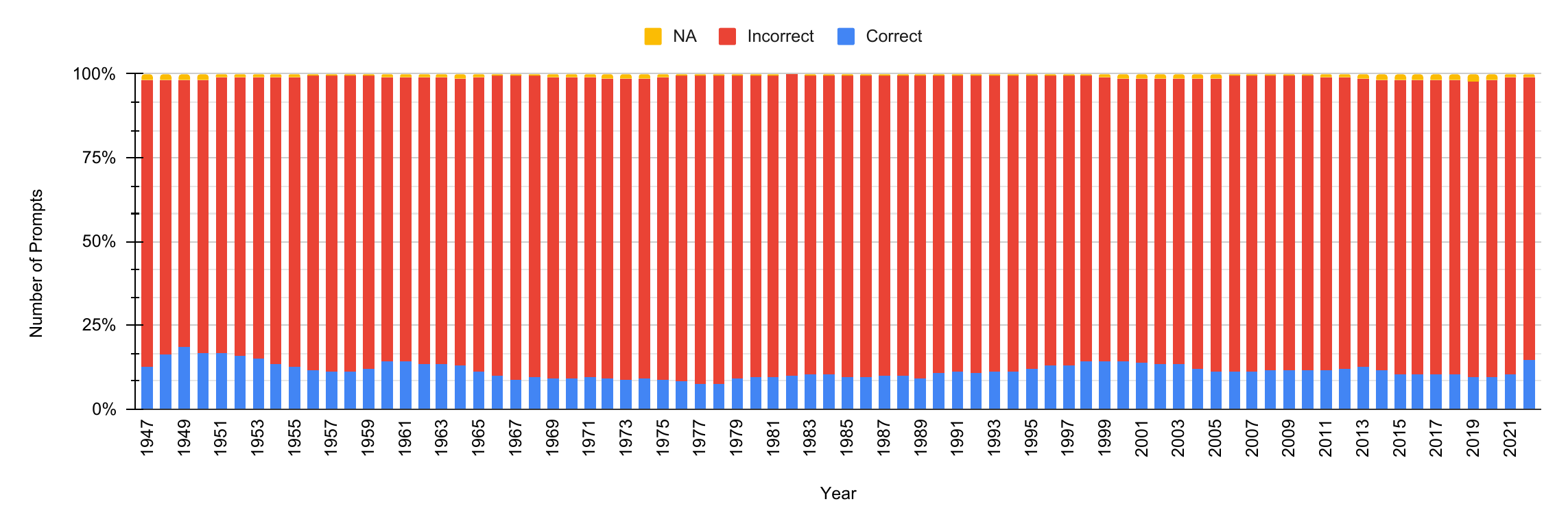}
\end{center}
\caption{Plot for the Window-based metric ($WB$) as year-wise count (In percentage) for \textbf{random fine-tuning} for \texttt{mistral-instruct}.}
\label{fig:window-based-rn-mistral}
\end{figure*}
\begin{figure*}
\begin{center}
\includegraphics[width=0.9\linewidth]{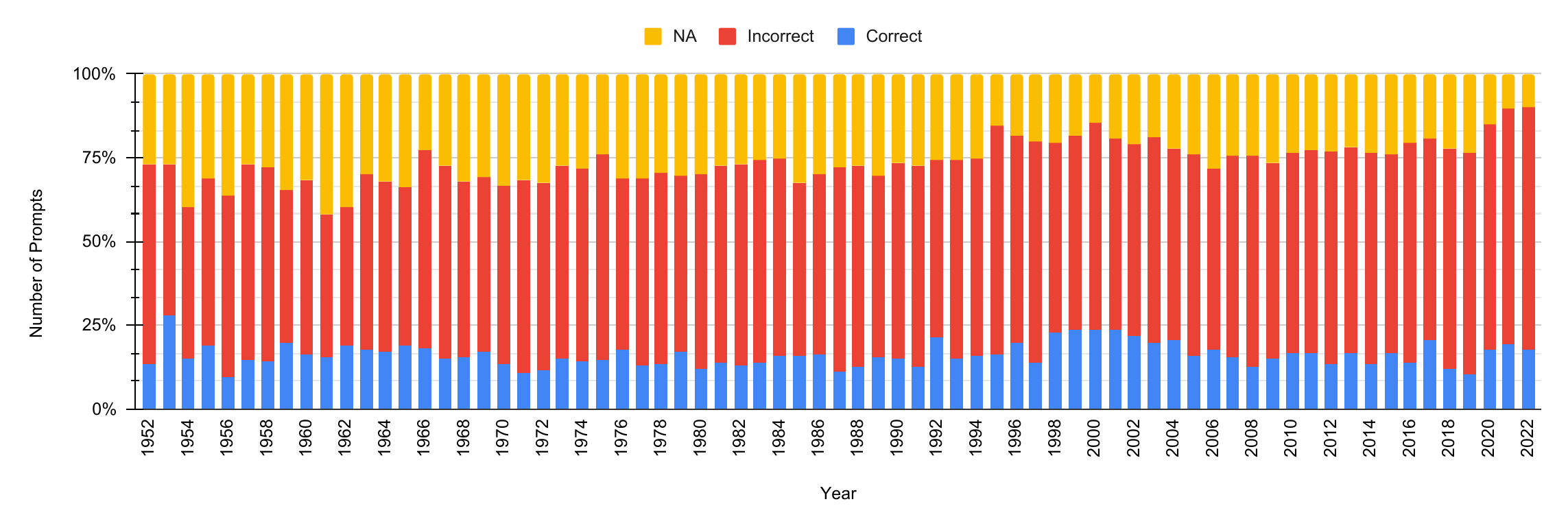}
\end{center}
\caption{Plot for the Min/Max-based metric ($MM$) as year-wise count (In percentage) for \textbf{random fine-tuning} for \texttt{mistral-instruct}.}
\label{fig:minmax-based-rn-mistral}
\end{figure*}

\begin{figure*}
\begin{center}
\includegraphics[width=0.9\linewidth]{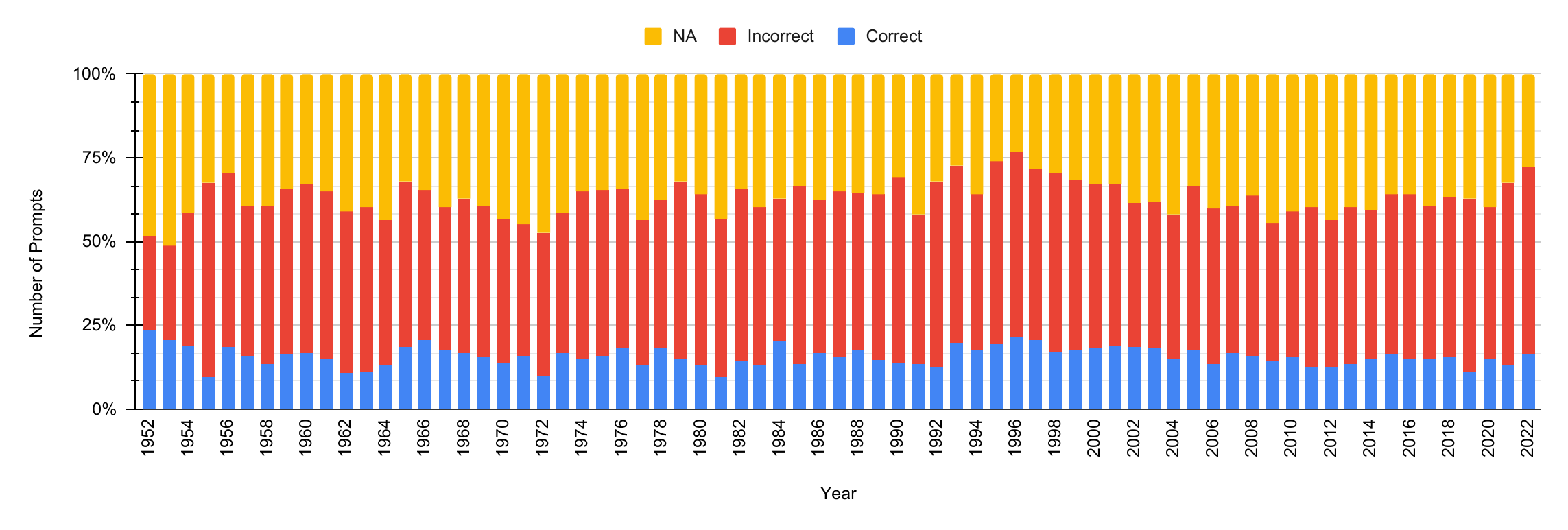}
\end{center}
\caption{Plot for the Range-based metric ($RB$) as year-wise count (In percentage) for \textbf{random fine-tuning} for \texttt{mistral-instruct}.}
\label{fig:rab-based-rn-mistral}
\end{figure*}

\begin{figure*}
\begin{center}
\includegraphics[width=0.9\linewidth]{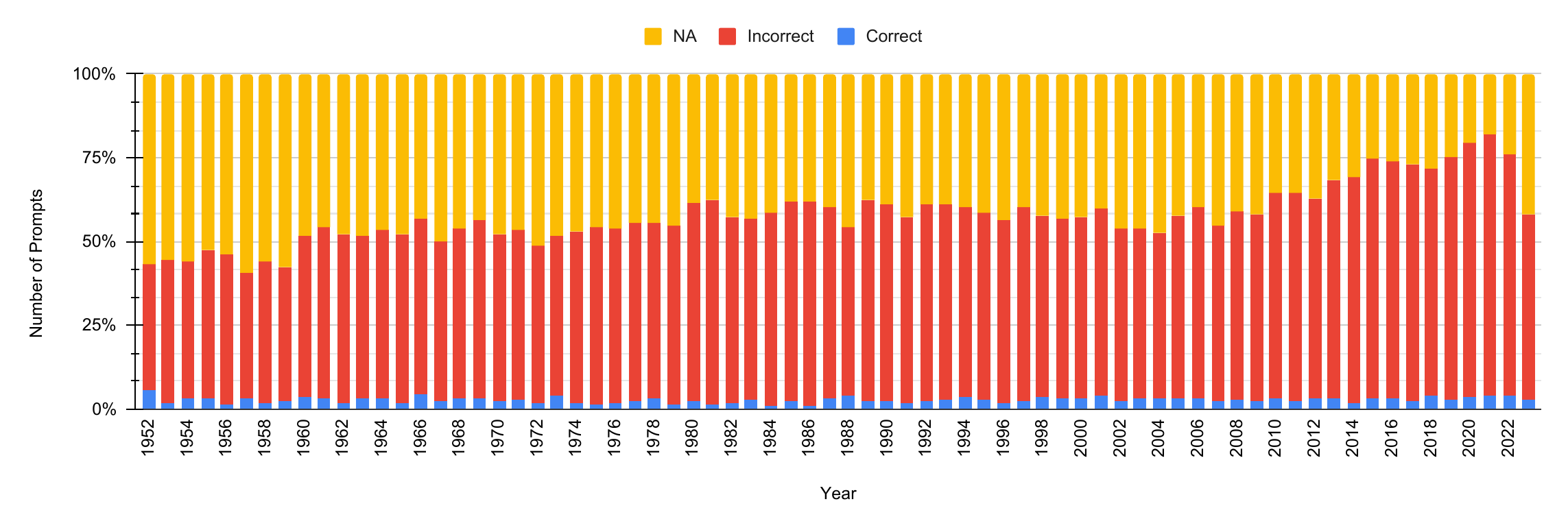}
\end{center}
\caption{Plot for the Trend-based metric ($TB$) as year-wise count (In percentage) for \textbf{random fine-tuning} for \texttt{mistral-instruct}.}
\label{fig:tb-based-rn-mistral}
\end{figure*}

\begin{figure*}
\begin{center}
\includegraphics[width=0.9\linewidth]{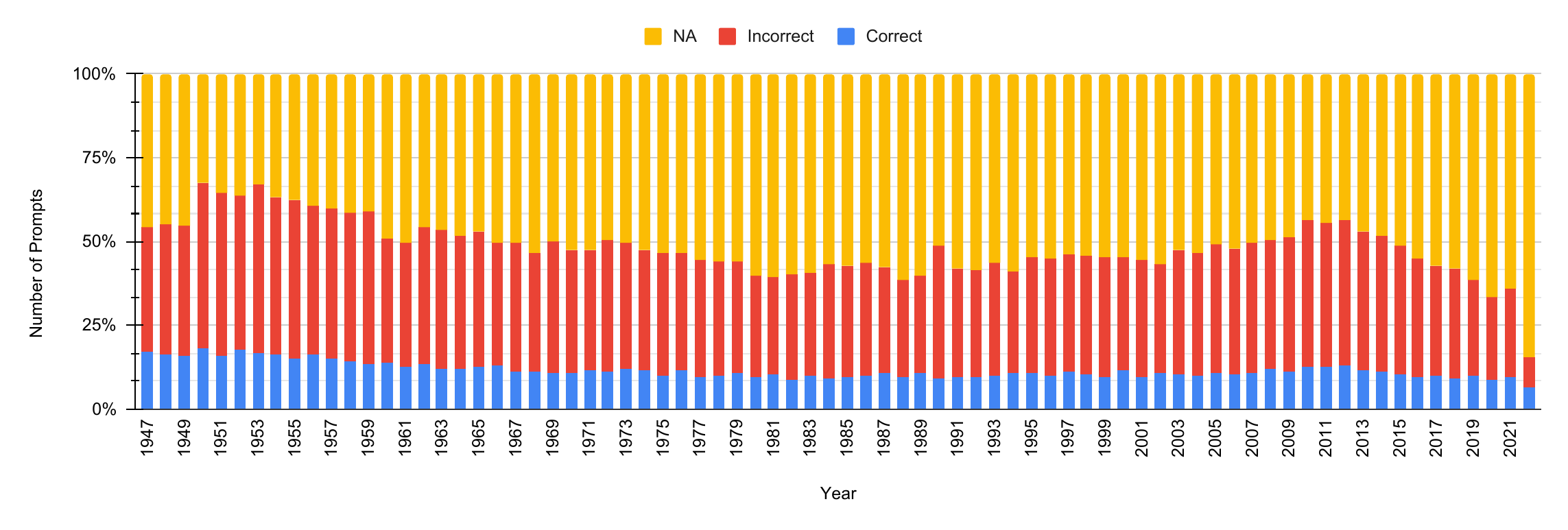}
\end{center}
\caption{Plot for the Date-based metric ($DB$) as year-wise count (In percentage) for \textbf{random fine-tuning} for \texttt{llama-2}.}
\label{fig:date-based-rn-llama}
\end{figure*}

\begin{figure*}
\begin{center}
\includegraphics[width=0.9\linewidth]{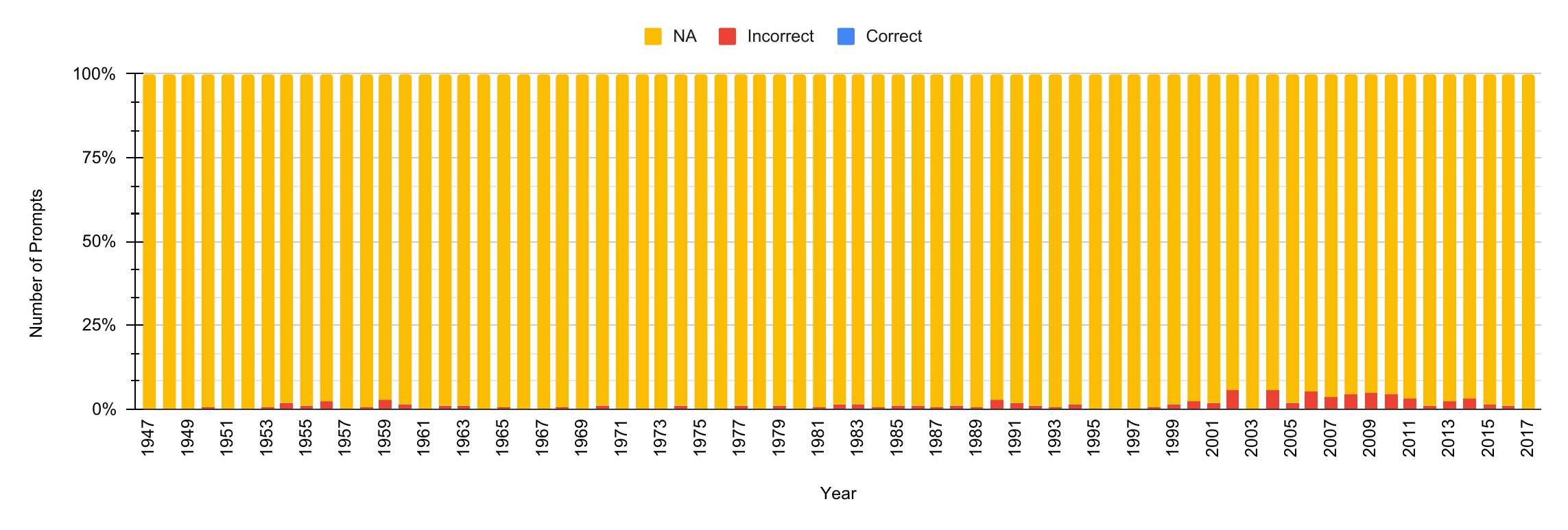}
\end{center}
\caption{Plot for the Comparative-based metric ($CP$) as year-wise count (In percentage) for \textbf{random fine-tuning} for \texttt{llama-2}.}
\label{fig:cp-rn-llama}
\end{figure*}

\begin{figure*}
\begin{center}
\includegraphics[width=0.9\linewidth]{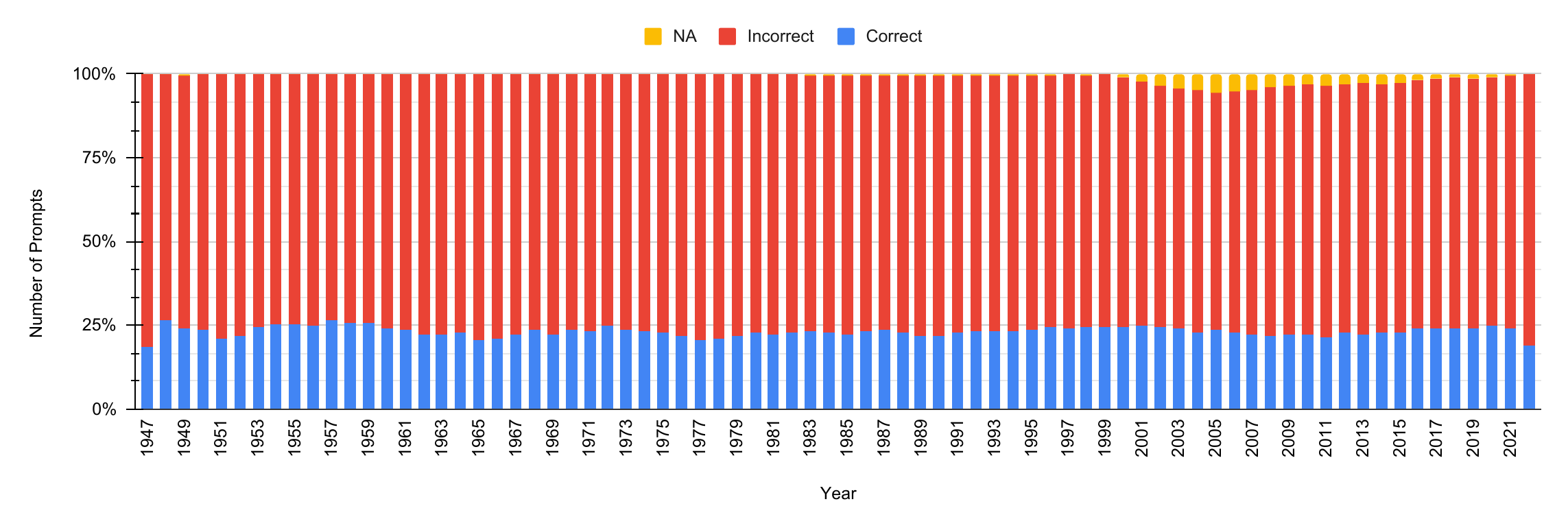}
\end{center}
\caption{Plot for the Window-based metric ($WB$) as year-wise count (In percentage) for \textbf{random fine-tuning} for \texttt{llama-2}.}
\label{fig:window-based-rn-llama}
\end{figure*}
\begin{figure*}
\begin{center}
\includegraphics[width=0.9\linewidth]{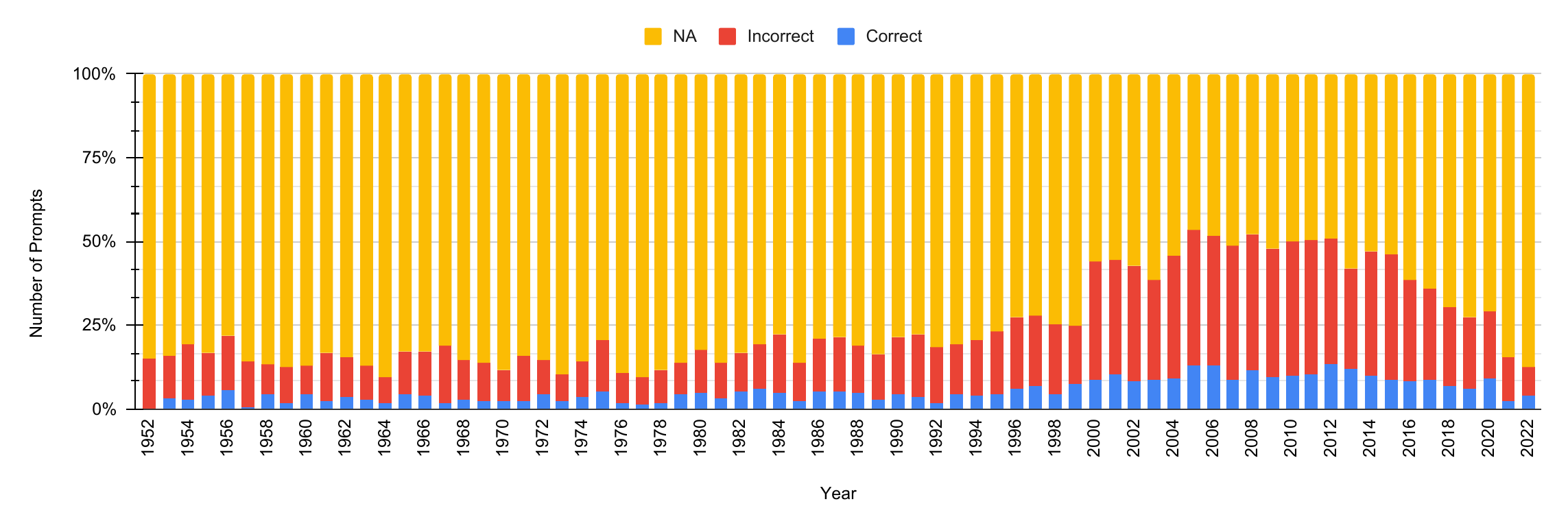}
\end{center}
\caption{Plot for the Min/Max-based metric ($MM$) as year-wise count (In percentage) for \textbf{random fine-tuning} for \texttt{llama-2}.}
\label{fig:minmax-based-rn-llama}
\end{figure*}

\begin{figure*}
\begin{center}
\includegraphics[width=0.9\linewidth]{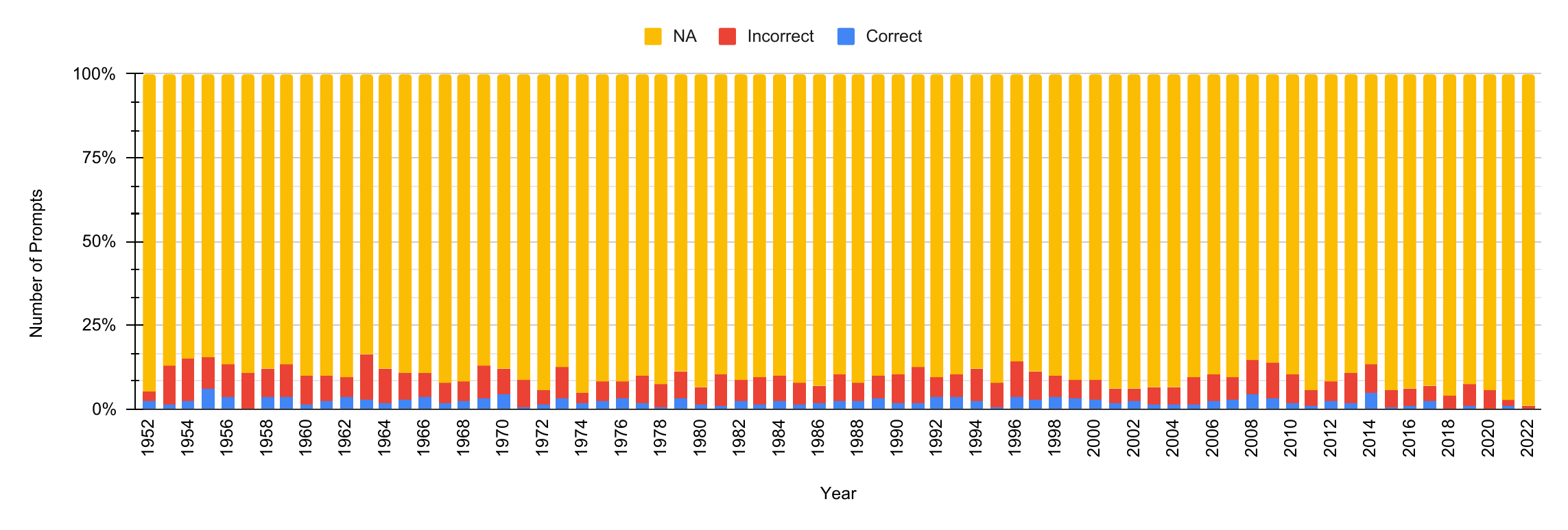}
\end{center}
\caption{Plot for the Range-based metric ($RB$) as year-wise count (In percentage) for \textbf{random fine-tuning} for \texttt{llama-2}.}
\label{fig:rab-based-rn-llama}
\end{figure*}

\begin{figure*}
\begin{center}
\includegraphics[width=0.9\linewidth]{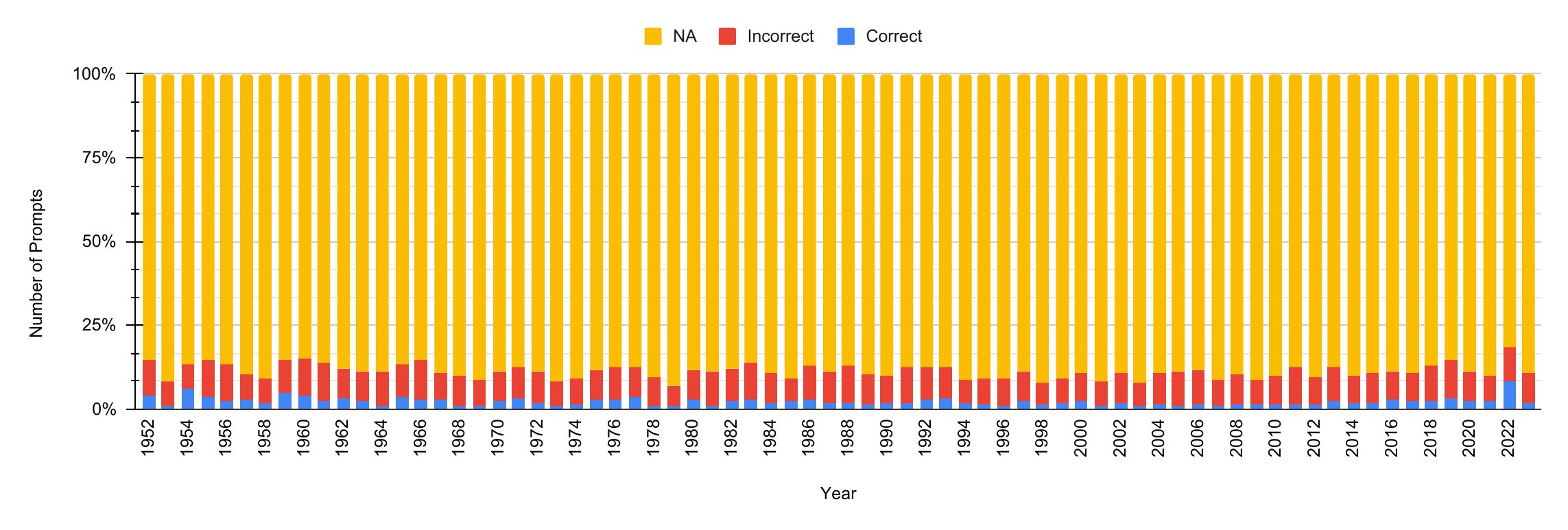}
\end{center}
\caption{Plot for the Trend-based metric ($TB$) as year-wise count (In percentage) for \textbf{random fine-tuning} for \texttt{llama-2}.}
\label{fig:tb-based-rn-llama}
\end{figure*}

\begin{figure*}
\begin{center}
\includegraphics[width=0.9\linewidth]{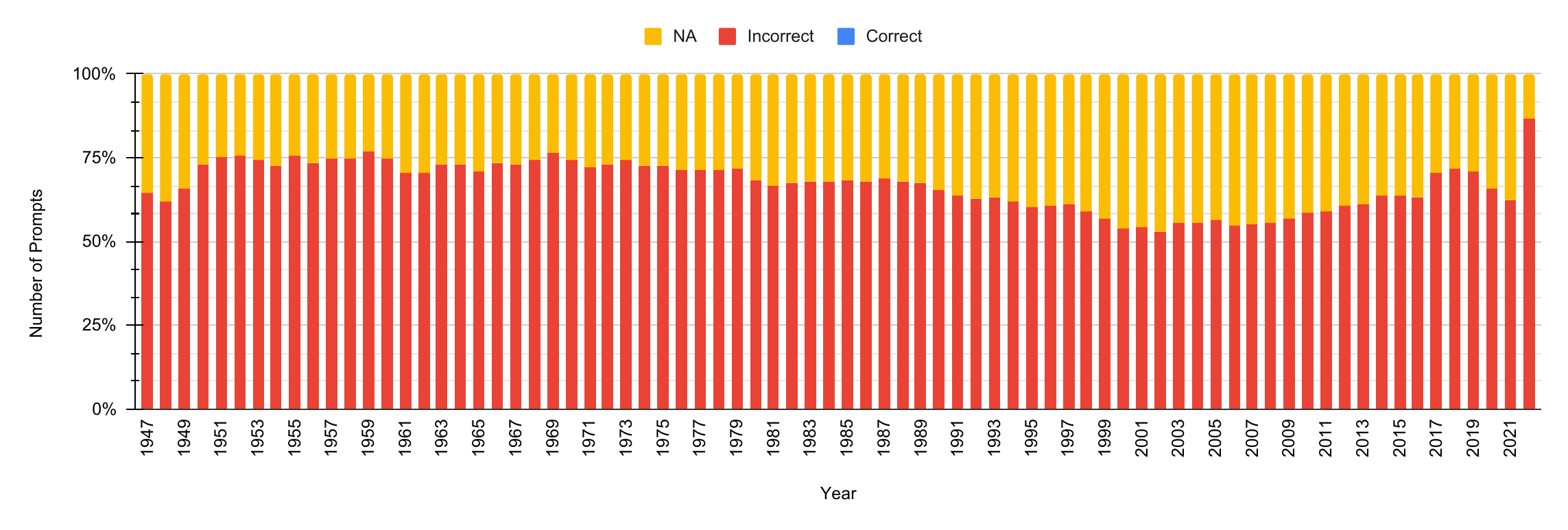}
\end{center}
\caption{Plot for the Date-based metric ($DB$) as year-wise count (In percentage) for \textbf{random fine-tuning} for \texttt{gemma-7b-it}.}
\label{fig:date-based-rn-gemma-7b-it}
\end{figure*}

\begin{figure*}
\begin{center}
\includegraphics[width=0.9\linewidth]{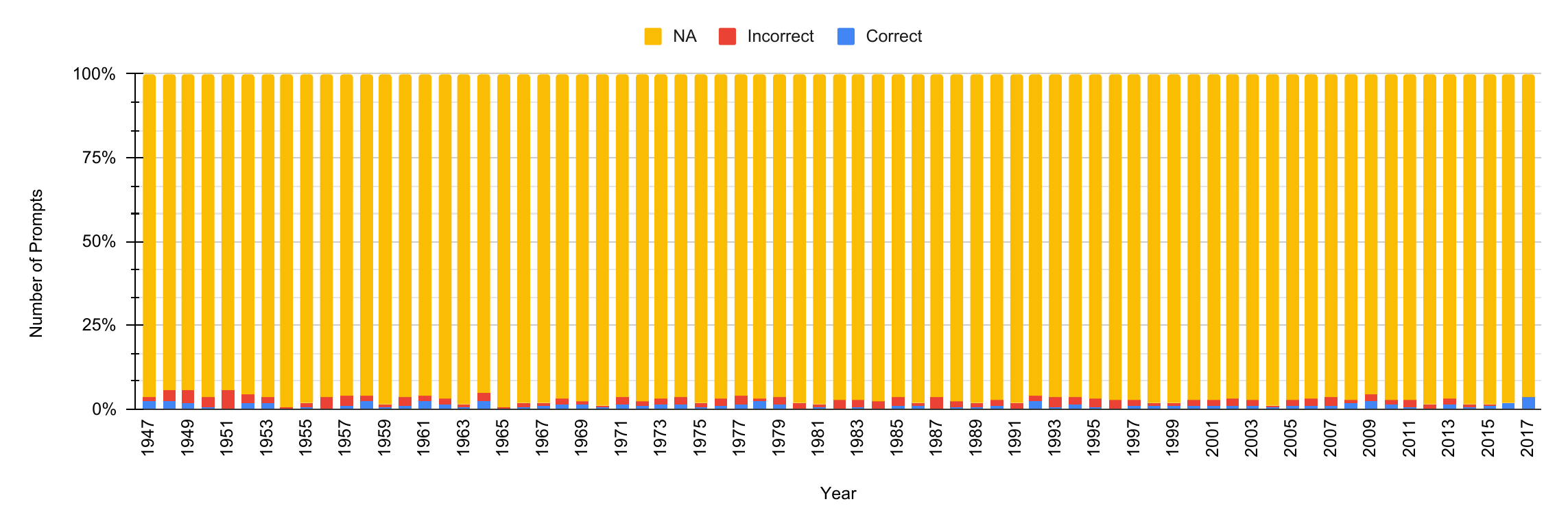}
\end{center}
\caption{Plot for the Comparative-based metric ($CP$) as year-wise count (In percentage) for \textbf{random fine-tuning} for \texttt{gemma-7b-it}.}
\label{fig:cp-rn-gemma-7b-it}
\end{figure*}

\begin{figure*}
\begin{center}
\includegraphics[width=0.9\linewidth]{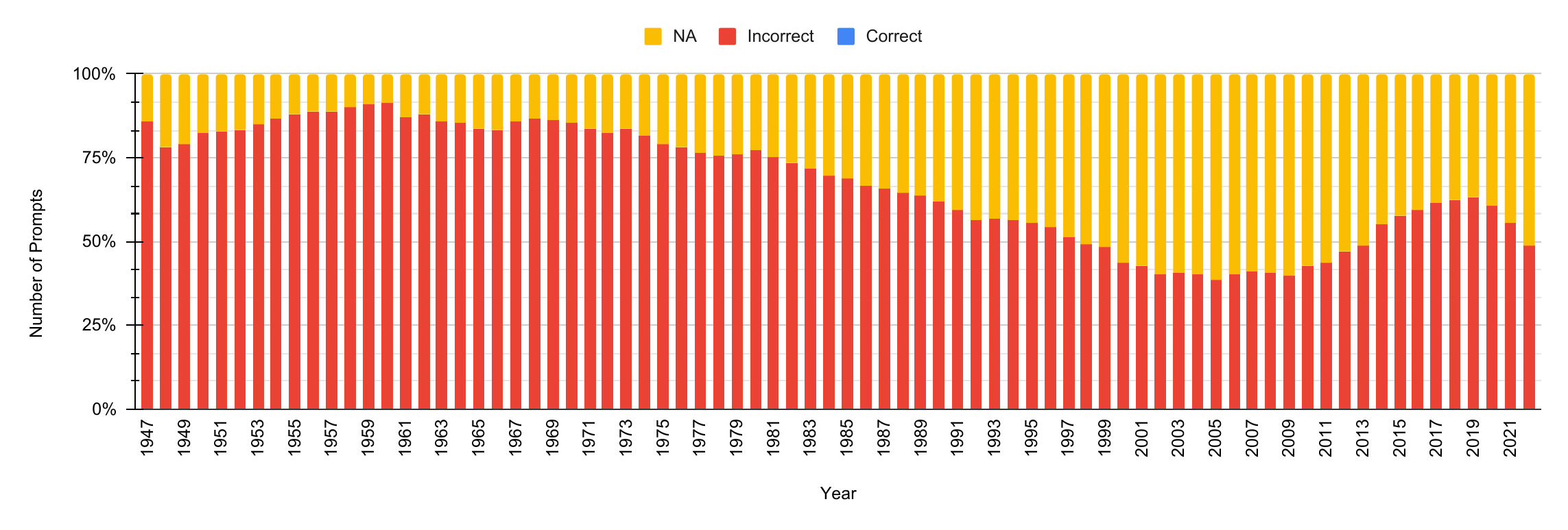}
\end{center}
\caption{Plot for the Window-based metric ($WB$) as year-wise count (In percentage) for \textbf{random fine-tuning} for \texttt{gemma-7b-it}.}
\label{fig:window-based-rn-gemma-7b-it}
\end{figure*}
\begin{figure*}
\begin{center}
\includegraphics[width=0.9\linewidth]{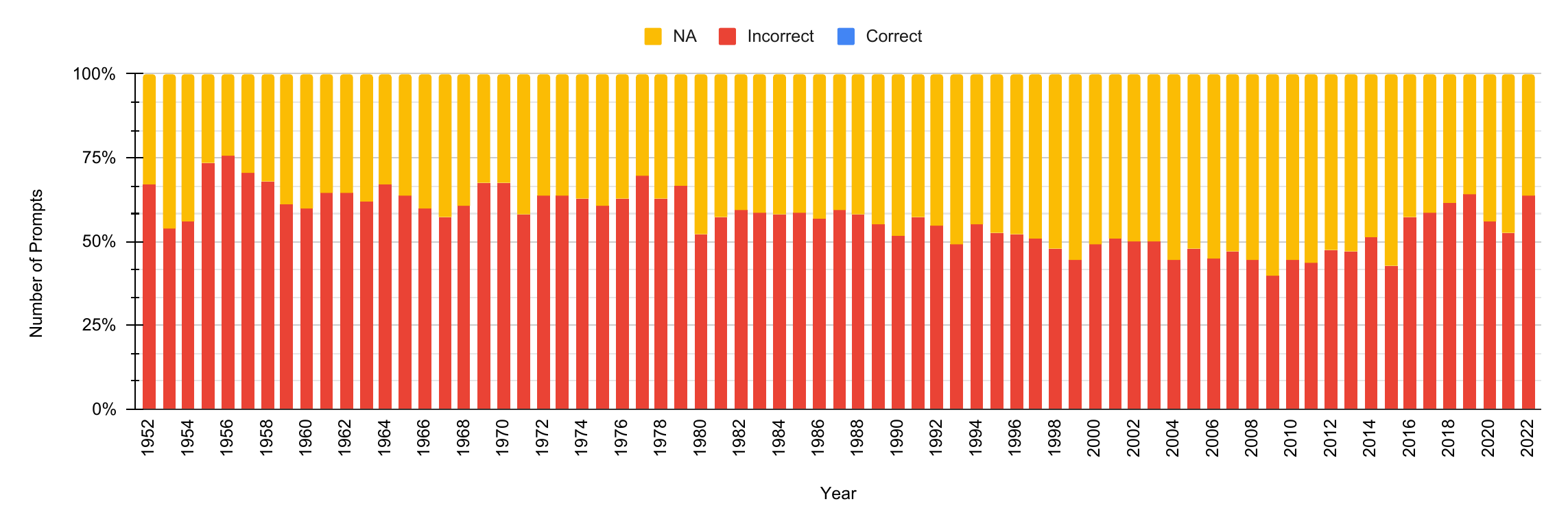}
\end{center}
\caption{Plot for the Min/Max-based metric ($MM$) as year-wise count (In percentage) for \textbf{random fine-tuning} for \texttt{gemma-7b-it}.}
\label{fig:minmax-based-rn-gemma-7b-it}
\end{figure*}

\begin{figure*}
\begin{center}
\includegraphics[width=0.9\linewidth]{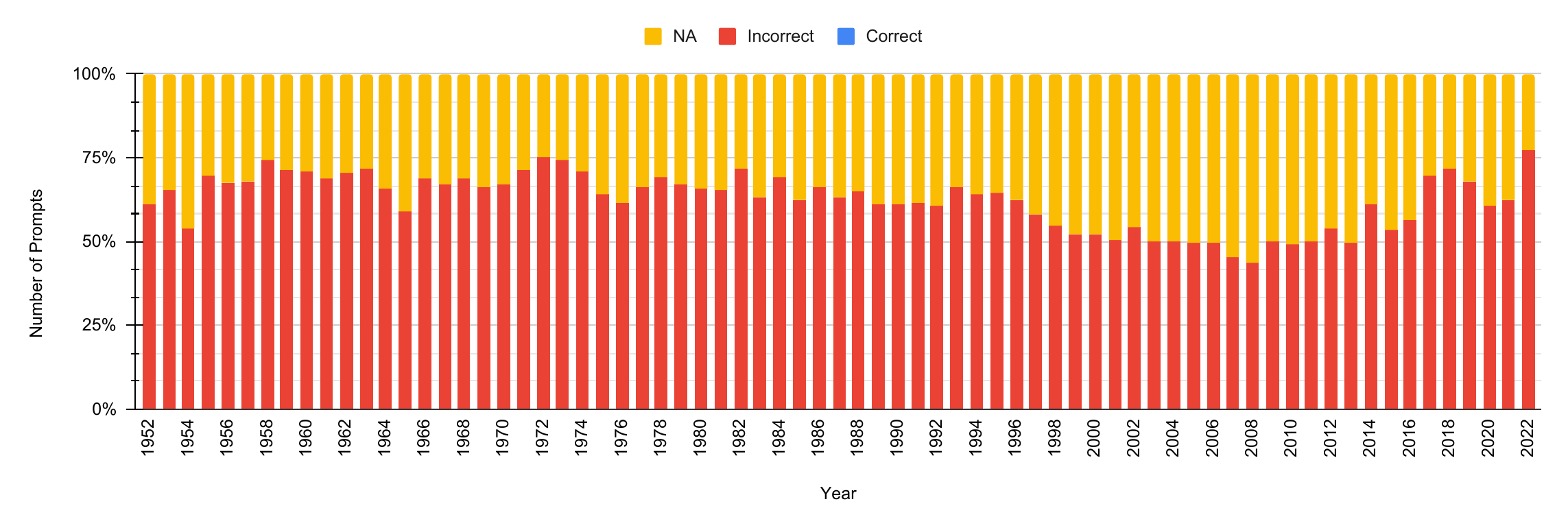}
\end{center}
\caption{Plot for the Range-based metric ($RB$) as year-wise count (In percentage) for \textbf{random fine-tuning} for \texttt{gemma-7b-it}.}
\label{fig:rab-based-rn-gemma-7b-it}
\end{figure*}

\begin{figure*}
\begin{center}
\includegraphics[width=0.9\linewidth]{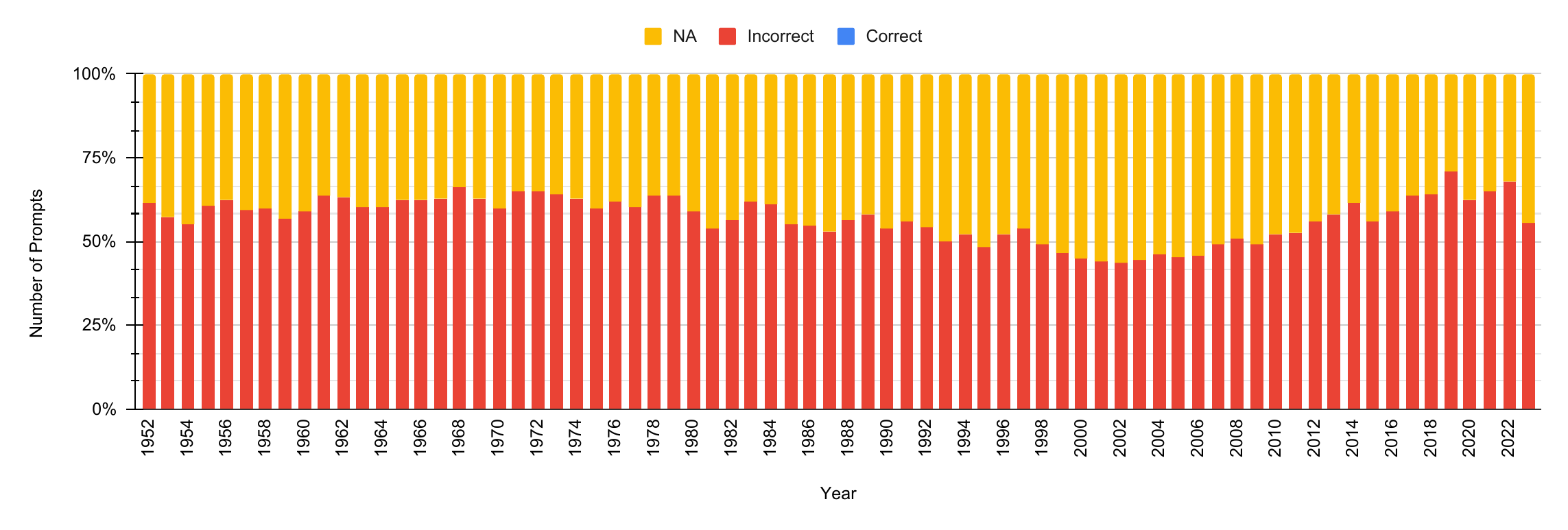}
\end{center}
\caption{Plot for the Trend-based metric ($TB$) as year-wise count (In percentage) for \textbf{random fine-tuning} for \texttt{gemma-7b-it}.}
\label{fig:tb-based-rn-gemma-7b-it}
\end{figure*}

\begin{figure*}
\begin{center}
\includegraphics[width=0.9\linewidth]{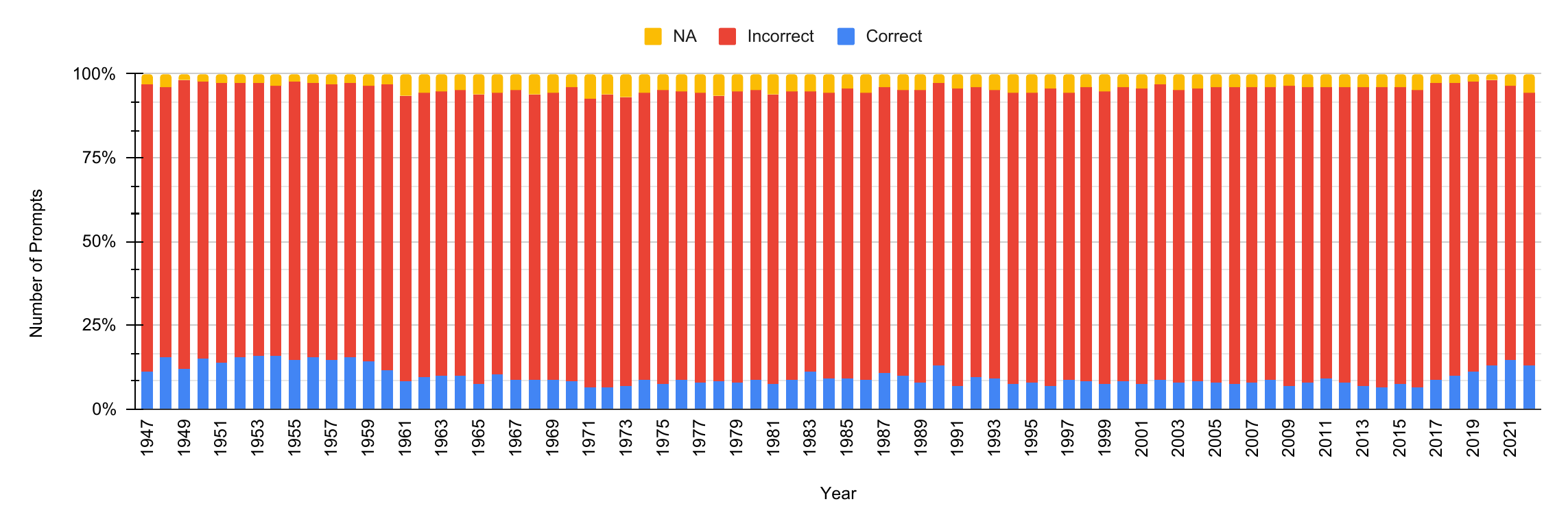}
\end{center}
\caption{Plot for the Date-based metric ($DB$) as year-wise count (In percentage) for \textbf{random fine-tuning} for \texttt{llama-3-8b}.}
\label{fig:date-based-rn-llama-3-8b}
\end{figure*}

\begin{figure*}
\begin{center}
\includegraphics[width=0.9\linewidth]{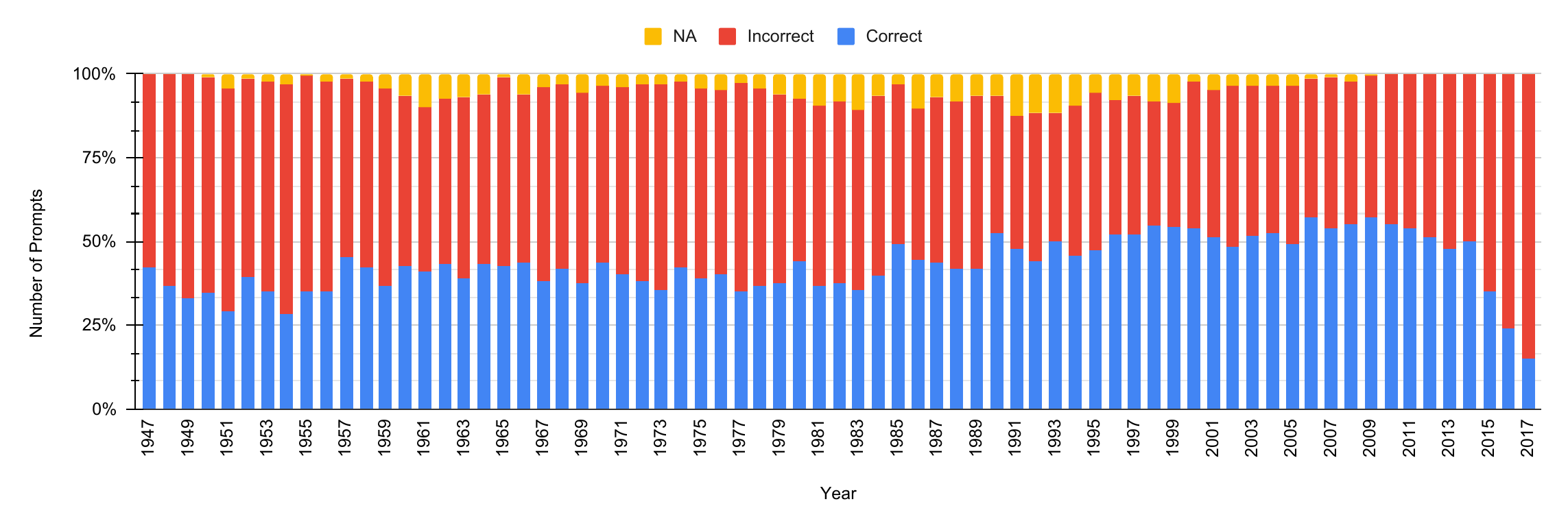}
\end{center}
\caption{Plot for the Comparative-based metric ($CP$) as year-wise count (In percentage) for \textbf{random fine-tuning} for \texttt{llama-3-8b}.}
\label{fig:cp-rn-llama-3-8b}
\end{figure*}

\begin{figure*}
\begin{center}
\includegraphics[width=0.9\linewidth]{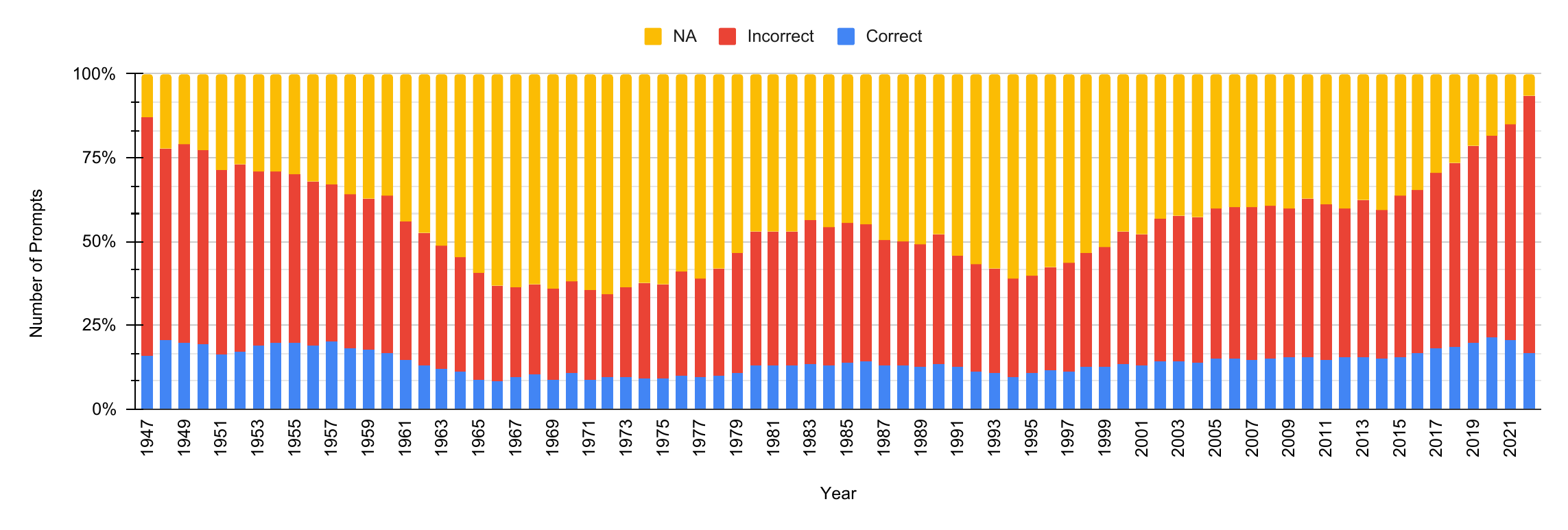}
\end{center}
\caption{Plot for the Window-based metric ($WB$) as year-wise count (In percentage) for \textbf{random fine-tuning} for \texttt{llama-3-8b}.}
\label{fig:window-based-rn-llama-3-8b}
\end{figure*}
\begin{figure*}
\begin{center}
\includegraphics[width=0.9\linewidth]{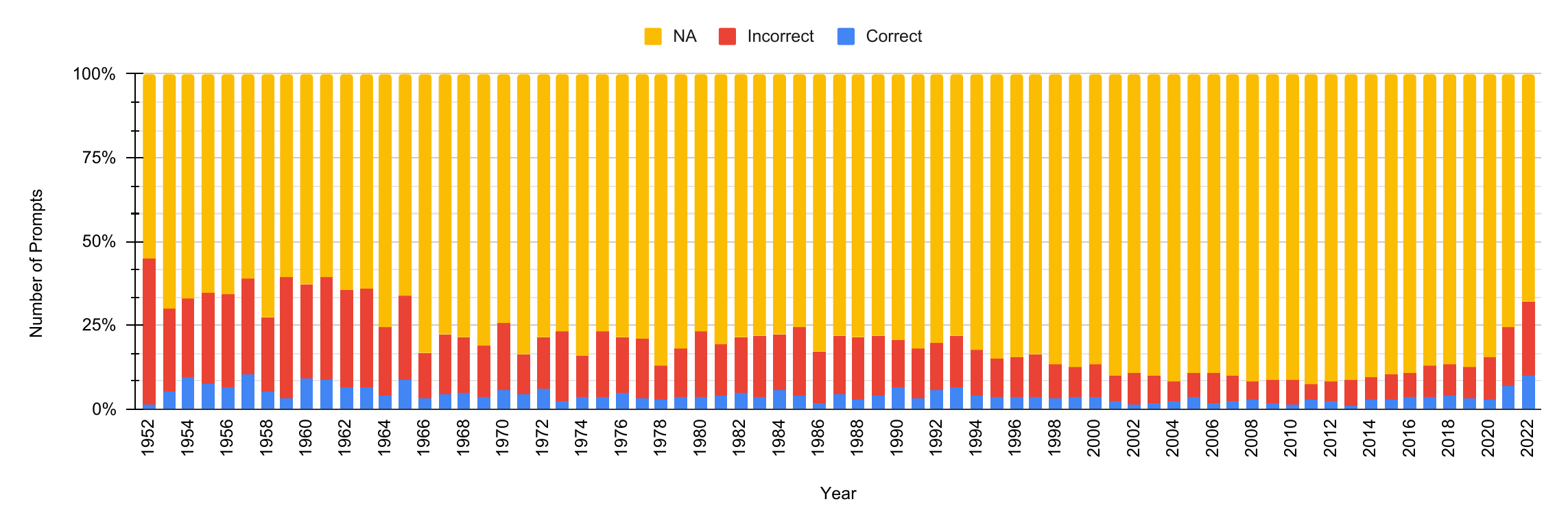}
\end{center}
\caption{Plot for the Min/Max-based metric ($MM$) as year-wise count (In percentage) for \textbf{random fine-tuning} for \texttt{llama-3-8b}.}
\label{fig:minmax-based-rn-llama-3-8b}
\end{figure*}

\begin{figure*}
\begin{center}
\includegraphics[width=0.9\linewidth]{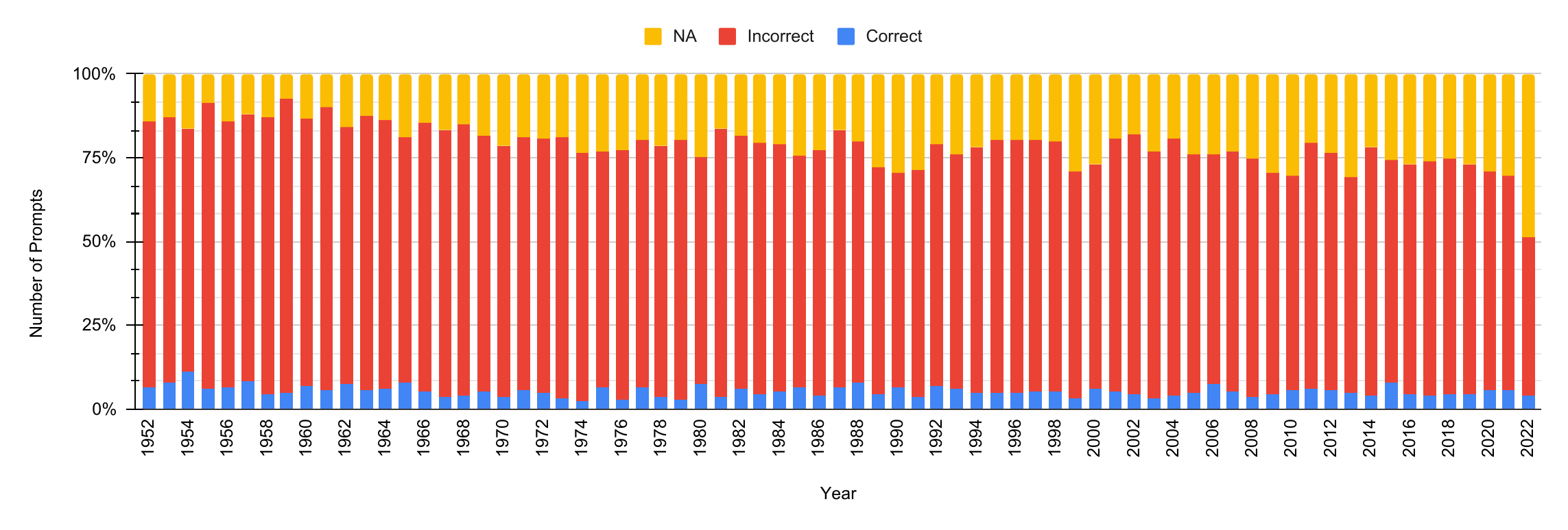}
\end{center}
\caption{Plot for the Range-based metric ($RB$) as year-wise count (In percentage) for \textbf{random fine-tuning} for \texttt{llama-3-8b}.}
\label{fig:rab-based-rn-llama-3-8b}
\end{figure*}

\begin{figure*}
\begin{center}
\includegraphics[width=0.9\linewidth]{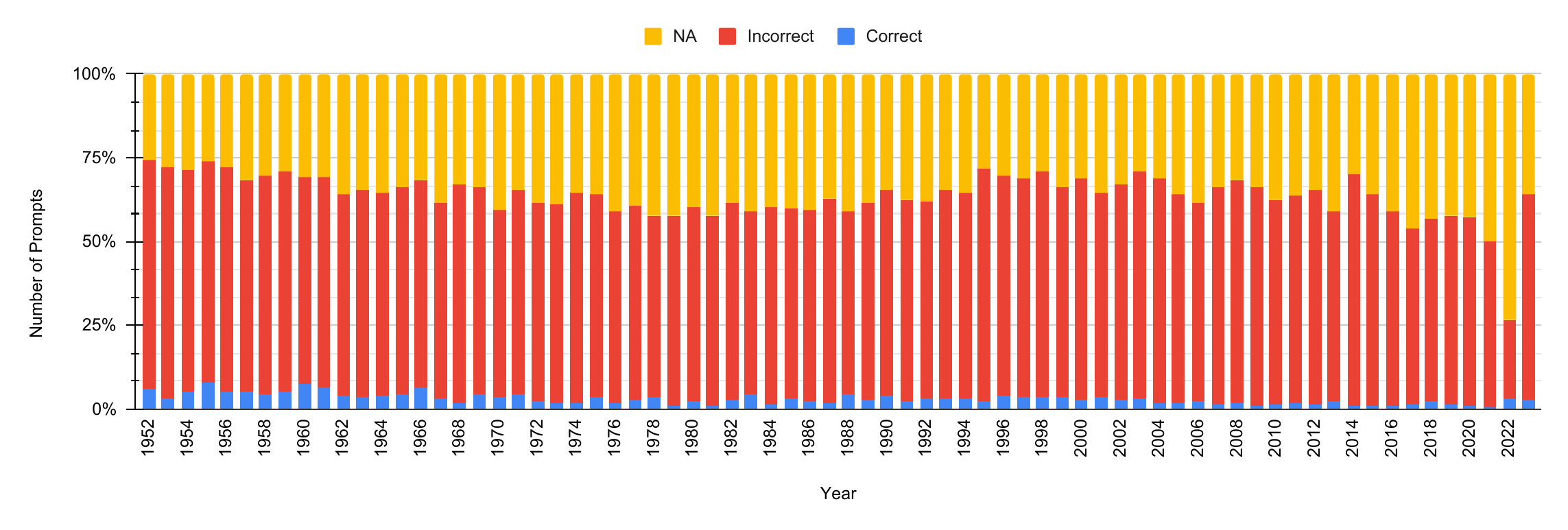}
\end{center}
\caption{Plot for the Trend-based metric ($TB$) as year-wise count (In percentage) for \textbf{random fine-tuning} for \texttt{llama-3-8b}.}
\label{fig:tb-based-rn-llama-3-8b}
\end{figure*}

\begin{figure*}
\begin{center}
\includegraphics[width=0.9\linewidth]{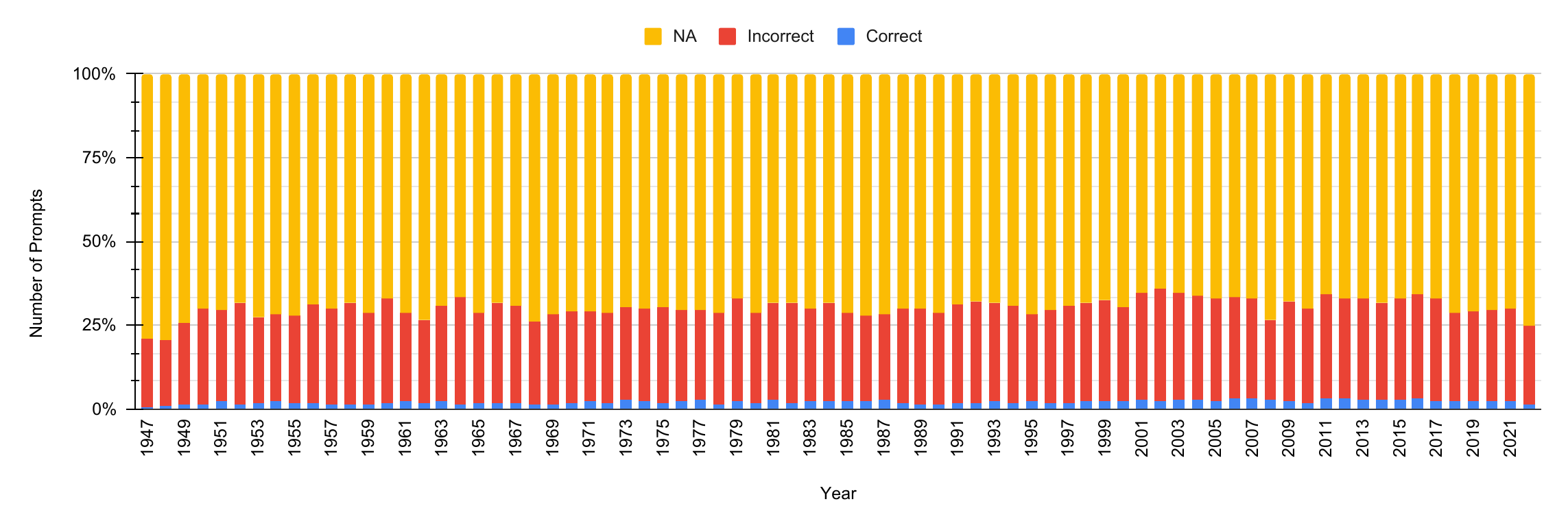}
\end{center}
\caption{Plot for the Date-based metric ($DB$) as year-wise count (In percentage) for \textbf{random fine-tuning} for \texttt{phi-3-medium}.}
\label{fig:date-based-rn-phi-3-medium}
\end{figure*}

\begin{figure*}
\begin{center}
\includegraphics[width=0.9\linewidth]{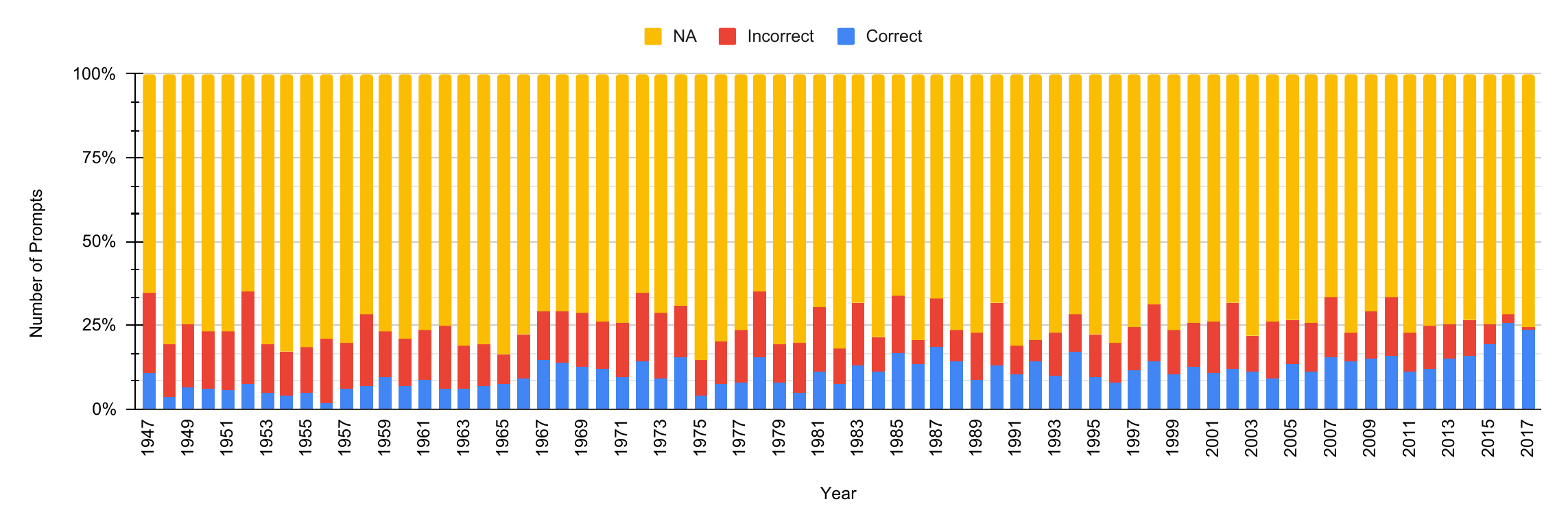}
\end{center}
\caption{Plot for the Comparative-based metric ($CP$) as year-wise count (In percentage) for \textbf{random fine-tuning} for \texttt{phi-3-medium}.}
\label{fig:cp-rn-phi-3-medium}
\end{figure*}

\begin{figure*}
\begin{center}
\includegraphics[width=0.9\linewidth]{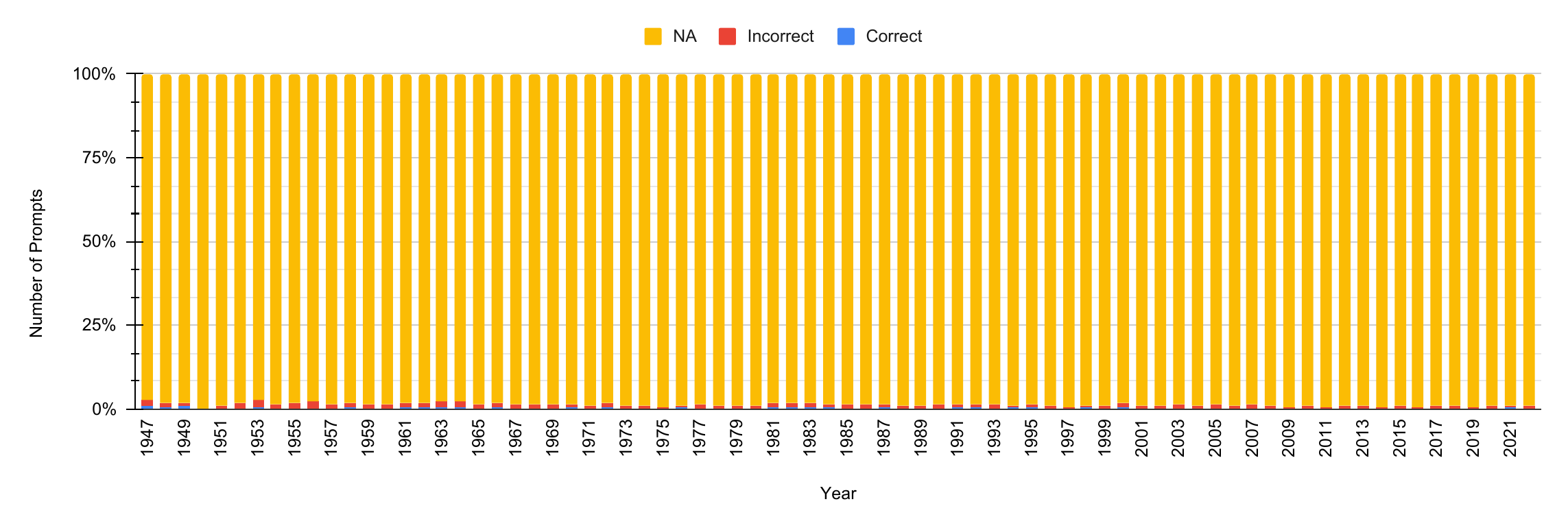}
\end{center}
\caption{Plot for the Window-based metric ($WB$) as year-wise count (In percentage) for \textbf{random fine-tuning} for \texttt{phi-3-medium}.}
\label{fig:window-based-rn-phi-3-medium}
\end{figure*}
\begin{figure*}
\begin{center}
\includegraphics[width=0.9\linewidth]{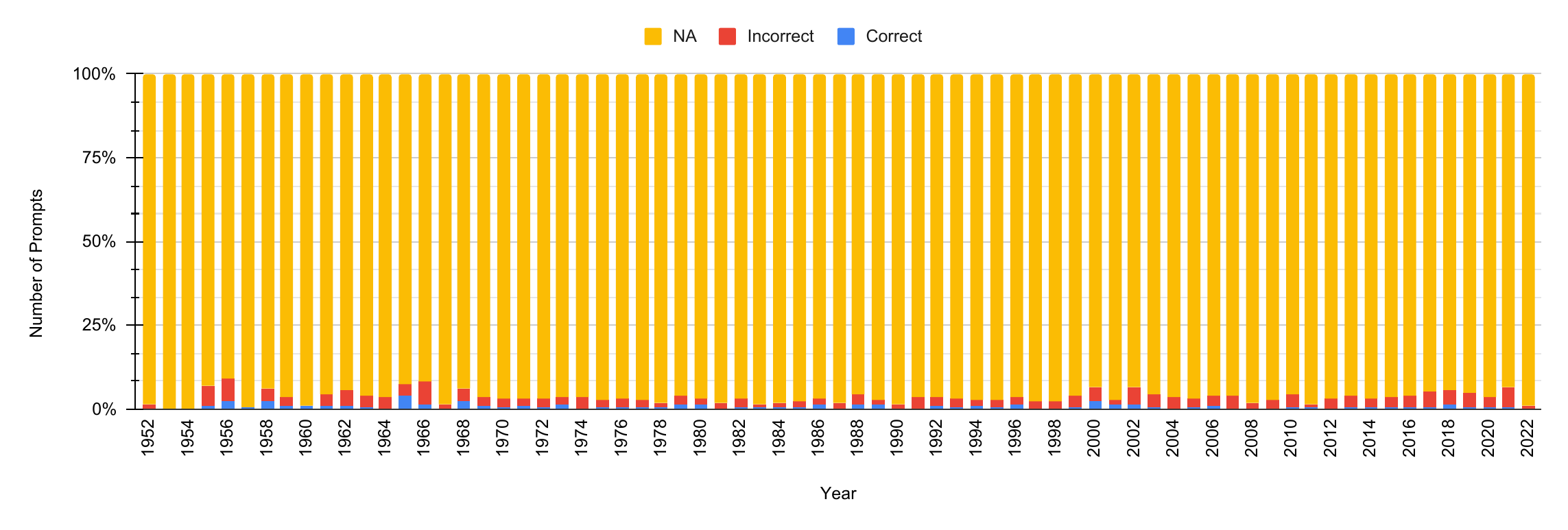}
\end{center}
\caption{Plot for the Min/Max-based metric ($MM$) as year-wise count (In percentage) for \textbf{random fine-tuning} for \texttt{phi-3-medium}.}
\label{fig:minmax-based-rn-phi-3-medium}
\end{figure*}

\begin{figure*}
\begin{center}
\includegraphics[width=0.9\linewidth]{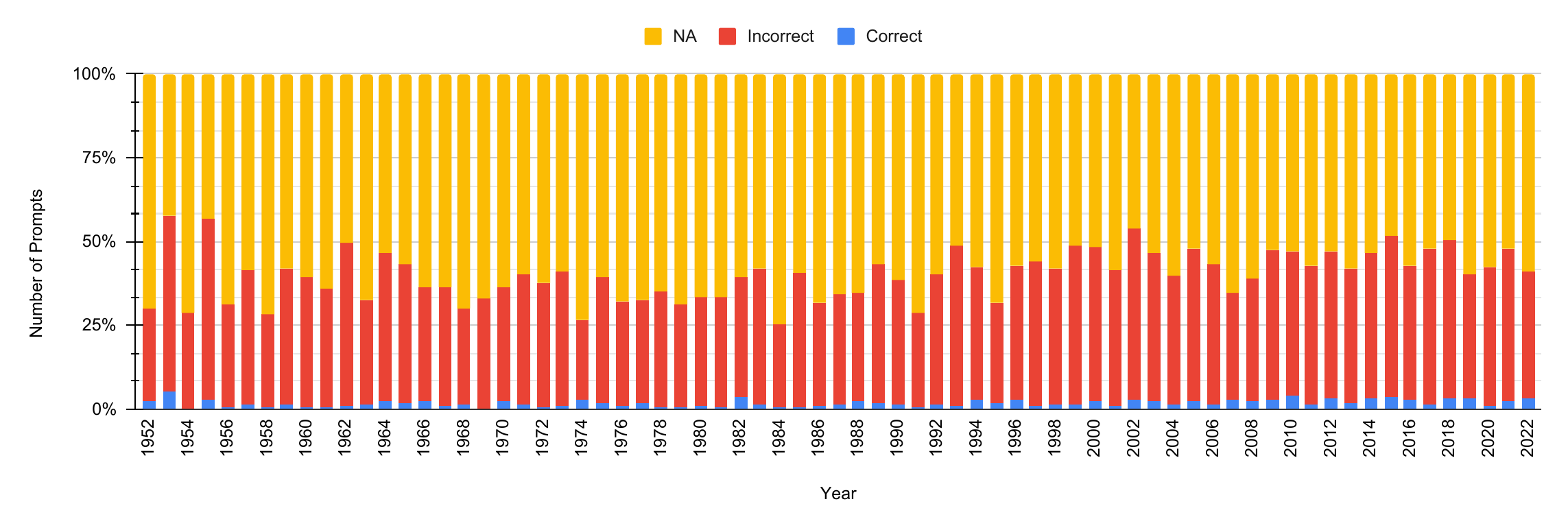}
\end{center}
\caption{Plot for the Range-based metric ($RB$) as year-wise count (In percentage) for \textbf{random fine-tuning} for \texttt{phi-3-medium}.}
\label{fig:rab-based-rn-phi-3-medium}
\end{figure*}

\begin{figure*}
\begin{center}
\includegraphics[width=0.9\linewidth]{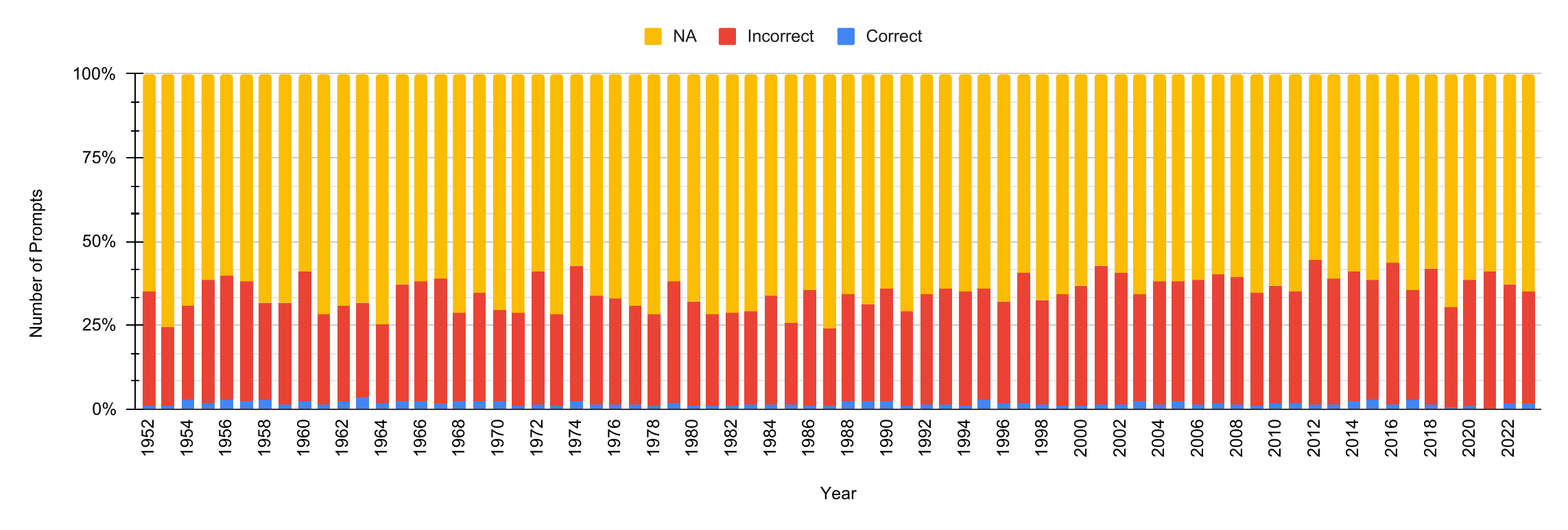}
\end{center}
\caption{Plot for the Trend-based metric ($TB$) as year-wise count (In percentage) for \textbf{random fine-tuning} for \texttt{phi-3-medium}.}
\label{fig:tb-based-rn-phi-3-medium}
\end{figure*}

\end{document}